\documentclass{article} % For LaTeX2e
\usepackage{arxiv,times}

\usepackage{graphicx}
\usepackage{amsmath,amsthm,amsfonts,amssymb}
\usepackage{booktabs}

\renewcommand{\eqref}[1]{(\ref{#1})}

\usepackage{hyperref}
\usepackage{url}
\hypersetup{hidelinks}
\usepackage{todonotes}
\title{Vanishing Initialization Turns Diagonal Tensor Gradient Flow into Greedy Matching}

\author{%
  Dmitry Yarotsky\\
  Applied AI Institute \& \\ Steklov Mathematical Institute\\
  Moscow, Russia \\
  \texttt{yarotsky@gmail.com} \\
    \And
  Eugene~Golikov\\
  Applied AI Institute\\
  Moscow, Russia \\
  \texttt{e.golikov@applied-ai.ru} \\
    \And
  Yaroslav~Gusev\\
  Applied AI Institute\\
  Moscow, Russia \\
  \texttt{i.gusev@applied-ai.ru} \\
}

\newtheorem{theorem}{Theorem}
\newtheorem{corollary}{Corollary}
\newtheorem{proposition}{Proposition}
\newtheorem{lemma}{Lemma}

\newcommand{\mainsectionwithappendix}[3]{%
    \section[#1]{%
        #1
        {\normalfont\small
        \hyperref[#3]{[Appendix~\ref*{#3}]}}%
    }%
    \label{#2}%
}

\newcommand{\appendixsectionwithmain}[3]{%
    \section[#1]{%
        #1
        {\normalfont\small
        \hyperref[#3]{[Section~\ref*{#3}]}}%
    }%
    \label{#2}%
}

\begin{document}

\maketitle

\begin{abstract}
We study gradient-flow (GF) learning of tensors from
vanishing initialization in canonical polyadic (CP)
factorization models. For tensors of order $\nu\geq3$ whose targets
are diagonal in mode-wise orthonormal bases, we prove that the limiting
learning dynamics follows a simple greedy bipartite
matching process. Each rank-one component of the
model, called an atom, can potentially learn each
rank-one component of the target, called a scalar
sector. Every atom--sector pair has an
initialization-dependent activation clock,
representing the time at which the atom would grow to macroscopic
size in that sector if the pair evolved in isolation.
Processing these clocks in increasing order and
selecting a pair whenever neither its atom nor
its sector has been selected before determines
the sequence of actual learning events. We prove
that the GF trajectory converges to the corresponding
staircase of learned target components. The task of
learning modular addition by a CP model with complex
weights falls exactly into this framework after
Fourier diagonalization. Its activation clocks are
independent and identically distributed, allowing
us to combine the GF reduction with known random
matching asymptotics and derive an explicit,
distribution-free large-system loss curve after time
reparametrization. We also obtain an asymptotic
formula relating this curve to the original GF
training time. For models with real weights,
conjugate Fourier modes instead form two-dimensional
blocks with richer internal structure. We characterize
their rank and border-rank cascade and conjecture
that the corresponding GF dynamics follows
interleaved multi-stage learning processes.
\end{abstract}

\mainsectionwithappendix
    {Introduction}
    {sec:introduction}
    {app:introduction}

Despite the remarkable practical success of neural
networks, our theoretical understanding of how
complex nonlinear models evolve under gradient-based
training remains limited. 
Detailed analytical descriptions of training trajectories
are available mainly in special settings. Important examples
include the linearized neural tangent kernel (NTK) regime
\citep{jacot2018neural,lee2019wide};
certain high-dimensional random models admitting
a dynamical mean-field description
\citep{mignacco2020dynamical,gerbelot2024rigorous};
and deep linear networks under suitable assumptions
on the data and initialization \citep{saxe2014exact}.
Extending such analytical understanding to broader
classes of nonlinear learning models remains a
major challenge.

An attractive complementary approach
is offered by the regime of vanishing weight initialization:
while allowing for nonlinear effects such as feature learning,
it can simplify training dynamics into a sequence of
incremental transitions between low-complexity states. 
Vanishing-initialization dynamics have been widely studied
in matrix factorization and deep linear networks
\citep{saxe2014exact,gissin2020implicit,li2021greedy,
jacot2021saddle,pesme2023saddle,jin2023incremental}.
For higher-order tensor models, there exist results
characterizing incremental and deflation-like learning
\citep{razin2021implicit,ge2021deflation}.
However, we are not aware of any rigorous descriptions
of \emph{complete learning trajectories of ordinary
gradient flow} for such models.

Higher-order tensor factorization presents additional
challenges compared with matrix factorization: general tensors
lack a canonical decomposition analogous to the matrix SVD
and have a substantially more complicated low-rank structure.
At the same time, important tensor-learning problems can possess
special algebraic structure.
For example, learning an $n$-ary arithmetic operation on a finite
set can naturally be formulated as learning an order-$(n+1)$
tensor, with one mode for each input and one for the output.

A prominent example is \emph{modular addition}, whose learning by
gradient-based models has attracted considerable attention
in recent years, particularly in connection with grokking
and Fourier feature learning
\citep{power2022grokking,nanda2023progress,
mohamadi2024grok,he2026modular}.
Despite this extensive research, we are not aware of
a rigorous analytical description of the complete
learning trajectory for modular addition under ordinary
gradient flow.

In this paper, we solve the problem of analytically
characterizing the complete learning trajectory in the
vanishing-initialization limit for an important class
of higher-order tensor targets: diagonal tensors.
Remarkably, this class includes modular addition
learned by a CP model with complex weights.
Our analysis proceeds in three steps.
First, for general diagonal targets, we prove that
the limiting gradient-flow dynamics is governed by
greedy bipartite matching between the rank-one
components of the model and those of the target. 
Each possible pairing has an initialization-dependent
potential activation time, called its \emph{activation
clock}. Processing these clocks in increasing order
and accepting a pair only if neither component has
been matched before determines the successive
learning events and the resulting loss trajectory. 
Second, for equal-amplitude target components,
we derive an explicit universal large-system loss
curve after a suitable time reparametrization.
Finally, we show that Fourier diagonalization
reduces complex modular addition exactly to this
equal-amplitude setting, allowing us to obtain
explicit asymptotic formulas for its loss dynamics.

An important feature of our result is its contrast with
matrix factorization. In the matrix case, small-initialization
dynamics is often governed by the singular-value structure.
In our higher-order tensor setting, by contrast, random
information from the microscopic initialization survives
in the vanishing-initialization limit and determines the
activation times of individual components. Thus, even
when all target components have equal amplitudes,
the order in which they are learned remains random.
Nevertheless, in the large-system limit, the resulting
loss trajectory approaches a universal deterministic
curve after a suitable time reparametrization.

The choice between complex and real weights is crucial
for modular addition. With complex weights, Fourier
transformation separates the target into one-dimensional
components, making the preceding theory directly
applicable. 
With real weights, conjugate Fourier modes instead
form two-dimensional blocks. We characterize their
rank and border-rank properties and conjecture
interleaved multi-stage GF cascades.

\paragraph{Relation to prior work.}
Previous studies of small-initialization tensor factorization
established greedy low-rank and deflation-like learning
\citep{razin2021implicit,ge2021deflation}.
Related work describes incremental learning through
auxiliary event-driven dynamics
\citep{kunin2025alternating}, while studies of modular
addition have characterized Fourier representations
and competition between frequencies
\citep{nanda2023progress,he2026modular}.
Our contribution differs in providing a rigorous
description of the complete limiting learning trajectory
of ordinary asymmetric CP gradient flow for diagonal
targets, including the ordering and timing of individual
learning events. For equal-amplitude targets, we further
connect these dynamics to known Random Edge matching
asymptotics \citep{arnosti2022greedy}.
A detailed review and comparison with previous work is given in
Appendix~\ref{app:introduction}.

\paragraph{Our contributions.}
\begin{itemize}
    \item \textbf{Complete learning trajectory for diagonal tensors.}
    We prove that vanishing-initialization gradient flow for
    diagonal order-$\nu$ CP tensor factorization, $\nu\geq3$,
    is governed by greedy bipartite matching with
    initialization-dependent activation clocks
    (Theorem~\ref{thm:greedy-matching-limit}).

    \item \textbf{Universal large-system loss dynamics.}
    For equal-amplitude target components, we derive an
    explicit deterministic large-system loss curve that,
    after time reparametrization, depends only on the
    ratio of model components to target components
    (Theorem~\ref{thm:universal-large-p}).

    \item \textbf{Learning modular addition with complex weights.}
    We show that Fourier diagonalization reduces this
    problem exactly to the equal-amplitude setting,
    giving a complete description of the limiting
    learning trajectory and explicit asymptotic formulas
    for its loss dynamics
    (Corollary~\ref{cor:addition-greedy} and
    Theorem~\ref{thm:explicit-addition-loss}).

    \item \textbf{Rank cascades for real-weight modular addition.}
    We characterize the rank and border-rank structure
    of the two-dimensional Fourier blocks arising in
    real-weight modular addition
    (Proposition~\ref{prop:real-block-rank-cascade})
    and conjecture that its gradient-flow dynamics
    follows interleaved multi-stage cascades.
\end{itemize}

A \hyperref[app:glossary]{glossary} of the terminology
used throughout the paper is provided at the end
of the appendices.

\mainsectionwithappendix{Setup and diagonal targets}{sec:setup}{app:setup}

\paragraph{Model and diagonal targets.}
Let the field $\mathbb{K}\in\{\mathbb{R},\mathbb{C}\}$ and $\nu\geq3$. We consider the task of fitting an order-$\nu$ tensor $T\in(\mathbb K^p)^{\otimes\nu}$ by a CP model with $H$ terms (\emph{atoms}):
\begin{equation}\label{eq:cp-model}
    \widehat T(U)
    =
    \sum_{k=1}^{H}
    \mathbf u_k^{(1)}\otimes\cdots\otimes \mathbf u_k^{(\nu)},
    \qquad
    U=(\mathbf u_k^{(a)})_{a\in[\nu],\,k\in[H]}
\in(\mathbb K^p)^{\nu H}\cong\mathbb K^{\nu\times H\times p}.
\end{equation}
The weights $U$ are learned by gradient flow (GF) with the squared Frobenius loss:
\begin{equation}
    L(U)
    =
    \frac12\|T-\widehat T(U)\|_F^2,
    \qquad
    \frac{d\mathbf u_k^{(a)}}{dt}
    =
    -\nabla_{\mathbf u_k^{(a)}}L(U).
    \label{eq:loss-gf}
\end{equation}
Here $\|\cdot\|^2_F$ denotes the squared Frobenius norm,
defined as the sum of squared
absolute values of the tensor entries. For $\mathbb K=\mathbb C$, gradients are taken with respect to the
underlying real Euclidean structure on $\mathbb C^p$; the equivalent
Wirtinger formulation is given in Appendix~\ref{app:setup}.

The summand
$\mathbf u_k^{(1)}\otimes\cdots\otimes\mathbf u_k^{(\nu)}$
will be called the $k$th \emph{atom}, and each $a\in[\nu]$ indexes a
\emph{mode}. We initialize the weights independently as
$\mathbf u_k^{(a)}(0)=\sigma\mathbf g_k^{(a)}$, with
$\mathbf g_k^{(a)}\sim\mathcal N_{\mathbb K}(0,I_p)$. For
$\mathbb K=\mathbb C$, we use the standard circularly symmetric convention
$Z=(X+iY)/\sqrt2$, where $X,Y\sim\mathcal N(0,1)$ independently, so that
$\mathbb E|Z|^2=1$.

We study diagonal targets
\begin{equation}
    T
    =
    \sum_{i=1}^{p}
    \lambda_i
    \mathbf e_i^{(1)}\otimes\cdots\otimes\mathbf e_i^{(\nu)},
    \qquad
    \lambda_i>0,
    \label{eq:diagonal-target}
\end{equation}
where $\{\mathbf e_i^{(a)}\}_{i=1}^p$ is an orthonormal basis in each
mode $a$. We assume all diagonal coefficients are nonzero; their signs or
complex phases can be absorbed into the corresponding basis vectors, giving
$\lambda_i>0$ without loss of generality. The orthonormal bases
$\{\mathbf e_i^{(a)}\}$ need not coincide with the original coordinate
bases; this will be important for modular addition, where they are Fourier
bases. 

We refer to the $i$th summand in \eqref{eq:diagonal-target} as the $i$th
\emph{scalar sector}. For each atom $k$ and sector $i$, we denote the
coordinates in the diagonalizing bases
as
$
    x_{ki}^{(a)}(t)
    =
    \langle \mathbf e_i^{(a)},\mathbf u_k^{(a)}(t)\rangle$ 
and 
$   
    g_{ki}^{(a)}
    =
    \langle \mathbf e_i^{(a)},\mathbf g_k^{(a)}\rangle.
$
By isotropy, the $g_{ki}^{(a)}$ are
independent standard scalar Gaussians over $\mathbb K$.

We first study the limit $\sigma\to0$ at fixed
$p,H,\nu$ (Section~\ref{sec:greedy}), then apply
the theory to complex modular addition
(Section~\ref{sec:modular-addition}) and the
proportional large-system limit $H/p\to\alpha$
(Section~\ref{sec:large-p}).
Finally, we examine non-scalar real Fourier sectors
(Section~\ref{sec:real-addition}).

\paragraph{Activation clocks.}
In the vanishing-initialization regime, while all factor
coordinates remain microscopic, the leading GF equations
take the approximate form  (see Appendix~\ref{app:setup})
\begin{equation}
    \frac{d x_{ki}^{(a)}}{dt}
    \approx
    \lambda_i
    \prod_{b\neq a}\overline{x_{ki}^{(b)}}.
    \label{eq:approx-microscopic-gf}
\end{equation}
Thus, the leading dynamics decouples across
atom--sector pairs.
 Since the coordinates are initially
of order $\sigma$ and their GF velocities are of order
$\sigma^{\nu-1}$, nontrivial learning occurs on physical
times $t$ of order $\sigma^{-(\nu-2)}$.
To obtain a nontrivial limiting loss trajectory as
$\sigma\to0$, we therefore introduce the \emph{slow time}
$\theta=\sigma^{\nu-2}t$, where $t$ denotes the original
\emph{physical time} of gradient flow.

Rescaling the factor coordinates by $\sigma^{-1}$
and introducing $s=\lambda_i\theta$ in
\eqref{eq:approx-microscopic-gf} leads to the
unit-amplitude auxiliary flow.
For $\mathbf g=(g^{(1)},\ldots,g^{(\nu)})
\in\mathbb K^\nu$, let
$\mathbf z(s;\mathbf g)
=(z^{(1)}(s),\ldots,z^{(\nu)}(s))$ solve
\begin{equation}
    \frac{d z^{(a)}}{ds}
    =
    \prod_{b\neq a}\overline{z^{(b)}},
    \qquad
    z^{(a)}(0)=g^{(a)},
    \qquad
    \tau_\nu(\mathbf g)
    :=
    \sup\{s>0:\|\mathbf z(s;\mathbf g)\|<\infty\},
    \label{eq:scalar-clock}
\end{equation}
where complex conjugation is immaterial over $\mathbb R$.

The blow-up time $\tau_\nu(\mathbf g)$ indicates
when the isolated pair leaves the microscopic regime.
In the original GF dynamics, an \emph{activation}
is a rapid transition from microscopic factor
coordinates, of order $\sigma$, to macroscopic
ones, of order one.
Writing
$\mathbf g_{ki}
=(g_{ki}^{(1)},\ldots,g_{ki}^{(\nu)})$,
we therefore define the \emph{activation clock}
of the atom--sector pair $(k,i)$ by
\begin{equation}
    \tau_{ki}
    =
    \frac{\tau_\nu(\mathbf g_{ki})}{\lambda_i}.
    \label{eq:candidate-clock}
\end{equation}
This is a potential activation time, expressed in
slow time: not every clock necessarily produces
an actual learning event in the full GF dynamics.

For Gaussian $\mathbf g$, $\tau_\nu(\mathbf g)$
is finite almost surely and has an atomless
distribution (Appendix~\ref{app:setup}).
Consequently, the $Hp$ candidate clocks are
independent and almost surely finite and distinct;
when the amplitudes $\lambda_i$ are equal,
they are i.i.d.

The pairs $(k,i)$ can now be viewed as edges of
the complete bipartite graph $K_{H,p}$
between model atoms and target sectors, each
carrying its activation clock
\eqref{eq:candidate-clock}.
Section~\ref{sec:greedy} shows that GF processes
these clocks greedily as $\sigma\to0$.

\mainsectionwithappendix{Vanishing initialization yields greedy matching}{sec:greedy}{app:greedy}

\paragraph{Greedy clock process.}
To describe the limiting GF dynamics, we introduce
the following auxiliary \emph{greedy matching} process
based on the activation clocks defined above. 
Recall that the candidate pairs $(k,i)$ form the edges of $K_{H,p}$, each
carrying the activation clock $\tau_{ki}$ from
\eqref{eq:candidate-clock}. Process these edges in increasing order of their
clocks. An edge is accepted if neither endpoint is incident to an earlier
accepted edge; otherwise it is discarded. Let $\mathcal M(\theta)$ be the
matching formed by accepted edges with clock at most $\theta$, and let
$\mathcal S(\theta)=\{i:(k,i)\in\mathcal M(\theta)\text{ for some }k\}$ be
the corresponding set of selected target sectors. The greedy loss trajectory
is
\begin{equation}
    L_*(\theta)
    =
    \frac12
    \sum_{i\notin\mathcal S(\theta)}
    \lambda_i^2 .
    \label{eq:greedy-loss}
\end{equation}
Each accepted edge represents the learning of one
target sector, decreasing the limiting loss by
$\lambda_i^2/2$.

For $\sigma>0$, write
$\widehat T^\sigma(\theta)
=\widehat T(U^\sigma(\sigma^{-(\nu-2)}\theta))$
and
$L^\sigma(\theta)
=L(U^\sigma(\sigma^{-(\nu-2)}\theta))$,
where $U^\sigma$ is the GF solution initialized at scale $\sigma$, and $t=\sigma^{-(\nu-2)}\theta$
is the physical time corresponding to slow time $\theta$.
Let $\mathcal T_*$ denote the finite set of clocks of the edges accepted by
the greedy process.

\begin{figure}[t]
    \centering
    \includegraphics[scale=0.8]{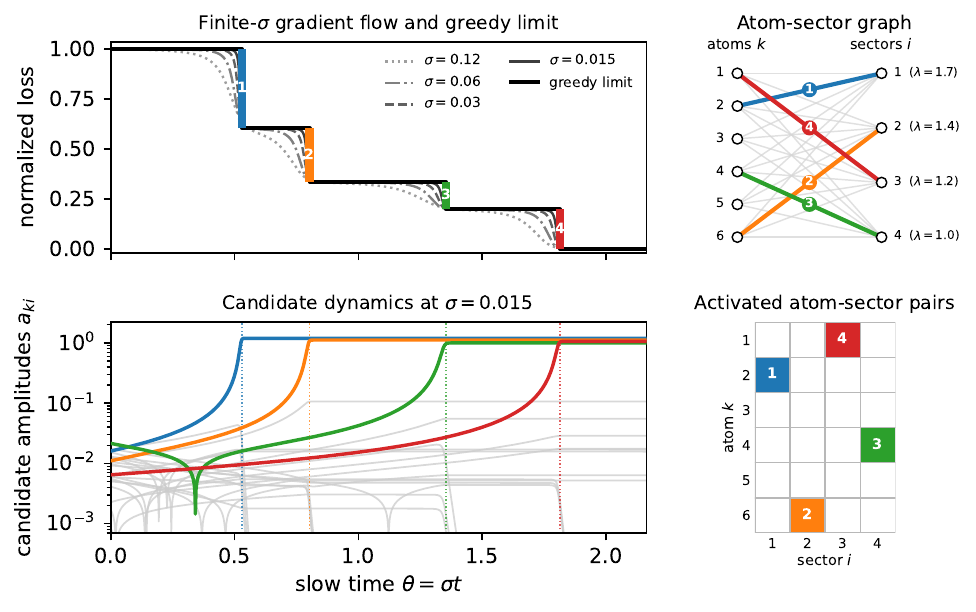}
    \caption{
\textbf{Vanishing-initialization learning as greedy matching for a diagonal tensor.}
Example with $p=4$, $H=6$, $\nu=3$, and unequal sector amplitudes
$\lambda=(1.7,1.4,1.2,1.0)$.
\emph{Top left:} normalized GF loss for several initialization scales
$\sigma$, plotted against the slow time $\theta=\sigma t$, together with the
predicted greedy-matching staircase. As $\sigma$ decreases, the finite-$\sigma$
transitions sharpen around the greedy activation times.
\emph{Bottom left:} candidate amplitudes
$a_{ki}=\big|\prod_{\mu=1}^\nu x_{ki}^{(\mu)}\big|^{1/\nu}$
for $\sigma=0.015$. Selected atom--sector pairs are highlighted in color and
losing candidates in gray. The third selected trajectory initially decreases because $A_0=P(0)<0$ in
its isolated scalar flow, producing the turning behavior described in
\eqref{eq:q-definition}--\eqref{eq:clock-integral-negative}.
\emph{Top right:} atom--sector graph, with accepted greedy matches colored and
numbered according to their activation order.
\emph{Bottom right:} accepted pairs in the
$H\times p$ atom--sector grid, using the same
colors and event numbers.
}    \label{fig:scalar-sectors}
\end{figure}

\begin{theorem}[Greedy matching limit]
\label{thm:greedy-matching-limit}
Under the setup of Section~\ref{sec:setup}, for almost every realization of
the Gaussian initialization, the following holds. For every compact
$K\subset[0,\infty)\setminus\mathcal T_*$,
\begin{equation}
    \sup_{\theta\in K}
    \Big\|
        \widehat T^\sigma(\theta)
        -
        \sum_{i\in\mathcal S(\theta)}
        \lambda_i
        \mathbf e_i^{(1)}\otimes\cdots\otimes\mathbf e_i^{(\nu)}
    \Big\|_F
    \longrightarrow 0
    \qquad
    \text{as }\sigma\to0,
    \label{eq:greedy-output-limit}
\end{equation}
and consequently
\begin{equation}
    \sup_{\theta\in K}
    \Big|
        L^\sigma(\theta)-L_*(\theta)
    \Big|
    \longrightarrow 0.
    \label{eq:greedy-loss-limit}
\end{equation}
Hence, away from activation times, the vanishing-initialization loss converges
to the staircase \eqref{eq:greedy-loss} determined by greedy matching of the
activation clocks.
\end{theorem}

Figure~\ref{fig:scalar-sectors} illustrates this limiting process.
Since all candidate clocks are finite almost surely, the greedy process
eventually produces a maximal matching of $K_{H,p}$. Hence, if $H\geq p$, it
selects every sector and $L_*(\theta)$ eventually vanishes; if $H<p$, it
selects exactly $H$ sectors. For unequal $\lambda_i$, these need not be the
$H$ largest coefficients because the outcome depends on the random initialization.

\paragraph{Proof idea.}
Before the first activation, all atom coordinates remain
microscopic and follow, to leading order, the decoupled
dynamics \eqref{eq:approx-microscopic-gf}.
Thus, all candidate pairs evolve according to the
auxiliary scalar flows \eqref{eq:scalar-clock} until
the smallest activation clock is reached.
Near that clock, the winning atom becomes macroscopic
over a vanishing interval of slow time and aligns with
the corresponding scalar sector, while the other
unactivated atoms remain microscopic. The learned
sector is thereby removed from the residual, and
the winning atom becomes asymptotically committed
to that sector.

Because the diagonal sectors are orthogonal, this
event does not change the leading microscopic dynamics
in the still-unclaimed sectors. Crucially, candidates
whose atom and sector remain available continue along
their original isolated trajectories: their clocks
are not reset. The next activation is therefore
the smallest remaining clock whose two endpoints
are still available. Iterating gives exactly the
greedy matching process above.
The full argument, including uniform control of the
microscopic regime and the fast activation transitions,
is given in Appendix~\ref{app:greedy}.

\mainsectionwithappendix
    {Complex modular addition}
    {sec:modular-addition}
    {app:modular-addition}

Let $\mathbb Z_p=\{0,\ldots,p-1\}$ and consider the modular-addition tensor
$T_p^{\mathrm{add}}\in\mathbb R^{p\times p\times p}$,
\[
    (T_p^{\mathrm{add}})_{x,y,z}
    =
    \mathbf 1\{x+y=z\pmod p\}.
\]
Writing $\omega=e^{2\pi i/p}$ and
$\boldsymbol{\varphi}_r(x)=p^{-1/2}\omega^{rx}$, the Fourier representation
of the modular Kronecker delta gives
\begin{equation}
    T_p^{\mathrm{add}}
    =
    \sqrt p
    \sum_{r\in\mathbb Z_p}
    \boldsymbol{\varphi}_r
    \otimes
    \boldsymbol{\varphi}_r
    \otimes
    \overline{\boldsymbol{\varphi}_r}.
    \label{eq:addition-fourier}
\end{equation}
For $\nu=3$, CP model \eqref{eq:cp-model} can equivalently be viewed as a shallow network
with $H$ multiplicative hidden units,
$f_U(x,y)_z=\sum_{k=1}^H
u_k^{(1)}(x)u_k^{(2)}(y)u_k^{(3)}(z)$,
trained full-batch on all $p^2$ input pairs $(x,y)$; its summed squared output loss is
exactly $L(U)$. Thus our results concern the training dynamics of this full-data model,
rather than train/test generalization or grokking.

Over $\mathbb C$, \eqref{eq:addition-fourier} is exactly an \emph{equal-amplitude}
diagonal order-$3$ tensor in the sense of Section~\ref{sec:setup}, with
$\mathbf e_r^{(1)}=\mathbf e_r^{(2)}=\boldsymbol{\varphi}_r$,
$\mathbf e_r^{(3)}=\overline{\boldsymbol{\varphi}_r}$, and
$\lambda_r=\sqrt p$.

By unitary invariance of the complex Gaussian initialization, the microscopic
atom--sector coordinates are i.i.d.\
$\mathbf g_{kr}\sim\mathcal N_{\mathbb C}(0,I_3)$. Setting
$\xi_{kr}:=\tau_3(\mathbf g_{kr})$, their activation clocks are
$\tau_{kr}=\xi_{kr}/\sqrt p$. We therefore use the rescaled time
$\vartheta=\sqrt p\,\theta=\sqrt p\,\sigma t$, in which the candidate
activation times are $\xi_{kr}$. Let $S_p(\vartheta)$ be the set of selected
Fourier sectors and $M_p(\vartheta)=|S_p(\vartheta)|$.

The following corollary is a direct application of
Theorem~\ref{thm:greedy-matching-limit} to complex
modular addition. It describes the limiting GF
output and loss for fixed $p,H$ as $\sigma\to0$. Write $\widehat T_p^{\mathrm{add},\sigma}(\vartheta)$ and
$L_p^{\mathrm{add},\sigma}(\vartheta)$ for the GF output and loss at
$\theta=\vartheta/\sqrt p$.

\begin{corollary}[Greedy Fourier-sector learning]
\label{cor:addition-greedy}
Fix $p$ and $H$. For almost every complex Gaussian initialization, away from
the accepted activation times,
\begin{equation}
    \lim_{\sigma\to0}
    \widehat T_p^{\mathrm{add},\sigma}(\vartheta)
    =
    \sqrt p
    \sum_{r\in S_p(\vartheta)}
    \boldsymbol{\varphi}_r
    \otimes
    \boldsymbol{\varphi}_r
    \otimes
    \overline{\boldsymbol{\varphi}_r},
    \qquad
    \frac{2L_{*,p}^{\mathrm{add}}(\vartheta)}{p^2}
    =
    1-\frac{M_p(\vartheta)}{p},
    \label{eq:addition-greedy-loss}
\end{equation}
where
$L_{*,p}^{\mathrm{add}}(\vartheta)
=\lim_{\sigma\to0}L_p^{\mathrm{add},\sigma}(\vartheta)$.
\end{corollary}

Since $\|T_p^{\mathrm{add}}\|_F^2=p^2$,
each learned Fourier sector decreases the
normalized loss by $1/p$.
We next study the large-system loss dynamics
of equal-amplitude diagonal targets as
$p,H\to\infty$ with $H/p\to\alpha\in(0,\infty)$.

\mainsectionwithappendix
    {Large-system limit}
    {sec:large-p}
    {app:large-p}

\paragraph{Random Edge matching.}
For equal-amplitude diagonal targets, the activation
clocks of all atom--sector pairs are i.i.d.\ and
continuous. Consequently, their ordering is uniformly
random. To describe the resulting greedy matching
process in the large-system limit, we use the
\emph{Random Edge} algorithm studied by
\citet{arnosti2022greedy} and defined as follows.

Let $p\to\infty$ with $H_p/p\to\alpha\in(0,\infty)$. Consider
the complete bipartite graph $K_{H_p,p}$
between $H_p$ atoms and $p$ sectors.
Assign each edge an independent uniform priority
in $[0,1]$, process the edges in increasing order
of priority, and accept an edge whenever both
endpoints are still unmatched.

This is a random-graph matching algorithm defined
independently of gradient flow; here we use it
to describe the limiting GF dynamics.

It is convenient to parametrize the matching process
by the \emph{edge exposure} $s$: at exposure $s$,
all edges with priority at most $s/p$ have been processed.
Each atom has $p$ incident edges and each sector has
$H_p$ incident edges. Thus, the expected numbers
of exposed edges per atom and per sector are
$s$ and $(H_p/p)s\to\alpha s$, respectively.
This scaling produces a nontrivial matching process
as $p\to\infty$.

Let $M_p^{\mathrm{RE}}(s)$ denote the number of
edges accepted by exposure $s$.
The following known result describes its
large-system limit.

\begin{proposition}[Random Edge matching limit]
\label{prop:random-edge-limit}
Suppose that $H_p/p\to\alpha\in(0,\infty)$. Then, for every fixed
$S<\infty$,
\begin{equation}
    \sup_{0\leq s\leq S}
    \left|
        \frac{M_p^{\mathrm{RE}}(s)}{p}
        -
        \bigl(1-\ell_\alpha(s)\bigr)
    \right|
    \xrightarrow{\mathbb P}
    0,
    \qquad
    \ell_\alpha(s)
    =
    \begin{cases}
        \displaystyle
        \frac{(\alpha-1)e^{-(\alpha-1)s}}
        {\alpha-e^{-(\alpha-1)s}},
        & \alpha\neq1, \\[3mm]
        \displaystyle
        \frac{1}{1+s},
        & \alpha=1.
    \end{cases}
    \label{eq:random-edge-limit}
\end{equation}
\end{proposition}
Here and below, $\xrightarrow{\mathbb P}$ denotes
convergence in probability.

Thus, $\ell_\alpha(s)$ is the limiting fraction of
unmatched sectors. The formula has a simple
interpretation. Since each accepted edge matches
one atom and one sector, the number of unmatched
atoms, normalized by $p$, is
$\alpha-1+\ell_\alpha(s)$.
A newly exposed edge produces a match when both
its endpoints are unmatched. In the large-system
limit, this leads to the differential equation
\begin{equation}
    \ell_\alpha'(s)
    =
    -\ell_\alpha(s)
    \bigl(\alpha-1+\ell_\alpha(s)\bigr),
    \qquad
    \ell_\alpha(0)=1,
    \label{eq:random-edge-ode}
\end{equation}
whose solution is precisely
\eqref{eq:random-edge-limit}.

\paragraph{Equal diagonal sectors.}
We now connect Random Edge matching to tensor gradient flow.
Consider the equal-amplitude diagonal target
\begin{equation}
    T_p
    =
    \lambda_p
    \sum_{i=1}^{p}
    \mathbf e_i^{(1)}
    \otimes\cdots\otimes
    \mathbf e_i^{(\nu)},
    \qquad
    \frac{H_p}{p}\to\alpha,
    \label{eq:equal-diagonal-target}
\end{equation}
where $\nu\geq3$ is fixed and $\lambda_p>0$.
Let $U_{p,\sigma}(t)$ denote the GF solution
initialized at scale $\sigma$.

Recall that the candidate activation clocks are
$\tau_{ki}=\xi_{ki}/\lambda_p$, where
$\xi_{ki}=\tau_\nu(\mathbf g_{ki})$ are i.i.d.
Let
\[
    F_\nu(x)
    :=
    \mathbb P\!\left(\tau_\nu(\mathbf g)\leq x\right),
    \qquad
    \mathbf g\sim\mathcal N_{\mathbb K}(0,I_\nu),
\]
be their common CDF.
Since this distribution is continuous, the
variables $F_\nu(\xi_{ki})$ are independent
uniform priorities in $[0,1]$.
Consequently, the greedy matching process of
Section~\ref{sec:greedy} is exactly the Random Edge
process when parametrized by these priorities.

At physical training time $t$, the slow time is
$\theta=\sigma^{\nu-2}t$, so all pairs with
$\xi_{ki}\leq\lambda_p\theta$ have clocks that
have been reached. 
The corresponding relation between physical training
time $t$ and edge exposure $s$, together with its
inverse, is 
\begin{equation}
    s_{p,\sigma}(t)
    =
    pF_\nu\!\left(
        \lambda_p\sigma^{\nu-2}t
    \right),
    \qquad
    t_{p,\sigma}(s)
    =
    \frac{1}{\sigma^{\nu-2}\lambda_p}
    F_\nu^{-1}\!\left(\frac{s}{p}\right),
    \label{eq:exposure-training-time}
\end{equation}
where $F_\nu^{-1}$ denotes the inverse clock CDF.

Combining the greedy matching limit of
Theorem~\ref{thm:greedy-matching-limit} with
Proposition~\ref{prop:random-edge-limit}
gives the following result.

\begin{theorem}[Large-system loss trajectory for equal diagonal sectors]
\label{thm:universal-large-p}
Let $H_p/p\to\alpha\in(0,\infty)$ and fix $\nu\geq3$. For every compact
$K\subset[0,\infty)$ and every $\varepsilon>0$,
\begin{equation}
    \lim_{p\to\infty}
    \limsup_{\sigma\downarrow0}
    \mathbb P\!\left(
        \sup_{s\in K}
        \left|
            \frac{
                2L\!\left(U_{p,\sigma}(t_{p,\sigma}(s))\right)
            }{
                p\lambda_p^2
            }
            -
            \ell_\alpha(s)
        \right|
        >\varepsilon
    \right)
    =
    0,
    \label{eq:equal-diagonal-large-system-limit}
\end{equation}
where $\ell_\alpha$ is given in
\eqref{eq:random-edge-limit}.
\end{theorem}

The theorem establishes a universal large-system loss
trajectory for equal-amplitude diagonal targets:
after parametrizing training by edge exposure $s$,
the normalized loss converges to $\ell_\alpha(s)$,
which depends only on the width ratio $\alpha$.
The tensor order $\nu$ and the microscopic activation
clock distribution affect only the conversion
\eqref{eq:exposure-training-time} to physical time.
Importantly, the limits are taken sequentially:
first $\sigma\to0$ at fixed $p,H_p$, and then
$p\to\infty$.

\begin{figure}[t]
    \centering
    \includegraphics[scale=0.74]{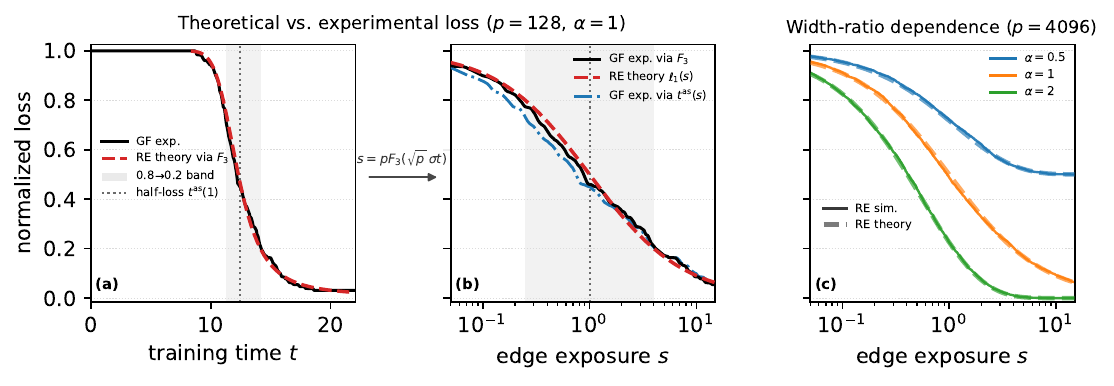}
    \caption{
\textbf{Large-system loss dynamics for complex modular addition.}
\emph{(a)} Normalized GF loss for $p=128$, $H=p$ ($\alpha=1$), and
$\sigma=0.005$ in physical training time $t$. The dashed curve is the
Random Edge (RE) prediction $\ell_1(s)=(1+s)^{-1}$ transported to physical time
using the numerically evaluated clock CDF $F_3$. The shaded region is the
explicit-asymptotic $0.8\to0.2$ loss-drop interval, and the dotted line marks
the corresponding half-loss time $t^{\mathrm{as}}_{p,\sigma}(1)$.
\emph{(b)} The same experiment in edge exposure
$s=pF_3(\sqrt p\,\sigma t)$. The reparameterized trajectory closely follows
the limiting Random Edge curve $\ell_1(s)$; using the explicit asymptotic
inverse $t^{\mathrm{as}}_{p,\sigma}(s)$ gives a similarly close
reparameterization. The shaded region is $s\in[1/4,4]$, where
$\ell_1(s)$ decreases from $0.8$ to $0.2$, and the dotted line marks $s=1$.
\emph{(c)} Dependence on the width ratio at $p=4096$. Simulated finite-$p$ Random Edge
trajectories for $\alpha=1/2,1,2$ are compared with the limiting curves
$\ell_\alpha(s)$.
}
    \label{fig:large-system}
\end{figure}

Figure~\ref{fig:large-system} illustrates these
results for complex modular addition. Panel (a)
compares GF dynamics with the prediction in physical
time, panel (b) shows the same dynamics in edge
exposure, and panel (c) illustrates the dependence
of the Random Edge limit on $\alpha$.

\paragraph{Explicit training time for complex modular addition.}
For complex modular addition, $\nu=3$ and every Fourier sector has amplitude
$\lambda_p=\sqrt p$. Hence \eqref{eq:exposure-training-time} specializes to
$s_{p,\sigma}(t)=pF_3(\sqrt p\,\sigma t)$. Although this exact conversion
involves the full cubic clock distribution, in the large-system limit it
admits an explicit asymptotic inverse. Indeed, for fixed edge exposure $s$, the relevant
CDF values are $s/p\to0$, so this inverse is
determined by the lower-tail asymptotics of $F_3$. 

\begin{theorem}[Explicit large-system loss trajectory for complex modular addition]
\label{thm:explicit-addition-loss}
Assume $H_p/p\to\alpha\in(0,\infty)$ and define
\begin{equation}
    t^{\mathrm{as}}_{p,\sigma}(s)
    =
    \frac{\sqrt3}{\sigma\sqrt p}
    \left[
        \log p
        +\frac12\log\log p
        +\log A_3
        -\log s
    \right]^{-1/2},
    \qquad
    A_3
    =
    \frac{5\sqrt{21\pi}}{27}.
    \label{eq:addition-explicit-training-time}
\end{equation}
Then, uniformly for $s$ in compact subsets of $(0,\infty)$,
\begin{equation}
    s_{p,\sigma}\!\left(
        t^{\mathrm{as}}_{p,\sigma}(s)
    \right)
    \longrightarrow s
    \qquad
    \text{as }p\to\infty.
    \label{eq:addition-asymptotic-inverse}
\end{equation}
Consequently, for every compact $K\subset(0,\infty)$ and every
$\varepsilon>0$,
\begin{equation}
    \lim_{p\to\infty}
    \limsup_{\sigma\downarrow0}
    \mathbb P\!\left(
        \sup_{s\in K}
        \left|
            \frac{
                2L\!\left(
                    U_{p,\sigma}\!\left(
                        t^{\mathrm{as}}_{p,\sigma}(s)
                    \right)
                \right)
            }{p^2}
            -
            \ell_\alpha(s)
        \right|
        >\varepsilon
    \right)
    =
    0,
    \label{eq:addition-explicit-loss-limit}
\end{equation}
where $\ell_\alpha$ is the Random Edge loss curve from
Proposition~\ref{prop:random-edge-limit}.
\end{theorem}

The proof is given in Appendix~\ref{app:large-p}.
The explicit asymptotic conversion
\eqref{eq:addition-explicit-training-time},
together with the universal loss curve $\ell_\alpha(s)$,
allows us to describe the large-system loss dynamics
in physical training time without numerically
evaluating the clock CDF $F_3$.

For every fixed $s\in(0,\infty)$, the corresponding
physical training time satisfies
\begin{equation}
    t^{\mathrm{as}}_{p,\sigma}(s)
    \sim
    \frac{\sqrt3}{
        \sigma\sqrt{p\log p}
    }.
    \label{eq:addition-learning-time}
\end{equation}
Interestingly, this leading-order expression does
not depend on $s$: all fixed exposure levels
correspond to approximately the same physical
training time. Their separation appears at a
smaller scale. More precisely, for fixed
$0<s_-<s_+<\infty$,
\begin{equation}
    t^{\mathrm{as}}_{p,\sigma}(s_+)
    -
    t^{\mathrm{as}}_{p,\sigma}(s_-)
    =
    \frac{
        \sqrt3\,\log(s_+/s_-)
    }{
        2\sigma\sqrt p\,(\log p)^{3/2}
    }
    \bigl(1+o(1)\bigr).
    \label{eq:addition-transition-width}
\end{equation}
Thus, the loss transition becomes increasingly
sharp relative to the characteristic learning time:
its duration is smaller by a factor of order
$1/\log p$.
At balanced width, $\ell_1(s)=(1+s)^{-1}$,
so $s=1$ corresponds to the asymptotic
half-loss time (see Figure~\ref{fig:large-system}).

\mainsectionwithappendix
    {Beyond scalar sectors: real Fourier blocks}
    {sec:real-addition}
    {app:real-addition}

The use of complex weights is crucial for our exact treatment of modular
addition in Sections~\ref{sec:modular-addition}--\ref{sec:large-p}.
Over $\mathbb C$, Fourier decomposition separates the target into scalar
rank-one sectors. 
Over $\mathbb R$, each pair of conjugate frequencies
$r$ and $-r$ gives a two-dimensional real
cosine--sine subspace $V_r^{(a)}$ in each mode $a$. Their tensor product
$V_r^{(1)}\otimes V_r^{(2)}\otimes V_r^{(3)}$
forms a \emph{non-scalar sector}. These sectors are mutually orthogonal,
so the target remains block-diagonal with respect to this coarser
decomposition. However, real modular addition falls outside the
scalar-sector theory of Theorem~\ref{thm:greedy-matching-limit}.

The restriction of the modular addition target to each such non-scalar sector is,
up to a factor $\sqrt{p/2}$, the tensor
\begin{equation}
    B
    =
    eee-eff-fef-ffe
    =
    \frac12
    \left[
        (\mathbf e+i\mathbf f)^{\otimes3}
        +
        (\mathbf e-i\mathbf f)^{\otimes3}
    \right],
    \label{eq:real-fourier-block}
\end{equation}
written in suitable orthonormal cosine--sine bases
$\{\mathbf e^{(a)},\mathbf f^{(a)}\}$ for the three modes.
Here, for example, $eff$ abbreviates
$\mathbf e^{(1)}\otimes\mathbf f^{(2)}\otimes\mathbf f^{(3)}$.
Together with the scalar sectors corresponding to self-conjugate
frequencies, these blocks give the full decomposition of the target.
Over $\mathbb C$, $B$ is a sum of two conjugate rank-one tensors;
over $\mathbb R$, it has CP rank three.

For a real tensor $X$, let $\operatorname{rank}_{\mathbb R}(X)$
denote its real CP rank. We define the
\emph{rank-$q$ approximation error} of $B$ by
\[
    d_q(B)^2
    :=
    \inf_{\operatorname{rank}_{\mathbb R}(X)\leq q}
    \|B-X\|_F^2 .
\]

\begin{figure}[t]
    \centering
    \includegraphics[scale=0.7]{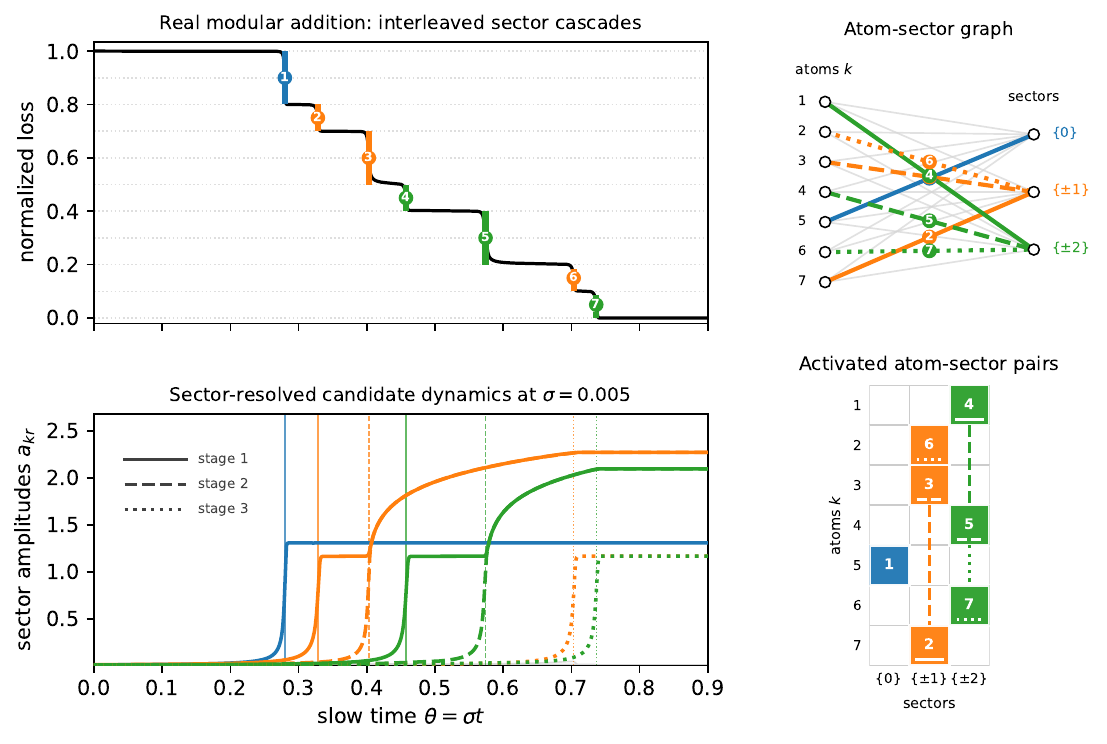}
    \caption{
\textbf{Interleaved learning of non-scalar Fourier sectors in real modular addition.}
Gradient flow for $p=5$, $H=7$, and $\sigma=0.005$, with real Fourier
sectors $\{0\}$, $\{\pm1\}$, and $\{\pm2\}$.
\emph{Top left:} normalized loss; numbered markers show the empirically
identified activation sequence, with drops occurring near multiples of $0.1$.
\emph{Bottom left:} sector-resolved amplitudes
$a_{kr}=\big(\prod_{\mu=1}^3
\|P_r^{(\mu)}\mathbf u_k^{(\mu)}\|_2\big)^{1/3}$.
Colors identify sectors and line styles distinguish successive stages.
Unlike in a scalar sector, previously activated atoms continue to evolve as
later atoms enter the same sector.
\emph{Top right:} atom--sector graph showing the corresponding activation
assignments.
\emph{Bottom right:} the same assignments as an atom--sector grid; numbers
give the global activation order, while dashed and dotted connections link
the first--second and second--third stages of each non-scalar cascade.
The two three-stage cascades are visibly interleaved.
}
    \label{fig:cascades}
\end{figure}

\begin{proposition}[Rank cascade of a real Fourier block]
\label{prop:real-block-rank-cascade}
For the real Fourier block $B$ defined in
\eqref{eq:real-fourier-block},
\begin{equation}
    d_0(B)^2=4,
    \qquad
    d_1(B)^2=3,
    \qquad
    d_2(B)^2=1,
    \qquad
    d_3(B)^2=0.
    \label{eq:real-block-rank-errors}
\end{equation}
The rank-two infimum is not attained.

More precisely, there is a phase-dependent family
$(E_\Theta,F_\Theta)$ such that
\begin{equation}
    B=E_\Theta-F_\Theta,
    \qquad
    \|E_\Theta\|_F^2=1,
    \qquad
    \|F_\Theta\|_F^2=3,
    \label{eq:real-block-EF}
\end{equation}
where $E_\Theta$ is rank one and $F_\Theta$ is a sum of three mutually
orthogonal rank-one tensors. Moreover, $-F_\Theta$ lies in the closure of
the real rank-two tensors and can be approached only through degenerating
rank-two representations with diverging factors.
\end{proposition}

The construction of $E_\Theta$ and $F_\Theta$
and the proof of the proposition are given in
Appendix~\ref{app:real-addition}. 
The proposition identifies the following sequence
of optimal residual levels as the allowed rank
increases, with the rank-two stage understood
as a limiting approximation:
\begin{equation}
    B
    \longrightarrow
    -F_{\Theta_1}
    \longrightarrow
    E_{\Theta_2}
    \longrightarrow
    0,
    \qquad
    \frac{\|R\|_F^2}{\|B\|_F^2}
    :
    1
    \longrightarrow
    \frac34
    \longrightarrow
    \frac14
    \longrightarrow
    0.
    \label{eq:real-sector-cascade}
\end{equation}
The first intermediate residual $-F_{\Theta_1}$ has
real rank three but border rank two: approaching
it with rank-two tensors requires diverging
factors even though the represented tensor
remains bounded.

Figure~\ref{fig:cascades} shows one representative
small-initialization GF trajectory for real modular
addition. Motivated by this behavior, we conjecture
that GF forms interleaved multi-stage cascades:
successive atoms enter a sector while previously
activated atoms reorganize, and cascades from
different Fourier sectors can interleave.
For $p=5$, each of the two nonzero sectors carries
$2/5$ of the target energy, so the relative levels
in \eqref{eq:real-sector-cascade} correspond to
successive normalized-loss drops of $1/10$,
$2/10$, and $1/10$, as seen in the figure.

\section{Discussion}
\label{sec:discussion}

\paragraph{Results and significance.}
We rigorously characterize the vanishing-initialization dynamics of diagonal
order-$\nu$ CP tensor factorization for $\nu\geq3$. We prove that the limiting
learning trajectory is governed by greedy bipartite matching between model
atoms and target sectors. This gives an explicit trajectory-level description
of a nonlinear higher-order factorization problem. For equal-amplitude
sectors, our results further yield an explicit large-system loss law: after
reparametrization by edge exposure, the limiting normalized loss depends only
on the atom-to-sector ratio. To our knowledge, such an exact matching
representation of the GF trajectory, together with the resulting large-system loss law, has not appeared in previous analyses. As a concrete
application, complex modular addition falls exactly within this framework,
yielding an explicit description of its large-$p$ loss dynamics.

\paragraph{Why tensor learning differs from the matrix case.}
In matrix and deep-linear settings, small-initialization dynamics is typically
organized by spectral structure. In our tensor case, the elementary learning events are
instead indexed by atom--sector pairs, and their ordering retains microscopic
initialization randomness in the vanishing-initialization limit. The resulting
global trajectory is therefore combinatorial rather than purely spectral.

\paragraph{Microscopic randomness and universality in edge exposure.}
Despite the microscopic randomness, equal-amplitude sectors have a
distribution-free large-system limit after reparametrization by the common
activation-clock CDF. The normalized loss converges to the Random Edge curve
$\ell_\alpha$, which depends only on the asymptotic atom-to-sector ratio
$\alpha$. The tensor order $\nu$ affects only the conversion from physical
time to edge exposure through the activation-clock distribution. More
broadly, we expect the same matching law in other factorized models with
i.i.d.\ continuous atom--sector activation times and the same
one-atom/one-sector exclusivity structure.

\paragraph{Beyond scalar sectors.}
Our analysis reveals a sharp distinction between complex and real modular
addition. With complex weights, Fourier diagonalization reduces the problem
exactly to the scalar-sector theory; with real weights, conjugate modes form
two-dimensional sectors with richer internal geometry. We characterize their
rank and border-rank cascade and conjecture that gradient flow follows these
stages through successive atom activations, reorganization of earlier atoms,
and potentially large parameter growth while the represented tensor remains
bounded. This suggests that a broader theory should replace one clock per
sector by a description of its internal cascade and its interleaving with
other sectors.

\paragraph{Limitations and open directions.}
Our exact theory assumes diagonal scalar sectors, gradient flow, and squared
tensor loss, and takes the vanishing-initialization limit before the
large-system limit. This iterated limit gives a clean formulation while
retaining the nontrivial macroscopic loss law, but it does not quantify how
small $\sigma$ must be as $p$ grows. Obtaining such a condition would require
estimates uniform in $p$, in particular control of increasingly close
activation events. Moreover, exact agreement of every event order may be
stronger than necessary for the large-system loss curve: finite-$\sigma$
reorderings of near-tied events may disappear macroscopically. Determining
the weakest scaling $\sigma=\sigma(p)$ sufficient for the same loss limit is
an interesting open problem, along with discrete-time optimization and
multidimensional-sector dynamics.

\subsection*{AI use statement}

In this work, we used generative AI tools to assist with research
brainstorming, the formulation and refinement of mathematical arguments,
mathematical derivations and proof development, the design and analysis of
numerical experiments, literature search, and the drafting and revision of
the manuscript. Generative AI was also used to assist with experimental and
figure-generation code and to improve the clarity, organization, and
presentation of the text.

AI-generated mathematical arguments were treated as provisional: the authors
independently checked the derivations, assumptions, theorem statements, and
proofs used in the final manuscript. Numerical results and AI-assisted code
were likewise inspected and verified by the authors. References suggested by
generative AI were checked against the original publications before
inclusion. We reviewed all AI-assisted work and take full responsibility for
the final content of this paper, including all mathematical claims,
experimental results, citations, and conclusions.

\phantomsection
\addcontentsline{toc}{section}{References}

\bibliography{iclr2027_conference}
\bibliographystyle{iclr2027_conference}

\newpage

\tableofcontents

\newpage

\appendix

The appendix mirrors the organization of the main text:
Appendix~A$j$ provides proofs, technical details, and additional material
for Section~$j$.

\renewcommand{\thesection}{A\arabic{section}}
\renewcommand{\thesubsection}{\thesection.\arabic{subsection}}
\renewcommand{\thesubsubsection}{\thesubsection.\arabic{subsubsection}}
\setcounter{section}{0}

\appendixsectionwithmain
    {Introduction: context and related work}
    {app:introduction}
    {sec:introduction}

This appendix places our results in the literature on small-initialization
dynamics, incremental low-rank learning, tensor factorization, and structured
algorithmic tasks. The closest precedents establish several ingredients of
the picture studied here: small initialization can produce incremental or
saddle-to-saddle learning; matrix and tensor factorizations can exhibit
greedy low-rank behavior; and Fourier features can emerge sequentially in
models trained on modular and group operations. Our contribution is not any
of these phenomena in isolation. The distinctive point is that, for
diagonal order-$\nu$ CP tensor learning with $\nu\geq3$, we resolve the full
vanishing-initialization trajectory into random microscopic activation
events, identify their global competition exactly with greedy bipartite
matching, and then solve the corresponding equal-amplitude sector large-system loss
trajectory.

\subsection{Small initialization and incremental trajectories}
\label{app:related-matrix}

\paragraph{Prior work.}
Factorized matrix and linear models have long provided tractable settings
for studying optimization dynamics and implicit regularization. Exact
solutions for deep linear networks already exhibit plateaus, rapid
transitions, and sequential learning of spectral modes
\citep{saxe2014exact}. Related work on deep matrix factorization connected gradient-based
optimization to implicit low-rank structure
\citep{arora2019implicit}. Small initialization was subsequently connected
more directly to incremental learning.  
\citet{gissin2020implicit} study the role of depth in producing incremental
learning, while \citet{li2021greedy} interpret infinitesimal-initialization
matrix factorization through \emph{Greedy Low-Rank Learning}, in which
optimization proceeds through solutions of increasing rank.
A particularly close dynamical comparison is
\citet{jacot2021saddle}, who analyze deep linear networks near vanishing
initialization. They rigorously characterize the first escape from the zero
saddle along an effectively low-rank path and propose a subsequent
saddle-to-saddle dynamics of increasing bottleneck rank. Importantly for our
setting, they also identify the qualitative distinction between depth two
and larger depth: for $L=2$ the escape time is logarithmic in the inverse
initialization scale, whereas for $L>2$ the higher-order homogeneous
degeneracy produces a polynomial scale of order
$\alpha^{-(L-2)}$.

More recent work gives complete incremental trajectories in several
degree-two settings. For diagonal linear regression,
\citet{berthier2023incremental} establish successive coordinate activation,
and \citet{pesme2023saddle} characterize the full
vanishing-initialization limiting trajectory by an explicit recursive
saddle-to-saddle algorithm, including the limiting jump times. In matrix
sensing, \citet{jin2023incremental} rigorously track all successive
rank-learning phases and show that gradient descent approaches solutions of
increasing rank until the target rank is reached. Thus neither incremental
learning nor a complete small-initialization trajectory is new in itself.

\paragraph{Our contribution and differences.}
The local higher-order escape mechanism in our tensor model is closely
related to that of deep linear networks of depth greater than two, rather
than being specific to tensors. For an order-$\nu$ CP model, the leading
balanced scalar dynamics near the origin has the form
$\dot a\propto a^{\nu-1}$, so activations occur on the scale
$\sigma^{-(\nu-2)}$, directly analogous to the
$\alpha^{-(L-2)}$ scale in depth-$L$ linear networks
\citep{jacot2021saddle}. The distinction therefore lies not in the
higher-order homogeneity itself, but in how the resulting escape processes
are organized globally.

\emph{Microscopic candidate clocks.}
For a diagonal tensor target, the near-origin dynamics decomposes into
candidate atom--sector pairs. After rescaling time by
$\sigma^{\nu-2}$, the normalized Gaussian initialization of each pair
survives as a finite random activation clock. Hence the singular limit
retains an entire family of competing random candidate times. This differs,
for example, from the first-escape result of
\citet{jacot2021saddle}, where under a simple leading singular value the
globally fastest escape direction is selected by the top singular mode,
up to the internal symmetries of the factorization.

\emph{Competition becomes matching.}
The candidate events are not independent at the macroscopic level: once an
atom learns a sector, both that atom and that sector become unavailable to
later candidates. We prove that these two exclusivity constraints resolve
the complete sequence of activations exactly as greedy bipartite matching.
The limiting trajectory is therefore not only an ordered sequence of ranks,
coordinates, or saddles, but an explicit combinatorial process generated by
the microscopic initialization.

\emph{From a finite event process to an explicit large-system loss trajectory.}
Finally, in the equal-amplitude sector regime the candidate clocks are i.i.d., and
their ordering turns the matching process into Random Edge on the complete
atom--sector bipartite graph. Taking the large-system limit then produces an
explicit deterministic loss curve depending only on the asymptotic
atom-to-sector ratio. Neither the matrix and diagonal-linear trajectory
results above nor the deep-linear saddle analysis derive an analogous
random matching representation together with a solvable large-system loss
trajectory.

\subsection{Greedy and deflation dynamics in tensor factorization}
\label{app:related-tensor}

\paragraph{Prior work.} The diagonal targets in \eqref{eq:diagonal-target} belong to the standard
orthogonally decomposable (\emph{odeco}) class over $\mathbb R$ and to its
unitary analogue (\emph{udeco}) over $\mathbb C$; our setting allows the
orthonormal factors to depend on the mode
\citep{boralevi2017orthogonal}. The symmetric odeco case underlies classical
tensor-power and designed deflation methods
\citep{anandkumar2014tensor}. 
A closely related line of work studies how small initialization can induce
incremental low-rank behavior in tensor factorization.
\citet{razin2021implicit} analyze CP tensor factorization dynamically and
show that small components evolve much more slowly than larger ones, leading
to a greedy low tensor-rank bias. Under suitable conditions, they prove that
the induced tensor trajectory remains close to a rank-one trajectory over
prescribed finite time or distance scales, and support a broader
component-by-component learning picture experimentally. Their later work
extends this implicit low-rank perspective to hierarchical tensor
factorizations and deep convolutional networks
\citep{razin2022hierarchical}.

A complementary line interprets overparameterized tensor decomposition as an
implicit deflation process. \citet{wang2020beyond} show that suitably
designed gradient-based methods can exploit low-rank tensor structure beyond
the lazy regime. More directly, \citet{ge2021deflation} study the symmetric odeco
specialization with a symmetric CP parameterization. They show that a
slightly modified gradient flow follows a deflation-like process and recovers
the target components, providing a rigorous tensor analogue of greedy
low-rank learning.

These works already establish the main qualitative phenomenon that motivates
our analysis: small initialization can cause tensor models to learn
low-rank structure incrementally rather than fitting all components
simultaneously. We therefore do not claim novelty for greedy tensor learning,
for sequential recovery of tensor components, or for deflation-like
behavior by itself.

\paragraph{Our contribution and differences.}
Our result instead treats the nonsymmetric odeco/udeco setting with
mode-dependent orthonormal factors and resolves the complete event structure
of ordinary asymmetric CP gradient flow.

\emph{From a dominant component to all activation events.}
For each atom--sector pair, the microscopic initialization determines a
candidate activation clock. We prove the singular-limit dynamics through
every accepted activation, including the persistence of already learned
pairs and the suppression of candidates made inadmissible by earlier events.
Thus the theorem determines the entire finite-system sequence rather than
only the emergence of an initial low-rank component or a qualitative
deflation order.

\emph{Random ordering instead of amplitude ordering.}
The distinction is especially clear for equal-amplitude targets. In a
classical deflation picture, a larger target component naturally has an
advantage and may be learned first. In our equal-amplitude sector setting there is no
such deterministic hierarchy: all target sectors have the same amplitude,
and the order remains random in the limit because it is inherited from the
microscopic Gaussian initialization. The diagonal structure makes this
randomness explicit through the collection of activation clocks.

\emph{Atom--sector competition rather than rank growth alone.}
The model contains multiple atoms competing for multiple target sectors.
Once an atom has captured one sector, that atom cannot independently capture
another, while a learned sector is no longer available to other atoms. These
two-sided resource constraints are not described by a scalar notion of
tensor rank or by ordinary deflation. We prove that they organize the
complete trajectory exactly as greedy bipartite matching.

\emph{A solvable many-sector limit.}
For equal-amplitude sectors, the matching representation can itself be
analyzed as the number of atoms and sectors grows. After transforming the
i.i.d.\ clocks to uniform priorities, the finite-system process becomes
Random Edge on a complete bipartite graph, and its large-system limit yields
an explicit deterministic loss trajectory. To our knowledge, prior tensor
factorization results do not derive such a random matching representation of
the full gradient-flow trajectory or an explicit corresponding 
large-system loss law.

\subsection{Event-driven feature learning and structured operations}
\label{app:related-feature-learning}

\paragraph{Prior work.}
A broader line of work studies small-initialization training as a sequence of
discrete feature-learning events. Most directly,
\citet{kunin2025alternating} introduce \emph{Alternating Gradient Flows}
(AGF), an effective description in which dormant neurons first compete for
useful directions and activation is followed by a faster loss-minimization
stage. This framework predicts successive activation times and loss drops
across several models. For diagonal linear networks, the authors rigorously
relate AGF to the vanishing-initialization limit of gradient flow, while in
more general architectures AGF serves as an effective approximation to the
observed staircase dynamics.

Related work studies representation-theoretic structure in learned group
operations. \citet{chughtai2023toy} characterize circuits based on irreducible
representations for finite-group composition, while
\citet{marchetti2026sequential} obtain an irrep-by-irrep activation picture
within an AGF-based analysis. Other work shows that random initialization can
drive individual neurons toward spectral representations of finite groups
\citep{he2026spectral}. These results are particularly relevant to our
discussion beyond scalar sectors, since non-Abelian irreducible
representations naturally lead to multidimensional Fourier blocks.

Modular addition itself has become a standard setting for studying grokking,
Fourier representations, and learning dynamics
\citep{power2022grokking,nanda2023progress,mohamadi2024grok}.
\citet{gromov2023grokking} derive analytic feature maps and weight
configurations solving modular-arithmetic tasks in simple two-layer networks
and show that gradient-based training can recover such representations.
\citet{morwani2024feature} theoretically characterize the emergence of
Fourier features in stylized models through margin maximization, while
\citet{he2026modular} show dynamically that Fourier frequencies compete
within neurons and that the winning frequency depends on
initialization-dependent spectral magnitude and phase. At the same time,
\citet{zhong2023clock} show that neural networks can learn qualitatively
different algorithms for modular addition under changes of initialization
and hyperparameters. Thus neither Fourier learning nor
initialization-dependent competition is specific to our model, and the
Fourier-sector mechanism need not describe arbitrary architectures.

\paragraph{Our contribution and differences.}
Our use of structured operations is narrower but more explicit: modular
addition is represented directly as a target tensor, so its Fourier
structure can be inserted into the exact CP gradient-flow analysis.

\emph{An exact reduction of ordinary gradient flow.}
The event-driven viewpoint itself is not new, and AGF provides a general
framework for it. Our result instead proves, for the diagonal CP model, that
the original gradient flow converges in the vanishing-initialization limit
to a specific discrete process through all activation events. No auxiliary
alternating dynamics is introduced: the activation clocks and the subsequent
matching rule are derived directly from the GF equations.

\emph{Fourier sectors as an exactly diagonal tensor problem.}
For complex modular addition, Fourier transformation does more than reveal
features learned by the network: it diagonalizes the target tensor exactly
into equal-amplitude scalar rank-one sectors. The modular-addition result is
therefore a direct corollary of the general diagonal-tensor theorem. In this
representation, the competition among Fourier modes is not merely observed
or inferred from trained weights; it is identified with the competition
among explicitly defined atom--sector clocks.

\emph{From feature competition to a solvable large-system trajectory.}
Initialization-dependent Fourier competition appears in prior work, but in
our setting the equal amplitudes make the microscopic ordering purely random
and the atom--sector exclusivity turns it into Random Edge matching. This
additional combinatorial structure allows us to pass from individual feature
selection events to an explicit large-system loss trajectory. To our
knowledge, prior work on modular or group operations does not derive an
analogous explicit large-system loss law from a random matching
representation of the training dynamics.

\emph{Complex versus real representations.}
The exact scalar-sector reduction also clarifies why the choice of field
matters. Over $\mathbb C$, the Fourier characters of modular addition give
independent one-dimensional sectors. Over $\mathbb R$, conjugate frequencies
must be combined into two-dimensional blocks, leading to the multi-stage
rank and border-rank cascades of Section~\ref{sec:real-addition}. This
provides a concrete example of the multidimensional internal sector dynamics
that also arise naturally in representation-theoretic analyses of more
general group operations.

\subsection{Greedy matching and the large-system limit}
\label{app:related-matching}

\paragraph{Prior work.}
Once the tensor dynamics has been reduced to a matching process, the
remaining large-system problem connects to classical probabilistic
combinatorics. The \emph{Random Edge} algorithm processes graph edges in a
uniformly random order and accepts an edge whenever both endpoints are still
unmatched. Average-case behavior of random-order greedy matching was studied
already by \citet{dyer1993average}, and sparse bipartite asymptotics for
Random Edge and related algorithms were derived by
\citet{arnosti2022greedy}. A standard way to obtain such deterministic
large-system trajectories is the differential-equation method for random
processes developed by
\citet{wormald1995differential,wormald1999differential}.

Accordingly, we do not claim novelty for Random Edge itself, for random-order
greedy matching, or for the sparse bipartite matching asymptotic used in
Section~\ref{sec:large-p}. Proposition~\ref{prop:random-edge-limit} is stated
and proved in the normalization needed for our application, but its closed
form agrees with the known sparse bipartite Random Edge formula of
\citet{arnosti2022greedy}.

\paragraph{Our contribution and differences.}
The novel connection runs in the opposite direction: we show that a singular
limit of nonlinear tensor gradient flow generates this random combinatorial
process.

\emph{From activation clocks to Random Edge.}
For equal-amplitude tensor sectors, the candidate atom--sector clocks are
i.i.d. Applying their common CDF converts them into independent uniform edge
priorities without changing their ordering. The full vanishing-initialization
tensor trajectory is therefore exactly Random Edge on the complete
atom--sector bipartite graph, observed as the priority threshold increases.
Thus the random matching process is not introduced as an approximation or
external algorithm; it is derived from the microscopic GF dynamics.

\emph{An explicit large-system loss trajectory.}
When $H_p/p\to\alpha$, the fraction of unmatched sectors converges to a
deterministic function $\ell_\alpha(s)$ of the edge exposure parameter.
Consequently, the normalized tensor loss has an explicit limiting trajectory
that depends only on the asymptotic atom-to-sector ratio. The tensor order
and the microscopic clock distribution disappear from the matching law
itself and enter only through the conversion between physical training time
and edge exposure.

\emph{Explicit physical time for modular addition.}
For complex modular addition, we go one step further. The Fourier sectors are
equal-amplitude and the scalar clocks are cubic. By deriving the lower tail
of this clock distribution and asymptotically inverting the exposure map, we
obtain an explicit physical learning-time scale in addition to the universal
matching trajectory.

\emph{Novelty boundary.}
To our knowledge, previous work does not characterize a
vanishing-initialization CP tensor gradient-flow trajectory as a random greedy
matching process, nor derive from such a reduction an explicit 
large-system loss law. Our contribution is therefore the chain
\[
    \text{CP GF}
    \;\longrightarrow\;
    \text{random activation clocks}
    \;\longrightarrow\;
    \text{greedy matching}
    \;\longrightarrow\;
    \text{large-system loss},
\]
rather than any of these ingredients individually.

\appendixsectionwithmain{Setup and diagonal targets: technical preliminaries}{app:setup}{sec:setup}

This appendix provides the technical facts underlying the setup of
Section~\ref{sec:setup}. We first specify the GF convention for complex
parameters, then derive the microscopic rescaling and the decoupled scalar
flows, and finally establish the properties of the activation clocks used in
Section~\ref{sec:greedy}.

\subsection{Gradient flow with complex weights}
\label{app:complex-gf}

\paragraph{Real Euclidean convention.}
When $\mathbb K=\mathbb C$, the loss is real valued but the learned weights
are complex. We define GF by regarding a complex vector as a real vector of
twice the dimension. Namely, if
$\mathbf u=\mathbf x+i\mathbf y\in\mathbb C^p$, with
$\mathbf x,\mathbf y\in\mathbb R^p$, we identify
\[
    \mathbb C^p
    \cong
    \mathbb R^{2p},
    \qquad
    \mathbf u=\mathbf x+i\mathbf y
    \longleftrightarrow
    (\mathbf x,\mathbf y).
\]
For $\mathbf u,\mathbf v\in\mathbb C^p$, let
\[
    \langle \mathbf u,\mathbf v\rangle_{\mathbb C}
    :=
    \mathbf u^*\mathbf v
    =
    \sum_{j=1}^p \overline{u_j}v_j,
\]
where $\mathbf u^*$ denotes the conjugate transpose of $\mathbf u$. The real
Euclidean inner product under the above identification is therefore
\[
    \langle \mathbf u,\mathbf v\rangle_{\mathbb R}
    :=
    \operatorname{Re}
    \langle \mathbf u,\mathbf v\rangle_{\mathbb C}.
\]
Equivalently, if $u_j=x_j+iy_j$ and $v_j=r_j+is_j$, then
\[
    \langle \mathbf u,\mathbf v\rangle_{\mathbb R}
    =
    \sum_{j=1}^p (x_jr_j+y_js_j).
\]

Let $f:\mathbb C^p\to\mathbb R$ be differentiable when regarded as a
function of the $2p$ real variables $(\mathbf x,\mathbf y)$. For a direction
$\mathbf h\in\mathbb C^p$, its directional derivative is
\[
    Df(\mathbf u)[\mathbf h]
    :=
    \left.
    \frac{d}{d\varepsilon}
    f(\mathbf u+\varepsilon\mathbf h)
    \right|_{\varepsilon=0},
    \qquad
    \varepsilon\in\mathbb R.
\]
The gradient $\nabla_{\mathbf u}f\in\mathbb C^p$ is the unique vector
satisfying
\begin{equation}
    Df(\mathbf u)[\mathbf h]
    =
    \operatorname{Re}
    \langle \nabla_{\mathbf u}f,\mathbf h\rangle_{\mathbb C}
    \qquad
    \text{for every }\mathbf h\in\mathbb C^p.
    \label{eq:complex-gradient-definition}
\end{equation}
Writing $u_j=x_j+iy_j$, this is simply
\[
    (\nabla_{\mathbf u}f)_j
    =
    \frac{\partial f}{\partial x_j}
    +
    i\frac{\partial f}{\partial y_j}.
\]
Hence
\[
    \frac{d\mathbf u}{dt}
    =
    -\nabla_{\mathbf u}f
\]
is exactly the ordinary real GF for
$(\mathbf x,\mathbf y)\in\mathbb R^{2p}$, written in complex notation.

For comparison with the commonly used Wirtinger notation, define
\[
    \frac{\partial}{\partial u_j}
    :=
    \frac12
    \left(
        \frac{\partial}{\partial x_j}
        -
        i\frac{\partial}{\partial y_j}
    \right),
    \qquad
    \frac{\partial}{\partial\overline{u_j}}
    :=
    \frac12
    \left(
        \frac{\partial}{\partial x_j}
        +
        i\frac{\partial}{\partial y_j}
    \right).
\]
Then, for a real-valued function $f$,
\begin{equation}
    \nabla_{\mathbf u}f
    =
    2\frac{\partial f}{\partial\overline{\mathbf u}},
    \qquad
    \frac{\partial f}{\partial\overline{\mathbf u}}
    :=
    \left(
        \frac{\partial f}{\partial\overline{u_j}}
    \right)_{j=1}^p.
    \label{eq:gradient-wirtinger}
\end{equation}
We use the real-gradient definition
\eqref{eq:complex-gradient-definition} throughout; no additional complex
optimization convention is required.

\paragraph{Basic identities.}
We record a few elementary consequences that will be used below. First, for
$f(\mathbf u)=\frac12\|\mathbf u-\mathbf v\|_2^2$, where
$\mathbf v\in\mathbb C^p$ is fixed,
\[
    Df(\mathbf u)[\mathbf h]
    =
    \operatorname{Re}
    \langle \mathbf u-\mathbf v,\mathbf h\rangle_{\mathbb C},
    \qquad
    \nabla_{\mathbf u}f
    =
    \mathbf u-\mathbf v.
\]
Thus the familiar gradient of a squared Euclidean distance is unchanged when
the variables are complex, provided the gradient is understood in the real
sense above.

The same convention applies to tensors. For complex tensors $A,B$ of the
same shape, we use the Hermitian Frobenius inner product
\[
    \langle A,B\rangle_F
    :=
    \sum_{i_1,\ldots,i_\nu}
    \overline{A_{i_1\cdots i_\nu}}
    B_{i_1\cdots i_\nu},
    \qquad
    \|A\|_F^2
    =
    \langle A,A\rangle_F.
\]
Consequently, if $R(\varepsilon)$ is a differentiable tensor-valued curve,
then
\begin{equation}
    \left.
    \frac{d}{d\varepsilon}
    \frac12\|R(\varepsilon)\|_F^2
    \right|_{\varepsilon=0}
    =
    \operatorname{Re}
    \left\langle
        R(0),
        \left.\frac{dR}{d\varepsilon}\right|_{\varepsilon=0}
    \right\rangle_F.
    \label{eq:frobenius-differential}
\end{equation}

We will also repeatedly differentiate a rank-one tensor with respect to one
of its factors. If
\[
    \Phi(\mathbf u^{(1)},\ldots,\mathbf u^{(\nu)})
    =
    \mathbf u^{(1)}\otimes\cdots\otimes\mathbf u^{(\nu)},
\]
then perturbing only the mode-$a$ factor gives
\begin{equation}
    D_{\mathbf u^{(a)}}\Phi[\mathbf h]
    =
    \mathbf u^{(1)}\otimes\cdots\otimes
    \mathbf u^{(a-1)}\otimes\mathbf h\otimes
    \mathbf u^{(a+1)}\otimes\cdots\otimes\mathbf u^{(\nu)}.
    \label{eq:rank-one-directional-derivative}
\end{equation}

Finally, for an order-$\nu$ tensor
$A\in(\mathbb C^p)^{\otimes\nu}$ and vectors
$\mathbf w^{(b)}\in\mathbb C^p$, define the mode-$a$ contraction
\[
    A\times_{b\neq a}\overline{\mathbf w^{(b)}}\in\mathbb C^p
\]
by
\begin{equation}
    \left[
        A\times_{b\neq a}\overline{\mathbf w^{(b)}}
    \right]_{i_a}
    :=
    \sum_{\{i_b:\,b\neq a\}}
    A_{i_1\cdots i_\nu}
    \prod_{b\neq a}\overline{w^{(b)}_{i_b}}.
    \label{eq:tensor-contraction-definition}
\end{equation}
With this definition,
\begin{equation}
    \left\langle
        A,\,
        \mathbf w^{(1)}\otimes\cdots\otimes
        \mathbf w^{(a-1)}\otimes\mathbf h\otimes
        \mathbf w^{(a+1)}\otimes\cdots\otimes\mathbf w^{(\nu)}
    \right\rangle_F
    =
    \left\langle
        A\times_{b\neq a}\overline{\mathbf w^{(b)}},
        \mathbf h
    \right\rangle_{\mathbb C}.
    \label{eq:contraction-inner-product}
\end{equation}
Equations~\eqref{eq:frobenius-differential},
\eqref{eq:rank-one-directional-derivative}, and
\eqref{eq:contraction-inner-product} are the only complex-calculus identities
needed to derive the CP GF equations below.

\subsection{Exact GF equations and microscopic scaling}
\label{app:microscopic-scaling}

Using the contraction defined in
\eqref{eq:tensor-contraction-definition}, differentiating the loss in
\eqref{eq:loss-gf} gives the exact GF equation
\begin{equation}
    \frac{d\mathbf u_k^{(a)}}{dt}
    =
    T\times_{b\neq a}\overline{\mathbf u_k^{(b)}}
    -
    \sum_{j=1}^{H}
    \mathbf u_j^{(a)}
    \prod_{b\neq a}
    \left\langle
        \mathbf u_k^{(b)},\mathbf u_j^{(b)}
    \right\rangle_{\mathbb K}.
    \label{eq:exact-gf}
\end{equation}
Here the first term is the contraction of the target with the other
$\nu-1$ factors of atom $k$, while the second is the corresponding
contraction of the current model output. We use
$\langle\cdot,\cdot\rangle_{\mathbb K}$ for the Euclidean inner product over
$\mathbb R$ and the Hermitian inner product over $\mathbb C$.

Set
\[
    \mathbf u_k^{(a)}(t)
    =
    \sigma\mathbf v_k^{(a)}(\theta),
    \qquad
    \theta=\sigma^{\nu-2}t.
\]
Substituting into \eqref{eq:exact-gf} and dividing by
$\sigma^{\nu-1}$ yields
\begin{equation}
    \frac{d\mathbf v_k^{(a)}}{d\theta}
    =
    T\times_{b\neq a}\overline{\mathbf v_k^{(b)}}
    -
    \sigma^\nu
    \sum_{j=1}^{H}
    \mathbf v_j^{(a)}
    \prod_{b\neq a}
    \left\langle
        \mathbf v_k^{(b)},\mathbf v_j^{(b)}
    \right\rangle_{\mathbb K}.
    \label{eq:rescaled-exact-gf}
\end{equation}
This explains the slow-time scaling used in the main text: while the
rescaled factors remain $O(1)$, the target-driven dynamics is $O(1)$ in
$\theta$, whereas interaction through the current model output is suppressed
by $\sigma^\nu$.

For the diagonal target \eqref{eq:diagonal-target}, define the rescaled
sector coordinates
\[
    z_{ki}^{(a)}(\theta)
    =
    \left\langle
        \mathbf e_i^{(a)},\mathbf v_k^{(a)}(\theta)
    \right\rangle_{\mathbb K}.
\]
Projecting \eqref{eq:rescaled-exact-gf} onto
$\mathbf e_i^{(a)}$ gives the exact coordinate equation
\begin{equation}
    \frac{d z_{ki}^{(a)}}{d\theta}
    =
    \lambda_i
    \prod_{b\neq a}\overline{z_{ki}^{(b)}}
    -
    \sigma^\nu
    \sum_{j=1}^{H}
    z_{ji}^{(a)}
    \prod_{b\neq a}
    \left\langle
        \mathbf v_k^{(b)},\mathbf v_j^{(b)}
    \right\rangle_{\mathbb K}.
    \label{eq:exact-sector-rescaled}
\end{equation}
Thus, before any rescaled factor becomes large, the second term vanishes as
$\sigma\to0$, and different atom--sector pairs evolve independently to
leading order.

More precisely, let $\mathbf z(s;\mathbf g)$ denote the isolated flow
\eqref{eq:scalar-clock} and define
\begin{align}
    \widetilde z_{ki}^{(a)}(\theta)
    &=
    z^{(a)}(\lambda_i\theta;\mathbf g_{ki}),
    \label{eq:tilde_z}\\
    \tau_{\min}
    &:=
    \min_{k,i}
    \frac{\tau_\nu(\mathbf g_{ki})}{\lambda_i}.
\end{align}

\begin{proposition}[Microscopic decoupling]
\label{prop:microscopic-decoupling}
Fix $p,H,\nu$ and a realization of the initialization for which
$\tau_{\min}>0$. For every $\Theta<\tau_{\min}$,
\begin{equation}
    \max_{k,i,a}
    \sup_{0\leq\theta\leq\Theta}
    \left|
        z_{ki}^{(a)}(\theta)
        -
        \widetilde z_{ki}^{(a)}(\theta)
    \right|
    \longrightarrow 0
    \qquad
    \text{as }\sigma\to0.
    \label{eq:microscopic-decoupling-app}
\end{equation}
\end{proposition}

\begin{proof}
On $[0,\Theta]$, all isolated trajectories
$\widetilde{\mathbf z}_{ki}$ remain bounded because
$\Theta<\tau_{\min}$. Since the mode-wise bases are complete, define
\[
    \widetilde{\mathbf v}_k^{(a)}(\theta)
    =
    \sum_{i=1}^p
    \widetilde z_{ki}^{(a)}(\theta)\mathbf e_i^{(a)}.
\]
These vectors remain bounded on $[0,\Theta]$ and satisfy the
$\sigma=0$ version of \eqref{eq:rescaled-exact-gf}.

Collect all rescaled factors into $\mathbf v^\sigma$ and the corresponding
limiting factors into $\widetilde{\mathbf v}$. The right-hand side of
\eqref{eq:rescaled-exact-gf} has the form
\[
    \mathcal F(\mathbf v)
    +
    \sigma^\nu\mathcal G(\mathbf v),
\]
where $\mathcal F$ and $\mathcal G$ are polynomial vector fields. On a fixed
bounded neighborhood of the limiting trajectory, $\mathcal F$ is Lipschitz
and $\mathcal G$ is uniformly bounded. Standard continuous dependence of ODE
solutions, or directly Gronwall's inequality, therefore gives
\[
    \sup_{0\leq\theta\leq\Theta}
    \left\|
        \mathbf v^\sigma(\theta)
        -
        \widetilde{\mathbf v}(\theta)
    \right\|
    =
    O(\sigma^\nu).
\]
Projecting onto the diagonalizing basis vectors proves
\eqref{eq:microscopic-decoupling-app}.
\end{proof}

Proposition~\ref{prop:microscopic-decoupling} shows directly why the
candidate clock associated with $(k,i)$ is
$\tau_\nu(\mathbf g_{ki})/\lambda_i$: after the change of variable
$s=\lambda_i\theta$, its limiting microscopic dynamics is precisely the
unit-amplitude flow \eqref{eq:scalar-clock}. The proposition applies only
strictly before the first activation. Through and after an activation, the
rescaled factors become large enough that the $\sigma^\nu$ interaction term
can no longer be neglected; this regime is analyzed in
Appendix~\ref{app:greedy}.

\paragraph{Atom-wise balance invariant.}
The full interacting CP flow has an exact balance invariant that will be
used later. For every atom $k$ and every pair of modes $a,b\in[\nu]$,
\begin{equation}
    \left\|
        \mathbf u_k^{(a)}(t)
    \right\|^2
    -
    \left\|
        \mathbf u_k^{(b)}(t)
    \right\|^2
    =
    \left\|
        \mathbf u_k^{(a)}(0)
    \right\|^2
    -
    \left\|
        \mathbf u_k^{(b)}(0)
    \right\|^2.
    \label{eq:full-flow-balance-invariant}
\end{equation}
Indeed, the exact GF equation gives
\begin{equation}
    \frac{d}{dt}
    \left\|
        \mathbf u_k^{(a)}
    \right\|^2
    =
    2\operatorname{Re}
    \left\langle
        T-\widehat T,
        \bigotimes_{c=1}^{\nu}
        \mathbf u_k^{(c)}
    \right\rangle_F,
    \label{eq:full-flow-norm-rate}
\end{equation}
whose right-hand side is independent of $a$. Subtracting the equations for
two modes proves \eqref{eq:full-flow-balance-invariant}. Under the Gaussian
initialization,
\begin{equation}
    \left\|
        \mathbf u_k^{(a)}(t)
    \right\|^2
    -
    \left\|
        \mathbf u_k^{(b)}(t)
    \right\|^2
    =
    \sigma^2
    \left(
        \left\|\mathbf g_k^{(a)}\right\|^2
        -
        \left\|\mathbf g_k^{(b)}\right\|^2
    \right).
    \label{eq:full-flow-balance-invariant-initialized}
\end{equation}

\subsection{The isolated scalar flow}
\label{app:scalar-flow}

We next record the properties of the isolated flow
\begin{equation}
    \frac{d z^{(a)}}{ds}
    =
    \prod_{b\neq a}\overline{z^{(b)}},
    \qquad
    z^{(a)}(0)=g^{(a)},
    \label{eq:scalar-flow-app}
\end{equation}
that are needed in the main argument. Let
$\mathbf z(s)=(z^{(1)}(s),\ldots,z^{(\nu)}(s))$,
$\mathbf g=(g^{(1)},\ldots,g^{(\nu)})$, and
\[
    P(s)=\prod_{a=1}^{\nu}z^{(a)}(s).
\]
A direct differentiation gives, for every $a$,
\begin{equation}
    \frac{d}{ds}|z^{(a)}|^2
    =
    2\operatorname{Re}P(s),
    \qquad
    \frac{dP}{ds}
    =
    \sum_{a=1}^{\nu}
    \prod_{b\neq a}|z^{(b)}|^2.
    \label{eq:scalar-invariants}
\end{equation}
In particular,
\begin{equation}
    |z^{(a)}(s)|^2-|z^{(b)}(s)|^2
    =
    |g^{(a)}|^2-|g^{(b)}|^2
    \label{eq:balanced-invariants}
\end{equation}
for every $a,b$, while $\operatorname{Im}P(s)$ is constant and
$\operatorname{Re}P(s)$ is nondecreasing.

These identities reduce the evolution of the magnitudes to one scalar
variable. Write
\[
    \rho_a=|g^{(a)}|^2,
    \qquad
    P(0)=A_0+iB.
\]
By \eqref{eq:balanced-invariants}, there is a scalar function $q(s)$ such that
\begin{equation}
    |z^{(a)}(s)|^2
    =
    \rho_a+q(s)
    \qquad
    \text{for all }a,
    \label{eq:q-definition}
\end{equation}
with $q(0)=0$. Since $q'=2\operatorname{Re}P$ and
$|P|^2=\prod_a(\rho_a+q)$,
\begin{equation}
    \bigl(q'(s)\bigr)^2
    =
    4D_{\mathbf g}(q(s)),
    \qquad
    D_{\mathbf g}(q)
    :=
    \prod_{a=1}^{\nu}(\rho_a+q)-B^2.
    \label{eq:q-equation}
\end{equation}

For generic initial data, this gives an explicit integral representation of
the activation clock. Let $\rho_{\min}=\min_a\rho_a$. If $A_0>0$, then $q$
increases from zero to infinity and
\begin{equation}
    \tau_\nu(\mathbf g)
    =
    \frac12
    \int_0^\infty
    \frac{dq}{\sqrt{D_{\mathbf g}(q)}}.
    \label{eq:clock-integral-positive}
\end{equation}
If $A_0<0$, then $q$ first decreases to the unique turning point
$q_*\in[-\rho_{\min},0)$ satisfying $D_{\mathbf g}(q_*)=0$, and then
increases to infinity, so
\begin{equation}
    \tau_\nu(\mathbf g)
    =
    \frac12
    \int_{q_*}^{0}
    \frac{dq}{\sqrt{D_{\mathbf g}(q)}}
    +
    \frac12
    \int_{q_*}^{\infty}
    \frac{dq}{\sqrt{D_{\mathbf g}(q)}}.
    \label{eq:clock-integral-negative}
\end{equation}
For complex Gaussian initialization,
$\operatorname{Im}P(0)\neq0$ almost surely, so
$q_*>-\rho_{\min}$ in the second case. For real Gaussian initialization,
$B=0$ and, almost surely, the minimum $\rho_{\min}$ is attained at a unique
mode, giving $q_*=-\rho_{\min}$.

\begin{proposition}[Properties of the scalar clock]
\label{prop:scalar-clock-properties}
Let $\nu\geq3$ and
$\mathbf g\sim\mathcal N_{\mathbb K}(0,I_\nu)$. Then
\[
    0<\tau_\nu(\mathbf g)<\infty
    \qquad
    \text{almost surely}.
\]
Moreover, for every $c>0$,
\begin{equation}
    \tau_\nu(c\mathbf g)
    =
    c^{-(\nu-2)}\tau_\nu(\mathbf g).
    \label{eq:clock-homogeneity}
\end{equation}
\end{proposition}

\begin{proof}
Gaussian initialization almost surely has nonzero coordinates and pairwise
distinct values of $|g^{(a)}|^2$. In the complex case, $A_0=0$ is also a
null event. Hence the integral representations
\eqref{eq:clock-integral-positive}--\eqref{eq:clock-integral-negative}
apply almost surely.

The only possible divergence relevant to forward blow-up occurs as
$q\to\infty$. Since
\[
    D_{\mathbf g}(q)
    =
    q^\nu(1+o(1)),
\]
the integrand is asymptotic to $q^{-\nu/2}$. Therefore
\[
    \int^\infty q^{-\nu/2}\,dq<\infty
\]
precisely when $\nu>2$. The turning-point integral in
\eqref{eq:clock-integral-negative} is also finite: almost surely the root
$q_*$ is simple, so
$D_{\mathbf g}(q)\asymp q-q_*$ near $q_*$. Thus $q(s)$ diverges to infinity
at a finite forward time, and hence
$\|\mathbf z(s)\|\to\infty$ at the same time. Local existence gives
$\tau_\nu(\mathbf g)>0$.

For homogeneity, if $\mathbf z(s;\mathbf g)$ solves
\eqref{eq:scalar-flow-app}, then
\[
    \widetilde{\mathbf z}(s)
    =
    c\,\mathbf z(c^{\nu-2}s;\mathbf g)
\]
solves the same equation with initial condition $c\mathbf g$. Its blow-up
time is therefore
$c^{-(\nu-2)}\tau_\nu(\mathbf g)$.
\end{proof}

\paragraph{Why $\nu\geq3$ is different from matrix factorization.}
For $\nu=2$, the isolated system is linear:
$\dot z^{(1)}=\overline{z^{(2)}}$ and
$\dot z^{(2)}=\overline{z^{(1)}}$. Equivalently, the large-$q$ integrand in
\eqref{eq:clock-integral-positive} behaves as $q^{-1}$, whose integral
diverges. Thus the isolated matrix dynamics grows exponentially rather than
blowing up at a finite rescaled time, and escape from a small initialization
occurs on a logarithmic scale. The finite random activation clock underlying
our tensor reduction is therefore specific to $\nu\geq3$.

\subsection{Distribution of the candidate clocks}
\label{app:clock-distribution}

It remains to justify the probabilistic properties stated after
\eqref{eq:candidate-clock}. By Gaussian isotropy and independence of the
initial weights, the vectors
\[
    \mathbf g_{ki}
    =
    \bigl(g_{ki}^{(1)},\ldots,g_{ki}^{(\nu)}\bigr),
    \qquad
    k\in[H],\quad i\in[p],
\]
are independent standard Gaussian vectors in $\mathbb K^\nu$.

\begin{proposition}[Distribution of the scalar and candidate clocks]
\label{prop:clock-atomless}
For $\nu\geq3$, the CDF
\[
    F_\nu(x)
    :=
    \mathbb P\!\left(
        \tau_\nu(\mathbf g)\leq x
    \right),
    \qquad
    \mathbf g\sim\mathcal N_{\mathbb K}(0,I_\nu),
\]
is a continuous strictly increasing bijection
\[
    F_\nu:[0,\infty)\longrightarrow[0,1).
\]
Consequently, the candidate clocks
\[
    \tau_{ki}
    =
    \frac{\tau_\nu(\mathbf g_{ki})}{\lambda_i}
\]
are independent, finite, and almost surely pairwise distinct. If all
$\lambda_i$ are equal, they are i.i.d.
\end{proposition}

\begin{proof}
Write the Gaussian vector in radial form
\[
    \mathbf g
    =
    R\boldsymbol{\omega},
    \qquad
    R=\|\mathbf g\|,
    \qquad
    \boldsymbol{\omega}
    =
    \frac{\mathbf g}{\|\mathbf g\|}.
\]
For both real and complex standard Gaussian vectors, $R$ has a density that
is strictly positive on $(0,\infty)$, and $R$ is independent of
$\boldsymbol{\omega}$.

By Proposition~\ref{prop:scalar-clock-properties},
\[
    0<
    \tau_\nu(\mathbf g)
    <
    \infty
    \qquad
    \text{almost surely},
\]
and the clock homogeneity
\eqref{eq:clock-homogeneity} gives
\begin{equation}
    \tau_\nu(\mathbf g)
    =
    R^{-(\nu-2)}
    \tau_\nu(\boldsymbol{\omega}).
    \label{eq:clock-radial}
\end{equation}
Since $R\in(0,\infty)$ almost surely, it follows that
\begin{equation}
    0<
    \tau_\nu(\boldsymbol{\omega})
    <
    \infty
    \qquad
    \text{for almost every }\boldsymbol{\omega}.
    \label{eq:clock-direction-finite}
\end{equation}

We now establish the claimed properties of the CDF
\[
    F_\nu(x)
    =
    \mathbb P\!\left(
        \tau_\nu(\mathbf g)\leq x
    \right).
\]

\paragraph{Step 1: atomlessness and continuity.}
Fix $x\in(0,\infty)$. Conditional on any direction
$\boldsymbol{\omega}$ satisfying
\eqref{eq:clock-direction-finite}, the radial representation
\eqref{eq:clock-radial} gives
\[
    \tau_\nu(\mathbf g)
    =
    R^{-(\nu-2)}
    \tau_\nu(\boldsymbol{\omega}).
\]
As a function of $R\in(0,\infty)$, the right-hand side is continuous and
strictly decreasing. Hence
\[
    \tau_\nu(\mathbf g)=x
\]
can occur only at the single radial value
\[
    R
    =
    \left(
        \frac{\tau_\nu(\boldsymbol{\omega})}{x}
    \right)^{1/(\nu-2)}.
\]
Because $R$ has an absolutely continuous distribution,
\[
    \mathbb P\!\left(
        \tau_\nu(\mathbf g)=x
        \,\middle|\,
        \boldsymbol{\omega}
    \right)
    =
    0
\]
for almost every $\boldsymbol{\omega}$. Averaging over
$\boldsymbol{\omega}$ gives
\begin{equation}
    \mathbb P\!\left(
        \tau_\nu(\mathbf g)=x
    \right)
    =
    0.
    \label{eq:clock-no-atoms}
\end{equation}
Also,
$\mathbb P(\tau_\nu(\mathbf g)=0)=0$ because the clock is strictly positive
almost surely. Thus the distribution of $\tau_\nu(\mathbf g)$ is atomless,
and consequently its CDF $F_\nu$ is continuous on $[0,\infty)$.

\paragraph{Step 2: strict monotonicity.}
Let
\[
    0\leq x<y<\infty.
\]
We show that
\[
    F_\nu(y)>F_\nu(x).
\]

First suppose $x>0$. Conditional on a direction
$\boldsymbol{\omega}$ satisfying
\eqref{eq:clock-direction-finite}, the event
\[
    x<
    \tau_\nu(\mathbf g)
    \leq y
\]
is, by the radial representation
\eqref{eq:clock-radial}, equivalent to
\begin{equation}
    \left(
        \frac{\tau_\nu(\boldsymbol{\omega})}{y}
    \right)^{1/(\nu-2)}
    \leq
    R
    <
    \left(
        \frac{\tau_\nu(\boldsymbol{\omega})}{x}
    \right)^{1/(\nu-2)}.
    \label{eq:clock-radial-interval}
\end{equation}
Because $x<y$ and
$\tau_\nu(\boldsymbol{\omega})\in(0,\infty)$, this is a nonempty interval
of positive length contained in $(0,\infty)$. Since the density of $R$ is
strictly positive on $(0,\infty)$,
\begin{equation}
    \mathbb P\!\left(
        x<
        \tau_\nu(\mathbf g)
        \leq y
        \,\middle|\,
        \boldsymbol{\omega}
    \right)
    >
    0
    \label{eq:clock-positive-conditional-interval}
\end{equation}
for almost every $\boldsymbol{\omega}$. Averaging over the direction gives
\[
    F_\nu(y)-F_\nu(x)
    =
    \mathbb P\!\left(
        x<
        \tau_\nu(\mathbf g)
        \leq y
    \right)
    >
    0.
\]

If $x=0$, then for almost every admissible direction,
\[
    \tau_\nu(\mathbf g)\leq y
    \quad\Longleftrightarrow\quad
    R
    \geq
    \left(
        \frac{\tau_\nu(\boldsymbol{\omega})}{y}
    \right)^{1/(\nu-2)}.
\]
The right-hand event has strictly positive conditional probability because
the radial Gaussian distribution has positive mass above every finite
threshold. Hence
\[
    F_\nu(y)>F_\nu(0).
\]
Since the clock is strictly positive almost surely,
\begin{equation}
    F_\nu(0)=0.
    \label{eq:clock-cdf-zero}
\end{equation}
Thus $F_\nu$ is strictly increasing on $[0,\infty)$.

\paragraph{Step 3: the range of $F_\nu$.}
Since
$\tau_\nu(\mathbf g)<\infty$ almost surely,
\begin{equation}
    F_\nu(x)
    \longrightarrow
    1
    \qquad
    \text{as }x\to\infty.
    \label{eq:clock-cdf-limit-one}
\end{equation}
On the other hand, $F_\nu(x)<1$ for every finite $x$. Indeed, for any
$x>0$ and almost every direction satisfying
\eqref{eq:clock-direction-finite},
\[
    \tau_\nu(\mathbf g)>x
\]
whenever
\begin{equation}
    0<R<
    \left(
        \frac{\tau_\nu(\boldsymbol{\omega})}{x}
    \right)^{1/(\nu-2)}.
    \label{eq:clock-large-time-small-radius}
\end{equation}
This radial interval has positive probability because the density of $R$ is
strictly positive on $(0,\infty)$. Hence
\[
    \mathbb P\!\left(
        \tau_\nu(\mathbf g)>x
    \right)
    >
    0,
\]
and therefore
\begin{equation}
    F_\nu(x)<1
    \qquad
    \text{for every }x<\infty.
    \label{eq:clock-cdf-below-one}
\end{equation}

Combining
\eqref{eq:clock-cdf-zero},
the continuity and strict monotonicity established above,
\eqref{eq:clock-cdf-limit-one}, and
\eqref{eq:clock-cdf-below-one}, we conclude that
\begin{equation}
    F_\nu:
    [0,\infty)
    \longrightarrow
    [0,1)
    \label{eq:clock-cdf-bijection}
\end{equation}
is a continuous strictly increasing bijection. In particular, its ordinary
inverse
\[
    F_\nu^{-1}:[0,1)\to[0,\infty)
\]
is well defined.

\paragraph{Step 4: candidate clocks.}
By Gaussian isotropy and independence of the initialization, the vectors
\[
    \mathbf g_{ki},
    \qquad
    k\in[H],\quad i\in[p],
\]
are independent standard Gaussian vectors in $\mathbb K^\nu$. Hence the
scalar clocks
$\tau_\nu(\mathbf g_{ki})$ are independent. Since
$\lambda_i>0$ is deterministic,
\[
    \tau_{ki}
    =
    \frac{
        \tau_\nu(\mathbf g_{ki})
    }{
        \lambda_i
    }
\]
are also independent and finite almost surely.

Each $\tau_{ki}$ has an atomless distribution, because multiplication by
the positive deterministic factor $\lambda_i^{-1}$ preserves
atomlessness. Therefore, for any two distinct pairs
$(k,i)\neq(\ell,j)$,
\[
    \mathbb P\!\left(
        \tau_{ki}=\tau_{\ell j}
    \right)
    =
    0.
\]
Since there are only finitely many candidate pairs, a union bound gives that
all $Hp$ candidate clocks are almost surely pairwise distinct.

Finally, if all sector amplitudes $\lambda_i$ are equal, the deterministic
rescaling is the same for every pair. Since the vectors
$\mathbf g_{ki}$ are i.i.d., the candidate clocks $\tau_{ki}$ are then
i.i.d. as well.
\end{proof}

Given a sector $i$ with amplitude $\lambda_i$, the clock of an edge incident to this sector has CDF
$F_\nu(\lambda_i\theta)$. For equal-amplitude sectors, all edges therefore
carry i.i.d.\ continuous clocks. This is the property that will later
connect the tensor dynamics to Random Edge matching.

\appendixsectionwithmain
    {Vanishing initialization yields greedy matching: proof and details}
    {app:greedy}
    {sec:greedy}

We now prove Theorem~\ref{thm:greedy-matching-limit}. Throughout this
appendix, we fix a realization of the Gaussian initialization belonging to
the probability-one event on which all candidate clocks are finite and
pairwise distinct, as established in Appendix~\ref{app:setup}. All constants
below may depend on this fixed realization, as well as on $p,H,\nu$ and the
target coefficients $\lambda_i$, but not on $\sigma$.

\subsection{Greedy event sequence and induction scheme}
\label{app:greedy-induction}

\paragraph{Exact and limiting dynamical variables.}
For each $\sigma>0$, let
$U^\sigma(t)=(\mathbf u_k^{\sigma,(a)}(t))_{a\in[\nu],\,k\in[H]}$
denote the physical-time GF solution of \eqref{eq:loss-gf}. Since the proof is
carried out in the slow time $\theta=\sigma^{\nu-2}t$, we write
\begin{equation}
    \mathbf u_{k,\sigma}^{(a)}(\theta)
    :=
    \mathbf u_k^{\sigma,(a)}
    \bigl(\sigma^{-(\nu-2)}\theta\bigr).
    \label{eq:a3-slow-weight}
\end{equation}
Thus $\mathbf u_{k,\sigma}^{(a)}(\theta)\in\mathbb K^p$ is the exact learned
weight vector, with only the time variable reparametrized.

Its exact coordinate in target sector $i$ is
\begin{equation}
    x_{ki,\sigma}^{(a)}(\theta)
    :=
    \bigl\langle
        \mathbf e_i^{(a)},
        \mathbf u_{k,\sigma}^{(a)}(\theta)
    \bigr\rangle .
    \label{eq:a3-sector-coordinate}
\end{equation}
While this coordinate is microscopic, it is natural to divide by the
initialization scale and define
\begin{equation}
    y_{ki,\sigma}^{(a)}(\theta)
    :=
    \frac{x_{ki,\sigma}^{(a)}(\theta)}{\sigma}.
    \label{eq:a3-microscopic-coordinate}
\end{equation}
The quantities $\mathbf u_{k,\sigma}^{(a)}$,
$x_{ki,\sigma}^{(a)}$, and $y_{ki,\sigma}^{(a)}$ are all exact
$\sigma$-dependent GF quantities.

We reserve the letter $z$ for the $\sigma$-independent isolated scalar flow.
For
$\mathbf g=(g^{(1)},\ldots,g^{(\nu)})\in\mathbb K^\nu$, write
\[
    \mathbf z(s;\mathbf g)
    =
    \bigl(
        z^{(1)}(s;\mathbf g),\ldots,z^{(\nu)}(s;\mathbf g)
    \bigr)
\]
for the solution of
\begin{equation}
    \frac{d z^{(a)}}{ds}
    =
    \prod_{b\neq a}\overline{z^{(b)}},
    \qquad
    z^{(a)}(0;\mathbf g)=g^{(a)},
    \qquad
    0\leq s<\tau_\nu(\mathbf g).
    \label{eq:a3-isolated-flow}
\end{equation}
For a candidate edge $(k,i)$, let
\[
    \mathbf g_{ki}
    =
    \bigl(
        g_{ki}^{(1)},\ldots,g_{ki}^{(\nu)}
    \bigr)
    \in\mathbb K^\nu
\]
and define its isolated profile in the global slow time by
\begin{equation}
    \boldsymbol{\zeta}_{ki}(\theta)
    :=
    \mathbf z(\lambda_i\theta;\mathbf g_{ki}),
    \qquad
    0\leq\theta<\tau_{ki},
    \qquad
    \tau_{ki}
    =
    \frac{\tau_\nu(\mathbf g_{ki})}{\lambda_i}.
    \label{eq:a3-candidate-profile}
\end{equation}
Thus $\boldsymbol{\zeta}_{ki}$ is independent of $\sigma$. Before the first
activation, Proposition~\ref{prop:microscopic-decoupling} states precisely
that the exact microscopic coordinates
$y_{ki,\sigma}^{(a)}$ converge to the corresponding components
$\zeta_{ki}^{(a)}$ of these isolated profiles.

\paragraph{Greedy event sequence.}
Let
\begin{equation}
    (k_1,i_1),\ldots,(k_M,i_M),
    \qquad
    M=\min\{H,p\},
    \label{eq:a3-accepted-edges}
\end{equation}
be the edges accepted by the greedy clock process, listed in their order of
acceptance. Denote their clock times by
\begin{equation}
    \theta_m^*
    :=
    \tau_{k_m i_m},
    \qquad
    0<\theta_1^*<\cdots<\theta_M^*,
    \qquad
    \theta_0^*:=0,
    \qquad
    \theta_{M+1}^*:=+\infty .
    \label{eq:a3-accepted-times}
\end{equation}

After $m$ accepted events, define the claimed atoms and sectors by
\begin{equation}
    \mathcal A_m
    :=
    \{k_1,\ldots,k_m\},
    \qquad
    \mathcal S_m
    :=
    \{i_1,\ldots,i_m\},
    \qquad
    \mathcal A_0=\mathcal S_0=\varnothing,
    \label{eq:a3-claimed-sets}
\end{equation}
and the candidate edges whose two endpoints are still available by
\begin{equation}
    \mathcal E_m
    :=
    \bigl([H]\setminus\mathcal A_m\bigr)
    \times
    \bigl([p]\setminus\mathcal S_m\bigr).
    \label{eq:a3-available-edges}
\end{equation}
By the definition of greedy matching,
\begin{equation}
    (k_{m+1},i_{m+1})
    =
    \operatorname*{arg\,min}_{(k,i)\in\mathcal E_m}\tau_{ki},
    \qquad
    \theta_{m+1}^*
    =
    \min_{(k,i)\in\mathcal E_m}\tau_{ki},
    \qquad
    0\leq m<M.
    \label{eq:a3-next-greedy-edge}
\end{equation}
The minimizer is unique because all candidate clocks are distinct.

A useful consequence of the greedy construction is
\begin{equation}
    \tau_{ki}>\theta_m^*
    \qquad
    \text{for every }(k,i)\in\mathcal E_m.
    \label{eq:a3-available-clocks-future}
\end{equation}
Indeed, if an edge in $\mathcal E_m$ had clock at most $\theta_m^*$, then
neither of its endpoints had been claimed when that edge was processed.
The greedy algorithm would therefore have accepted it, a contradiction.
In particular, every profile $\boldsymbol{\zeta}_{ki}$ with
$(k,i)\in\mathcal E_m$ remains finite in a right neighborhood of
$\theta_m^*$.

The tensor represented by the first $m$ accepted sectors and the
corresponding residual target are
\begin{equation}
    T_m
    :=
    \sum_{r=1}^{m}
    \lambda_{i_r}
    \mathbf e_{i_r}^{(1)}
    \otimes\cdots\otimes
    \mathbf e_{i_r}^{(\nu)},
    \qquad
    R_m
    :=
    T-T_m.
    \label{eq:a3-partial-target}
\end{equation}
The limiting greedy trajectory predicts the tensor $T_m$ throughout the
plateau between the $m$th and $(m+1)$st accepted events.

\paragraph{Captured-atom notation and the exact balance invariant.}
For an accepted edge $(k_r,i_r)$, define its main coordinates, their product,
and the component orthogonal to its selected sector by
\begin{equation}
    q_{r,\sigma}^{(a)}(\theta)
    :=
    x_{k_r i_r,\sigma}^{(a)}(\theta),
    \qquad
    \pi_{r,\sigma}(\theta)
    :=
    \prod_{a=1}^{\nu}
    q_{r,\sigma}^{(a)}(\theta),
    \label{eq:a3-locked-q-pi}
\end{equation}
and
\begin{equation}
    \mathbf h_{r,\sigma}^{(a)}(\theta)
    :=
    \mathbf u_{k_r,\sigma}^{(a)}(\theta)
    -
    q_{r,\sigma}^{(a)}(\theta)
    \mathbf e_{i_r}^{(a)}.
    \label{eq:a3-locked-h}
\end{equation}
By construction,
\begin{equation}
    \bigl\langle
        \mathbf e_{i_r}^{(a)},
        \mathbf h_{r,\sigma}^{(a)}
    \bigr\rangle
    =
    0.
    \label{eq:a3-locked-h-orthogonality}
\end{equation}

A key tool throughout the proof is the exact atom-wise balance invariant
derived in \eqref{eq:full-flow-balance-invariant-initialized}. In the present
slow-time notation it reads
\begin{equation}
    \left\|
        \mathbf u_{k,\sigma}^{(a)}(\theta)
    \right\|^2
    -
    \left\|
        \mathbf u_{k,\sigma}^{(b)}(\theta)
    \right\|^2
    =
    \sigma^2
    \left(
        \left\|\mathbf g_k^{(a)}\right\|^2
        -
        \left\|\mathbf g_k^{(b)}\right\|^2
    \right)
    \label{eq:a3-exact-balance-invariant}
\end{equation}
for every atom $k$, every $a,b\in[\nu]$, and every time for which the GF
solution exists. Since the initialization realization is fixed, there is a
constant $C_{\mathrm{bal}}<\infty$ such that
\begin{equation}
    \max_{\substack{k\in[H]\\a,b\in[\nu]}}
    \left|
        \left\|
            \mathbf u_{k,\sigma}^{(a)}(\theta)
        \right\|^2
        -
        \left\|
            \mathbf u_{k,\sigma}^{(b)}(\theta)
        \right\|^2
    \right|
    \leq
    C_{\mathrm{bal}}\sigma^2
    \label{eq:a3-exact-balance-bound}
\end{equation}
uniformly in time.

For a captured atom, the orthogonal decomposition
\eqref{eq:a3-locked-h} therefore gives
\begin{equation}
    \begin{aligned}
    \left|
        |q_{r,\sigma}^{(a)}|^2
        -
        |q_{r,\sigma}^{(b)}|^2
    \right|
    &\leq
    C_{\mathrm{bal}}\sigma^2
    +
    \left\|
        \mathbf h_{r,\sigma}^{(a)}
    \right\|^2
    +
    \left\|
        \mathbf h_{r,\sigma}^{(b)}
    \right\|^2 .
    \end{aligned}
    \label{eq:a3-main-coordinate-balance-from-invariant}
\end{equation}
This relation will be used throughout the proof to control the balance of
the mode-wise factors.

\paragraph{Reset times.}
The proof proceeds inductively through the accepted edges. Between
successive accepted clock times, the unclaimed atoms remain microscopic and
their coordinates in unclaimed sectors follow the original isolated profiles
$\boldsymbol{\zeta}_{ki}$. Near the next accepted clock, the winning profile
becomes large and the microscopic approximation breaks down. The associated
atom then undergoes a fast transition and captures the corresponding target
sector.

For each $m\geq1$, we will construct a \emph{reset time}
$\rho_m^\sigma$ after the $m$th fast transition such that
\begin{equation}
    \rho_m^\sigma
    \longrightarrow
    \theta_m^*
    \qquad
    \text{as }\sigma\to0,
    \label{eq:a3-reset-times}
\end{equation}
and set
\begin{equation}
    \rho_0^\sigma:=0.
    \label{eq:a3-initial-reset-time}
\end{equation}
Since the accepted clock times are distinct, for every fixed
$m\in\{1,\ldots,M\}$ and all sufficiently small $\sigma$,
\begin{equation}
    \theta_{m-1}^*
    <
    \rho_m^\sigma
    <
    \theta_{m+1}^*.
    \label{eq:a3-reset-between-neighboring-clocks}
\end{equation}

The reset times are checkpoints in the proof; the GF dynamics itself is not
restarted. In particular, surviving microscopic candidates continue to be
compared with the isolated profiles
$\boldsymbol{\zeta}_{ki}(\theta)$ that started from the original
initialization at $\theta=0$.

\paragraph{Notation used in the proof.}
Table~\ref{tab:greedy-proof-notation} summarizes the principal notation that
persists throughout the argument. Quantities specific to the entrance and
fast-transition layers will be introduced locally when needed.

\begin{table}[ht]
\centering
\small
\renewcommand{\arraystretch}{1.16}
\begin{tabular}{
    p{0.21\textwidth}
    p{0.56\textwidth}
    p{0.11\textwidth}
}
\hline
\textbf{Quantity}
&
\textbf{Meaning}
&
\textbf{Definition}
\\
\hline

$\mathbf u_{k,\sigma}^{(a)}(\theta)$
&
Exact mode-$a$ weight vector of atom $k$, expressed in slow time.
&
\eqref{eq:a3-slow-weight}
\\[1mm]

$x_{ki,\sigma}^{(a)}(\theta)$
&
Exact scalar coordinate of atom $k$ in target sector $i$.
&
\eqref{eq:a3-sector-coordinate}
\\[1mm]

$y_{ki,\sigma}^{(a)}(\theta)$
&
Microscopic coordinate
$x_{ki,\sigma}^{(a)}(\theta)/\sigma$.
&
\eqref{eq:a3-microscopic-coordinate}
\\[1mm]

$\mathbf z(s;\mathbf g)$
&
Generic $\sigma$-independent isolated unit-amplitude scalar-sector flow.
&
\eqref{eq:a3-isolated-flow}
\\[1mm]

$\boldsymbol{\zeta}_{ki}(\theta)$
&
Isolated profile of candidate $(k,i)$ in global slow time.
&
\eqref{eq:a3-candidate-profile}
\\[1mm]

$\theta_m^*$
&
Clock time of the $m$th edge accepted by the greedy process.
&
\eqref{eq:a3-accepted-times}
\\[1mm]

$\mathcal A_m,\mathcal S_m$
&
Atoms and sectors claimed after the first $m$ accepted events.
&
\eqref{eq:a3-claimed-sets}
\\[1mm]

$\mathcal E_m$
&
Candidate edges whose atom and sector are both still available.
&
\eqref{eq:a3-available-edges}
\\[1mm]

$T_m,\ R_m$
&
Tensor already learned after $m$ events and the corresponding residual
target.
&
\eqref{eq:a3-partial-target}
\\[1mm]

$q_{r,\sigma}^{(a)},\ \pi_{r,\sigma}$
&
Main coordinate of captured atom $k_r$ in sector $i_r$, and the product of
its $\nu$ main coordinates.
&
\eqref{eq:a3-locked-q-pi}
\\[1mm]

$\mathbf h_{r,\sigma}^{(a)}$
&
Off-sector component of captured atom $k_r$ in mode $a$.
&
\eqref{eq:a3-locked-h}
\\[1mm]

$\rho_m^\sigma$
&
Checkpoint after the $m$th fast transition, converging to
$\theta_m^*$.
&
\eqref{eq:a3-reset-times}
\\
\hline
\end{tabular}
\caption{
Principal notation used in the proof of
Theorem~\ref{thm:greedy-matching-limit}.
}
\label{tab:greedy-proof-notation}
\end{table}

\paragraph{Quantitative induction hypothesis.}
For each fixed $m\in\{0,\ldots,M\}$, we will construct a reset time
$\rho_m^\sigma$, numbers $\varepsilon_{m,\sigma}>0$, and a constant
$C_m<\infty$, independent of $\sigma$, such that
\begin{equation}
    \varepsilon_{m,\sigma}\longrightarrow0
    \qquad
    \text{as }\sigma\to0,
    \label{eq:a3-induction-error}
\end{equation}
and the following four properties hold at $\rho_m^\sigma$.

\medskip
\noindent
\textbf{(I$_m$) Captured coefficients.}
For every accepted edge $(k_r,i_r)$ with $r\leq m$, the product of its main
coordinates represents the corresponding target coefficient with the
quantitative accuracy
\begin{equation}
    \max_{1\leq r\leq m}
    \left|
        \pi_{r,\sigma}(\rho_m^\sigma)
        -
        \lambda_{i_r}
    \right|
    \leq
    C_m\sigma^\nu.
    \label{eq:a3-induction-captured-product}
\end{equation}

\medskip
\noindent
\textbf{(II$_m$) Microscopic surviving atoms.}
Every atom whose index has not yet been claimed remains at the initialization
scale:
\begin{equation}
    \max_{\substack{k\notin\mathcal A_m\\a\in[\nu]}}
    \left\|
        \mathbf u_{k,\sigma}^{(a)}(\rho_m^\sigma)
    \right\|
    \leq
    C_m\sigma.
    \label{eq:a3-induction-small-atoms}
\end{equation}

\medskip
\noindent
\textbf{(III$_m$) Preservation of the available candidate profiles.}
For every edge whose atom and sector are both still available, the exact
microscopic coordinate remains close to its original isolated profile:
\begin{equation}
    \max_{\substack{(k,i)\in\mathcal E_m\\a\in[\nu]}}
    \left|
        y_{ki,\sigma}^{(a)}(\rho_m^\sigma)
        -
        \zeta_{ki}^{(a)}(\rho_m^\sigma)
    \right|
    \leq
    \varepsilon_{m,\sigma}.
    \label{eq:a3-induction-profile-preservation}
\end{equation}
Thus surviving candidates are always compared with the profiles started from
the original Gaussian initialization at $\theta=0$; their clocks are not
reset after earlier captures.

\medskip
\noindent
\textbf{(IV$_m$) Blocking of claimed endpoints.}
Every captured atom is quantitatively confined to its assigned sector:
\begin{equation}
    \max_{\substack{1\leq r\leq m\\a\in[\nu]}}
    \left\|
        \mathbf h_{r,\sigma}^{(a)}(\rho_m^\sigma)
    \right\|
    \leq
    C_m\sigma^\nu.
    \label{eq:a3-induction-captured-locking}
\end{equation}
At the same time, every surviving atom remains microscopic inside every
already claimed sector:
\begin{equation}
    \max_{\substack{
        k\notin\mathcal A_m,\,
        i\in\mathcal S_m\\
        a\in[\nu]
    }}
    \left|
        x_{ki,\sigma}^{(a)}(\rho_m^\sigma)
    \right|
    \leq
    C_m\sigma,
    \label{eq:a3-induction-blocked-coordinates}
\end{equation}
or equivalently,
\begin{equation}
    \max_{\substack{
        k\notin\mathcal A_m,\,
        i\in\mathcal S_m\\
        a\in[\nu]
    }}
    \left|
        y_{ki,\sigma}^{(a)}(\rho_m^\sigma)
    \right|
    \leq
    C_m.
    \label{eq:a3-induction-blocked-rescaled}
\end{equation}

These four properties constitute the induction hypothesis at
$\rho_m^\sigma$.

\paragraph{Consequences of the induction hypotheses.}
We record two consequences that will be used repeatedly. First, the captured
factors are asymptotically balanced. Indeed,
\eqref{eq:a3-main-coordinate-balance-from-invariant} and
$\mathrm{(IV_m)}$ give
\begin{equation}
    \max_{\substack{
        1\leq r\leq m\\
        a,b\in[\nu]
    }}
    \left|
        |q_{r,\sigma}^{(a)}(\rho_m^\sigma)|^2
        -
        |q_{r,\sigma}^{(b)}(\rho_m^\sigma)|^2
    \right|
    =
    O(\sigma^2).
    \label{eq:a3-induction-squared-balance}
\end{equation}
On the other hand, $\mathrm{(I_m)}$ implies
\begin{equation}
    \prod_{a=1}^{\nu}
    \left|
        q_{r,\sigma}^{(a)}(\rho_m^\sigma)
    \right|
    =
    \lambda_{i_r}+o(1).
    \label{eq:a3-induction-magnitude-product}
\end{equation}
Combining
\eqref{eq:a3-induction-squared-balance} and
\eqref{eq:a3-induction-magnitude-product} yields
\begin{equation}
    \max_{\substack{
        1\leq r\leq m\\
        a\in[\nu]
    }}
    \left|
        \left|
            q_{r,\sigma}^{(a)}(\rho_m^\sigma)
        \right|
        -
        \lambda_{i_r}^{1/\nu}
    \right|
    \longrightarrow0.
    \label{eq:a3-induction-captured-balance}
\end{equation}

Second, the represented tensor at the reset time converges to the partial
target $T_m$. For every captured atom, $\mathrm{(IV_m)}$ and
\eqref{eq:a3-induction-captured-balance} give
\begin{equation}
    \mathbf u_{k_r,\sigma}^{(1)}(\rho_m^\sigma)
    \otimes\cdots\otimes
    \mathbf u_{k_r,\sigma}^{(\nu)}(\rho_m^\sigma)
    \longrightarrow
    \lambda_{i_r}
    \mathbf e_{i_r}^{(1)}
    \otimes\cdots\otimes
    \mathbf e_{i_r}^{(\nu)}.
    \label{eq:a3-captured-tensor-limit}
\end{equation}
At the same time, $\mathrm{(II_m)}$ implies that every surviving atom
contributes $O(\sigma^\nu)$ in Frobenius norm. Since $H$ is fixed,
\begin{equation}
    \widehat T\!\left(
        U^\sigma\!\left(
            \sigma^{-(\nu-2)}\rho_m^\sigma
        \right)
    \right)
    \longrightarrow
    T_m.
    \label{eq:a3-reset-output-limit}
\end{equation}

\paragraph{Base case.}
For $m=0$, $\mathrm{(I_0)}$ and the captured-atom part of
$\mathrm{(IV_0)}$ are vacuous. Property $\mathrm{(II_0)}$ follows directly
from the Gaussian initialization because $p,H,\nu$ are fixed, while
$\mathrm{(III_0)}$ holds exactly since
\[
    y_{ki,\sigma}^{(a)}(0)
    =
    g_{ki}^{(a)}
    =
    \zeta_{ki}^{(a)}(0).
\]
The claimed-sector part of $\mathrm{(IV_0)}$ is also vacuous.

\paragraph{Proof roadmap.}
The induction step $m\to m+1$ is organized as follows.

\begin{enumerate}
    \item
    \emph{Stability of the captured background.}
    In Subsection~\ref{app:locked-background}, we show that, as long as the
    surviving atoms remain microscopic, the previously captured atoms stay
    locked to their sectors, available atom--sector pairs retain their
    isolated dynamics to leading order, and claimed sectors exert no
    leading-order drive on surviving atoms. These statements are summarized
    in Proposition~\ref{prop:a3-locked-background}.

    \item
    \emph{Propagation up to the next activation.}
    In Subsection~\ref{app:slow-propagation}, we close the microscopic
    bootstrap and propagate the induction state through every fixed interval
    lying strictly before the next accepted clock $\theta_{m+1}^*$.
    Proposition~\ref{prop:a3-slow-propagation} shows in particular that the
    available candidates continue to follow their original isolated profiles
    $\boldsymbol{\zeta}_{ki}$, so their clocks are not restarted after earlier
    captures.

    \item
    \emph{Entrance into the fast layer.}
    In Subsection~\ref{app:fast-entrance}, we approach
    $\theta_{m+1}^*$ on a $\sigma$-dependent mesoscopic scale.
    Proposition~\ref{prop:a3-mesoscopic-entrance} constructs an entrance time
    $\eta_{m+1}^\sigma\to\theta_{m+1}^*$ at which the winning atom is much
    larger than the initialization scale but still vanishing in the original
    variables, while all competing atoms remain microscopic.

    \item
    \emph{Fast capture and renewal.}
    In Subsection~\ref{app:fast-capture},
    Lemma~\ref{lem:a3-fast-growth} carries the winning atom from the
    mesoscopic scale to a fixed macroscopic scale, and
    Lemma~\ref{lem:a3-fast-locking} then establishes locking of the enlarged
    captured set. Proposition~\ref{prop:a3-fast-capture} combines these
    estimates to construct a new reset time
    $\rho_{m+1}^\sigma\to\theta_{m+1}^*$ at which
    $\mathrm{(I_{m+1})}$--$\mathrm{(IV_{m+1})}$ hold.
\end{enumerate}

Throughout the induction step,
\eqref{eq:a3-exact-balance-invariant} controls the relative mode-wise
magnitudes, while
\eqref{eq:a3-induction-profile-preservation} carries the original candidate
profiles through successive captures.

\medskip
\noindent
\emph{Completion of the proof.}
After iterating the induction through all accepted edges,
Subsection~\ref{app:a3-completion} controls the plateau following the final
capture. Lemma~\ref{lem:a3-terminal-propagation} gives the required terminal
propagation estimate, and the resulting plateau bounds yield the tensor and
loss convergence stated in Theorem~\ref{thm:greedy-matching-limit}.

\subsection{Dynamics in a locked background}
\label{app:locked-background}

We next isolate the estimates that describe the dynamics after some atoms
have already captured distinct target sectors. The role of this subsection is
deliberately conditional. Suppose that, on a given slow-time interval, the
previously captured atoms start well locked to their selected sectors and the
remaining atoms stay microscopic, of size $O(\sigma)$. We show that the
captured atoms then form a stable background: they remain locked, available
atom--sector pairs continue to follow their original isolated dynamics up to a
vanishing perturbation, and already claimed sectors exert no leading-order
drive on the surviving atoms.

The assumption that the surviving atoms remain microscopic will be proved in
the next subsection by a separate slow-propagation argument. Keeping these two
steps separate avoids mixing the stability of the captured background with the
bootstrap that controls the next generation of candidate atoms.

\paragraph{Notation and exact equations.}
For $k,j\in[H]$ and $a\in[\nu]$, define the mode-wise Gram entries
\begin{equation}
    G_{kj,\sigma}^{(a)}(\theta)
    :=
    \Big\langle
        \mathbf u_{k,\sigma}^{(a)}(\theta),
        \mathbf u_{j,\sigma}^{(a)}(\theta)
    \Big\rangle .
    \label{eq:a3-locked-gram}
\end{equation}
In the slow time $\theta=\sigma^{\nu-2}t$, the exact GF equation can be
written, using the diagonal form of the target, as
\begin{equation}
    \frac{d\mathbf u_{k,\sigma}^{(a)}}{d\theta}
    =
    \sigma^{-(\nu-2)}
    \left[
        \sum_{i=1}^{p}
        \lambda_i\mathbf e_i^{(a)}
        \prod_{b\neq a}
        \overline{x_{ki,\sigma}^{(b)}}
        -
        \sum_{j=1}^{H}
        \mathbf u_{j,\sigma}^{(a)}
        \prod_{b\neq a}
        G_{kj,\sigma}^{(b)}
    \right].
    \label{eq:a3-locked-vector-gf}
\end{equation}
Projecting the exact vector equation
\eqref{eq:a3-locked-vector-gf} onto the sector direction
$\mathbf e_i^{(a)}$ gives
\begin{equation}
    \frac{d x_{ki,\sigma}^{(a)}}{d\theta}
    =
    \sigma^{-(\nu-2)}
    \left[
        \lambda_i
        \prod_{b\neq a}
        \overline{x_{ki,\sigma}^{(b)}}
        -
        \sum_{j=1}^{H}
        x_{ji,\sigma}^{(a)}
        \prod_{b\neq a}
        G_{kj,\sigma}^{(b)}
    \right].
    \label{eq:a3-locked-coordinate-gf}
\end{equation}

Finally, for convenience, we denote the exact predicted tensor in slow time by
\begin{equation}
    \widehat T_\sigma(\theta)
    :=
    \sum_{k=1}^{H}
    \mathbf u_{k,\sigma}^{(1)}(\theta)
    \otimes\cdots\otimes
    \mathbf u_{k,\sigma}^{(\nu)}(\theta).
    \label{eq:a3-slow-output}
\end{equation}

\paragraph{A bootstrap continuation principle.}
We will use the same elementary continuation argument several times in the
proof. We record it once here.

\begin{lemma}[Bootstrap continuation]
\label{lem:a3-bootstrap-continuation}
Let
$I_\sigma=[\theta_0^\sigma,\theta_1^\sigma]$, let
$E_\sigma:I_\sigma\to[0,\infty)$ be continuous, and let $a_\sigma>0$.
Fix constants
\begin{equation}
    B_{\mathrm{init}}>0,
    \qquad
    B_{\mathrm{imp}}>0,
    \qquad
    B_{\mathrm{boot}}>0,
    \label{eq:a3-bootstrap-constants}
\end{equation}
independent of $\sigma$, such that
\begin{equation}
    B_{\mathrm{init}}<B_{\mathrm{boot}},
    \qquad
    B_{\mathrm{imp}}<B_{\mathrm{boot}}.
    \label{eq:a3-bootstrap-strict-slack}
\end{equation}
No relation between $B_{\mathrm{init}}$ and $B_{\mathrm{imp}}$ is assumed.

Suppose that the initial error satisfies
\begin{equation}
    E_\sigma(\theta_0^\sigma)
    \leq
    B_{\mathrm{init}}a_\sigma.
    \label{eq:a3-bootstrap-initial}
\end{equation}
Assume moreover that, for every
$\theta\in[\theta_0^\sigma,\theta_1^\sigma]$ for which
\begin{equation}
    \sup_{\vartheta\in[\theta_0^\sigma,\theta]}
    E_\sigma(\vartheta)
    \leq
    B_{\mathrm{boot}}a_\sigma,
    \label{eq:a3-bootstrap-assumption}
\end{equation}
the dynamical estimates imply the sharper bound
\begin{equation}
    \sup_{\vartheta\in[\theta_0^\sigma,\theta]}
    E_\sigma(\vartheta)
    \leq
    B_{\mathrm{imp}}a_\sigma.
    \label{eq:a3-bootstrap-improvement}
\end{equation}
Then
\begin{equation}
    \sup_{\theta\in I_\sigma}
    E_\sigma(\theta)
    \leq
    B_{\mathrm{imp}}a_\sigma.
    \label{eq:a3-bootstrap-conclusion}
\end{equation}
\end{lemma}

\begin{proof}
Define the maximal bootstrap time
\begin{equation}
    \theta_{\mathrm{boot}}^\sigma
    :=
    \sup\left\{
        \theta\in I_\sigma:
        \sup_{\vartheta\in[\theta_0^\sigma,\theta]}
        E_\sigma(\vartheta)
        \leq
        B_{\mathrm{boot}}a_\sigma
    \right\}.
    \label{eq:a3-bootstrap-time}
\end{equation}
By the initial bound \eqref{eq:a3-bootstrap-initial} and the strict slack
$B_{\mathrm{init}}<B_{\mathrm{boot}}$ in
\eqref{eq:a3-bootstrap-strict-slack}, continuity of $E_\sigma$ implies
\begin{equation}
    \theta_{\mathrm{boot}}^\sigma
    >
    \theta_0^\sigma.
    \label{eq:a3-bootstrap-nontrivial}
\end{equation}

On every compact subinterval of
$[\theta_0^\sigma,\theta_{\mathrm{boot}}^\sigma)$, the bootstrap assumption
\eqref{eq:a3-bootstrap-assumption} holds. Hence the improvement estimate
\eqref{eq:a3-bootstrap-improvement}, followed by continuity at the endpoint,
gives
\begin{equation}
    \sup_{\vartheta\in
        [\theta_0^\sigma,\theta_{\mathrm{boot}}^\sigma]}
    E_\sigma(\vartheta)
    \leq
    B_{\mathrm{imp}}a_\sigma
    <
    B_{\mathrm{boot}}a_\sigma.
    \label{eq:a3-bootstrap-endpoint-slack}
\end{equation}

If
$\theta_{\mathrm{boot}}^\sigma<\theta_1^\sigma$, the strict inequality in
\eqref{eq:a3-bootstrap-endpoint-slack} and continuity would allow the
bootstrap bound to remain valid on a slightly larger interval. This
contradicts the definition of
$\theta_{\mathrm{boot}}^\sigma$. Therefore
\begin{equation}
    \theta_{\mathrm{boot}}^\sigma
    =
    \theta_1^\sigma.
    \label{eq:a3-bootstrap-full-interval}
\end{equation}
Applying the improvement estimate
\eqref{eq:a3-bootstrap-improvement} on the full interval gives
\eqref{eq:a3-bootstrap-conclusion}.
\end{proof}

In the present application, the quantity to be bootstrapped will be the
off-sector locking error
\begin{equation}
    E_{\mathrm{lock},\sigma}(\theta)
    :=
    \max_{\substack{r\leq m\\a\in[\nu]}}
    \bigl\|
        \mathbf h_{r,\sigma}^{(a)}(\theta)
    \bigr\|.
    \label{eq:a3-locking-error}
\end{equation}
The continuation principle applies because GF is a smooth finite-dimensional
ODE, hence its solution and all quantities in
\eqref{eq:a3-locking-error} are continuous as long as the weights remain
bounded. On the bootstrap intervals considered below, boundedness follows
from the coarse bounds on the captured coordinates, the microscopic bound on
the surviving atoms, and the bootstrap bound itself. The constants in the
estimates below depend only on these coarse bounds and not on the endpoint of
the bootstrap interval.

\paragraph{Main locked-background statement.}

\begin{proposition}[Reduction in a locked background]
\label{prop:a3-locked-background}
Fix $m\in\{0,\ldots,M\}$ and an interval
$I_\sigma=[\theta_0^\sigma,\theta_1^\sigma]$ whose length is bounded
uniformly in $\sigma$. Suppose that the following conditions hold.

First, all surviving atoms remain microscopic on $I_\sigma$:
\begin{equation}
    \max_{\substack{k\notin\mathcal A_m\\a\in[\nu]}}
    \sup_{\theta\in I_\sigma}
    \bigl\|
        \mathbf u_{k,\sigma}^{(a)}(\theta)
    \bigr\|
    \leq
    B\sigma.
    \label{eq:a3-prop-survivors-small}
\end{equation}
Second, the main coordinates of the captured atoms stay in a fixed compact
annulus:
\begin{equation}
    c
    \leq
    \bigl|
        q_{r,\sigma}^{(a)}(\theta)
    \bigr|
    \leq
    C,
    \qquad
    r\leq m,\quad
    a\in[\nu],\quad
    \theta\in I_\sigma,
    \label{eq:a3-prop-captured-coarse}
\end{equation}
for constants $0<c<C<\infty$ independent of $\sigma$. Third, at the left
endpoint the captured atoms are quantitatively locked:
\begin{equation}
    \max_{\substack{r\leq m\\a\in[\nu]}}
    \bigl\|
        \mathbf h_{r,\sigma}^{(a)}(\theta_0^\sigma)
    \bigr\|
    +
    \max_{r\leq m}
    \bigl|
        \pi_{r,\sigma}(\theta_0^\sigma)-\lambda_{i_r}
    \bigr|
    \leq
    B\sigma^\nu.
    \label{eq:a3-prop-initial-lock}
\end{equation}

Then there is a constant $C_{\mathrm{bg}}<\infty$, independent of $\sigma$,
such that the following conclusions hold for all sufficiently small
$\sigma$.

\begin{enumerate}
\item
\emph{Stability of captured atoms.}
The off-sector and product errors remain of order $\sigma^\nu$:
\begin{equation}
    \max_{\substack{r\leq m\\a\in[\nu]}}
    \sup_{\theta\in I_\sigma}
    \bigl\|
        \mathbf h_{r,\sigma}^{(a)}(\theta)
    \bigr\|
    +
    \max_{r\leq m}
    \sup_{\theta\in I_\sigma}
    \bigl|
        \pi_{r,\sigma}(\theta)-\lambda_{i_r}
    \bigr|
    \leq
    C_{\mathrm{bg}}\sigma^\nu.
    \label{eq:a3-prop-captured-stability}
\end{equation}
Moreover, the exact full-flow balance invariant gives
\begin{equation}
    \max_{\substack{r\leq m\\a,b\in[\nu]}}
    \sup_{\theta\in I_\sigma}
    \left|
        |q_{r,\sigma}^{(a)}(\theta)|^2
        -
        |q_{r,\sigma}^{(b)}(\theta)|^2
    \right|
    \leq
    C_{\mathrm{bg}}\sigma^2.
    \label{eq:a3-prop-balance-stability}
\end{equation}

\item
\emph{Available sectors retain the isolated dynamics.}
For every available edge $(k,i)\in\mathcal E_m$,
\begin{equation}
    \frac{d y_{ki,\sigma}^{(a)}}{d\theta}
    =
    \lambda_i
    \prod_{b\neq a}
    \overline{y_{ki,\sigma}^{(b)}}
    +
    r_{ki,\sigma}^{(a)}(\theta),
    \label{eq:a3-prop-available-flow}
\end{equation}
where
\begin{equation}
    \max_{\substack{(k,i)\in\mathcal E_m\\a\in[\nu]}}
    \sup_{\theta\in I_\sigma}
    \bigl|
        r_{ki,\sigma}^{(a)}(\theta)
    \bigr|
    \leq
    C_{\mathrm{bg}}\sigma^\nu.
    \label{eq:a3-prop-available-remainder}
\end{equation}

\item
\emph{Claimed sectors are dynamically cancelled.}
If $k\notin\mathcal A_m$ and $i=i_r\in\mathcal S_m$, then
\begin{equation}
    \frac{d y_{k i_r,\sigma}^{(a)}}{d\theta}
    =
    \bigl(
        \lambda_{i_r}-\pi_{r,\sigma}
    \bigr)
    \prod_{b\neq a}
    \overline{y_{k i_r,\sigma}^{(b)}}
    +
    \widehat r_{k i_r,\sigma}^{(a)}(\theta),
    \label{eq:a3-prop-claimed-flow}
\end{equation}
with
\begin{equation}
    \max_{\substack{k\notin\mathcal A_m,\ i_r\in\mathcal S_m\\a\in[\nu]}}
    \sup_{\theta\in I_\sigma}
    \bigl|
        \widehat r_{k i_r,\sigma}^{(a)}(\theta)
    \bigr|
    \leq
    C_{\mathrm{bg}}\sigma^\nu.
    \label{eq:a3-prop-claimed-remainder}
\end{equation}
Consequently,
\begin{equation}
    \max_{\substack{k\notin\mathcal A_m,\ i\in\mathcal S_m\\a\in[\nu]}}
    \sup_{\theta\in I_\sigma}
    \left|
        \frac{d y_{ki,\sigma}^{(a)}}{d\theta}
    \right|
    \leq
    C_{\mathrm{bg}}\sigma^\nu.
    \label{eq:a3-prop-claimed-frozen}
\end{equation}

\item
\emph{Tensor-level approximation.}
The locked background represents the already selected part of the target:
\begin{equation}
    \sup_{\theta\in I_\sigma}
    \bigl\|
        \widehat T_\sigma(\theta)-T_m
    \bigr\|_F
    \leq
    C_{\mathrm{bg}}\sigma^\nu.
    \label{eq:a3-prop-output}
\end{equation}
\end{enumerate}
\end{proposition}

We prove Proposition~\ref{prop:a3-locked-background} through three elementary
lemmas. The first is an exact decomposition of the dynamics of a captured
coordinate. The second uses this decomposition and the bootstrap principle to
prove stability of the captured atoms. The third derives the effective
dynamics of the surviving atoms.

\paragraph{Captured-coordinate decomposition.}

\begin{lemma}[Exact equation for a captured coordinate]
\label{lem:a3-captured-decomposition}
Assume the hypotheses
\eqref{eq:a3-prop-survivors-small} and
\eqref{eq:a3-prop-captured-coarse}. Let
$J_\sigma\subseteq I_\sigma$ be an interval beginning at
$\theta_0^\sigma$ on which, for some fixed $B_*>0$,
\begin{equation}
    \max_{\substack{r\leq m\\a\in[\nu]}}
    \sup_{\theta\in J_\sigma}
    \bigl\|
        \mathbf h_{r,\sigma}^{(a)}(\theta)
    \bigr\|
    \leq
    B_*\sigma^\nu.
    \label{eq:a3-decomp-bootstrap}
\end{equation}
Then every captured main coordinate satisfies
\begin{equation}
    \frac{d q_{r,\sigma}^{(a)}}{d\theta}
    =
    \sigma^{-(\nu-2)}
    \bigl(
        \lambda_{i_r}-\pi_{r,\sigma}
    \bigr)
    \prod_{b\neq a}
    \overline{q_{r,\sigma}^{(b)}}
    +
    R_{r,\sigma}^{(a)}(\theta),
    \label{eq:a3-decomp-main}
\end{equation}
where
\begin{equation}
    \max_{\substack{r\leq m\\a\in[\nu]}}
    \sup_{\theta\in J_\sigma}
    \bigl|
        R_{r,\sigma}^{(a)}(\theta)
    \bigr|
    =
    O(\sigma^2).
    \label{eq:a3-decomp-remainder-bound}
\end{equation}
\end{lemma}

\begin{proof}
Fix $r\leq m$, set $i=i_r$, and abbreviate
$q^{(a)}=q_{r,\sigma}^{(a)}$,
$\pi=\pi_{r,\sigma}$, and
$\mathbf h^{(a)}=\mathbf h_{r,\sigma}^{(a)}$.

Taking $k=k_r$ and $i=i_r$ in the exact sector-coordinate equation
\eqref{eq:a3-locked-coordinate-gf} gives
\begin{equation}
    \frac{d q^{(a)}}{d\theta}
    =
    \sigma^{-(\nu-2)}
    \left[
        \lambda_i
        \prod_{b\neq a}\overline{q^{(b)}}
        -
        \sum_{j=1}^{H}
        x_{ji,\sigma}^{(a)}
        \prod_{b\neq a}
        G_{k_rj,\sigma}^{(b)}
    \right].
    \label{eq:a3-decomp-start}
\end{equation}
We separate the self-interaction $j=k_r$ from the remaining terms.

\paragraph{Step 1: the self-interaction.}
Since the captured-atom decomposition
\eqref{eq:a3-locked-h} gives
$\mathbf u_{k_r,\sigma}^{(b)}
=q^{(b)}\mathbf e_i^{(b)}+\mathbf h^{(b)}$,
and the two components are orthogonal by
\eqref{eq:a3-locked-h-orthogonality},
\begin{equation}
    G_{k_rk_r,\sigma}^{(b)}
    =
    \bigl\|
        \mathbf u_{k_r,\sigma}^{(b)}
    \bigr\|^2
    =
    |q^{(b)}|^2+\|\mathbf h^{(b)}\|^2.
    \label{eq:a3-decomp-self-gram}
\end{equation}
Hence the self-interaction equals
\begin{equation}
    q^{(a)}
    \prod_{b\neq a}
    \left(
        |q^{(b)}|^2+\|\mathbf h^{(b)}\|^2
    \right).
    \label{eq:a3-decomp-self-exact}
\end{equation}
Its leading part can be rewritten using
$\pi=\prod_c q^{(c)}$ as
\begin{equation}
    q^{(a)}
    \prod_{b\neq a}|q^{(b)}|^2
    =
    \pi
    \prod_{b\neq a}\overline{q^{(b)}}.
    \label{eq:a3-decomp-product-identity}
\end{equation}
Therefore the correction caused by the off-sector components is
\begin{equation}
    R_{r,\sigma}^{\mathrm{self},(a)}
    :=
    -
    \sigma^{-(\nu-2)}
    q^{(a)}
    \left[
        \prod_{b\neq a}
        \left(
            |q^{(b)}|^2+\|\mathbf h^{(b)}\|^2
        \right)
        -
        \prod_{b\neq a}|q^{(b)}|^2
    \right].
    \label{eq:a3-decomp-r-self}
\end{equation}
By the off-sector bootstrap
\eqref{eq:a3-decomp-bootstrap} and the coarse boundedness of the captured
coordinates in \eqref{eq:a3-prop-captured-coarse},
\begin{equation}
    R_{r,\sigma}^{\mathrm{self},(a)}
    =
    O(\sigma^{\nu+2}).
    \label{eq:a3-decomp-r-self-bound}
\end{equation}

\paragraph{Step 2: interactions with surviving atoms.}
The contribution of all $j\notin\mathcal A_m$ is
\begin{equation}
    R_{r,\sigma}^{\mathrm{surv},(a)}
    :=
    -
    \sigma^{-(\nu-2)}
    \sum_{j\notin\mathcal A_m}
    x_{ji,\sigma}^{(a)}
    \prod_{b\neq a}
    G_{k_rj,\sigma}^{(b)}.
    \label{eq:a3-decomp-r-surv}
\end{equation}
By the assumed microscopic bound on surviving atoms
\eqref{eq:a3-prop-survivors-small},
\begin{equation}
    |x_{ji,\sigma}^{(a)}|
    =
    O(\sigma),
    \qquad
    |G_{k_rj,\sigma}^{(b)}|
    =
    O(\sigma).
    \label{eq:a3-decomp-surv-scales}
\end{equation}
Each summand before the slow-time prefactor is therefore $O(\sigma^\nu)$.
Since $H$ is fixed,
\begin{equation}
    R_{r,\sigma}^{\mathrm{surv},(a)}
    =
    O(\sigma^2).
    \label{eq:a3-decomp-r-surv-bound}
\end{equation}

\paragraph{Step 3: interactions with the other captured atoms.}
For $s\leq m$, $s\neq r$, we have $i_s\neq i_r=i$. Hence the coordinate of
captured atom $k_s$ in the wrong sector is contained entirely in its
off-sector component:
\begin{equation}
    x_{k_s i,\sigma}^{(a)}
    =
    \Big\langle
        \mathbf e_i^{(a)},
        \mathbf h_{s,\sigma}^{(a)}
    \Big\rangle
    =
    O(\sigma^\nu).
    \label{eq:a3-decomp-other-wrong-coordinate}
\end{equation}
Moreover, orthogonality of the sector directions
$\mathbf e_{i_r}^{(b)}$ and $\mathbf e_{i_s}^{(b)}$, together with the
decomposition \eqref{eq:a3-locked-h}, gives
\begin{equation}
    \begin{aligned}
    G_{k_rk_s,\sigma}^{(b)}
    &=
    \overline{q_{r,\sigma}^{(b)}}
    \Big\langle
        \mathbf e_{i_r}^{(b)},
        \mathbf h_{s,\sigma}^{(b)}
    \Big\rangle
    +
    q_{s,\sigma}^{(b)}
    \Big\langle
        \mathbf h_{r,\sigma}^{(b)},
        \mathbf e_{i_s}^{(b)}
    \Big\rangle
    \\
    &\hspace{1.5cm}
    +
    \Big\langle
        \mathbf h_{r,\sigma}^{(b)},
        \mathbf h_{s,\sigma}^{(b)}
    \Big\rangle
    =
    O(\sigma^\nu).
    \end{aligned}
    \label{eq:a3-decomp-other-gram}
\end{equation}
Thus the interaction with the other captured atoms,
\begin{equation}
    R_{r,\sigma}^{\mathrm{cap},(a)}
    :=
    -
    \sigma^{-(\nu-2)}
    \sum_{\substack{s\leq m\\s\neq r}}
    x_{k_s i,\sigma}^{(a)}
    \prod_{b\neq a}
    G_{k_rk_s,\sigma}^{(b)},
    \label{eq:a3-decomp-r-cap}
\end{equation}
satisfies
\begin{equation}
    R_{r,\sigma}^{\mathrm{cap},(a)}
    =
    O\!\left(
        \sigma^{\nu^2-\nu+2}
    \right).
    \label{eq:a3-decomp-r-cap-bound}
\end{equation}

Combining the three contributions, define
\begin{equation}
    R_{r,\sigma}^{(a)}
    :=
    R_{r,\sigma}^{\mathrm{self},(a)}
    +
    R_{r,\sigma}^{\mathrm{surv},(a)}
    +
    R_{r,\sigma}^{\mathrm{cap},(a)}.
    \label{eq:a3-decomp-r-total}
\end{equation}
Substituting the leading self-interaction identity
\eqref{eq:a3-decomp-product-identity} into the exact equation
\eqref{eq:a3-decomp-start} now gives
\eqref{eq:a3-decomp-main}. Since $\nu\geq3$, the bounds
\eqref{eq:a3-decomp-r-self-bound},
\eqref{eq:a3-decomp-r-surv-bound}, and
\eqref{eq:a3-decomp-r-cap-bound} imply the remainder estimate
\eqref{eq:a3-decomp-remainder-bound}. The dominant perturbation is the
interaction with the still-microscopic atoms.
\end{proof}

\paragraph{Why the captured state is stable.}
Before proving the estimates, it is useful to describe the underlying
vector-field geometry. Ignoring the vanishing interactions with the other
atoms, the configurations
\begin{equation}
    \mathbf u^{(a)}
    =
    q^{(a)}\mathbf e_i^{(a)},
    \qquad
    \prod_{a=1}^{\nu}q^{(a)}=\lambda_i,
    \label{eq:a3-captured-equilibrium-set}
\end{equation}
form a manifold of equilibria for a captured sector. The vector field is
contracting in the two directions relevant to locking. The product error
$\pi-\lambda_i$ obeys, to leading order,
\begin{equation}
    \frac{d}{d\theta}
    (\pi-\lambda_i)
    =
    -
    \sigma^{-(\nu-2)}
    S(\mathbf q)(\pi-\lambda_i),
    \qquad
    S(\mathbf q)
    =
    \sum_{a=1}^{\nu}
    \prod_{b\neq a}|q^{(b)}|^2,
    \label{eq:a3-captured-product-contraction}
\end{equation}
while an off-sector perturbation satisfies
\begin{equation}
    \frac{d\mathbf h^{(a)}}{d\theta}
    =
    -
    \sigma^{-(\nu-2)}
    \gamma^{(a)}(\mathbf q)\mathbf h^{(a)}
    +
    \text{perturbing terms},
    \qquad
    \gamma^{(a)}(\mathbf q)
    =
    \prod_{b\neq a}|q^{(b)}|^2.
    \label{eq:a3-captured-transverse-contraction}
\end{equation}
The coarse lower bound
$|q^{(a)}|\geq c$ from
\eqref{eq:a3-prop-captured-coarse} therefore gives
\begin{equation}
    S(\mathbf q)\geq\nu c^{2(\nu-1)},
    \qquad
    \gamma^{(a)}(\mathbf q)\geq c^{2(\nu-1)}=:c_\gamma>0.
    \label{eq:a3-captured-uniform-contraction}
\end{equation}
Thus the equilibrium manifold is uniformly attracting in the product and
off-sector directions. The remaining directions tangent to the manifold
correspond to redistributing scale and phase among the mode-wise factors.
The relative factor magnitudes are controlled globally by the exact
atom-wise balance invariant
\eqref{eq:a3-exact-balance-invariant}, while the phase directions remain
neutral.

\paragraph{Stability of the captured atoms.}

\begin{lemma}[Stability of a captured background]
\label{lem:a3-captured-stability}
Under the assumptions of
Proposition~\ref{prop:a3-locked-background}, the locking estimate
\eqref{eq:a3-prop-captured-stability} and the balance estimate
\eqref{eq:a3-prop-balance-stability} hold.
\end{lemma}

\begin{proof}
We use Lemma~\ref{lem:a3-bootstrap-continuation} with the locking error
\eqref{eq:a3-locking-error} and scale $a_\sigma=\sigma^\nu$.

Choose a fixed bootstrap constant $B_*>B$. On the maximal interval on which
\begin{equation}
    E_{\mathrm{lock},\sigma}(\theta)
    \leq
    B_*\sigma^\nu,
    \label{eq:a3-stability-bootstrap}
\end{equation}
the captured-coordinate decomposition of
Lemma~\ref{lem:a3-captured-decomposition} applies. We derive estimates that
improve this bound.

\paragraph{Step 1: stability of the captured product.}
For a fixed $r$, differentiate
$\pi_{r,\sigma}=\prod_a q_{r,\sigma}^{(a)}$ and use the effective captured
coordinate equation \eqref{eq:a3-decomp-main}. This gives
\begin{equation}
    \frac{d\pi_{r,\sigma}}{d\theta}
    =
    \sigma^{-(\nu-2)}
    \bigl(
        \lambda_{i_r}-\pi_{r,\sigma}
    \bigr)
    S_{r,\sigma}(\theta)
    +
    R_{r,\sigma}^{\pi}(\theta),
    \label{eq:a3-stability-product-equation}
\end{equation}
where
\begin{equation}
    S_{r,\sigma}(\theta)
    :=
    \sum_{a=1}^{\nu}
    \prod_{b\neq a}
    |q_{r,\sigma}^{(b)}(\theta)|^2
    \label{eq:a3-stability-product-coefficient}
\end{equation}
and
\begin{equation}
    R_{r,\sigma}^{\pi}(\theta)
    :=
    \sum_{a=1}^{\nu}
    R_{r,\sigma}^{(a)}(\theta)
    \prod_{c\neq a}
    q_{r,\sigma}^{(c)}(\theta)
    =
    O(\sigma^2).
    \label{eq:a3-stability-product-remainder}
\end{equation}
The coarse annulus bounds
\eqref{eq:a3-prop-captured-coarse} imply
\begin{equation}
    0<c_S
    \leq
    S_{r,\sigma}(\theta)
    \leq
    C_S<\infty
    \label{eq:a3-stability-product-coercivity}
\end{equation}
uniformly on the bootstrap interval.

Writing
$e_{r,\sigma}^{\pi}:=\pi_{r,\sigma}-\lambda_{i_r}$,
variation of constants in the contracting product equation
\eqref{eq:a3-stability-product-equation} yields
\begin{equation}
    \bigl|
        e_{r,\sigma}^{\pi}(\theta)
    \bigr|
    \leq
    e^{-c_S(\theta-\theta_0^\sigma)/\sigma^{\nu-2}}
    \bigl|
        e_{r,\sigma}^{\pi}(\theta_0^\sigma)
    \bigr|
    +
    C\sigma^2
    \int_{\theta_0^\sigma}^{\theta}
    e^{-c_S(\theta-s)/\sigma^{\nu-2}}\,ds .
    \label{eq:a3-stability-product-voc}
\end{equation}
Since the fast exponential kernel satisfies
\begin{equation}
    \int_{\theta_0^\sigma}^{\theta}
    e^{-c_S(\theta-s)/\sigma^{\nu-2}}\,ds
    \leq
    \frac{\sigma^{\nu-2}}{c_S},
    \label{eq:a3-stability-fast-kernel}
\end{equation}
the initial quantitative locking assumption
\eqref{eq:a3-prop-initial-lock} gives
\begin{equation}
    \sup_{\theta}
    \bigl|
        \pi_{r,\sigma}(\theta)-\lambda_{i_r}
    \bigr|
    \leq
    C_\pi\sigma^\nu
    \label{eq:a3-stability-product-bound}
\end{equation}
on the bootstrap interval.

\paragraph{Step 2: stability of the off-sector components.}
Let
\begin{equation}
    Q_{i_r}^{(a)}
    :=
    I-
    \mathbf e_{i_r}^{(a)}
    \bigl(\mathbf e_{i_r}^{(a)}\bigr)^*
    \label{eq:a3-stability-projector}
\end{equation}
be the orthogonal projector onto the complement of the captured sector.
By the definition of the off-sector component
\eqref{eq:a3-locked-h},
\begin{equation}
    \mathbf h_{r,\sigma}^{(a)}
    =
    Q_{i_r}^{(a)}
    \mathbf u_{k_r,\sigma}^{(a)}.
\end{equation}
Applying this projector to the exact vector GF equation
\eqref{eq:a3-locked-vector-gf} gives
\begin{equation}
    \frac{d\mathbf h_{r,\sigma}^{(a)}}{d\theta}
    =
    -
    \sigma^{-(\nu-2)}
    \gamma_{r,\sigma}^{(a)}(\theta)
    \mathbf h_{r,\sigma}^{(a)}
    +
    \mathbf F_{r,\sigma}^{(a)}(\theta),
    \label{eq:a3-stability-h-equation}
\end{equation}
where
\begin{equation}
    \gamma_{r,\sigma}^{(a)}(\theta)
    :=
    \prod_{b\neq a}
    \bigl\|
        \mathbf u_{k_r,\sigma}^{(b)}(\theta)
    \bigr\|^2
    \label{eq:a3-stability-h-damping}
\end{equation}
and
\begin{equation}
    \begin{aligned}
    \mathbf F_{r,\sigma}^{(a)}
    &:=
    \sigma^{-(\nu-2)}
    \sum_{\ell\neq i_r}
    \lambda_\ell\mathbf e_\ell^{(a)}
    \prod_{b\neq a}
    \overline{x_{k_r\ell,\sigma}^{(b)}}
    \\
    &\quad
    -
    \sigma^{-(\nu-2)}
    \sum_{j\neq k_r}
    Q_{i_r}^{(a)}
    \mathbf u_{j,\sigma}^{(a)}
    \prod_{b\neq a}
    G_{k_rj,\sigma}^{(b)} .
    \end{aligned}
    \label{eq:a3-stability-h-forcing}
\end{equation}
The coarse lower bound on the captured main coordinates in
\eqref{eq:a3-prop-captured-coarse} implies
\begin{equation}
    \gamma_{r,\sigma}^{(a)}(\theta)
    \geq
    c^{2(\nu-1)}
    =:
    c_\gamma>0.
    \label{eq:a3-stability-h-coercivity}
\end{equation}

We next estimate the forcing in
\eqref{eq:a3-stability-h-forcing}. The interaction with surviving atoms is
$O(\sigma^2)$ in slow time, by the same scaling as in the surviving-atom
remainder estimate \eqref{eq:a3-decomp-r-surv-bound}. The target components
orthogonal to $i_r$ contain $\nu-1$ off-sector coordinates of the same
captured atom, while the interaction with another captured atom contains
$\nu-1$ small cross-Gram factors. Hence, on the locking bootstrap
\eqref{eq:a3-stability-bootstrap},
\begin{equation}
    \max_{\substack{r\leq m\\a\in[\nu]}}
    \bigl\|
        \mathbf F_{r,\sigma}^{(a)}(\theta)
    \bigr\|
    \leq
    C_0\sigma^2
    +
    C_1(B_*)\,
    \sigma^{\nu^2-2\nu+2}.
    \label{eq:a3-stability-h-forcing-bound}
\end{equation}
For $\nu\geq3$, the second exponent is strictly larger than $2$.

Variation of constants in the damped off-sector equation
\eqref{eq:a3-stability-h-equation}, together with the uniform damping bound
\eqref{eq:a3-stability-h-coercivity}, gives
\begin{equation}
    \begin{aligned}
    \bigl\|
        \mathbf h_{r,\sigma}^{(a)}(\theta)
    \bigr\|
    &\leq
    e^{-c_\gamma(\theta-\theta_0^\sigma)/\sigma^{\nu-2}}
    \bigl\|
        \mathbf h_{r,\sigma}^{(a)}(\theta_0^\sigma)
    \bigr\|
    \\
    &\quad
    +
    \int_{\theta_0^\sigma}^{\theta}
    e^{-c_\gamma(\theta-s)/\sigma^{\nu-2}}
    \bigl\|
        \mathbf F_{r,\sigma}^{(a)}(s)
    \bigr\|\,ds .
    \end{aligned}
    \label{eq:a3-stability-h-voc}
\end{equation}
Using the fast-kernel bound
\eqref{eq:a3-stability-fast-kernel}, with $c_\gamma$ in place of $c_S$,
gives
\begin{equation}
    \sup_{\theta}
    \bigl\|
        \mathbf h_{r,\sigma}^{(a)}(\theta)
    \bigr\|
    \leq
    \left(
        B+C_h+o(1)
    \right)\sigma^\nu,
    \label{eq:a3-stability-h-improved}
\end{equation}
where $C_h$ is independent of the bootstrap constant $B_*$.

\paragraph{Step 3: closing the locking bootstrap.}
Choose
\begin{equation}
    B_*
    >
    B+C_h+1.
    \label{eq:a3-stability-bootstrap-choice}
\end{equation}
For sufficiently small $\sigma$, the improved off-sector estimate
\eqref{eq:a3-stability-h-improved} is then a strict improvement over the
bootstrap assumption \eqref{eq:a3-stability-bootstrap}. The continuation
principle of Lemma~\ref{lem:a3-bootstrap-continuation} therefore shows that
the bootstrap interval is all of $I_\sigma$. Combining this conclusion with
the product estimate \eqref{eq:a3-stability-product-bound} proves the locking
bound \eqref{eq:a3-prop-captured-stability}.

\paragraph{Step 4: balance from the exact full-flow invariant.}
It remains to prove the balance estimate
\eqref{eq:a3-prop-balance-stability}. The exact atom-wise invariant
\eqref{eq:a3-exact-balance-invariant} gives, for the captured atom $k_r$,
a difference of squared full mode norms of order $O(\sigma^2)$ at every
time. Using the orthogonal decomposition
\eqref{eq:a3-locked-h}, this was rewritten in
\eqref{eq:a3-main-coordinate-balance-from-invariant} as
\[
    \left|
        |q_{r,\sigma}^{(a)}|^2
        -
        |q_{r,\sigma}^{(b)}|^2
    \right|
    \leq
    C_{\mathrm{bal}}\sigma^2
    +
    \left\|
        \mathbf h_{r,\sigma}^{(a)}
    \right\|^2
    +
    \left\|
        \mathbf h_{r,\sigma}^{(b)}
    \right\|^2.
\]
The off-sector stability already proved in
\eqref{eq:a3-prop-captured-stability} gives
$\|\mathbf h_{r,\sigma}^{(a)}\|=O(\sigma^\nu)$ uniformly on $I_\sigma$.
Since $\nu\geq3$, the last two terms are $O(\sigma^{2\nu})$ and can be
absorbed into the $O(\sigma^2)$ bound. This proves
\eqref{eq:a3-prop-balance-stability}.
\end{proof}

\paragraph{Effective dynamics of the surviving atoms.}

\begin{lemma}[Surviving atoms in a locked background]
\label{lem:a3-survivor-dynamics}
Under the assumptions of
Proposition~\ref{prop:a3-locked-background}, the available-sector estimates
\eqref{eq:a3-prop-available-flow}--\eqref{eq:a3-prop-available-remainder}
and the claimed-sector estimates
\eqref{eq:a3-prop-claimed-flow}--\eqref{eq:a3-prop-claimed-frozen}
hold.
\end{lemma}

\begin{proof}
For $k\notin\mathcal A_m$, divide the exact coordinate equation
\eqref{eq:a3-locked-coordinate-gf} by $\sigma$ and use the microscopic
rescaling
$x_{ki,\sigma}^{(a)}=\sigma y_{ki,\sigma}^{(a)}$. This gives
\begin{equation}
    \frac{d y_{ki,\sigma}^{(a)}}{d\theta}
    =
    \lambda_i
    \prod_{b\neq a}
    \overline{y_{ki,\sigma}^{(b)}}
    -
    \sigma^{-(\nu-1)}
    \sum_{j=1}^{H}
    x_{ji,\sigma}^{(a)}
    \prod_{b\neq a}
    G_{kj,\sigma}^{(b)}.
    \label{eq:a3-survivor-exact}
\end{equation}

\paragraph{Step 1: an unclaimed sector.}
Suppose $(k,i)\in\mathcal E_m$, so $i\notin\mathcal S_m$.

If $j\notin\mathcal A_m$, then both atoms $k$ and $j$ are microscopic by
the survivor hypothesis \eqref{eq:a3-prop-survivors-small}. Hence
\begin{equation}
    |x_{ji,\sigma}^{(a)}|
    =
    O(\sigma),
    \qquad
    |G_{kj,\sigma}^{(b)}|
    =
    O(\sigma^2),
    \label{eq:a3-survivor-unclaimed-small-small}
\end{equation}
and the corresponding term in the exact microscopic equation
\eqref{eq:a3-survivor-exact}, after multiplication by
$\sigma^{-(\nu-1)}$, is $O(\sigma^\nu)$.

If instead $j=k_r$ is captured, then $i\neq i_r$. The off-sector locking
estimate \eqref{eq:a3-prop-captured-stability} therefore gives
\begin{equation}
    |x_{k_r i,\sigma}^{(a)}|
    =
    O(\sigma^\nu),
    \qquad
    |G_{k k_r,\sigma}^{(b)}|
    =
    O(\sigma).
    \label{eq:a3-survivor-unclaimed-small-captured}
\end{equation}
The corresponding rescaled interaction is again $O(\sigma^\nu)$.
Summing over the fixed number of atoms proves the isolated-flow equation
\eqref{eq:a3-prop-available-flow} and its remainder estimate
\eqref{eq:a3-prop-available-remainder}.

\paragraph{Step 2: a claimed sector.}
Now let $i=i_r\in\mathcal S_m$. The leading interaction in the exact
microscopic equation \eqref{eq:a3-survivor-exact} comes from the captured
atom $k_r$.

Using the captured-atom decomposition
\eqref{eq:a3-locked-h} and its locking estimate
\eqref{eq:a3-prop-captured-stability},
\begin{equation}
    \begin{aligned}
    G_{k k_r,\sigma}^{(b)}
    &=
    \Big\langle
        \mathbf u_{k,\sigma}^{(b)},
        q_{r,\sigma}^{(b)}
        \mathbf e_{i_r}^{(b)}
        +
        \mathbf h_{r,\sigma}^{(b)}
    \Big\rangle
    \\
    &=
    \sigma q_{r,\sigma}^{(b)}
    \overline{y_{k i_r,\sigma}^{(b)}}
    +
    O(\sigma^{\nu+1}).
    \end{aligned}
    \label{eq:a3-survivor-claimed-gram}
\end{equation}
Consequently,
\begin{equation}
    \sigma^{-(\nu-1)}
    x_{k_r i_r,\sigma}^{(a)}
    \prod_{b\neq a}
    G_{k k_r,\sigma}^{(b)}
    =
    \pi_{r,\sigma}
    \prod_{b\neq a}
    \overline{y_{k i_r,\sigma}^{(b)}}
    +
    O(\sigma^\nu).
    \label{eq:a3-survivor-claimed-cancellation}
\end{equation}
Thus the model contribution of the captured atom cancels the target drive in
its claimed sector up to the product error
$\lambda_{i_r}-\pi_{r,\sigma}$.

The remaining surviving atoms contribute $O(\sigma^\nu)$ by the
small--small estimate
\eqref{eq:a3-survivor-unclaimed-small-small}, while the other captured atoms
contribute $O(\sigma^\nu)$ by the same cross-sector estimate as in
\eqref{eq:a3-survivor-unclaimed-small-captured}. Substituting these bounds
into the exact microscopic equation \eqref{eq:a3-survivor-exact} yields the
claimed-sector equation \eqref{eq:a3-prop-claimed-flow} and the remainder
bound \eqref{eq:a3-prop-claimed-remainder}.

Finally, the microscopic survivor hypothesis
\eqref{eq:a3-prop-survivors-small} implies
\begin{equation}
    \max_{\substack{k\notin\mathcal A_m\\i\in[p]\\a\in[\nu]}}
    |y_{ki,\sigma}^{(a)}|
    \leq
    B.
    \label{eq:a3-survivor-y-bounded}
\end{equation}
Together with the product locking estimate
\eqref{eq:a3-prop-captured-stability}, the claimed-sector equation
\eqref{eq:a3-prop-claimed-flow}, and its remainder estimate
\eqref{eq:a3-prop-claimed-remainder}, this gives the near-freezing bound
\eqref{eq:a3-prop-claimed-frozen}.
\end{proof}

\paragraph{Proof of the locked-background proposition.}

\begin{proof}[Proof of Proposition~\ref{prop:a3-locked-background}]
We verify the four conclusions in the order in which they were stated.

\paragraph{1. Captured atoms.}
Lemma~\ref{lem:a3-captured-stability} gives the quantitative locking estimate
\eqref{eq:a3-prop-captured-stability} and, using the exact full-flow
invariant, the balance estimate \eqref{eq:a3-prop-balance-stability}.

\paragraph{2. Available sectors.}
The first part of Lemma~\ref{lem:a3-survivor-dynamics} gives the perturbed
isolated-flow equation \eqref{eq:a3-prop-available-flow} and its
$O(\sigma^\nu)$ remainder estimate
\eqref{eq:a3-prop-available-remainder}.

\paragraph{3. Claimed sectors.}
The second part of Lemma~\ref{lem:a3-survivor-dynamics} gives the
target--model cancellation equation
\eqref{eq:a3-prop-claimed-flow}, the corresponding remainder bound
\eqref{eq:a3-prop-claimed-remainder}, and hence the near-freezing estimate
\eqref{eq:a3-prop-claimed-frozen}.

\paragraph{4. Tensor-level approximation.}
For a captured atom $k_r$, expand its tensor product using the decomposition
\eqref{eq:a3-locked-h}:
\begin{equation}
    \bigotimes_{a=1}^{\nu}
    \mathbf u_{k_r,\sigma}^{(a)}
    =
    \bigotimes_{a=1}^{\nu}
    \left(
        q_{r,\sigma}^{(a)}
        \mathbf e_{i_r}^{(a)}
        +
        \mathbf h_{r,\sigma}^{(a)}
    \right).
    \label{eq:a3-output-captured-expansion}
\end{equation}
The term containing no off-sector factor equals
\begin{equation}
    \pi_{r,\sigma}
    \mathbf e_{i_r}^{(1)}
    \otimes\cdots\otimes
    \mathbf e_{i_r}^{(\nu)}.
    \label{eq:a3-output-captured-leading}
\end{equation}
Every other term contains at least one
$\mathbf h_{r,\sigma}^{(a)}$ and is therefore $O(\sigma^\nu)$ in Frobenius
norm by the captured-atom locking estimate
\eqref{eq:a3-prop-captured-stability}. Combining this with the product
estimate in the same equation gives
\begin{equation}
    \left\|
        \bigotimes_{a=1}^{\nu}
        \mathbf u_{k_r,\sigma}^{(a)}
        -
        \lambda_{i_r}
        \mathbf e_{i_r}^{(1)}
        \otimes\cdots\otimes
        \mathbf e_{i_r}^{(\nu)}
    \right\|_F
    =
    O(\sigma^\nu).
    \label{eq:a3-output-captured-error}
\end{equation}
For a surviving atom, the microscopic bound
\eqref{eq:a3-prop-survivors-small} gives
\begin{equation}
    \left\|
        \mathbf u_{k,\sigma}^{(1)}
        \otimes\cdots\otimes
        \mathbf u_{k,\sigma}^{(\nu)}
    \right\|_F
    =
    O(\sigma^\nu).
    \label{eq:a3-output-survivor-error}
\end{equation}
Since $H$ is fixed in the present $\sigma\downarrow0$ limit, summing the
captured-atom errors \eqref{eq:a3-output-captured-error} and the
surviving-atom errors \eqref{eq:a3-output-survivor-error} over the finitely
many atoms proves the tensor-level approximation
\eqref{eq:a3-prop-output}.
\end{proof}

Proposition~\ref{prop:a3-locked-background} is conditional on the surviving
atoms remaining $O(\sigma)$ and on the already captured main coordinates
remaining inside the coarse annulus
\eqref{eq:a3-prop-captured-coarse}. The next subsection closes precisely
these conditions. Starting from a reset time $\rho_m^\sigma$, we use another
continuation argument together with the available-sector dynamics
\eqref{eq:a3-prop-available-flow} to show that the surviving microscopic
coordinates remain close to their original isolated profiles
$\boldsymbol{\zeta}_{ki}$ throughout the slow interval preceding the next
accepted clock. At the same time, the captured-atom stability
\eqref{eq:a3-prop-captured-stability} and the exact balance estimate
\eqref{eq:a3-prop-balance-stability} keep the previously captured atoms
inside the required coarse bounds.

\subsection{Slow propagation between activation events}
\label{app:slow-propagation}

We now remove the conditional assumption in
Proposition~\ref{prop:a3-locked-background}. Starting from a reset time
$\rho_m^\sigma$, we show that the surviving atoms remain microscopic and
continue to follow their original isolated profiles throughout every compact
subinterval lying strictly before the next accepted clock
$\theta_{m+1}^*$. At the same time, the previously captured atoms remain
locked. This provides the uniform control on the plateaus between successive
activation events.

For convenience, collect the microscopic coordinates of a candidate
$(k,i)$ into the vector
\begin{equation}
    \mathbf y_{ki,\sigma}(\theta)
    :=
    \bigl(
        y_{ki,\sigma}^{(1)}(\theta),
        \ldots,
        y_{ki,\sigma}^{(\nu)}(\theta)
    \bigr)
    \in\mathbb K^\nu.
    \label{eq:a3-slow-y-vector}
\end{equation}
Recall that its limiting isolated profile is
$\boldsymbol{\zeta}_{ki}(\theta)$ from
\eqref{eq:a3-candidate-profile}. Define the scalar-sector vector field
$\mathbf F_i:\mathbb K^\nu\to\mathbb K^\nu$ by
\begin{equation}
    \bigl[\mathbf F_i(\mathbf z)\bigr]^{(a)}
    :=
    \lambda_i
    \prod_{b\neq a}\overline{z^{(b)}}.
    \label{eq:a3-slow-vector-field}
\end{equation}
Then the isolated profile satisfies
\begin{equation}
    \frac{d}{d\theta}
    \boldsymbol{\zeta}_{ki}(\theta)
    =
    \mathbf F_i\bigl(
        \boldsymbol{\zeta}_{ki}(\theta)
    \bigr),
    \qquad
    0\leq\theta<\tau_{ki}.
    \label{eq:a3-slow-isolated-profile-equation}
\end{equation}

\begin{proposition}[Slow propagation before the next clock]
\label{prop:a3-slow-propagation}
Fix $m\in\{0,\ldots,M-1\}$ and suppose that the induction hypotheses
$\mathrm{(I_m)}$--$\mathrm{(IV_m)}$ hold at the reset time
$\rho_m^\sigma$. In particular, we will use the captured-product estimate
\eqref{eq:a3-induction-captured-product}, the microscopic survivor bound
\eqref{eq:a3-induction-small-atoms}, the preservation of the available
profiles \eqref{eq:a3-induction-profile-preservation}, and the two
claimed-endpoint bounds
\eqref{eq:a3-induction-captured-locking} and
\eqref{eq:a3-induction-blocked-rescaled}. We will also use the derived
balance consequence \eqref{eq:a3-induction-captured-balance}.

Fix a constant $\delta$ satisfying
\begin{equation}
    0<\delta<
    \frac12
    \bigl(
        \theta_{m+1}^*-\theta_m^*
    \bigr).
    \label{eq:a3-slow-delta}
\end{equation}
Define
\begin{equation}
    I_{m,\delta}^\sigma
    :=
    \bigl[
        \rho_m^\sigma,
        \theta_{m+1}^*-\delta
    \bigr].
    \label{eq:a3-slow-interval}
\end{equation}
Since $\rho_m^\sigma\to\theta_m^*$, this interval is nonempty for all
sufficiently small $\sigma$.

Then the following statements hold.

\begin{enumerate}
\item
\emph{Uniform preservation of the candidate profiles.}
For every available edge,
\begin{equation}
    \max_{\substack{(k,i)\in\mathcal E_m\\a\in[\nu]}}
    \sup_{\theta\in I_{m,\delta}^\sigma}
    \left|
        y_{ki,\sigma}^{(a)}(\theta)
        -
        \zeta_{ki}^{(a)}(\theta)
    \right|
    \longrightarrow0
    \qquad
    \text{as }\sigma\to0.
    \label{eq:a3-slow-profile-convergence}
\end{equation}

\item
\emph{Surviving atoms remain microscopic.}
There exists a constant $C_{m,\delta}<\infty$, independent of $\sigma$,
such that
\begin{equation}
    \max_{\substack{k\notin\mathcal A_m\\a\in[\nu]}}
    \sup_{\theta\in I_{m,\delta}^\sigma}
    \bigl\|
        \mathbf u_{k,\sigma}^{(a)}(\theta)
    \bigr\|
    \leq
    C_{m,\delta}\sigma.
    \label{eq:a3-slow-survivors-remain-small}
\end{equation}
In already claimed sectors, their rescaled coordinates change only by
$O(\sigma^\nu)$:
\begin{equation}
    \max_{\substack{k\notin\mathcal A_m,\ i\in\mathcal S_m\\a\in[\nu]}}
    \sup_{\theta\in I_{m,\delta}^\sigma}
    \left|
        y_{ki,\sigma}^{(a)}(\theta)
        -
        y_{ki,\sigma}^{(a)}(\rho_m^\sigma)
    \right|
    \leq
    C_{m,\delta}\sigma^\nu.
    \label{eq:a3-slow-blocked-coordinate-control}
\end{equation}

\item
\emph{Captured atoms remain locked and balanced.}
Uniformly on $I_{m,\delta}^\sigma$,
\begin{equation}
    \max_{\substack{r\leq m\\a\in[\nu]}}
    \sup_{\theta\in I_{m,\delta}^\sigma}
    \bigl\|
        \mathbf h_{r,\sigma}^{(a)}(\theta)
    \bigr\|
    +
    \max_{r\leq m}
    \sup_{\theta\in I_{m,\delta}^\sigma}
    \bigl|
        \pi_{r,\sigma}(\theta)-\lambda_{i_r}
    \bigr|
    =
    O(\sigma^\nu),
    \label{eq:a3-slow-captured-lock}
\end{equation}
and
\begin{equation}
    \max_{\substack{r\leq m\\a\in[\nu]}}
    \sup_{\theta\in I_{m,\delta}^\sigma}
    \left|
        \bigl|
            q_{r,\sigma}^{(a)}(\theta)
        \bigr|
        -
        \lambda_{i_r}^{1/\nu}
    \right|
    \longrightarrow0.
    \label{eq:a3-slow-captured-balance}
\end{equation}

\item
\emph{Uniform plateau approximation.}
The exact tensor output satisfies
\begin{equation}
    \sup_{\theta\in I_{m,\delta}^\sigma}
    \bigl\|
        \widehat T_\sigma(\theta)-T_m
    \bigr\|_F
    \longrightarrow0,
    \label{eq:a3-slow-output-convergence}
\end{equation}
and consequently
\begin{equation}
    \sup_{\theta\in I_{m,\delta}^\sigma}
    \left|
        L\!\left(
            U^\sigma\!\left(
                \sigma^{-(\nu-2)}\theta
            \right)
        \right)
        -
        \frac12
        \sum_{i\notin\mathcal S_m}\lambda_i^2
    \right|
    \longrightarrow0.
    \label{eq:a3-slow-loss-convergence}
\end{equation}
\end{enumerate}
\end{proposition}

\begin{proof}
The main point is to justify, rather than assume, the microscopic survivor
bound required by Proposition~\ref{prop:a3-locked-background}. We do this by
a continuation argument with three simultaneous bootstrap conditions: one
for the available candidate profiles, one for coordinates in already claimed
sectors, and one for the main coordinates of the captured atoms.

\paragraph{Step 1: boundedness of the isolated profiles.}
By the greedy characterization of the next accepted edge
\eqref{eq:a3-next-greedy-edge},
\begin{equation}
    \tau_{ki}
    \geq
    \theta_{m+1}^*
    \qquad
    \text{for every }(k,i)\in\mathcal E_m,
    \label{eq:a3-slow-available-clock-lower-bound}
\end{equation}
with equality only for the next accepted edge
$(k_{m+1},i_{m+1})$. Thus every available isolated profile is regular on a
fixed neighborhood of the interval between $\theta_m^*$ and
$\theta_{m+1}^*-\delta$.

To also cover the possibly $\sigma$-dependent starting point
$\rho_m^\sigma$, define
\begin{equation}
    \underline\theta_{m,\delta}
    :=
    \max\{0,\theta_m^*-\delta\}.
    \label{eq:a3-slow-profile-left-endpoint}
\end{equation}
Since $\rho_m^\sigma\to\theta_m^*$, for all sufficiently small $\sigma$,
\begin{equation}
    \rho_m^\sigma
    \geq
    \underline\theta_{m,\delta}.
    \label{eq:a3-slow-reset-in-fixed-interval}
\end{equation}
Because there are only finitely many available candidates, the quantity
\begin{equation}
    Z_{m,\delta}
    :=
    \max_{\substack{(k,i)\in\mathcal E_m\\a\in[\nu]}}
    \sup_{\theta\in
        [\underline\theta_{m,\delta},\,
         \theta_{m+1}^*-\delta]}
    \left|
        \zeta_{ki}^{(a)}(\theta)
    \right|
    \label{eq:a3-slow-profile-bound}
\end{equation}
is finite. This fixed bound will control the available microscopic
coordinates throughout the bootstrap interval.

\paragraph{Step 2: bootstrap interval.}
We introduce three bootstrap conditions. First, every available candidate is
required to remain within unit distance of its isolated profile:
\begin{equation}
    \max_{\substack{(k,i)\in\mathcal E_m\\a\in[\nu]}}
    \left|
        y_{ki,\sigma}^{(a)}(\theta)
        -
        \zeta_{ki}^{(a)}(\theta)
    \right|
    \leq
    1.
    \label{eq:a3-slow-bootstrap-profile}
\end{equation}
Second, in every already claimed sector we require
\begin{equation}
    \max_{\substack{k\notin\mathcal A_m,\ i\in\mathcal S_m\\a\in[\nu]}}
    \left|
        y_{ki,\sigma}^{(a)}(\theta)
    \right|
    \leq
    C_m+1.
    \label{eq:a3-slow-bootstrap-claimed}
\end{equation}
Third, every captured main coordinate is required to remain in the coarse
annulus
\begin{equation}
    \frac12\lambda_{i_r}^{1/\nu}
    \leq
    \bigl|
        q_{r,\sigma}^{(a)}(\theta)
    \bigr|
    \leq
    2\lambda_{i_r}^{1/\nu},
    \qquad
    r\leq m,\quad a\in[\nu].
    \label{eq:a3-slow-bootstrap-captured}
\end{equation}
For $m=0$, the last two conditions are vacuous.

At $\theta=\rho_m^\sigma$, the available-profile preservation hypothesis
\eqref{eq:a3-induction-profile-preservation}, the claimed-sector bound
\eqref{eq:a3-induction-blocked-rescaled}, and the derived captured-factor
balance
\eqref{eq:a3-induction-captured-balance} show that all three bootstrap
conditions hold with strict slack for sufficiently small $\sigma$. By
continuity, they therefore hold on a nontrivial interval beginning at
$\rho_m^\sigma$.

Let
\begin{equation}
    \theta_{\mathrm{boot}}^\sigma
    :=
    \sup\left\{
        \theta\in I_{m,\delta}^\sigma:
        \eqref{eq:a3-slow-bootstrap-profile},
        \eqref{eq:a3-slow-bootstrap-claimed},
        \eqref{eq:a3-slow-bootstrap-captured}
        \text{ hold on }
        [\rho_m^\sigma,\theta]
    \right\}.
    \label{eq:a3-slow-bootstrap-time}
\end{equation}
We will show that each of the three bootstrap bounds is improved strictly on
$[\rho_m^\sigma,\theta_{\mathrm{boot}}^\sigma]$. The same continuation
argument as in Lemma~\ref{lem:a3-bootstrap-continuation} will then force
$\theta_{\mathrm{boot}}^\sigma$ to equal the right endpoint of
$I_{m,\delta}^\sigma$.

\paragraph{Step 3: the bootstrap conditions imply microscopic surviving
atoms.}
For an unclaimed sector $i\notin\mathcal S_m$, the profile bootstrap
\eqref{eq:a3-slow-bootstrap-profile} and the fixed isolated-profile bound
\eqref{eq:a3-slow-profile-bound} give
\begin{equation}
    \left|
        y_{ki,\sigma}^{(a)}(\theta)
    \right|
    \leq
    Z_{m,\delta}+1.
    \label{eq:a3-slow-unclaimed-y-bootstrap-bound}
\end{equation}
For a claimed sector, the claimed-coordinate bootstrap
\eqref{eq:a3-slow-bootstrap-claimed} gives the corresponding bound
$C_m+1$. Since
$x_{ki,\sigma}^{(a)}=\sigma y_{ki,\sigma}^{(a)}$ and
$\{\mathbf e_i^{(a)}\}_{i=1}^{p}$ is an orthonormal basis,
\begin{equation}
    \begin{aligned}
    \bigl\|
        \mathbf u_{k,\sigma}^{(a)}(\theta)
    \bigr\|^2
    &=
    \sum_{i=1}^{p}
    \left|
        x_{ki,\sigma}^{(a)}(\theta)
    \right|^2
    \\
    &=
    \sigma^2
    \sum_{i=1}^{p}
    \left|
        y_{ki,\sigma}^{(a)}(\theta)
    \right|^2
    \leq
    B_{m,\delta}^2\sigma^2
    \end{aligned}
    \label{eq:a3-slow-bootstrap-implies-small}
\end{equation}
for a constant $B_{m,\delta}<\infty$ independent of $\sigma$.

The captured-coordinate bootstrap
\eqref{eq:a3-slow-bootstrap-captured} supplies the uniform lower and upper
bounds required by the coarse annulus hypothesis
\eqref{eq:a3-prop-captured-coarse} of
Proposition~\ref{prop:a3-locked-background}. Its quantitative locking
hypothesis at the left endpoint follows directly from the induction
hypotheses. Indeed, the captured-product condition
\eqref{eq:a3-induction-captured-product} gives
\begin{equation}
    \max_{r\leq m}
    \left|
        \pi_{r,\sigma}(\rho_m^\sigma)-\lambda_{i_r}
    \right|
    \leq
    C_m\sigma^\nu,
    \label{eq:a3-slow-initial-product-lock}
\end{equation}
while the captured off-sector condition
\eqref{eq:a3-induction-captured-locking} gives
\begin{equation}
    \max_{\substack{r\leq m\\a\in[\nu]}}
    \left\|
        \mathbf h_{r,\sigma}^{(a)}(\rho_m^\sigma)
    \right\|
    \leq
    C_m\sigma^\nu.
    \label{eq:a3-slow-initial-off-sector-lock}
\end{equation}
Finally, the bootstrap interval has uniformly bounded length because it is
contained in $I_{m,\delta}^\sigma$ and
$\rho_m^\sigma\to\theta_m^*$. Thus all assumptions of
Proposition~\ref{prop:a3-locked-background} hold on
$[\rho_m^\sigma,\theta_{\mathrm{boot}}^\sigma]$.

\paragraph{Step 4: available candidates remain close to their original
profiles.}
By the available-sector conclusion
\eqref{eq:a3-prop-available-flow}--\eqref{eq:a3-prop-available-remainder}
of Proposition~\ref{prop:a3-locked-background}, every available candidate
satisfies
\begin{equation}
    \frac{d\mathbf y_{ki,\sigma}}{d\theta}
    =
    \mathbf F_i(\mathbf y_{ki,\sigma})
    +
    \mathbf r_{ki,\sigma}(\theta),
    \qquad
    \sup_{\theta}
    \bigl\|
        \mathbf r_{ki,\sigma}(\theta)
    \bigr\|
    \leq
    C_{m,\delta}\sigma^\nu.
    \label{eq:a3-slow-available-vector-equation}
\end{equation}
The corresponding isolated profile solves the unperturbed equation
\eqref{eq:a3-slow-isolated-profile-equation}.

Both $\mathbf y_{ki,\sigma}$ and
$\boldsymbol{\zeta}_{ki}$ remain in a fixed bounded subset of
$\mathbb K^\nu$ on the bootstrap interval. Since the scalar-sector vector
field $\mathbf F_i$ in \eqref{eq:a3-slow-vector-field} is polynomial,
viewed as a smooth real vector field when $\mathbb K=\mathbb C$, there is a
constant $L_{m,\delta}<\infty$ such that
\begin{equation}
    \bigl\|
        \mathbf F_i(\mathbf v)
        -
        \mathbf F_i(\mathbf w)
    \bigr\|
    \leq
    L_{m,\delta}
    \|\mathbf v-\mathbf w\|
    \label{eq:a3-slow-lipschitz}
\end{equation}
for all $\mathbf v,\mathbf w$ in this bounded set and all
$i\notin\mathcal S_m$.

Set
\begin{equation}
    \mathbf d_{ki,\sigma}(\theta)
    :=
    \mathbf y_{ki,\sigma}(\theta)
    -
    \boldsymbol{\zeta}_{ki}(\theta).
    \label{eq:a3-slow-profile-error-vector}
\end{equation}
Subtracting the isolated profile equation
\eqref{eq:a3-slow-isolated-profile-equation} from the perturbed exact
equation \eqref{eq:a3-slow-available-vector-equation} and using the
Lipschitz bound \eqref{eq:a3-slow-lipschitz} gives
\begin{equation}
    \frac{d}{d\theta}
    \|\mathbf d_{ki,\sigma}(\theta)\|
    \leq
    L_{m,\delta}
    \|\mathbf d_{ki,\sigma}(\theta)\|
    +
    C_{m,\delta}\sigma^\nu
    \label{eq:a3-slow-profile-error-differential}
\end{equation}
in the standard upper-Dini-derivative sense at points where the norm
vanishes.

Gronwall's inequality and the initial profile error
\eqref{eq:a3-induction-profile-preservation} therefore give
\begin{equation}
    \sup_{\theta\in
        [\rho_m^\sigma,\theta_{\mathrm{boot}}^\sigma]}
    \|\mathbf d_{ki,\sigma}(\theta)\|
    \leq
    C_{m,\delta}'
    \left(
        \varepsilon_{m,\sigma}
        +
        \sigma^\nu
    \right).
    \label{eq:a3-slow-profile-gronwall}
\end{equation}
Since $\varepsilon_{m,\sigma}\to0$, the right-hand side tends to zero
uniformly over the finitely many available candidates. In particular, for
all sufficiently small $\sigma$, it is strictly smaller than the unit
threshold in the profile bootstrap
\eqref{eq:a3-slow-bootstrap-profile}.

\paragraph{Step 5: claimed coordinates remain nearly frozen.}
The claimed-sector near-freezing estimate
\eqref{eq:a3-prop-claimed-frozen} from
Proposition~\ref{prop:a3-locked-background} gives
\begin{equation}
    \max_{\substack{k\notin\mathcal A_m,\ i\in\mathcal S_m\\a\in[\nu]}}
    \left|
        \frac{d y_{ki,\sigma}^{(a)}}{d\theta}
    \right|
    \leq
    C_{m,\delta}\sigma^\nu
    \label{eq:a3-slow-claimed-derivative}
\end{equation}
throughout the bootstrap interval. Since
$I_{m,\delta}^\sigma$ has uniformly bounded length, integration gives
\begin{equation}
    \max_{\substack{k\notin\mathcal A_m,\ i\in\mathcal S_m\\a\in[\nu]}}
    \sup_{\theta\in
        [\rho_m^\sigma,\theta_{\mathrm{boot}}^\sigma]}
    \left|
        y_{ki,\sigma}^{(a)}(\theta)
        -
        y_{ki,\sigma}^{(a)}(\rho_m^\sigma)
    \right|
    \leq
    C_{m,\delta}'\sigma^\nu.
    \label{eq:a3-slow-claimed-integrated}
\end{equation}
Together with the claimed-sector reset bound
\eqref{eq:a3-induction-blocked-rescaled}, this is a strict improvement of
the bootstrap condition \eqref{eq:a3-slow-bootstrap-claimed} for
sufficiently small $\sigma$.

\paragraph{Step 6: captured coordinates remain inside the coarse annulus.}
The captured-product stability
\eqref{eq:a3-prop-captured-stability} from
Proposition~\ref{prop:a3-locked-background} gives
\begin{equation}
    \max_{r\leq m}
    \sup_{\theta\in
        [\rho_m^\sigma,\theta_{\mathrm{boot}}^\sigma]}
    \left|
        \pi_{r,\sigma}(\theta)-\lambda_{i_r}
    \right|
    =
    O(\sigma^\nu).
    \label{eq:a3-slow-product-control}
\end{equation}
At the same time, the exact balance consequence
\eqref{eq:a3-prop-balance-stability} gives
\begin{equation}
    \max_{\substack{r\leq m\\a,b\in[\nu]}}
    \sup_{\theta\in
        [\rho_m^\sigma,\theta_{\mathrm{boot}}^\sigma]}
    \left|
        |q_{r,\sigma}^{(a)}(\theta)|^2
        -
        |q_{r,\sigma}^{(b)}(\theta)|^2
    \right|
    =
    O(\sigma^2).
    \label{eq:a3-slow-all-magnitudes-equal}
\end{equation}
Thus the mode-wise magnitudes of each captured atom are uniformly
asymptotically equal throughout the bootstrap interval.

On the other hand, the product control
\eqref{eq:a3-slow-product-control} implies
\begin{equation}
    \prod_{a=1}^{\nu}
    \bigl|
        q_{r,\sigma}^{(a)}(\theta)
    \bigr|
    =
    \bigl|
        \pi_{r,\sigma}(\theta)
    \bigr|
    =
    \lambda_{i_r}+o(1)
    \label{eq:a3-slow-magnitude-product}
\end{equation}
uniformly on the same interval. Combining the asymptotic equality of the
mode-wise magnitudes
\eqref{eq:a3-slow-all-magnitudes-equal} with the product identity
\eqref{eq:a3-slow-magnitude-product} yields
\begin{equation}
    \max_{\substack{r\leq m\\a\in[\nu]}}
    \sup_{\theta\in
        [\rho_m^\sigma,\theta_{\mathrm{boot}}^\sigma]}
    \left|
        \bigl|
            q_{r,\sigma}^{(a)}(\theta)
        \bigr|
        -
        \lambda_{i_r}^{1/\nu}
    \right|
    \longrightarrow0.
    \label{eq:a3-slow-balanced-magnitudes}
\end{equation}
For sufficiently small $\sigma$, every captured main coordinate therefore
lies strictly inside the coarse annulus
\eqref{eq:a3-slow-bootstrap-captured}.

\paragraph{Step 7: closure of the continuation argument.}
The profile estimate \eqref{eq:a3-slow-profile-gronwall}, the claimed-sector
estimate \eqref{eq:a3-slow-claimed-integrated}, and the captured-factor
estimate \eqref{eq:a3-slow-balanced-magnitudes} strictly improve,
respectively, the three bootstrap conditions
\eqref{eq:a3-slow-bootstrap-profile},
\eqref{eq:a3-slow-bootstrap-claimed}, and
\eqref{eq:a3-slow-bootstrap-captured}. By the same continuity argument as
in Lemma~\ref{lem:a3-bootstrap-continuation}, the bootstrap interval cannot
terminate before the right endpoint of $I_{m,\delta}^\sigma$. Hence
\begin{equation}
    \theta_{\mathrm{boot}}^\sigma
    =
    \theta_{m+1}^*-\delta.
    \label{eq:a3-slow-bootstrap-closes}
\end{equation}

The Gronwall estimate \eqref{eq:a3-slow-profile-gronwall} now gives the
uniform profile convergence
\eqref{eq:a3-slow-profile-convergence} on the full interval, while the
integrated near-freezing estimate
\eqref{eq:a3-slow-claimed-integrated} gives
\eqref{eq:a3-slow-blocked-coordinate-control}. The microscopic coordinate
bounds used in \eqref{eq:a3-slow-bootstrap-implies-small} yield the
survivor estimate \eqref{eq:a3-slow-survivors-remain-small}. Finally, the
captured-background stability
\eqref{eq:a3-prop-captured-stability} and the balanced-magnitude estimate
\eqref{eq:a3-slow-balanced-magnitudes} give
\eqref{eq:a3-slow-captured-lock} and
\eqref{eq:a3-slow-captured-balance}.

\paragraph{Step 8: tensor and loss convergence.}
The tensor-level conclusion
\eqref{eq:a3-prop-output} of
Proposition~\ref{prop:a3-locked-background}, now valid on the full interval
$I_{m,\delta}^\sigma$, gives
\begin{equation}
    \sup_{\theta\in I_{m,\delta}^\sigma}
    \bigl\|
        \widehat T_\sigma(\theta)-T_m
    \bigr\|_F
    =
    O(\sigma^\nu).
    \label{eq:a3-slow-output-rate}
\end{equation}
This proves the uniform tensor convergence
\eqref{eq:a3-slow-output-convergence}.

Since the loss is one half of the squared Frobenius residual, the tensor
estimate \eqref{eq:a3-slow-output-rate} implies
\begin{equation}
    \sup_{\theta\in I_{m,\delta}^\sigma}
    \left|
        \frac12
        \bigl\|
            T-\widehat T_\sigma(\theta)
        \bigr\|_F^2
        -
        \frac12
        \|T-T_m\|_F^2
    \right|
    \longrightarrow0.
    \label{eq:a3-slow-loss-from-output}
\end{equation}
Finally, orthonormality of the diagonal target components and the definition
of the partial target $T_m$ in \eqref{eq:a3-partial-target} give
\begin{equation}
    \|T-T_m\|_F^2
    =
    \sum_{i\notin\mathcal S_m}\lambda_i^2.
    \label{eq:a3-slow-residual-target-norm}
\end{equation}
Substitution into \eqref{eq:a3-slow-loss-from-output} proves the loss
convergence \eqref{eq:a3-slow-loss-convergence}.
\end{proof}

Proposition~\ref{prop:a3-slow-propagation} controls the dynamics uniformly
on every fixed interval that stays a positive distance before the next
accepted clock. It does not yet describe a shrinking neighborhood of
$\theta_{m+1}^*$, where the winning isolated profile
$\boldsymbol{\zeta}_{k_{m+1}i_{m+1}}$ becomes large and the microscopic
approximation eventually ceases to be valid. The next subsection therefore
constructs a $\sigma$-dependent entrance time approaching
$\theta_{m+1}^*$ from below. At that time, the winning candidate will already
be much larger than the initialization scale while still vanishing in the
original variables, whereas all competing candidates remain microscopic.
This provides the initial state for the subsequent fast-capture analysis.

\subsection{Entrance into the next fast layer}
\label{app:fast-entrance}

Proposition~\ref{prop:a3-slow-propagation} controls the dynamics on every
fixed compact interval lying strictly before the next accepted clock. We now
extend this control to a $\sigma$-dependent time approaching that clock.

The main idea is a diagonal argument. For every fixed large amplitude $R$,
the winning isolated profile reaches amplitude $R$ at a time strictly before
its blow-up, so Proposition~\ref{prop:a3-slow-propagation} applies up to that
time. We then let the entrance level diverge sufficiently slowly as
$\sigma\to0$. The resulting exact winning atom is much larger than the
initialization scale but still vanishing in the original variables. This
provides the entrance state for the fast-capture analysis of the next
subsection.

We first record the asymptotic form of an isolated profile near its
activation clock.

\begin{lemma}[Approach to an isolated activation clock]
\label{lem:a3-isolated-blowup}
Let $(k,i)$ be any candidate and let
$\boldsymbol{\zeta}_{ki}(\theta)$ be its isolated profile
\eqref{eq:a3-candidate-profile}, with activation clock $\tau_{ki}$. Define
its geometric-mean amplitude by
\begin{equation}
    A_{ki}(\theta)
    :=
    \left(
        \prod_{a=1}^{\nu}
        \left|
            \zeta_{ki}^{(a)}(\theta)
        \right|
    \right)^{1/\nu}.
    \label{eq:a3-isolated-amplitude}
\end{equation}
Then, as $\theta\uparrow\tau_{ki}$,
\begin{equation}
    A_{ki}(\theta)^{\nu-2}
    \bigl(
        \tau_{ki}-\theta
    \bigr)
    \longrightarrow
    \frac{1}{(\nu-2)\lambda_i}.
    \label{eq:a3-isolated-blowup-rate}
\end{equation}
Moreover, the factor magnitudes become asymptotically balanced:
\begin{equation}
    \frac{
        \left|
            \zeta_{ki}^{(a)}(\theta)
        \right|
    }{
        A_{ki}(\theta)
    }
    \longrightarrow
    1,
    \qquad
    a\in[\nu].
    \label{eq:a3-isolated-blowup-balance}
\end{equation}
There also exist phases
$\omega_{ki}^{(a)}\in\mathbb K$ with
$|\omega_{ki}^{(a)}|=1$ such that
\begin{equation}
    \frac{
        \zeta_{ki}^{(a)}(\theta)
    }{
        |\zeta_{ki}^{(a)}(\theta)|
    }
    \longrightarrow
    \omega_{ki}^{(a)},
    \qquad
    a\in[\nu],
    \label{eq:a3-isolated-phase-limit}
\end{equation}
and these limiting phases satisfy
\begin{equation}
    \prod_{a=1}^{\nu}
    \omega_{ki}^{(a)}
    =
    1.
    \label{eq:a3-isolated-phase-product}
\end{equation}
For $\mathbb K=\mathbb R$, the limiting phases
$\omega_{ki}^{(a)}$ belong to $\{-1,1\}$.
\end{lemma}

\begin{proof}
Recall from the scalar-flow analysis in Appendix~\ref{app:setup} that the
isolated unit-amplitude flow satisfies
\begin{equation}
    |z^{(a)}(s)|^2
    =
    \rho_a+q(s),
    \qquad
    \rho_a:=|g^{(a)}|^2,
    \label{eq:a3-blowup-q-representation}
\end{equation}
where, on the final increasing branch approaching the blow-up time,
\begin{equation}
    q'(s)
    =
    2
    \sqrt{
        \prod_{a=1}^{\nu}
        \bigl(
            \rho_a+q(s)
        \bigr)
        -
        B^2
    },
    \qquad
    B:=
    \operatorname{Im}
    \prod_{a=1}^{\nu}g^{(a)}.
    \label{eq:a3-blowup-q-equation}
\end{equation}
As $q\to\infty$, the scalar equation
\eqref{eq:a3-blowup-q-equation} gives
\begin{equation}
    q'(s)
    =
    2q(s)^{\nu/2}
    \left(
        1+O(q(s)^{-1})
    \right).
    \label{eq:a3-blowup-q-asymptotic}
\end{equation}
Writing
$r(s):=\sqrt{q(s)}$, the large-amplitude equation
\eqref{eq:a3-blowup-q-asymptotic} becomes
\begin{equation}
    r'(s)
    =
    r(s)^{\nu-1}
    \left(
        1+O(r(s)^{-2})
    \right).
    \label{eq:a3-blowup-r-equation}
\end{equation}
Equivalently,
\begin{equation}
    \frac{d}{ds}
    r(s)^{-(\nu-2)}
    =
    -(\nu-2)
    \left(
        1+O(r(s)^{-2})
    \right).
    \label{eq:a3-blowup-reciprocal-equation}
\end{equation}
Integrating the reciprocal-amplitude equation
\eqref{eq:a3-blowup-reciprocal-equation} up to the blow-up time
$\tau_\nu(\mathbf g)$ gives
\begin{equation}
    r(s)^{\nu-2}
    \bigl(
        \tau_\nu(\mathbf g)-s
    \bigr)
    \longrightarrow
    \frac{1}{\nu-2}.
    \label{eq:a3-blowup-unit-clock-rate}
\end{equation}

For the candidate $(k,i)$,
the representation \eqref{eq:a3-blowup-q-representation} reads
\[
    |\zeta_{ki}^{(a)}(\theta)|^2
    =
    \rho_a+q(\lambda_i\theta),
\]
and its activation clock satisfies
$\tau_\nu(\mathbf g_{ki})=\lambda_i\tau_{ki}$. Since the geometric-mean
amplitude is asymptotic to
$\sqrt{q(\lambda_i\theta)}$, the unit-flow blow-up law
\eqref{eq:a3-blowup-unit-clock-rate} gives the clock asymptotic
\eqref{eq:a3-isolated-blowup-rate}.

The same representation shows that
\begin{equation}
    |\zeta_{ki}^{(a)}(\theta)|^2
    -
    |\zeta_{ki}^{(b)}(\theta)|^2
    =
    \rho_a-\rho_b
    \label{eq:a3-blowup-isolated-balance-invariant}
\end{equation}
is constant. Since all factor magnitudes diverge as
$\theta\uparrow\tau_{ki}$, the fixed differences in
\eqref{eq:a3-blowup-isolated-balance-invariant} become negligible relative
to their common magnitude. This proves the asymptotic balance
\eqref{eq:a3-isolated-blowup-balance}.

It remains to identify the limiting phases. Define the scalar product
\begin{equation}
    P_{ki}(\theta)
    :=
    \prod_{a=1}^{\nu}
    \zeta_{ki}^{(a)}(\theta).
    \label{eq:a3-blowup-product}
\end{equation}
The isolated equations imply that
$\operatorname{Im}P_{ki}(\theta)$ is constant, while
$\operatorname{Re}P_{ki}(\theta)\to+\infty$ as the profile approaches its
clock. Therefore
\begin{equation}
    \frac{
        P_{ki}(\theta)
    }{
        |P_{ki}(\theta)|
    }
    \longrightarrow
    1.
    \label{eq:a3-blowup-product-phase}
\end{equation}

For any mode $a$, once
$\zeta_{ki}^{(a)}(\theta)\neq0$, the isolated flow gives
\begin{equation}
    \frac{d}{d\theta}
    \arg\zeta_{ki}^{(a)}(\theta)
    =
    -
    \lambda_i
    \frac{
        \operatorname{Im}P_{ki}(\theta)
    }{
        |\zeta_{ki}^{(a)}(\theta)|^2
    }.
    \label{eq:a3-blowup-phase-derivative}
\end{equation}
The numerator in the phase equation
\eqref{eq:a3-blowup-phase-derivative} is constant, whereas its denominator
diverges. Hence
\begin{equation}
    \frac{d}{d\theta}
    \arg\zeta_{ki}^{(a)}(\theta)
    \longrightarrow0
    \qquad
    \text{as }\theta\uparrow\tau_{ki}.
    \label{eq:a3-blowup-phase-derivative-vanishes}
\end{equation}
In particular, the phase derivative is bounded in a left neighborhood of
the finite time $\tau_{ki}$ and is therefore integrable there. Thus every
factor phase has a finite limit, proving
\eqref{eq:a3-isolated-phase-limit}. Taking the product of the limiting
phases and using the product-phase limit
\eqref{eq:a3-blowup-product-phase} gives
\eqref{eq:a3-isolated-phase-product}.
\end{proof}

We now specialize to the next accepted edge. Write
\begin{equation}
    k_*:=k_{m+1},
    \qquad
    i_*:=i_{m+1},
    \qquad
    \theta_*:=\theta_{m+1}^*
    =
    \tau_{k_*i_*}.
    \label{eq:a3-entrance-winner-notation}
\end{equation}
Since all candidate clocks are distinct and $(k_*,i_*)$ is the unique
minimum over the available set $\mathcal E_m$, every other available
candidate has a strictly later clock. Because $\mathcal E_m$ is finite,
there is a positive clock gap
\begin{equation}
    \Delta_m
    :=
    \min_{\substack{
        (k,i)\in\mathcal E_m\\
        (k,i)\neq(k_*,i_*)
    }}
    \bigl(
        \tau_{ki}-\theta_*
    \bigr)
    >
    0,
    \label{eq:a3-entrance-clock-gap}
\end{equation}
with the convention $\Delta_m=+\infty$ if
$\mathcal E_m=\{(k_*,i_*)\}$. Consequently, all nonwinning available
profiles remain uniformly bounded up to the winning clock:
\begin{equation}
    B_m^{\mathrm{nw}}
    :=
    \max_{\substack{
        (k,i)\in\mathcal E_m\\
        (k,i)\neq(k_*,i_*)
    }}
    \max_{a\in[\nu]}
    \sup_{0\leq\theta\leq\theta_*}
    \left|
        \zeta_{ki}^{(a)}(\theta)
    \right|
    <
    \infty,
    \label{eq:a3-entrance-nonwinner-bound}
\end{equation}
where the maximum over the empty set is interpreted as zero.

For the winning isolated profile, define
\begin{equation}
    A_*(\theta)
    :=
    A_{k_*i_*}(\theta).
    \label{eq:a3-entrance-winning-amplitude}
\end{equation}
By the isolated blow-up lemma,
Lemma~\ref{lem:a3-isolated-blowup},
$A_*(\theta)\to\infty$ as $\theta\uparrow\theta_*$. Hence, for every
sufficiently large $R$, there is a unique time on the final increasing
branch, denoted by $\eta_{m+1}(R)$, such that
\begin{equation}
    A_*\bigl(\eta_{m+1}(R)\bigr)
    =
    R.
    \label{eq:a3-entrance-fixed-level}
\end{equation}
Indeed, the scalar representation
\eqref{eq:a3-blowup-q-representation} gives
\begin{equation}
    A_*(\theta)
    =
    \left[
        \prod_{a=1}^{\nu}
        \bigl(
            \rho_a+q(\lambda_{i_*}\theta)
        \bigr)
    \right]^{1/(2\nu)},
    \label{eq:a3-entrance-amplitude-q}
\end{equation}
and the scalar equation
\eqref{eq:a3-blowup-q-equation} shows that $q$ is strictly increasing on
the final branch. Thus sufficiently large amplitude levels are crossed
exactly once.

The blow-up-rate estimate
\eqref{eq:a3-isolated-blowup-rate} furthermore gives
\begin{equation}
    \theta_*-\eta_{m+1}(R)
    =
    \frac{
        1+o_R(1)
    }{
        (\nu-2)\lambda_{i_*}R^{\nu-2}
    },
    \qquad
    R\to\infty.
    \label{eq:a3-entrance-fixed-level-gap}
\end{equation}

\begin{proposition}[Mesoscopic entrance into the next fast layer]
\label{prop:a3-mesoscopic-entrance}
Fix $m\in\{0,\ldots,M-1\}$ and suppose that the induction hypotheses
$\mathrm{(I_m)}$--$\mathrm{(IV_m)}$ hold at the reset time
$\rho_m^\sigma$.

Then one can choose an initialization-dependent scale
$R_{m,\sigma}$ and define
\begin{equation}
    \eta_{m+1}^\sigma
    :=
    \eta_{m+1}(R_{m,\sigma})
    \label{eq:a3-entrance-time}
\end{equation}
such that the following properties hold.

\begin{enumerate}
\item
\emph{Mesoscopic scale.}
The scale can be chosen so that
\begin{equation}
    R_{m,\sigma}\longrightarrow\infty,
    \qquad
    \sigma R_{m,\sigma}^q\longrightarrow0
    \quad
    \text{for every fixed }q>0.
    \label{eq:a3-entrance-mesoscopic-scale}
\end{equation}
Define the corresponding factor scale in the original variables by
\begin{equation}
    r_{m,\sigma}
    :=
    \sigma R_{m,\sigma}.
    \label{eq:a3-entrance-physical-amplitude}
\end{equation}
Then
\begin{equation}
    \sigma\ll r_{m,\sigma}\ll1.
    \label{eq:a3-entrance-scale-separation}
\end{equation}

\item
\emph{The entrance time approaches the next clock.}
We have
\begin{equation}
    \eta_{m+1}^\sigma
    \longrightarrow
    \theta_*,
    \label{eq:a3-entrance-time-limit}
\end{equation}
and, more precisely,
\begin{equation}
    R_{m,\sigma}^{\nu-2}
    \bigl(
        \theta_*-\eta_{m+1}^\sigma
    \bigr)
    \longrightarrow
    \frac{1}{(\nu-2)\lambda_{i_*}}.
    \label{eq:a3-entrance-time-asymptotic}
\end{equation}

\item
\emph{The winning atom has reached mesoscopic size.}
Let
\[
    \boldsymbol{\omega}_*
    =
    \bigl(
        \omega_*^{(1)},\ldots,\omega_*^{(\nu)}
    \bigr)
\]
be the limiting phase vector of the winning isolated profile from
Lemma~\ref{lem:a3-isolated-blowup}. Then
\begin{equation}
    \max_{a\in[\nu]}
    \left|
        \frac{
            x_{k_*i_*,\sigma}^{(a)}
            (\eta_{m+1}^\sigma)
        }{
            r_{m,\sigma}
        }
        -
        \omega_*^{(a)}
    \right|
    \longrightarrow0,
    \label{eq:a3-entrance-winning-main-coordinate}
\end{equation}
where
\begin{equation}
    |\omega_*^{(a)}|=1,
    \qquad
    \prod_{a=1}^{\nu}\omega_*^{(a)}=1.
    \label{eq:a3-entrance-winning-phase-condition}
\end{equation}
Thus the winning main coordinates are asymptotically balanced and phase
aligned.

\item
\emph{The winning atom is localized in the winning sector.}
There exists a constant
$C_m^{\mathrm{ent}}<\infty$, independent of $\sigma$, such that
\begin{equation}
    \max_{a\in[\nu]}
    \left\|
        \mathbf u_{k_*,\sigma}^{(a)}
        (\eta_{m+1}^\sigma)
        -
        x_{k_*i_*,\sigma}^{(a)}
        (\eta_{m+1}^\sigma)
        \mathbf e_{i_*}^{(a)}
    \right\|
    \leq
    C_m^{\mathrm{ent}}\sigma
    \label{eq:a3-entrance-winning-localization}
\end{equation}
for all sufficiently small $\sigma$. In view of the scale separation
\eqref{eq:a3-entrance-scale-separation}, this off-sector component is
$o(r_{m,\sigma})$.

\item
\emph{All other surviving atoms remain microscopic.}
After increasing $C_m^{\mathrm{ent}}$ if necessary,
\begin{equation}
    \max_{\substack{
        k\notin\mathcal A_m\\
        k\neq k_*\\
        a\in[\nu]
    }}
    \left\|
        \mathbf u_{k,\sigma}^{(a)}
        (\eta_{m+1}^\sigma)
    \right\|
    \leq
    C_m^{\mathrm{ent}}\sigma
    \label{eq:a3-entrance-other-survivors}
\end{equation}
for all sufficiently small $\sigma$.

\item
\emph{The previously captured background remains quantitatively locked.}
After increasing $C_m^{\mathrm{ent}}$ once more if necessary, the product
and off-sector errors of every previously captured atom satisfy
\begin{equation}
    \max_{r\leq m}
    \left|
        \pi_{r,\sigma}(\eta_{m+1}^\sigma)
        -
        \lambda_{i_r}
    \right|
    +
    \max_{\substack{r\leq m\\a\in[\nu]}}
    \left\|
        \mathbf h_{r,\sigma}^{(a)}
        (\eta_{m+1}^\sigma)
    \right\|
    \leq
    C_m^{\mathrm{ent}}\sigma.
    \label{eq:a3-entrance-old-quantitative-locking}
\end{equation}
Their factor magnitudes remain asymptotically balanced:
\begin{equation}
    \max_{\substack{r\leq m\\a\in[\nu]}}
    \left|
        \left|
            q_{r,\sigma}^{(a)}
            (\eta_{m+1}^\sigma)
        \right|
        -
        \lambda_{i_r}^{1/\nu}
    \right|
    \longrightarrow0.
    \label{eq:a3-entrance-old-balance}
\end{equation}
Consequently, the tensor output immediately before the fast transition
satisfies
\begin{equation}
    \widehat T_\sigma(\eta_{m+1}^\sigma)
    \longrightarrow
    T_m.
    \label{eq:a3-entrance-output-before-capture}
\end{equation}
\end{enumerate}
\end{proposition}

\begin{proof}
The proof is a diagonalization of the fixed-distance estimates from
Proposition~\ref{prop:a3-slow-propagation}. The important point is that the
constants in that proposition may depend on the distance from the next
clock. We therefore keep the entrance level fixed while taking
$\sigma\to0$, and only afterwards let the entrance level diverge. This
produces a single $\sigma$-dependent scale for which all estimates needed by
the subsequent fast analysis hold simultaneously.

\paragraph{Step 1: fixed entrance levels.}
Choose a strictly increasing sequence
\begin{equation}
    1<R_1<R_2<\cdots,
    \qquad
    R_n\longrightarrow\infty.
    \label{eq:a3-entrance-Rn}
\end{equation}
By the clock asymptotic and phase-balance conclusions of
Lemma~\ref{lem:a3-isolated-blowup}, we may choose this sequence so that, if
\begin{equation}
    \eta_n
    :=
    \eta_{m+1}(R_n),
    \label{eq:a3-entrance-eta-n}
\end{equation}
then
\begin{equation}
    \left|
        R_n^{\nu-2}
        \bigl(
            \theta_*-\eta_n
        \bigr)
        -
        \frac{1}{(\nu-2)\lambda_{i_*}}
    \right|
    \leq
    \frac1n
    \label{eq:a3-entrance-fixed-rate-accuracy}
\end{equation}
and
\begin{equation}
    \max_{a\in[\nu]}
    \left|
        \frac{
            \zeta_{k_*i_*}^{(a)}(\eta_n)
        }{
            R_n
        }
        -
        \omega_*^{(a)}
    \right|
    \leq
    \frac1n.
    \label{eq:a3-entrance-fixed-profile-accuracy}
\end{equation}
By increasing $R_n$ further if necessary, we may also ensure
\begin{equation}
    0<
    \theta_*-\eta_n
    <
    \frac12
    \bigl(
        \theta_{m+1}^*-\theta_m^*
    \bigr).
    \label{eq:a3-entrance-fixed-delta-admissible}
\end{equation}
Since $\theta_*=\theta_{m+1}^*$, the last inequality implies
$\eta_n>\theta_m^*$.

Define the fixed distance from the next clock by
\begin{equation}
    \delta_n
    :=
    \theta_*-\eta_n.
    \label{eq:a3-entrance-delta-n}
\end{equation}
For each fixed $n$, $\delta_n>0$ is independent of $\sigma$, so the
slow-propagation result
Proposition~\ref{prop:a3-slow-propagation} applies with
$\delta=\delta_n$. Moreover, since
$\rho_m^\sigma\to\theta_m^*$ and $\eta_n>\theta_m^*$, the interval
$[\rho_m^\sigma,\eta_n]$ is nonempty for all sufficiently small $\sigma$.

\paragraph{Step 2: estimates at a fixed entrance level.}
Fix $n$. Applying the available-profile conclusion
\eqref{eq:a3-slow-profile-convergence} of
Proposition~\ref{prop:a3-slow-propagation} with
$\delta=\delta_n$ gives
\begin{equation}
    \max_{\substack{(k,i)\in\mathcal E_m\\a\in[\nu]}}
    \sup_{\theta\in[\rho_m^\sigma,\eta_n]}
    \left|
        y_{ki,\sigma}^{(a)}(\theta)
        -
        \zeta_{ki}^{(a)}(\theta)
    \right|
    \longrightarrow0
    \qquad
    \text{as }\sigma\to0.
    \label{eq:a3-entrance-fixed-profile-convergence}
\end{equation}

For the microscopic coordinates of surviving atoms inside already claimed
sectors, the near-freezing conclusion
\eqref{eq:a3-slow-blocked-coordinate-control} gives a finite constant
$C_n^{\mathrm{cl}}<\infty$, depending on $n$ through $\delta_n$ but not on
$\sigma$, such that
\begin{equation}
    \max_{\substack{
        k\notin\mathcal A_m,\,
        i\in\mathcal S_m\\
        a\in[\nu]
    }}
    \sup_{\theta\in[\rho_m^\sigma,\eta_n]}
    \left|
        y_{ki,\sigma}^{(a)}(\theta)
        -
        y_{ki,\sigma}^{(a)}(\rho_m^\sigma)
    \right|
    \leq
    C_n^{\mathrm{cl}}\sigma^\nu
    \label{eq:a3-entrance-fixed-claimed-change}
\end{equation}
for all sufficiently small $\sigma$.

The captured-background part of
Proposition~\ref{prop:a3-slow-propagation} is also quantitative. In
particular, the locking estimate
\eqref{eq:a3-slow-captured-lock} gives, for some finite constant
$C_n^{\mathrm{old}}<\infty$,
\begin{equation}
    \max_{r\leq m}
    \left|
        \pi_{r,\sigma}(\eta_n)-\lambda_{i_r}
    \right|
    +
    \max_{\substack{r\leq m\\a\in[\nu]}}
    \left\|
        \mathbf h_{r,\sigma}^{(a)}(\eta_n)
    \right\|
    \leq
    C_n^{\mathrm{old}}\sigma^\nu
    \label{eq:a3-entrance-fixed-old-quantitative}
\end{equation}
for all sufficiently small $\sigma$.

Finally, the captured-factor balance conclusion
\eqref{eq:a3-slow-captured-balance} gives
\begin{equation}
    \max_{\substack{r\leq m\\a\in[\nu]}}
    \left|
        \left|
            q_{r,\sigma}^{(a)}(\eta_n)
        \right|
        -
        \lambda_{i_r}^{1/\nu}
    \right|
    \longrightarrow0
    \qquad
    \text{as }\sigma\to0.
    \label{eq:a3-entrance-fixed-old-balance}
\end{equation}

The constants $C_n^{\mathrm{cl}}$ and $C_n^{\mathrm{old}}$ may diverge as
$n\to\infty$, because $\delta_n\downarrow0$. We do not require any
uniformity in $n$. Instead, the next step absorbs these constants into the
diagonal choice of the initialization scale.

\paragraph{Step 3: diagonal choice of $\sigma_n$ and $R_{m,\sigma}$.}
We recursively choose a strictly decreasing sequence
$\sigma_n\downarrow0$ such that
\begin{equation}
    \sigma_{n+1}
    <
    \frac12\sigma_n,
    \qquad
    \sigma_n
    <
    \frac1n.
    \label{eq:a3-entrance-sigma-n-decreasing}
\end{equation}
For each fixed $n$, choose $\sigma_n$ sufficiently small that, whenever
$0<\sigma\leq\sigma_n$, all the fixed-$n$ estimates from Step~2 apply and
the following additional bounds hold.

First, the available-profile approximation at the fixed entrance time
$\eta_n$ satisfies
\begin{equation}
    \max_{\substack{(k,i)\in\mathcal E_m\\a\in[\nu]}}
    \left|
        y_{ki,\sigma}^{(a)}(\eta_n)
        -
        \zeta_{ki}^{(a)}(\eta_n)
    \right|
    \leq
    \frac1n.
    \label{eq:a3-entrance-diagonal-profile-control}
\end{equation}
Second, the balance error of the previously captured atoms satisfies
\begin{equation}
    \max_{\substack{r\leq m\\a\in[\nu]}}
    \left|
        \left|
            q_{r,\sigma}^{(a)}(\eta_n)
        \right|
        -
        \lambda_{i_r}^{1/\nu}
    \right|
    \leq
    \frac1n.
    \label{eq:a3-entrance-diagonal-old-balance}
\end{equation}

We also absorb the possibly $n$-dependent constants in the claimed-sector
and old-background estimates by requiring
\begin{equation}
    C_n^{\mathrm{cl}}\sigma_n^\nu
    \leq
    \frac1n
    \label{eq:a3-entrance-diagonal-claimed-control}
\end{equation}
and
\begin{equation}
    C_n^{\mathrm{old}}\sigma_n^{\nu-1}
    \leq
    1.
    \label{eq:a3-entrance-diagonal-old-quantitative}
\end{equation}
Finally, we require the entrance level to grow sufficiently slowly relative
to $\sigma_n$:
\begin{equation}
    \sigma_nR_n^n
    \leq
    \frac1n.
    \label{eq:a3-entrance-diagonal-slow-growth}
\end{equation}
All these requirements can be imposed simultaneously because, at this
stage of the construction, $n$, $R_n$, $\eta_n$,
$C_n^{\mathrm{cl}}$, and $C_n^{\mathrm{old}}$ are fixed and only
$\sigma_n$ is being made smaller.

For
\begin{equation}
    \sigma_{n+1}<\sigma\leq\sigma_n,
    \label{eq:a3-entrance-diagonal-block}
\end{equation}
define
\begin{equation}
    R_{m,\sigma}:=R_n,
    \qquad
    \eta_{m+1}^\sigma:=\eta_n.
    \label{eq:a3-entrance-diagonal-definition}
\end{equation}
Since the block index $n$ tends to infinity as $\sigma\to0$, we immediately
obtain
\begin{equation}
    R_{m,\sigma}\longrightarrow\infty.
    \label{eq:a3-entrance-Rsigma-diverges}
\end{equation}

Now fix $q>0$. For all sufficiently large $n$, we have $n\geq q$, and
since $R_n>1$,
\begin{equation}
    \sigma R_{m,\sigma}^q
    \leq
    \sigma_nR_n^q
    \leq
    \sigma_nR_n^n
    \leq
    \frac1n.
    \label{eq:a3-entrance-diagonal-polynomial}
\end{equation}
Thus
\begin{equation}
    \sigma R_{m,\sigma}^q
    \longrightarrow0
    \qquad
    \text{for every fixed }q>0,
    \label{eq:a3-entrance-diagonal-polynomial-limit}
\end{equation}
which proves the mesoscopic-scale condition
\eqref{eq:a3-entrance-mesoscopic-scale}. In particular, with
$r_{m,\sigma}=\sigma R_{m,\sigma}$,
\begin{equation}
    \frac{r_{m,\sigma}}{\sigma}
    =
    R_{m,\sigma}
    \longrightarrow\infty,
    \qquad
    r_{m,\sigma}
    =
    \sigma R_{m,\sigma}
    \longrightarrow0.
    \label{eq:a3-entrance-rsigma-separation}
\end{equation}
This is precisely the scale separation
\eqref{eq:a3-entrance-scale-separation}.

\paragraph{Step 4: location of the entrance time.}
If $\sigma$ belongs to the $n$th diagonal block
\eqref{eq:a3-entrance-diagonal-block}, then
$\eta_{m+1}^\sigma=\eta_n$ and $R_{m,\sigma}=R_n$. Therefore the
fixed-level clock estimate
\eqref{eq:a3-entrance-fixed-rate-accuracy} gives
\begin{equation}
    \left|
        R_{m,\sigma}^{\nu-2}
        \bigl(
            \theta_*-\eta_{m+1}^\sigma
        \bigr)
        -
        \frac{1}{(\nu-2)\lambda_{i_*}}
    \right|
    \leq
    \frac1n.
    \label{eq:a3-entrance-diagonal-time-accuracy}
\end{equation}
Since $n\to\infty$ as $\sigma\to0$, this proves the precise entrance-time
asymptotic \eqref{eq:a3-entrance-time-asymptotic}. Since
$R_{m,\sigma}\to\infty$, the same relation also gives
\begin{equation}
    \theta_*-\eta_{m+1}^\sigma
    \longrightarrow0,
    \label{eq:a3-entrance-diagonal-time-gap-zero}
\end{equation}
and hence the convergence of the entrance time
\eqref{eq:a3-entrance-time-limit}.

\paragraph{Step 5: the winning main coordinates.}
For $\sigma$ in the $n$th diagonal block, the diagonal profile estimate
\eqref{eq:a3-entrance-diagonal-profile-control} gives
\begin{equation}
    \max_{a\in[\nu]}
    \left|
        y_{k_*i_*,\sigma}^{(a)}
        (\eta_{m+1}^\sigma)
        -
        \zeta_{k_*i_*}^{(a)}(\eta_n)
    \right|
    \leq
    \frac1n.
    \label{eq:a3-entrance-winning-y-error}
\end{equation}
Using
$x_{k_*i_*,\sigma}^{(a)}
=\sigma y_{k_*i_*,\sigma}^{(a)}$ and
$r_{m,\sigma}=\sigma R_n$, we obtain
\begin{equation}
    \begin{aligned}
    \left|
        \frac{
            x_{k_*i_*,\sigma}^{(a)}
            (\eta_{m+1}^\sigma)
        }{
            r_{m,\sigma}
        }
        -
        \omega_*^{(a)}
    \right|
    &\leq
    \frac1{R_n}
    \left|
        y_{k_*i_*,\sigma}^{(a)}(\eta_n)
        -
        \zeta_{k_*i_*}^{(a)}(\eta_n)
    \right|
    \\
    &\quad
    +
    \left|
        \frac{
            \zeta_{k_*i_*}^{(a)}(\eta_n)
        }{
            R_n
        }
        -
        \omega_*^{(a)}
    \right|
    \\
    &\leq
    \frac{1}{nR_n}
    +
    \frac1n.
    \end{aligned}
    \label{eq:a3-entrance-winning-scaled-error}
\end{equation}
The right-hand side tends to zero. This proves the winning-coordinate
asymptotic \eqref{eq:a3-entrance-winning-main-coordinate}. The phase
condition \eqref{eq:a3-entrance-winning-phase-condition} is exactly the
phase-product conclusion of Lemma~\ref{lem:a3-isolated-blowup} applied to
the winning candidate.

\paragraph{Step 6: uniform control of the winning off-sector coordinates.}
We next show that all microscopic coordinates of the winning atom outside
sector $i_*$ remain uniformly bounded. This is where the diagonal control of
the $n$-dependent constants from Step~3 is needed.

First suppose that
$i\notin\mathcal S_m$ and $i\neq i_*$. Then
$(k_*,i)\in\mathcal E_m$ is a nonwinning available edge. By the positive
clock gap \eqref{eq:a3-entrance-clock-gap}, its isolated profile is regular
up to $\theta_*$, and the nonwinner bound
\eqref{eq:a3-entrance-nonwinner-bound} gives
\begin{equation}
    \left|
        \zeta_{k_*i}^{(a)}(\eta_n)
    \right|
    \leq
    B_m^{\mathrm{nw}}.
    \label{eq:a3-entrance-winning-unclaimed-isolated}
\end{equation}
Combining this bound with the diagonal profile approximation
\eqref{eq:a3-entrance-diagonal-profile-control} gives
\begin{equation}
    \left|
        y_{k_*i,\sigma}^{(a)}
        (\eta_{m+1}^\sigma)
    \right|
    \leq
    B_m^{\mathrm{nw}}+1
    \label{eq:a3-entrance-winning-unclaimed-bound}
\end{equation}
for all sufficiently small $\sigma$.

Now suppose that $i\in\mathcal S_m$. At the reset time, the blocked-sector
part of the induction hypothesis
\eqref{eq:a3-induction-blocked-rescaled} gives
\begin{equation}
    \left|
        y_{k_*i,\sigma}^{(a)}(\rho_m^\sigma)
    \right|
    \leq
    C_m.
    \label{eq:a3-entrance-winning-claimed-reset}
\end{equation}
If $\sigma$ belongs to the $n$th diagonal block, the fixed-level
near-freezing estimate \eqref{eq:a3-entrance-fixed-claimed-change} and the
diagonal choice \eqref{eq:a3-entrance-diagonal-claimed-control} imply
\begin{equation}
    \left|
        y_{k_*i,\sigma}^{(a)}(\eta_n)
        -
        y_{k_*i,\sigma}^{(a)}(\rho_m^\sigma)
    \right|
    \leq
    C_n^{\mathrm{cl}}\sigma^\nu
    \leq
    \frac1n.
    \label{eq:a3-entrance-winning-claimed-change}
\end{equation}
Together with the reset bound
\eqref{eq:a3-entrance-winning-claimed-reset}, this gives
\begin{equation}
    \left|
        y_{k_*i,\sigma}^{(a)}
        (\eta_{m+1}^\sigma)
    \right|
    \leq
    C_m+1.
    \label{eq:a3-entrance-winning-claimed-bound}
\end{equation}

Combining the unclaimed-sector estimate
\eqref{eq:a3-entrance-winning-unclaimed-bound} and the claimed-sector
estimate \eqref{eq:a3-entrance-winning-claimed-bound}, there is a constant
$C_m^{\mathrm{off}}<\infty$, independent of $\sigma$, such that
\begin{equation}
    \max_{\substack{i\neq i_*\\a\in[\nu]}}
    \left|
        y_{k_*i,\sigma}^{(a)}
        (\eta_{m+1}^\sigma)
    \right|
    \leq
    C_m^{\mathrm{off}}.
    \label{eq:a3-entrance-winning-offsector-y}
\end{equation}
Since the sector vectors
$\{\mathbf e_i^{(a)}\}_{i=1}^{p}$ form an orthonormal basis,
\begin{equation}
    \begin{aligned}
    &\left\|
        \mathbf u_{k_*,\sigma}^{(a)}
        (\eta_{m+1}^\sigma)
        -
        x_{k_*i_*,\sigma}^{(a)}
        (\eta_{m+1}^\sigma)
        \mathbf e_{i_*}^{(a)}
    \right\|^2
    \\
    &\qquad
    =
    \sum_{i\neq i_*}
    \left|
        x_{k_*i,\sigma}^{(a)}
        (\eta_{m+1}^\sigma)
    \right|^2
    \leq
    (p-1)
    \bigl(
        C_m^{\mathrm{off}}
    \bigr)^2
    \sigma^2.
    \end{aligned}
    \label{eq:a3-entrance-winning-localization-bound}
\end{equation}
Thus the winning off-sector component is $O(\sigma)$ uniformly in
$\sigma$. Since
$\sigma/r_{m,\sigma}=R_{m,\sigma}^{-1}\to0$, it is also
$o(r_{m,\sigma})$ relative to the winning main coordinates.

\paragraph{Step 7: uniform control of the other surviving atoms.}
Let
$k\notin\mathcal A_m$ with $k\neq k_*$. If
$i\notin\mathcal S_m$, then $(k,i)$ is an available edge and, because
$k\neq k_*$, it is not the winning edge. The same nonwinner-profile
argument used in
\eqref{eq:a3-entrance-winning-unclaimed-isolated}--%
\eqref{eq:a3-entrance-winning-unclaimed-bound} therefore gives
\begin{equation}
    \left|
        y_{ki,\sigma}^{(a)}
        (\eta_{m+1}^\sigma)
    \right|
    \leq
    B_m^{\mathrm{nw}}+1.
    \label{eq:a3-entrance-other-unclaimed-bound}
\end{equation}
If instead $i\in\mathcal S_m$, the reset bound
\eqref{eq:a3-induction-blocked-rescaled}, the fixed-level near-freezing
estimate \eqref{eq:a3-entrance-fixed-claimed-change}, and the diagonal
choice \eqref{eq:a3-entrance-diagonal-claimed-control} give
\begin{equation}
    \left|
        y_{ki,\sigma}^{(a)}
        (\eta_{m+1}^\sigma)
    \right|
    \leq
    C_m+1.
    \label{eq:a3-entrance-other-claimed-bound}
\end{equation}
Consequently, for a constant
$C_m^{\mathrm{surv}}<\infty$ independent of $\sigma$,
\begin{equation}
    \max_{\substack{
        k\notin\mathcal A_m,\ k\neq k_*\\
        i\in[p],\ a\in[\nu]
    }}
    \left|
        y_{ki,\sigma}^{(a)}
        (\eta_{m+1}^\sigma)
    \right|
    \leq
    C_m^{\mathrm{surv}}.
    \label{eq:a3-entrance-other-survivor-y}
\end{equation}
Using orthonormality of the sector basis once again,
\begin{equation}
    \left\|
        \mathbf u_{k,\sigma}^{(a)}
        (\eta_{m+1}^\sigma)
    \right\|
    \leq
    \sqrt p\,
    C_m^{\mathrm{surv}}\sigma
    \label{eq:a3-entrance-other-survivor-norm}
\end{equation}
uniformly over
$k\notin\mathcal A_m$, $k\neq k_*$, and $a\in[\nu]$.
Thus every surviving atom other than the winner is still microscopic at the
entrance time.

\paragraph{Step 8: the previously captured background.}
Suppose that $\sigma$ belongs to the $n$th diagonal block. The quantitative
fixed-level locking estimate
\eqref{eq:a3-entrance-fixed-old-quantitative} gives
\begin{equation}
    \begin{aligned}
    &\max_{r\leq m}
    \left|
        \pi_{r,\sigma}(\eta_{m+1}^\sigma)
        -
        \lambda_{i_r}
    \right|
    \\
    &\qquad
    +
    \max_{\substack{r\leq m\\a\in[\nu]}}
    \left\|
        \mathbf h_{r,\sigma}^{(a)}
        (\eta_{m+1}^\sigma)
    \right\|
    \leq
    C_n^{\mathrm{old}}\sigma^\nu.
    \end{aligned}
    \label{eq:a3-entrance-old-quantitative-block}
\end{equation}
Since $\sigma\leq\sigma_n$, the diagonal condition
\eqref{eq:a3-entrance-diagonal-old-quantitative} implies
\begin{equation}
    C_n^{\mathrm{old}}\sigma^\nu
    =
    \sigma
    C_n^{\mathrm{old}}\sigma^{\nu-1}
    \leq
    \sigma
    C_n^{\mathrm{old}}\sigma_n^{\nu-1}
    \leq
    \sigma.
    \label{eq:a3-entrance-old-quantitative-diagonal-bound}
\end{equation}
Thus the product and off-sector errors of the old captured atoms satisfy
\begin{equation}
    \max_{r\leq m}
    \left|
        \pi_{r,\sigma}(\eta_{m+1}^\sigma)
        -
        \lambda_{i_r}
    \right|
    +
    \max_{\substack{r\leq m\\a\in[\nu]}}
    \left\|
        \mathbf h_{r,\sigma}^{(a)}
        (\eta_{m+1}^\sigma)
    \right\|
    \leq
    \sigma.
    \label{eq:a3-entrance-old-quantitative-final}
\end{equation}

For their main-coordinate magnitudes, the diagonal balance estimate
\eqref{eq:a3-entrance-diagonal-old-balance} gives
\begin{equation}
    \max_{\substack{r\leq m\\a\in[\nu]}}
    \left|
        \left|
            q_{r,\sigma}^{(a)}
            (\eta_{m+1}^\sigma)
        \right|
        -
        \lambda_{i_r}^{1/\nu}
    \right|
    \leq
    \frac1n.
    \label{eq:a3-entrance-old-balance-final}
\end{equation}
Since the diagonal block index $n$ tends to infinity with
$\sigma\to0$, this proves the old-background balance conclusion
\eqref{eq:a3-entrance-old-balance}.

We can now choose one constant for the three quantitative entrance bounds.
Let $C_m^{\mathrm{off}}$ and $C_m^{\mathrm{surv}}$ be the constants from
Steps~6 and~7 and set
\begin{equation}
    C_m^{\mathrm{ent}}
    :=
    \max\left\{
        \sqrt{p}\,C_m^{\mathrm{off}},
        \sqrt{p}\,C_m^{\mathrm{surv}},
        1
    \right\}.
    \label{eq:a3-entrance-common-constant}
\end{equation}
Then the winning localization estimate
\eqref{eq:a3-entrance-winning-localization-bound}, the spectator norm bound
\eqref{eq:a3-entrance-other-survivor-norm}, and the old-background locking
estimate \eqref{eq:a3-entrance-old-quantitative-final} yield, respectively,
\eqref{eq:a3-entrance-winning-localization},
\eqref{eq:a3-entrance-other-survivors}, and
\eqref{eq:a3-entrance-old-quantitative-locking}.

\paragraph{Step 9: tensor output at the entrance time.}
It remains to prove the tensor-level conclusion
\eqref{eq:a3-entrance-output-before-capture}.

For a previously captured atom $k_r$, the decomposition
\eqref{eq:a3-locked-h} reads at the entrance time
\begin{equation}
    \mathbf u_{k_r,\sigma}^{(a)}
    =
    q_{r,\sigma}^{(a)}
    \mathbf e_{i_r}^{(a)}
    +
    \mathbf h_{r,\sigma}^{(a)}.
    \label{eq:a3-entrance-old-atom-decomposition}
\end{equation}
The term in its tensor product containing no off-sector component is
\begin{equation}
    \pi_{r,\sigma}
    \mathbf e_{i_r}^{(1)}
    \otimes\cdots\otimes
    \mathbf e_{i_r}^{(\nu)}.
    \label{eq:a3-entrance-old-atom-main-tensor}
\end{equation}
By the old-background product estimate
\eqref{eq:a3-entrance-old-quantitative-final}, this leading term converges
to
\begin{equation}
    \lambda_{i_r}
    \mathbf e_{i_r}^{(1)}
    \otimes\cdots\otimes
    \mathbf e_{i_r}^{(\nu)}.
    \label{eq:a3-entrance-old-main-tensor-limit}
\end{equation}
Every remaining term in the multilinear expansion contains at least one
off-sector factor
$\mathbf h_{r,\sigma}^{(a)}$. These factors are $O(\sigma)$ by
\eqref{eq:a3-entrance-old-quantitative-final}, while the balance estimate
\eqref{eq:a3-entrance-old-balance-final} keeps all main coordinates
$q_{r,\sigma}^{(a)}$ uniformly bounded. Consequently,
\begin{equation}
    \bigotimes_{a=1}^{\nu}
    \mathbf u_{k_r,\sigma}^{(a)}
    (\eta_{m+1}^\sigma)
    \longrightarrow
    \lambda_{i_r}
    \mathbf e_{i_r}^{(1)}
    \otimes\cdots\otimes
    \mathbf e_{i_r}^{(\nu)}.
    \label{eq:a3-entrance-old-atom-tensor-limit}
\end{equation}

For the winning atom, its mesoscopic main-coordinate estimate
\eqref{eq:a3-entrance-winning-main-coordinate} and off-sector localization
\eqref{eq:a3-entrance-winning-localization} imply
\begin{equation}
    \max_{a\in[\nu]}
    \left\|
        \mathbf u_{k_*,\sigma}^{(a)}
        (\eta_{m+1}^\sigma)
    \right\|
    =
    O(r_{m,\sigma}),
    \label{eq:a3-entrance-winning-vector-size}
\end{equation}
because $\sigma=o(r_{m,\sigma})$. Therefore the tensor represented by the
winner is still vanishing at the entrance:
\begin{equation}
    \left\|
        \bigotimes_{a=1}^{\nu}
        \mathbf u_{k_*,\sigma}^{(a)}
        (\eta_{m+1}^\sigma)
    \right\|_F
    =
    O(r_{m,\sigma}^\nu)
    =
    o(1).
    \label{eq:a3-entrance-winning-tensor-small}
\end{equation}

Finally, every other surviving atom is $O(\sigma)$ in every mode by the
spectator estimate \eqref{eq:a3-entrance-other-survivors}. Hence
\begin{equation}
    \left\|
        \bigotimes_{a=1}^{\nu}
        \mathbf u_{k,\sigma}^{(a)}
        (\eta_{m+1}^\sigma)
    \right\|_F
    =
    O(\sigma^\nu)
    \label{eq:a3-entrance-other-tensor-small}
\end{equation}
for
$k\notin\mathcal A_m\cup\{k_*\}$.

Since $H$ is fixed, summing the old captured-atom limits
\eqref{eq:a3-entrance-old-atom-tensor-limit}, the vanishing winning-atom
contribution \eqref{eq:a3-entrance-winning-tensor-small}, and the
microscopic spectator contributions
\eqref{eq:a3-entrance-other-tensor-small} gives
\begin{equation}
    \widehat T_\sigma
    (\eta_{m+1}^\sigma)
    \longrightarrow
    \sum_{r=1}^{m}
    \lambda_{i_r}
    \mathbf e_{i_r}^{(1)}
    \otimes\cdots\otimes
    \mathbf e_{i_r}^{(\nu)}
    =
    T_m.
    \label{eq:a3-entrance-output-limit-proof}
\end{equation}
This proves the tensor-level entrance statement
\eqref{eq:a3-entrance-output-before-capture} and completes the proof.
\end{proof}

The entrance state now has precisely the scale separation needed for the fast
transition. The winning atom has factor size $r_{m,\sigma}$, with
$r_{m,\sigma}\to0$ but
$r_{m,\sigma}/\sigma=R_{m,\sigma}\to\infty$, and its mass is concentrated
asymptotically in sector $i_*$. Its main coordinates are balanced and phase
aligned, whereas every other surviving atom remains only $O(\sigma)$. The
previously captured atoms remain asymptotically locked to their assigned
sectors.

Moreover, the entrance-time asymptotic
\eqref{eq:a3-entrance-time-asymptotic} shows that the remaining slow time to
the limiting clock is of order
$R_{m,\sigma}^{-(\nu-2)}\to0$. The next subsection analyzes the ensuing
fast growth and saturation of the winning atom inside this vanishing
slow-time layer and constructs the next reset time
$\rho_{m+1}^\sigma$.

\subsection{Fast capture and renewal of the induction hypotheses}
\label{app:fast-capture}

We now analyze the shrinking slow-time layer following the entrance time
$\eta_{m+1}^\sigma$ constructed in
Proposition~\ref{prop:a3-mesoscopic-entrance}. Throughout this subsection,
$m$ is fixed, and we write
\begin{equation}
    k_*:=k_{m+1},
    \qquad
    i_*:=i_{m+1},
    \qquad
    \lambda_*:=\lambda_{i_*},
    \qquad
    \theta_*:=\theta_{m+1}^*.
    \label{eq:a3-fast-winner-notation}
\end{equation}
For readability, we abbreviate the stage-dependent mesoscopic scales from
Proposition~\ref{prop:a3-mesoscopic-entrance} by
\begin{equation}
    R_\sigma:=R_{m,\sigma},
    \qquad
    r_\sigma:=r_{m,\sigma}
    =
    \sigma R_\sigma.
    \label{eq:a3-fast-r-sigma}
\end{equation}
Thus
\begin{equation}
    R_\sigma\longrightarrow\infty,
    \qquad
    \sigma\ll r_\sigma\ll1,
    \qquad
    \sigma R_\sigma^q\longrightarrow0
    \quad
    \text{for every fixed }q>0.
    \label{eq:a3-fast-mesoscopic-properties}
\end{equation}

The fast transition is most naturally studied in the original physical-time
scale. We therefore introduce a local physical time $\xi\geq0$ by
\begin{equation}
    \Theta_\sigma(\xi)
    :=
    \eta_{m+1}^\sigma
    +
    \sigma^{\nu-2}\xi.
    \label{eq:a3-fast-slow-time-map}
\end{equation}
Thus $\Theta_\sigma(\xi)$ is the slow time corresponding to a physical-time
increment $\xi$ after the entrance. Whenever quantities associated with the
previously captured atoms are written below as functions of $\xi$, they are
understood to be evaluated at the slow time $\Theta_\sigma(\xi)$.

For the winning atom, define its main coordinates and their product by
\begin{equation}
    q_\sigma^{(a)}(\xi)
    :=
    x_{k_*i_*,\sigma}^{(a)}
    \bigl(\Theta_\sigma(\xi)\bigr),
    \qquad
    \pi_\sigma(\xi)
    :=
    \prod_{a=1}^{\nu}
    q_\sigma^{(a)}(\xi),
    \label{eq:a3-fast-winning-main}
\end{equation}
and its off-sector component by
\begin{equation}
    \mathbf h_\sigma^{(a)}(\xi)
    :=
    \mathbf u_{k_*,\sigma}^{(a)}
    \bigl(\Theta_\sigma(\xi)\bigr)
    -
    q_\sigma^{(a)}(\xi)
    \mathbf e_{i_*}^{(a)}.
    \label{eq:a3-fast-winning-offsector}
\end{equation}
At $\xi=0$, the mesoscopic entrance estimate
\eqref{eq:a3-entrance-winning-main-coordinate} gives
\begin{equation}
    \frac{q_\sigma^{(a)}(0)}{r_\sigma}
    \longrightarrow
    \omega_*^{(a)},
    \qquad
    |\omega_*^{(a)}|=1,
    \qquad
    \prod_{a=1}^{\nu}\omega_*^{(a)}=1,
    \label{eq:a3-fast-entrance-main}
\end{equation}
while the winning-sector localization
\eqref{eq:a3-entrance-winning-localization} gives
\begin{equation}
    \max_{a\in[\nu]}
    \bigl\|
        \mathbf h_\sigma^{(a)}(0)
    \bigr\|
    =
    O(\sigma).
    \label{eq:a3-fast-entrance-offsector}
\end{equation}

The capture is divided into two stages. During the first stage, the winning
coordinates grow from the mesoscopic scale $r_\sigma$ to a fixed small
macroscopic amplitude. During the second stage, all mode-wise factors of the
winner are bounded away from zero, and the vector field becomes uniformly
contracting toward the captured equilibrium manifold.

\paragraph{Growth from the mesoscopic scale.}

Choose once and for all a constant $a_0>0$ sufficiently small that
\begin{equation}
    4^\nu a_0^\nu
    \leq
    \frac{\lambda_*}{4}.
    \label{eq:a3-fast-a0-choice}
\end{equation}
The numerical constant is unimportant. Its role is to ensure that, while
the winning factors have magnitude at most a fixed multiple of $a_0$, the
residual coefficient $\lambda_*-\pi_\sigma$ still drives the winning sector
strongly in the positive-growth direction.

Define the average squared magnitude and the maximal squared-magnitude
imbalance of the winning coordinates by
\begin{equation}
    Q_\sigma(\xi)
    :=
    \frac1\nu
    \sum_{a=1}^{\nu}
    |q_\sigma^{(a)}(\xi)|^2,
    \qquad
    D_\sigma(\xi)
    :=
    \max_{a,b\in[\nu]}
    \left|
        |q_\sigma^{(a)}(\xi)|^2
        -
        |q_\sigma^{(b)}(\xi)|^2
    \right|.
    \label{eq:a3-fast-Q-D}
\end{equation}

\begin{lemma}[Growth to a fixed macroscopic scale]
\label{lem:a3-fast-growth}
Under the assumptions of
Proposition~\ref{prop:a3-mesoscopic-entrance}, define
\begin{equation}
    E_{\mathrm{old},\sigma}(\xi)
    :=
    \max_{r\leq m}
    \left|
        \pi_{r,\sigma}
        \bigl(\Theta_\sigma(\xi)\bigr)
        -
        \lambda_{i_r}
    \right|
    +
    \max_{\substack{r\leq m\\a\in[\nu]}}
    \left\|
        \mathbf h_{r,\sigma}^{(a)}
        \bigl(\Theta_\sigma(\xi)\bigr)
    \right\|.
    \label{eq:a3-fast-old-background-error}
\end{equation}
For $m=0$, this quantity is understood to be identically zero.

Then, for all sufficiently small $\sigma$, there exists a finite first
hitting time
\begin{equation}
    \xi_{\mathrm{grow}}^\sigma
    :=
    \inf\left\{
        \xi>0:
        Q_\sigma(\xi)=a_0^2
    \right\},
    \label{eq:a3-fast-growth-hitting-time}
\end{equation}
satisfying
\begin{equation}
    \xi_{\mathrm{grow}}^\sigma
    \leq
    C_{\mathrm{grow}}
    r_\sigma^{-(\nu-2)}.
    \label{eq:a3-fast-growth-time}
\end{equation}
At the hitting time,
\begin{equation}
    \frac{a_0}{2}
    \leq
    |q_\sigma^{(a)}(\xi_{\mathrm{grow}}^\sigma)|
    \leq
    2a_0,
    \qquad
    a\in[\nu].
    \label{eq:a3-fast-growth-terminal}
\end{equation}

Moreover, throughout
$0\leq\xi\leq\xi_{\mathrm{grow}}^\sigma$,
\begin{equation}
    \max_{a\in[\nu]}
    \bigl\|
        \mathbf h_\sigma^{(a)}(\xi)
    \bigr\|
    \leq
    C_{\mathrm{grow}}\sigma,
    \label{eq:a3-fast-growth-h}
\end{equation}
and
\begin{equation}
    \max_{\substack{
        k\notin\mathcal A_m\cup\{k_*\}\\
        a\in[\nu]
    }}
    \left\|
        \mathbf u_{k,\sigma}^{(a)}
        \bigl(\Theta_\sigma(\xi)\bigr)
    \right\|
    \leq
    C_{\mathrm{grow}}\sigma.
    \label{eq:a3-fast-growth-spectators}
\end{equation}
The previously captured background satisfies
\begin{equation}
    \sup_{0\leq\xi\leq\xi_{\mathrm{grow}}^\sigma}
    E_{\mathrm{old},\sigma}(\xi)
    \leq
    C_{\mathrm{grow}}\sigma,
    \label{eq:a3-fast-growth-old-background}
\end{equation}
and the exact balance invariant gives the stronger estimate
\begin{equation}
    \sup_{0\leq\xi\leq\xi_{\mathrm{grow}}^\sigma}
    D_\sigma(\xi)
    =
    O(\sigma^2)
    =
    o(r_\sigma^2).
    \label{eq:a3-fast-growth-balance}
\end{equation}
Finally, the winning product remains asymptotically phase aligned:
\begin{equation}
    \sup_{0\leq\xi\leq\xi_{\mathrm{grow}}^\sigma}
    \left|
        \frac{\pi_\sigma(\xi)}
             {|\pi_\sigma(\xi)|}
        -
        1
    \right|
    \longrightarrow0.
    \label{eq:a3-fast-growth-phase}
\end{equation}
\end{lemma}

\begin{proof}
We use a stopping-time argument. The purpose of the bootstrap is to keep the
winner inside a small macroscopic neighborhood while controlling its
off-sector component, all spectator atoms, and the previously captured
background. The exact balance invariant will then supply balance of the
winner automatically.

\paragraph{Entrance estimates.}
Set
\begin{equation}
    C_{\mathrm{ent}}
    :=
    C_m^{\mathrm{ent}},
    \label{eq:a3-fast-entrance-constant}
\end{equation}
where $C_m^{\mathrm{ent}}$ is the constant from
Proposition~\ref{prop:a3-mesoscopic-entrance}. The quantitative old-background
locking estimate
\eqref{eq:a3-entrance-old-quantitative-locking} gives
\begin{equation}
    E_{\mathrm{old},\sigma}(0)
    \leq
    C_{\mathrm{ent}}\sigma.
    \label{eq:a3-fast-old-initial-quantitative}
\end{equation}
The winning localization estimate
\eqref{eq:a3-entrance-winning-localization} gives
\begin{equation}
    \max_{a\in[\nu]}
    \|\mathbf h_\sigma^{(a)}(0)\|
    \leq
    C_{\mathrm{ent}}\sigma,
    \label{eq:a3-fast-h-initial-uniform}
\end{equation}
while the spectator estimate
\eqref{eq:a3-entrance-other-survivors} gives
\begin{equation}
    \max_{\substack{
        k\notin\mathcal A_m\cup\{k_*\}\\
        a\in[\nu]
    }}
    \left\|
        \mathbf u_{k,\sigma}^{(a)}
        \bigl(\Theta_\sigma(0)\bigr)
    \right\|
    \leq
    C_{\mathrm{ent}}\sigma.
    \label{eq:a3-fast-spectator-initial-uniform}
\end{equation}

For the winning main coordinates, the entrance asymptotic
\eqref{eq:a3-fast-entrance-main} implies
\begin{equation}
    Q_\sigma(0)
    =
    r_\sigma^2(1+o(1)),
    \label{eq:a3-fast-Q-initial}
\end{equation}
and, using
$\prod_a\omega_*^{(a)}=1$,
\begin{equation}
    \frac{\pi_\sigma(0)}{r_\sigma^\nu}
    \longrightarrow1.
    \label{eq:a3-fast-pi-initial}
\end{equation}
Finally, the balance conclusion for the old captured background
\eqref{eq:a3-entrance-old-balance} gives
\begin{equation}
    \max_{\substack{r\leq m\\a\in[\nu]}}
    \left|
        \left|
            q_{r,\sigma}^{(a)}(0)
        \right|
        -
        \lambda_{i_r}^{1/\nu}
    \right|
    \longrightarrow0.
    \label{eq:a3-fast-old-balance-initial}
\end{equation}

\paragraph{Step 1: bootstrap interval.}
Choose constants
\begin{equation}
    B_{\mathrm{fast}}
    >
    C_{\mathrm{ent}}+2,
    \qquad
    0<\varepsilon_{\mathrm{old}}<1,
    \label{eq:a3-fast-bootstrap-constants-proof}
\end{equation}
where $\varepsilon_{\mathrm{old}}$ will be chosen sufficiently small below.

By the old-background balance at the entrance
\eqref{eq:a3-fast-old-balance-initial}, we may fix
\begin{equation}
    0<c_{\mathrm{old}}<C_{\mathrm{old}}<\infty
    \label{eq:a3-fast-old-annulus-constants}
\end{equation}
such that every $\lambda_{i_r}^{1/\nu}$ lies strictly inside
$(c_{\mathrm{old}},C_{\mathrm{old}})$ and, for all sufficiently small
$\sigma$,
\begin{equation}
    2c_{\mathrm{old}}
    \leq
    |q_{r,\sigma}^{(a)}(0)|
    \leq
    \frac12C_{\mathrm{old}},
    \qquad
    r\leq m,\quad a\in[\nu].
    \label{eq:a3-fast-old-annulus-initial}
\end{equation}

Choose a fixed constant $K>0$, to be specified in Step~9, and define the
comparison horizon
\begin{equation}
    T_{\mathrm{cmp},\sigma}
    :=
    K r_\sigma^{-(\nu-2)}.
    \label{eq:a3-fast-comparison-horizon}
\end{equation}

We consider the maximal interval
$[0,\xi_{\mathrm{boot}}^\sigma]$, stopped no later than
$T_{\mathrm{cmp},\sigma}$, on which either the level
$Q_\sigma=a_0^2$ has already been reached or all of the following strict
bootstrap conditions hold:
\begin{align}
    \frac12Q_\sigma(0)
    &<
    Q_\sigma(\xi)
    <
    a_0^2,
    \label{eq:a3-fast-bootstrap-Q-proof}
    \\
    \operatorname{Re}\pi_\sigma(\xi)
    &>
    \frac12|\pi_\sigma(\xi)|,
    \label{eq:a3-fast-bootstrap-phase-proof}
    \\
    \max_{a\in[\nu]}
    \|\mathbf h_\sigma^{(a)}(\xi)\|
    &<
    B_{\mathrm{fast}}\sigma,
    \label{eq:a3-fast-bootstrap-new-h-proof}
    \\
    \max_{\substack{
        k\notin\mathcal A_m\cup\{k_*\}\\
        a\in[\nu]
    }}
    \left\|
        \mathbf u_{k,\sigma}^{(a)}
        \bigl(\Theta_\sigma(\xi)\bigr)
    \right\|
    &<
    B_{\mathrm{fast}}\sigma,
    \label{eq:a3-fast-bootstrap-spectator-proof}
    \\
    E_{\mathrm{old},\sigma}(\xi)
    &<
    \varepsilon_{\mathrm{old}},
    \label{eq:a3-fast-bootstrap-old-proof}
    \\
    c_{\mathrm{old}}
    &<
    |q_{r,\sigma}^{(a)}(\xi)|
    <
    C_{\mathrm{old}},
    \qquad
    r\leq m,\quad a\in[\nu].
    \label{eq:a3-fast-bootstrap-old-annulus-proof}
\end{align}
The entrance estimates above show that all bootstrap conditions hold
initially with strict slack for sufficiently small $\sigma$.

\paragraph{Step 2: stability of the previously captured background.}
We first improve the two bootstrap conditions concerning the previously
captured atoms.

Define
\begin{equation}
    P_\sigma(\xi)
    :=
    \max_{r\leq m}
    \left|
        \pi_{r,\sigma}
        \bigl(\Theta_\sigma(\xi)\bigr)
        -
        \lambda_{i_r}
    \right|,
    \qquad
    H_\sigma(\xi)
    :=
    \max_{\substack{r\leq m\\a\in[\nu]}}
    \left\|
        \mathbf h_{r,\sigma}^{(a)}
        \bigl(\Theta_\sigma(\xi)\bigr)
    \right\|,
    \label{eq:a3-fast-old-P-H}
\end{equation}
so that
\begin{equation}
    E_{\mathrm{old},\sigma}
    =
    P_\sigma+H_\sigma.
    \label{eq:a3-fast-old-E-P-H}
\end{equation}

\noindent\textbf{Product stability.}
Fix a previously captured edge $(k_r,i_r)$ and set
\begin{equation}
    e_{r,\sigma}
    :=
    \pi_{r,\sigma}-\lambda_{i_r}.
    \label{eq:a3-fast-old-product-error}
\end{equation}
Repeating the captured-coordinate decomposition from
Lemma~\ref{lem:a3-captured-decomposition}, now in physical time, gives
\begin{equation}
    \frac{d q_{r,\sigma}^{(a)}}{d\xi}
    =
    -
    e_{r,\sigma}
    \prod_{b\neq a}
    \overline{q_{r,\sigma}^{(b)}}
    +
    \mathcal R_{r,\sigma}^{(a)}.
    \label{eq:a3-fast-old-q-equation}
\end{equation}
Uniformly on the bootstrap interval,
\begin{equation}
    \left|
        \mathcal R_{r,\sigma}^{(a)}
    \right|
    \leq
    C
    \left(
        H_\sigma^2+\sigma^\nu
    \right).
    \label{eq:a3-fast-old-q-remainder-bound}
\end{equation}

To see this, the exact self Gram factors satisfy
\begin{equation}
    G_{k_rk_r,\sigma}^{(b)}
    =
    |q_{r,\sigma}^{(b)}|^2
    +
    \|\mathbf h_{r,\sigma}^{(b)}\|^2,
    \label{eq:a3-fast-old-self-gram}
\end{equation}
so the error between the exact self-interaction and the pure one-sector
interaction is $O(H_\sigma^2)$ under the old-annulus bootstrap
\eqref{eq:a3-fast-bootstrap-old-annulus-proof}. If $s\neq r$ is another old
captured atom, orthogonality of the assigned sectors gives
\begin{equation}
    |x_{k_s i_r,\sigma}^{(a)}|
    \leq
    H_\sigma,
    \qquad
    |G_{k_rk_s,\sigma}^{(b)}|
    \leq
    C H_\sigma,
    \label{eq:a3-fast-old-old-cross-bounds}
\end{equation}
and therefore an $O(H_\sigma^\nu)$ contribution.

For the interaction with the new winner, $i_*\neq i_r$. The winner
off-sector bootstrap
\eqref{eq:a3-fast-bootstrap-new-h-proof} and the boundedness
$Q_\sigma<a_0^2$ give
\begin{equation}
    |x_{k_*i_r,\sigma}^{(a)}|
    \leq
    C\sigma,
    \qquad
    |G_{k_rk_*,\sigma}^{(b)}|
    \leq
    C(\sigma+H_\sigma),
    \label{eq:a3-fast-old-winner-cross-bounds}
\end{equation}
and hence an error bounded by
$C(H_\sigma^2+\sigma^\nu)$. Every other surviving atom is
$O(\sigma)$ by the spectator bootstrap and contributes only
$O(\sigma^\nu)$. This proves
\eqref{eq:a3-fast-old-q-remainder-bound}.

Differentiating the product
$\pi_{r,\sigma}=\prod_a q_{r,\sigma}^{(a)}$
in the effective old-coordinate equation
\eqref{eq:a3-fast-old-q-equation} gives
\begin{equation}
    \frac{d e_{r,\sigma}}{d\xi}
    =
    -
    S_{r,\sigma}e_{r,\sigma}
    +
    \mathcal R_{r,\sigma}^{\pi},
    \label{eq:a3-fast-old-product-equation}
\end{equation}
where
\begin{equation}
    S_{r,\sigma}
    :=
    \sum_{a=1}^{\nu}
    \prod_{b\neq a}
    |q_{r,\sigma}^{(b)}|^2,
    \label{eq:a3-fast-old-product-S}
\end{equation}
and the product remainder is defined by
\begin{equation}
    \mathcal R_{r,\sigma}^{\pi}
    :=
    \sum_{a=1}^{\nu}
    \left(
        \prod_{b\neq a}q_{r,\sigma}^{(b)}
    \right)
    \mathcal R_{r,\sigma}^{(a)}.
    \label{eq:a3-fast-old-product-remainder-definition}
\end{equation}
The old-annulus bootstrap
\eqref{eq:a3-fast-bootstrap-old-annulus-proof}
gives the uniform coercivity
\begin{equation}
    S_{r,\sigma}
    \geq
    \nu c_{\mathrm{old}}^{2(\nu-1)}
    =:
    c_\pi
    >
    0,
    \label{eq:a3-fast-old-product-coercivity}
\end{equation}
while the upper main-coordinate bound
$|q_{r,\sigma}^{(a)}|\leq C_{\mathrm{old}}$
in \eqref{eq:a3-fast-bootstrap-old-annulus-proof},
together with the coordinate-remainder estimate
\eqref{eq:a3-fast-old-q-remainder-bound}, implies
\begin{equation}
    |\mathcal R_{r,\sigma}^{\pi}|
    \leq
    \nu C_{\mathrm{old}}^{\nu-1}
    \max_{a\in[\nu]}|\mathcal R_{r,\sigma}^{(a)}|
    \leq
    C\left(
        H_\sigma^2+\sigma^\nu
    \right).
    \label{eq:a3-fast-old-product-remainder}
\end{equation}
Taking upper Dini derivatives and maximizing over $r$ therefore yields
\begin{equation}
    D^+P_\sigma
    \leq
    -c_\pi P_\sigma
    +
    C
    \left(
        H_\sigma^2+\sigma^\nu
    \right).
    \label{eq:a3-fast-old-P-differential}
\end{equation}

\noindent\textbf{Transverse stability.}
For an old captured edge $(k_r,i_r)$, let
\begin{equation}
    \mathcal P_r^{(a),\perp}
    :=
    I-
    \mathbf e_{i_r}^{(a)}
    \bigl(\mathbf e_{i_r}^{(a)}\bigr)^*
    \label{eq:a3-fast-old-perp-projection}
\end{equation}
be orthogonal projection onto the complement of its assigned sector. By the
definition of the off-sector residual,
\begin{equation}
    \mathbf h_{r,\sigma}^{(a)}
    =
    \mathcal P_r^{(a),\perp}
    \mathbf u_{k_r,\sigma}^{(a)}.
    \label{eq:a3-fast-old-h-as-projection}
\end{equation}
Because the local variable $\xi$ is physical time, applying
$\mathcal P_r^{(a),\perp}$ to the exact GF equation gives
\begin{equation}
    \frac{d\mathbf h_{r,\sigma}^{(a)}}{d\xi}
    =
    -
    \gamma_{r,\sigma}^{(a)}
    \mathbf h_{r,\sigma}^{(a)}
    +
    \mathbf N_{r,\sigma}^{(a)},
    \label{eq:a3-fast-old-h-equation}
\end{equation}
where
\begin{equation}
    \gamma_{r,\sigma}^{(a)}
    :=
    \prod_{b\neq a}
    \left\|
        \mathbf u_{k_r,\sigma}^{(b)}
    \right\|^2
    \geq
    c_{\mathrm{old}}^{2(\nu-1)}
    =:
    c_\gamma>0.
    \label{eq:a3-fast-old-h-coercivity}
\end{equation}
The remainder $\mathbf N_{r,\sigma}^{(a)}$ consists exactly of the
off-sector target terms and the interactions with atoms other than $k_r$.
The same estimates used above give
\begin{equation}
    \|\mathbf N_{r,\sigma}^{(a)}\|
    \leq
    C
    \left(
        H_\sigma^{\nu-1}
        +
        \sigma^{\nu-1}
    \right).
    \label{eq:a3-fast-old-h-remainder-bound}
\end{equation}
Hence
\begin{equation}
    D^+H_\sigma
    \leq
    -c_\gamma H_\sigma
    +
    C
    \left(
        H_\sigma^{\nu-1}
        +
        \sigma^{\nu-1}
    \right).
    \label{eq:a3-fast-old-H-differential}
\end{equation}

Set
\begin{equation}
    c_0
    :=
    \min\{c_\pi,c_\gamma\}>0.
    \label{eq:a3-fast-old-common-coercivity}
\end{equation}
Adding the product estimate
\eqref{eq:a3-fast-old-P-differential} and the transverse estimate
\eqref{eq:a3-fast-old-H-differential}, and using
$E_{\mathrm{old},\sigma}=P_\sigma+H_\sigma$, gives
\begin{equation}
    D^+E_{\mathrm{old},\sigma}
    \leq
    -c_0E_{\mathrm{old},\sigma}
    +
    C
    \left(
        H_\sigma^2
        +
        H_\sigma^{\nu-1}
    \right)
    +
    C\sigma^{\nu-1}.
    \label{eq:a3-fast-old-E-before-absorption}
\end{equation}
Choose $\varepsilon_{\mathrm{old}}$ in the bootstrap
\eqref{eq:a3-fast-bootstrap-old-proof} sufficiently small that
\begin{equation}
    C
    \left(
        \varepsilon_{\mathrm{old}}
        +
        \varepsilon_{\mathrm{old}}^{\nu-2}
    \right)
    \leq
    \frac12c_0.
    \label{eq:a3-fast-old-nonlinear-absorption-proof}
\end{equation}
Since
$H_\sigma\leq E_{\mathrm{old},\sigma}<\varepsilon_{\mathrm{old}}$,
we obtain
\begin{equation}
    D^+E_{\mathrm{old},\sigma}
    \leq
    -\frac12c_0E_{\mathrm{old},\sigma}
    +
    C\sigma^{\nu-1}.
    \label{eq:a3-fast-old-linear-proof}
\end{equation}
Variation of constants, together with the entrance estimate
\eqref{eq:a3-fast-old-initial-quantitative}, yields
\begin{equation}
    E_{\mathrm{old},\sigma}(\xi)
    \leq
    e^{-c_0\xi/2}
    E_{\mathrm{old},\sigma}(0)
    +
    C\sigma^{\nu-1}
    \leq
    C_{\mathrm{old}}'\sigma.
    \label{eq:a3-fast-old-O-sigma}
\end{equation}
Thus the old-background error bootstrap is strictly improved.

It remains to preserve the old captured annulus. Applying the exact
main-coordinate balance consequence
\eqref{eq:a3-main-coordinate-balance-from-invariant} to each old captured
atom, and using the just-established bound
$H_\sigma=O(\sigma)$, gives
\begin{equation}
    \max_{\substack{r\leq m\\a,b\in[\nu]}}
    \sup_{0\leq\xi\leq\xi_{\mathrm{boot}}^\sigma}
    \left|
        |q_{r,\sigma}^{(a)}(\xi)|^2
        -
        |q_{r,\sigma}^{(b)}(\xi)|^2
    \right|
    =
    O(\sigma^2).
    \label{eq:a3-fast-old-main-balance-uniform}
\end{equation}
At the same time, the product part of
\eqref{eq:a3-fast-old-O-sigma} gives
\begin{equation}
    \prod_{a=1}^{\nu}
    |q_{r,\sigma}^{(a)}(\xi)|
    =
    \lambda_{i_r}
    +
    O(\sigma)
    \label{eq:a3-fast-old-product-magnitude-uniform}
\end{equation}
uniformly on the bootstrap interval. Combining these two estimates yields
\begin{equation}
    \max_{\substack{r\leq m\\a\in[\nu]}}
    \sup_{0\leq\xi\leq\xi_{\mathrm{boot}}^\sigma}
    \left|
        |q_{r,\sigma}^{(a)}(\xi)|
        -
        \lambda_{i_r}^{1/\nu}
    \right|
    =
    O(\sigma).
    \label{eq:a3-fast-old-balanced-through-growth}
\end{equation}
Since each $\lambda_{i_r}^{1/\nu}$ lies strictly inside
$(c_{\mathrm{old}},C_{\mathrm{old}})$, this strictly improves the
old-annulus bootstrap
\eqref{eq:a3-fast-bootstrap-old-annulus-proof}.

\paragraph{Step 3: balance of the winning factors from the exact invariant.}
The winning off-sector bootstrap
\eqref{eq:a3-fast-bootstrap-new-h-proof} and the exact balance relation
\eqref{eq:a3-main-coordinate-balance-from-invariant}, now applied to atom
$k_*$ and sector $i_*$, give
\begin{equation}
    D_\sigma(\xi)
    \leq
    \left(
        C_{\mathrm{bal}}
        +
        2B_{\mathrm{fast}}^2
    \right)\sigma^2
    \label{eq:a3-fast-D-exact-bound}
\end{equation}
throughout the bootstrap interval.

On the other hand, the lower amplitude bootstrap
\eqref{eq:a3-fast-bootstrap-Q-proof} and the entrance asymptotic
\eqref{eq:a3-fast-Q-initial} give
\begin{equation}
    Q_\sigma(\xi)
    \geq
    \frac12Q_\sigma(0)
    =
    \frac12r_\sigma^2(1+o(1)).
    \label{eq:a3-fast-Q-mesoscopic-lower}
\end{equation}
Therefore
\begin{equation}
    \frac{D_\sigma(\xi)}{Q_\sigma(\xi)}
    \leq
    \frac{C}{R_\sigma^2}
    \longrightarrow0
    \label{eq:a3-fast-relative-balance}
\end{equation}
uniformly on the bootstrap interval. In particular, for sufficiently small
$\sigma$,
\begin{equation}
    \frac34Q_\sigma
    \leq
    |q_\sigma^{(a)}|^2
    \leq
    \frac54Q_\sigma,
    \qquad
    a\in[\nu].
    \label{eq:a3-fast-comparable-magnitudes-proof}
\end{equation}
This also already proves the balance conclusion
\eqref{eq:a3-fast-growth-balance}, conditional only on closing the
off-sector bootstrap later in the argument.

\paragraph{Step 4: the winning main-coordinate equation.}
Separating the self-interaction of the winning atom in the exact
physical-time coordinate equation gives
\begin{equation}
    \frac{d q_\sigma^{(a)}}{d\xi}
    =
    \bigl(
        \lambda_*-\pi_\sigma
    \bigr)
    \prod_{b\neq a}
    \overline{q_\sigma^{(b)}}
    +
    \mathcal R_\sigma^{(a)}(\xi).
    \label{eq:a3-fast-main-equation-proof}
\end{equation}
Let
\begin{equation}
    M_\sigma(\xi)
    :=
    \max_{a\in[\nu]}
    |q_\sigma^{(a)}(\xi)|.
    \label{eq:a3-fast-M-proof}
\end{equation}
Using the winner off-sector bootstrap, the spectator bootstrap, and the
old-background estimate \eqref{eq:a3-fast-old-O-sigma}, the remainder in
\eqref{eq:a3-fast-main-equation-proof} satisfies
\begin{equation}
    \left|
        \mathcal R_\sigma^{(a)}(\xi)
    \right|
    \leq
    C
    \left(
        \sigma^\nu
        +
        \sigma^2
        M_\sigma(\xi)^{2\nu-3}
    \right).
    \label{eq:a3-fast-main-remainder-proof}
\end{equation}

Indeed, the correction to the winning self-interaction caused by the
off-sector residual is
\begin{equation}
    O\!\left(
        \sigma^2M_\sigma^{2\nu-3}
    \right),
    \label{eq:a3-fast-self-remainder-proof}
\end{equation}
because
$\|\mathbf h_\sigma^{(b)}\|=O(\sigma)$. An interaction with another
surviving atom contains one $O(\sigma)$ coordinate and $\nu-1$ Gram factors
of size $O(\sigma M_\sigma+\sigma^2)$ and is therefore $O(\sigma^\nu)$
while $M_\sigma$ remains bounded. Finally, by the old-background bound
\eqref{eq:a3-fast-old-O-sigma}, every interaction with a previously captured
atom contains an $O(\sigma)$ wrong-sector coordinate and $\nu-1$ cross-Gram
factors of size $O(\sigma)$, giving another $O(\sigma^\nu)$ contribution.

\paragraph{Step 5: positive growth of the winning amplitude.}
The balance estimate
\eqref{eq:a3-fast-comparable-magnitudes-proof} implies
\begin{equation}
    c_\nu Q_\sigma^{\nu/2}
    \leq
    |\pi_\sigma|
    \leq
    C_\nu Q_\sigma^{\nu/2}.
    \label{eq:a3-fast-product-Q-proof}
\end{equation}
Averaging the squared-magnitude equations obtained from
\eqref{eq:a3-fast-main-equation-proof} gives
\begin{equation}
    \frac{dQ_\sigma}{d\xi}
    =
    2\operatorname{Re}
    \left[
        \bigl(
            \lambda_*-\pi_\sigma
        \bigr)
        \overline{\pi_\sigma}
    \right]
    +
    \mathcal R_\sigma^Q,
    \label{eq:a3-fast-Q-equation-proof}
\end{equation}
where the main-coordinate remainder estimate
\eqref{eq:a3-fast-main-remainder-proof} gives
\begin{equation}
    |\mathcal R_\sigma^Q|
    \leq
    C M_\sigma
    \left(
        \sigma^\nu
        +
        \sigma^2M_\sigma^{2\nu-3}
    \right).
    \label{eq:a3-fast-Q-remainder-proof}
\end{equation}

Under the phase bootstrap
\eqref{eq:a3-fast-bootstrap-phase-proof},
\begin{equation}
    \begin{aligned}
    2\operatorname{Re}
    \left[
        (\lambda_*-\pi_\sigma)
        \overline{\pi_\sigma}
    \right]
    &=
    2
    \left(
        \lambda_*\operatorname{Re}\pi_\sigma
        -
        |\pi_\sigma|^2
    \right)
    \\
    &\geq
    \lambda_*|\pi_\sigma|
    -
    2|\pi_\sigma|^2.
    \end{aligned}
    \label{eq:a3-fast-Q-leading-lower-proof}
\end{equation}
The amplitude bootstrap $Q_\sigma<a_0^2$, the factor comparability
\eqref{eq:a3-fast-comparable-magnitudes-proof}, and the fixed choice of
$a_0$ in \eqref{eq:a3-fast-a0-choice} imply
\begin{equation}
    |\pi_\sigma|
    \leq
    \frac{\lambda_*}{4}.
    \label{eq:a3-fast-product-small-proof}
\end{equation}
Hence
\begin{equation}
    2\operatorname{Re}
    \left[
        (\lambda_*-\pi_\sigma)
        \overline{\pi_\sigma}
    \right]
    \geq
    \frac{\lambda_*}{2}
    |\pi_\sigma|.
    \label{eq:a3-fast-Q-leading-positive-proof}
\end{equation}

The lower amplitude bound
\eqref{eq:a3-fast-Q-mesoscopic-lower} and factor comparability give
\begin{equation}
    M_\sigma
    \geq
    c r_\sigma.
    \label{eq:a3-fast-M-lower-proof}
\end{equation}
Therefore
\begin{equation}
    \frac{
        |\mathcal R_\sigma^Q|
    }{
        |\pi_\sigma|
    }
    \leq
    C
    \left(
        \frac{\sigma^\nu}{M_\sigma^{\nu-1}}
        +
        \sigma^2M_\sigma^{\nu-2}
    \right)
    \leq
    C
    \left(
        \frac{\sigma}{R_\sigma^{\nu-1}}
        +
        \sigma^2
    \right)
    \longrightarrow0.
    \label{eq:a3-fast-Q-remainder-relative-proof}
\end{equation}
For sufficiently small $\sigma$, the positive leading term therefore
dominates the perturbation, and
\begin{equation}
    \frac{dQ_\sigma}{d\xi}
    \geq
    c_{\mathrm{grow}}
    Q_\sigma^{\nu/2}.
    \label{eq:a3-fast-Q-growth-proof}
\end{equation}
In particular, $Q_\sigma$ is strictly increasing, so
\begin{equation}
    Q_\sigma(\xi)
    \geq
    Q_\sigma(0),
    \label{eq:a3-fast-Q-lower-improved}
\end{equation}
which strictly improves the lower bound in the amplitude bootstrap
\eqref{eq:a3-fast-bootstrap-Q-proof}.

\paragraph{Step 6: phase alignment is preserved.}
Differentiating the winning product
$\pi_\sigma=\prod_aq_\sigma^{(a)}$ in the effective main-coordinate equation
\eqref{eq:a3-fast-main-equation-proof} gives
\begin{equation}
    \frac{d\pi_\sigma}{d\xi}
    =
    S_\sigma(\xi)
    \bigl(
        \lambda_*-\pi_\sigma
    \bigr)
    +
    \mathcal E_\sigma^\pi(\xi),
    \label{eq:a3-fast-pi-equation-proof}
\end{equation}
where
\begin{equation}
    S_\sigma
    :=
    \sum_{a=1}^{\nu}
    \prod_{b\neq a}|q_\sigma^{(b)}|^2
    \label{eq:a3-fast-S-proof}
\end{equation}
satisfies, by factor comparability,
\begin{equation}
    c_\nu Q_\sigma^{\nu-1}
    \leq
    S_\sigma
    \leq
    C_\nu Q_\sigma^{\nu-1}.
    \label{eq:a3-fast-S-Q-proof}
\end{equation}
The product remainder obeys
\begin{equation}
    \left|
        \mathcal E_\sigma^\pi
    \right|
    \leq
    C
    M_\sigma^{\nu-1}
    \left(
        \sigma^\nu
        +
        \sigma^2M_\sigma^{2\nu-3}
    \right).
    \label{eq:a3-fast-pi-remainder-proof}
\end{equation}
Using the lower bound $M_\sigma\geq cr_\sigma$ from
\eqref{eq:a3-fast-M-lower-proof}, we obtain
\begin{equation}
    \frac{
        |\mathcal E_\sigma^\pi|
    }{
        \lambda_*S_\sigma
    }
    \leq
    C
    \left(
        \frac{\sigma^\nu}{M_\sigma^{\nu-1}}
        +
        \sigma^2M_\sigma^{\nu-2}
    \right)
    =:
    \varepsilon_\sigma,
    \qquad
    \varepsilon_\sigma\longrightarrow0.
    \label{eq:a3-fast-pi-relative-error-proof}
\end{equation}

Define
\begin{equation}
    A_\sigma(\xi)
    :=
    \int_0^\xi
    S_\sigma(s)\,ds.
    \label{eq:a3-fast-A-proof}
\end{equation}
Variation of constants in the product equation
\eqref{eq:a3-fast-pi-equation-proof} gives
\begin{equation}
    \begin{aligned}
    \pi_\sigma(\xi)
    &=
    e^{-A_\sigma(\xi)}
    \pi_\sigma(0)
    +
    \lambda_*
    \bigl(
        1-e^{-A_\sigma(\xi)}
    \bigr)
    \\
    &\quad
    +
    \int_0^\xi
    e^{-[A_\sigma(\xi)-A_\sigma(s)]}
    \mathcal E_\sigma^\pi(s)\,ds .
    \end{aligned}
    \label{eq:a3-fast-pi-voc-proof}
\end{equation}
The relative remainder bound
\eqref{eq:a3-fast-pi-relative-error-proof} implies
\begin{equation}
    \left|
        \int_0^\xi
        e^{-[A_\sigma(\xi)-A_\sigma(s)]}
        \mathcal E_\sigma^\pi(s)\,ds
    \right|
    \leq
    \varepsilon_\sigma\lambda_*
    \bigl(
        1-e^{-A_\sigma(\xi)}
    \bigr).
    \label{eq:a3-fast-pi-voc-error-proof}
\end{equation}

By the entrance product asymptotic
\eqref{eq:a3-fast-pi-initial}, there exists
$\delta_\sigma\to0$ such that
\begin{equation}
    \left|
        \pi_\sigma(0)-r_\sigma^\nu
    \right|
    \leq
    \delta_\sigma r_\sigma^\nu.
    \label{eq:a3-fast-pi-initial-error-proof}
\end{equation}
Define the positive real quantity
\begin{equation}
    B_\sigma(\xi)
    :=
    e^{-A_\sigma(\xi)}r_\sigma^\nu
    +
    \lambda_*
    \bigl(
        1-e^{-A_\sigma(\xi)}
    \bigr).
    \label{eq:a3-fast-B-positive-proof}
\end{equation}
Then
\eqref{eq:a3-fast-pi-voc-proof}--%
\eqref{eq:a3-fast-pi-initial-error-proof} give
\begin{equation}
    \left|
        \pi_\sigma(\xi)-B_\sigma(\xi)
    \right|
    \leq
    \bigl(
        \delta_\sigma+\varepsilon_\sigma
    \bigr)
    B_\sigma(\xi).
    \label{eq:a3-fast-pi-close-positive-proof}
\end{equation}
Since $B_\sigma(\xi)>0$, we conclude that
\begin{equation}
    \sup_{0\leq\xi\leq\xi_{\mathrm{boot}}^\sigma}
    \left|
        \frac{\pi_\sigma(\xi)}
             {|\pi_\sigma(\xi)|}
        -
        1
    \right|
    \longrightarrow0.
    \label{eq:a3-fast-phase-improved-proof}
\end{equation}
For sufficiently small $\sigma$, this strictly improves the phase bootstrap
\eqref{eq:a3-fast-bootstrap-phase-proof}.

\paragraph{Step 7: the winning off-sector component remains $O(\sigma)$.}
Let
\begin{equation}
    \mathcal P_*^{(a),\perp}
    :=
    I-
    \mathbf e_{i_*}^{(a)}
    \bigl(\mathbf e_{i_*}^{(a)}\bigr)^*.
    \label{eq:a3-fast-new-perp-projector}
\end{equation}
By the definition of the winning off-sector residual
\eqref{eq:a3-fast-winning-offsector},
\begin{equation}
    \mathbf h_\sigma^{(a)}
    =
    \mathcal P_*^{(a),\perp}
    \mathbf u_{k_*,\sigma}^{(a)}.
    \label{eq:a3-fast-new-h-projection}
\end{equation}
Applying this projector to the exact physical-time GF equation gives
\begin{equation}
    \frac{d\mathbf h_\sigma^{(a)}}{d\xi}
    =
    -
    \gamma_{*,\sigma}^{(a)}
    \mathbf h_\sigma^{(a)}
    +
    \mathbf N_{*,\sigma}^{(a)},
    \label{eq:a3-fast-new-h-equation}
\end{equation}
where
\begin{equation}
    \gamma_{*,\sigma}^{(a)}
    :=
    \prod_{b\neq a}
    \left\|
        \mathbf u_{k_*,\sigma}^{(b)}
    \right\|^2
    \geq0.
    \label{eq:a3-fast-new-h-damping}
\end{equation}
The remainder contains the off-sector target terms and all interactions with
atoms other than $k_*$. Under the winner off-sector bootstrap
\eqref{eq:a3-fast-bootstrap-new-h-proof}, the spectator bootstrap
\eqref{eq:a3-fast-bootstrap-spectator-proof}, and the old-background bound
\eqref{eq:a3-fast-old-O-sigma}, every such term is
$O(\sigma^{\nu-1})$. Hence
\begin{equation}
    \max_{a\in[\nu]}
    \|\mathbf N_{*,\sigma}^{(a)}\|
    \leq
    C\sigma^{\nu-1}.
    \label{eq:a3-fast-new-h-forcing-bound}
\end{equation}
Dropping the nonpositive self-damping term from the norm inequality gives
\begin{equation}
    D^+
    \left(
        \max_{a\in[\nu]}
        \|\mathbf h_\sigma^{(a)}\|
    \right)
    \leq
    C\sigma^{\nu-1}.
    \label{eq:a3-fast-new-h-differential-proof}
\end{equation}
Using the entrance bound
\eqref{eq:a3-fast-h-initial-uniform} and the comparison horizon
\eqref{eq:a3-fast-comparison-horizon},
\begin{equation}
    \max_{a\in[\nu]}
    \|\mathbf h_\sigma^{(a)}(\xi)\|
    \leq
    C_{\mathrm{ent}}\sigma
    +
    C\sigma^{\nu-1}T_{\mathrm{cmp},\sigma}.
    \label{eq:a3-fast-new-h-integrated-proof}
\end{equation}
The mesoscopic scale gives
\begin{equation}
    \sigma^{\nu-1}T_{\mathrm{cmp},\sigma}
    =
    K
    \frac{\sigma}{R_\sigma^{\nu-2}}
    =
    o(\sigma).
    \label{eq:a3-fast-new-h-small-change-proof}
\end{equation}
Therefore
\begin{equation}
    \max_{a\in[\nu]}
    \sup_{0\leq\xi\leq\xi_{\mathrm{boot}}^\sigma}
    \|\mathbf h_\sigma^{(a)}(\xi)\|
    \leq
    \bigl(
        C_{\mathrm{ent}}+o(1)
    \bigr)\sigma,
    \label{eq:a3-fast-new-h-improved-proof}
\end{equation}
which strictly improves the winner off-sector bootstrap
\eqref{eq:a3-fast-bootstrap-new-h-proof}. In particular, the balance
estimate derived conditionally in Step~3 is now valid throughout the full
bootstrap interval.

\paragraph{Step 8: the other surviving atoms remain microscopic.}
Let $k\notin\mathcal A_m\cup\{k_*\}$. The target part of its physical-time
GF is $O(\sigma^{\nu-1})$ under the spectator bootstrap. Interactions with
other microscopic atoms are smaller, while interactions with an old
captured atom contain $\nu-1$ Gram factors of size $O(\sigma)$ and are also
$O(\sigma^{\nu-1})$.

The only potentially larger interaction is with the growing winner:
\begin{equation}
    \mathbf u_{k_*,\sigma}^{(a)}
    \prod_{b\neq a}
    G_{k k_*,\sigma}^{(b)}.
    \label{eq:a3-fast-spectator-winner-term-proof}
\end{equation}
Since the spectator atom is $O(\sigma)$ and the winner has main magnitude
$M_\sigma$ and off-sector part $O(\sigma)$,
\begin{equation}
    \left|
        G_{k k_*,\sigma}^{(b)}
    \right|
    \leq
    C\sigma M_\sigma.
    \label{eq:a3-fast-spectator-winner-gram-proof}
\end{equation}
Therefore
\begin{equation}
    \left\|
        \mathbf u_{k_*,\sigma}^{(a)}
        \prod_{b\neq a}
        G_{k k_*,\sigma}^{(b)}
    \right\|
    \leq
    C\sigma^{\nu-1}M_\sigma^\nu
    \leq
    C\sigma^{\nu-1},
    \label{eq:a3-fast-spectator-winner-bound-proof}
\end{equation}
because $M_\sigma$ remains bounded while $Q_\sigma<a_0^2$.

Hence
\begin{equation}
    \max_{\substack{
        k\notin\mathcal A_m\cup\{k_*\}\\
        a\in[\nu]
    }}
    \left\|
        \frac{d}{d\xi}
        \mathbf u_{k,\sigma}^{(a)}
        \bigl(\Theta_\sigma(\xi)\bigr)
    \right\|
    \leq
    C\sigma^{\nu-1}.
    \label{eq:a3-fast-spectator-derivative-proof}
\end{equation}
Integrating over the comparison horizon and using the entrance estimate
\eqref{eq:a3-fast-spectator-initial-uniform} gives
\begin{equation}
    \max_{\substack{
        k\notin\mathcal A_m\cup\{k_*\}\\
        a\in[\nu]
    }}
    \left\|
        \mathbf u_{k,\sigma}^{(a)}
        \bigl(\Theta_\sigma(\xi)\bigr)
    \right\|
    \leq
    \bigl(
        C_{\mathrm{ent}}+o(1)
    \bigr)\sigma.
    \label{eq:a3-fast-spectator-improved-proof}
\end{equation}
This strictly improves the spectator bootstrap
\eqref{eq:a3-fast-bootstrap-spectator-proof}.

\paragraph{Step 9: existence of the hitting time.}
Before the level $Q_\sigma=a_0^2$ is reached, Steps~2--8 strictly improve
all auxiliary bootstrap conditions for sufficiently small $\sigma$.
Therefore the bootstrap interval cannot terminate except by reaching the
level $Q_\sigma=a_0^2$ or the comparison horizon
$T_{\mathrm{cmp},\sigma}$.

The growth inequality \eqref{eq:a3-fast-Q-growth-proof} gives
\begin{equation}
    \frac{d}{d\xi}
    Q_\sigma(\xi)^{-(\nu-2)/2}
    \leq
    -
    \frac{\nu-2}{2}
    c_{\mathrm{grow}}.
    \label{eq:a3-fast-reciprocal-growth-proof}
\end{equation}
Consequently, as long as $Q_\sigma<a_0^2$,
\begin{equation}
    Q_\sigma(\xi)^{-(\nu-2)/2}
    \leq
    Q_\sigma(0)^{-(\nu-2)/2}
    -
    \frac{\nu-2}{2}
    c_{\mathrm{grow}}\xi.
    \label{eq:a3-fast-reciprocal-integrated-proof}
\end{equation}
The target level must therefore be reached no later than
\begin{equation}
    \frac{
        2
    }{
        (\nu-2)c_{\mathrm{grow}}
    }
    \left[
        Q_\sigma(0)^{-(\nu-2)/2}
        -
        a_0^{-(\nu-2)}
    \right].
    \label{eq:a3-fast-hitting-upper-proof}
\end{equation}
Since the entrance amplitude satisfies
$Q_\sigma(0)=r_\sigma^2(1+o(1))$, there is a constant
$C_{\mathrm{grow}}<\infty$ such that this upper bound is at most
\begin{equation}
    C_{\mathrm{grow}}
    r_\sigma^{-(\nu-2)}
    \label{eq:a3-fast-hitting-order-proof}
\end{equation}
for sufficiently small $\sigma$. Choosing $K>C_{\mathrm{grow}}$ in the
comparison horizon
\eqref{eq:a3-fast-comparison-horizon} shows that the level
$Q_\sigma=a_0^2$ is reached before the horizon. This proves the existence
and time estimate
\eqref{eq:a3-fast-growth-hitting-time}--%
\eqref{eq:a3-fast-growth-time}.

\paragraph{Step 10: state at the hitting time.}
At
$\xi=\xi_{\mathrm{grow}}^\sigma$,
\begin{equation}
    Q_\sigma(\xi_{\mathrm{grow}}^\sigma)
    =
    a_0^2.
    \label{eq:a3-fast-Q-at-hitting-proof}
\end{equation}
The exact-invariant balance estimate
\eqref{eq:a3-fast-D-exact-bound} gives
\begin{equation}
    D_\sigma(\xi_{\mathrm{grow}}^\sigma)
    =
    O(\sigma^2)
    =
    o(1).
    \label{eq:a3-fast-D-at-hitting-proof}
\end{equation}
Because $Q_\sigma$ is the average of the squared factor magnitudes,
\begin{equation}
    \max_{a\in[\nu]}
    \left|
        |q_\sigma^{(a)}
        (\xi_{\mathrm{grow}}^\sigma)|^2
        -
        a_0^2
    \right|
    \leq
    D_\sigma(\xi_{\mathrm{grow}}^\sigma)
    =
    o(1).
    \label{eq:a3-fast-terminal-square-balance-proof}
\end{equation}
Hence, for sufficiently small $\sigma$,
\eqref{eq:a3-fast-growth-terminal} holds.

The improved winner off-sector estimate
\eqref{eq:a3-fast-new-h-improved-proof}, the spectator estimate
\eqref{eq:a3-fast-spectator-improved-proof}, and the old-background estimate
\eqref{eq:a3-fast-old-O-sigma} give
\eqref{eq:a3-fast-growth-h},
\eqref{eq:a3-fast-growth-spectators}, and
\eqref{eq:a3-fast-growth-old-background}, respectively. The exact-invariant
bound \eqref{eq:a3-fast-D-exact-bound} gives
\eqref{eq:a3-fast-growth-balance}, and the phase estimate
\eqref{eq:a3-fast-phase-improved-proof} gives
\eqref{eq:a3-fast-growth-phase}. This completes the proof.
\end{proof}

\paragraph{Uniform locking after the winning factors become macroscopic.}

Once the winning factors satisfy
\eqref{eq:a3-fast-growth-terminal}, their magnitudes are bounded away from
zero. The new captured sector therefore enters the same uniformly coercive
regime as the previously captured sectors. The next lemma records the
contraction needed to reach the quantitative $O(\sigma^\nu)$ locking scale
required by the induction hypothesis.

\begin{lemma}[Uniform locking of the enlarged captured set]
\label{lem:a3-fast-locking}
Suppose that at a physical time $\xi_0^\sigma$ the following conditions
hold.

\begin{enumerate}
\item
As $\sigma\to 0$, the first $m$ captured atoms satisfy
\begin{equation}
    \max_{r\leq m}
    |\pi_{r,\sigma}-\lambda_{i_r}|
    +
    \max_{\substack{r\leq m\\a\in[\nu]}}
    \|\mathbf h_{r,\sigma}^{(a)}\|
    =
    o(1),
    \label{eq:a3-fast-locking-old-initial}
\end{equation}
and
\begin{equation}
    \max_{\substack{r\leq m\\a\in[\nu]}}
    \left|
        |q_{r,\sigma}^{(a)}(\xi_0^\sigma)|
        -
        \lambda_{i_r}^{1/\nu}
    \right|
    =
    o(1).
    \label{eq:a3-fast-locking-old-main-initial}
\end{equation}

\item
The new winning atom satisfies
\begin{equation}
    c_0
    \leq
    |q_\sigma^{(a)}(\xi_0^\sigma)|
    \leq
    C_0,
    \qquad
    a\in[\nu],
    \label{eq:a3-fast-locking-main-bounds}
\end{equation}
for fixed $0<c_0<C_0<\infty$, and its product is asymptotically aligned
with the positive real axis:
\begin{equation}
    \frac{
        \pi_\sigma(\xi_0^\sigma)
    }{
        |\pi_\sigma(\xi_0^\sigma)|
    }
    \longrightarrow1.
    \label{eq:a3-fast-locking-phase-initial}
\end{equation}

\item
The new winner has off-sector residual $O(\sigma)$:
\begin{equation}
    \max_{a\in[\nu]}
    \|\mathbf h_\sigma^{(a)}(\xi_0^\sigma)\|
    =
    O(\sigma).
    \label{eq:a3-fast-locking-new-h-initial}
\end{equation}

\item
Every atom whose index is outside
$\mathcal A_m\cup\{k_*\}$ is $O(\sigma)$.
\end{enumerate}

Then there exists a constant $C_{\mathrm{lock}}>0$, independent of
$\sigma$, such that after the additional physical time
\begin{equation}
    \xi_{\mathrm{lock}}^\sigma
    :=
    C_{\mathrm{lock}}
    \log\frac1\sigma
    \label{eq:a3-fast-locking-time}
\end{equation}
all $m+1$ captured atoms satisfy
\begin{equation}
    \max_{1\leq r\leq m+1}
    \left|
        \pi_{r,\sigma}-\lambda_{i_r}
    \right|
    \leq
    C\sigma^\nu,
    \label{eq:a3-fast-locking-product}
\end{equation}
and
\begin{equation}
    \max_{\substack{1\leq r\leq m+1\\a\in[\nu]}}
    \left\|
        \mathbf h_{r,\sigma}^{(a)}
    \right\|
    \leq
    C\sigma^\nu.
    \label{eq:a3-fast-locking-offsector}
\end{equation}
Every atom whose index is outside $\mathcal A_{m+1}$ remains
$O(\sigma)$ throughout this interval.
\end{lemma}

\begin{proof}
The proof is another continuation argument, now in a fixed macroscopic
neighborhood of the enlarged captured manifold.

\paragraph{Step 1: preservation of a coarse captured annulus.}
We first specify the fixed neighborhood in which the
continuation argument will take place. For the new winner, set
$k_{m+1}=k_*$, $i_{m+1}=i_*$, and use the notation
\[
    q_{m+1,\sigma}^{(a)}=q_\sigma^{(a)},
    \qquad
    \pi_{m+1,\sigma}=\pi_\sigma,
    \qquad
    \mathbf h_{m+1,\sigma}^{(a)}
    =\mathbf h_\sigma^{(a)},
\]
consistent with
\eqref{eq:a3-locked-q-pi}--\eqref{eq:a3-locked-h}
and
\eqref{eq:a3-fast-winning-main}--%
\eqref{eq:a3-fast-winning-offsector}.
For every captured atom $r\leq m+1$, write
\[
    e_{r,\sigma}
    :=
    \pi_{r,\sigma}-\lambda_{i_r},
\]
as in \eqref{eq:a3-fast-old-product-error}, and define
the maximal off-sector error by
\begin{equation}
    E_{\perp,\sigma}(\xi)
    :=
    \max_{\substack{1\leq r\leq m+1\\a\in[\nu]}}
    \left\|
        \mathbf h_{r,\sigma}^{(a)}(\xi)
    \right\|.
    \label{eq:a3-fast-locking-Eperp}
\end{equation}
By the initial off-sector bounds
\eqref{eq:a3-fast-locking-old-initial} and
\eqref{eq:a3-fast-locking-new-h-initial},
\[
    E_{\perp,\sigma}(\xi_0^\sigma)=o(1).
\]

To choose the coarse annulus, set
\[
    q_{\min}
    :=
    \min\left\{
        c_0,\,
        \min_{1\leq r\leq m+1}
        \lambda_{i_r}^{1/\nu}
    \right\},
    \qquad
    q_{\max}
    :=
    \max\left\{
        C_0,\,
        \max_{1\leq r\leq m+1}
        \lambda_{i_r}^{1/\nu}
    \right\},
\]
and define the fixed constants
\[
    c_*:=\frac{q_{\min}}4,
    \qquad
    C_*:=4q_{\max}.
\]
The initial convergence of the old main-coordinate
magnitudes in
\eqref{eq:a3-fast-locking-old-main-initial},
together with the bounds on the new winning factors
in \eqref{eq:a3-fast-locking-main-bounds},
implies that, for sufficiently small $\sigma$,
\[
    c_*<
    |q_{r,\sigma}^{(a)}(\xi_0^\sigma)|
    <C_*,
    \qquad
    1\leq r\leq m+1,\quad a\in[\nu].
\]

We bootstrap that, throughout the locking interval,
\begin{equation}
    c_*
    \leq
    |q_{r,\sigma}^{(a)}(\xi)|
    \leq
    C_*,
    \qquad
    1\leq r\leq m+1,\quad a\in[\nu],
    \label{eq:a3-fast-locking-coarse-annulus}
\end{equation}
together with
$E_{\perp,\sigma}(\xi)\leq\delta$
for a sufficiently small fixed $\delta>0$, and an
$O(\sigma)$ bound on every still-unclaimed atom.
The latter two bootstrap bounds will be verified in
Steps 2 and 4, respectively.

We next derive the product equation for the enlarged
captured set. The same calculation as in
\eqref{eq:a3-fast-old-q-equation} and
\eqref{eq:a3-fast-old-product-equation}, now applied
to all $m+1$ captured atoms, gives
\begin{equation}
    \frac{d e_{r,\sigma}}{d\xi}
    =
    -
    S_{r,\sigma}(\xi)e_{r,\sigma}
    +
    \mathcal R_{r,\sigma}^{\pi}(\xi),
    \qquad 1\leq r\leq m+1,
    \label{eq:a3-fast-locking-product-equation}
\end{equation}
where the product coefficient $S_{r,\sigma}$ is defined
in \eqref{eq:a3-fast-old-product-S}:
\[
    S_{r,\sigma}(\xi)
    =
    \sum_{a=1}^{\nu}
    \prod_{b\neq a}
    |q_{r,\sigma}^{(b)}(\xi)|^2.
\]
In particular, the coarse lower bound
$|q_{r,\sigma}^{(a)}|\geq c_*$
in \eqref{eq:a3-fast-locking-coarse-annulus}
implies
\begin{equation}
    S_{r,\sigma}(\xi)
    \geq
    \nu c_*^{2(\nu-1)}
    =:
    c_S>0,
    \label{eq:a3-fast-locking-product-coercivity}
\end{equation}
uniformly for all captured atoms.

We now estimate the product remainder
$\mathcal R_{r,\sigma}^{\pi}$.
Recall from
\eqref{eq:a3-fast-old-product-remainder-definition}
that it is obtained from the coordinate remainders
$\mathcal R_{r,\sigma}^{(a)}$ by
\[
    \mathcal R_{r,\sigma}^{\pi}
    =
    \sum_{a=1}^{\nu}
    \left(
        \prod_{b\neq a}q_{r,\sigma}^{(b)}
    \right)
    \mathcal R_{r,\sigma}^{(a)}.
\]
For each captured atom, the exact coordinate equation
\eqref{eq:a3-locked-coordinate-gf}, expressed in
physical time, gives
\[
    \frac{d q_{r,\sigma}^{(a)}}{d\xi}
    =
    -
    e_{r,\sigma}
    \prod_{b\neq a}
    \overline{q_{r,\sigma}^{(b)}}
    +
    \mathcal R_{r,\sigma}^{(a)}.
\]
Using the orthogonal decomposition
\eqref{eq:a3-locked-h}, the coordinate remainder
can be written explicitly as
\[
\begin{aligned}
    \mathcal R_{r,\sigma}^{(a)}
    ={}&
    -q_{r,\sigma}^{(a)}
    \left[
        \prod_{b\neq a}
        \left(
            |q_{r,\sigma}^{(b)}|^2
            +
            \|\mathbf h_{r,\sigma}^{(b)}\|^2
        \right)
        -
        \prod_{b\neq a}|q_{r,\sigma}^{(b)}|^2
    \right]
    \\
    &-
    \sum_{k\neq k_r}
    x_{ki_r,\sigma}^{(a)}
    \prod_{b\neq a}
    G_{k_rk,\sigma}^{(b)}.
\end{aligned}
\]
The first term represents the correction to the
self-interaction caused by the off-sector components
of atom $r$. By the upper main-coordinate bound
in \eqref{eq:a3-fast-locking-coarse-annulus}
and $E_{\perp,\sigma}\leq\delta<1$,
this term is $O(E_{\perp,\sigma}^2)$.

For any other captured atom $s\neq r$,
orthogonality of the assigned sectors and the
decomposition \eqref{eq:a3-locked-h} give
\[
    |x_{k_s i_r,\sigma}^{(a)}|
    \leq E_{\perp,\sigma},
    \qquad
    |G_{k_rk_s,\sigma}^{(b)}|
    \leq C E_{\perp,\sigma}.
\]
Hence the interaction with atom $s$ contributes
$O(E_{\perp,\sigma}^{\nu})$.
Each still-unclaimed atom contributes
$O(\sigma^\nu)$ under the microscopic-atom
bootstrap bound. Since $\nu\geq3$ and
$E_{\perp,\sigma}\leq\delta<1$, we conclude that
\[
    |\mathcal R_{r,\sigma}^{(a)}|
    \leq
    C\left(
        \sigma^\nu+E_{\perp,\sigma}^2
    \right).
\]
Using the product-remainder identity above
and the upper bound
$|q_{r,\sigma}^{(a)}|\leq C_*$
in \eqref{eq:a3-fast-locking-coarse-annulus},
we obtain
\begin{equation}
    |\mathcal R_{r,\sigma}^{\pi}(\xi)|
    \leq
    C\left(
        \sigma^\nu
        +
        E_{\perp,\sigma}(\xi)^2
    \right).
    \label{eq:a3-fast-locking-product-forcing}
\end{equation}

We next establish uniform upper and lower bounds
on the captured products. For the old captured atoms,
the initial product estimate
\eqref{eq:a3-fast-locking-old-initial} gives
\[
    \pi_{r,\sigma}(\xi_0^\sigma)
    =
    \lambda_{i_r}+o(1),
    \qquad r\leq m.
\]
For the new winner, the initial main-coordinate
bounds \eqref{eq:a3-fast-locking-main-bounds}
and the phase condition
\eqref{eq:a3-fast-locking-phase-initial}
give
\[
    \operatorname{Re}
    \pi_{m+1,\sigma}(\xi_0^\sigma)
    \geq
    \frac12 c_0^\nu
\]
for sufficiently small $\sigma$.
Consequently, all $m+1$ initial products satisfy
\[
    \operatorname{Re}
    \pi_{r,\sigma}(\xi_0^\sigma)
    \geq
    \frac12 q_{\min}^\nu,
    \qquad
    |\pi_{r,\sigma}(\xi_0^\sigma)|
    \leq
    \frac32 q_{\max}^\nu
\]
for sufficiently small $\sigma$.

Writing
\[
    I_r(\xi)
    :=
    \int_{\xi_0^\sigma}^{\xi}
    S_{r,\sigma}(s)\,ds,
\]
variation of constants in the product equation
\eqref{eq:a3-fast-locking-product-equation} gives
\[
\begin{aligned}
    \pi_{r,\sigma}(\xi)
    ={}&
    e^{-I_r(\xi)}
    \pi_{r,\sigma}(\xi_0^\sigma)
    +
    \bigl(1-e^{-I_r(\xi)}\bigr)\lambda_{i_r}\\
    &+
    \int_{\xi_0^\sigma}^{\xi}
    e^{-\int_s^\xi S_{r,\sigma}(v)\,dv}
    \mathcal R_{r,\sigma}^{\pi}(s)\,ds.
\end{aligned}
\]
By the positive lower bound on $S_{r,\sigma}$
in \eqref{eq:a3-fast-locking-product-coercivity}
and the remainder estimate
\eqref{eq:a3-fast-locking-product-forcing},
the final integral has magnitude at most
\[
    \frac{C}{c_S}
    \left(
        \sigma^\nu+\delta^2
    \right).
\]
The first two terms form a convex combination
of the initial product and $\lambda_{i_r}$.
Since
$q_{\min}^\nu\leq\lambda_{i_r}\leq q_{\max}^\nu$,
choosing $\delta$ sufficiently small and then
$\sigma$ sufficiently small yields
\[
    \operatorname{Re}\pi_{r,\sigma}(\xi)
    \geq
    \pi_-,
    \qquad
    |\pi_{r,\sigma}(\xi)|
    \leq
    \pi_+,
\]
where
\[
    \pi_-:=\frac14q_{\min}^\nu,
    \qquad
    \pi_+:=2q_{\max}^\nu.
\]
These bounds hold uniformly for every captured
atom $r=1,\ldots,m+1$ throughout the bootstrap
interval.

Finally, we show that the product bounds prevent
any main-coordinate magnitude from leaving the
coarse annulus
\eqref{eq:a3-fast-locking-coarse-annulus}.
The exact main-coordinate balance estimate
\eqref{eq:a3-main-coordinate-balance-from-invariant},
together with the transverse bootstrap bound
$E_{\perp,\sigma}\leq\delta$, gives
\[
    \left|
        |q_{r,\sigma}^{(a)}|^2
        -
        |q_{r,\sigma}^{(b)}|^2
    \right|
    \leq
    C_{\mathrm{bal}}\sigma^2+2\delta^2
\]
for all $r\leq m+1$ and $a,b\in[\nu]$.
Since
\[
    |\pi_{r,\sigma}|^{2/\nu}
    =
    \left(
        \prod_{a=1}^{\nu}
        |q_{r,\sigma}^{(a)}|^2
    \right)^{1/\nu}
\]
lies between the minimum and maximum of the
squared main-coordinate magnitudes, we obtain
\[
    \left|
        |q_{r,\sigma}^{(a)}|^2
        -
        |\pi_{r,\sigma}|^{2/\nu}
    \right|
    \leq
    C_{\mathrm{bal}}\sigma^2+2\delta^2.
\]
Together with the product bounds
$\pi_-\leq|\pi_{r,\sigma}|\leq\pi_+$,
this yields
\[
\begin{aligned}
    |q_{r,\sigma}^{(a)}|^2
    &\geq
    \pi_-^{2/\nu}
    -
    C_{\mathrm{bal}}\sigma^2
    -
    2\delta^2,\\
    |q_{r,\sigma}^{(a)}|^2
    &\leq
    \pi_+^{2/\nu}
    +
    C_{\mathrm{bal}}\sigma^2
    +
    2\delta^2.
\end{aligned}
\]
By the choices
$c_*=q_{\min}/4$ and $C_*=4q_{\max}$,
\[
    c_*<\pi_-^{1/\nu},
    \qquad
    \pi_+^{1/\nu}<C_*.
\]
Hence, for sufficiently small $\delta$ and $\sigma$,
the preceding estimates place every
$|q_{r,\sigma}^{(a)}(\xi)|$ strictly inside
$(c_*,C_*)$, with a margin independent of $\sigma$.

Thus no main coordinate of any of the $m+1$
captured atoms can exit through the boundary of
the coarse annulus
\eqref{eq:a3-fast-locking-coarse-annulus}.
The transverse and still-unclaimed-atom bootstrap
bounds are verified in Steps 2 and 4, respectively,
completing the continuation argument on the full
locking interval.

\paragraph{Step 2: transverse contraction.}
As in the projected captured-atom equation
\eqref{eq:a3-fast-old-h-equation}, orthogonal projection
of the exact GF equation \eqref{eq:exact-gf} onto
the complement of the assigned sector gives
\[
    \frac{d\mathbf h_{r,\sigma}^{(a)}}{d\xi}
    =
    -
    \gamma_{r,\sigma}^{(a)}
    \mathbf h_{r,\sigma}^{(a)}
    +
    \mathbf N_{r,\sigma}^{(a)},
\]
where
\[
    \gamma_{r,\sigma}^{(a)}
    :=
    \prod_{b\neq a}
    \|\mathbf u_{k_r,\sigma}^{(b)}\|^2
    \geq
    c_*^{2(\nu-1)}
\]
by the coarse-annulus bound
\eqref{eq:a3-fast-locking-coarse-annulus}.
The off-sector target terms and interactions with
other captured atoms are
$O(E_{\perp,\sigma}^{\nu-1})$, since each
contains at least $\nu-1$ off-sector coordinates
or cross-Gram factors. Interactions with
still-unclaimed atoms are $O(\sigma^\nu)$
under the microscopic-atom bootstrap.
Consequently,
\[
    \|\mathbf N_{r,\sigma}^{(a)}\|
    \leq
    C\left(
        E_{\perp,\sigma}^{\nu-1}
        +
        \sigma^\nu
    \right).
\]
Taking upper Dini derivatives and maximizing over
the captured atoms and modes gives
\begin{equation}
    D^+E_{\perp,\sigma}
    \leq
    -c_\gamma E_{\perp,\sigma}
    +
    C E_{\perp,\sigma}^{\nu-1}
    +
    C\sigma^\nu
    \label{eq:a3-fast-locking-transverse-inequality}
\end{equation}
for some $c_\gamma>0$ independent of $\sigma$. The positive damping constant
comes from the coarse lower bounds on all captured main coordinates.

Since $\nu\geq3$, the nonlinear term is of higher order. Choosing $\delta>0$ sufficiently small gives
\begin{equation}
    C E_{\perp,\sigma}^{\nu-1}
    \leq
    \frac{c_\gamma}{2}
    E_{\perp,\sigma},
    \label{eq:a3-fast-locking-nonlinear-absorbed}
\end{equation}
and therefore
\begin{equation}
    D^+E_{\perp,\sigma}
    \leq
    -\frac{c_\gamma}{2}
    E_{\perp,\sigma}
    +
    C\sigma^\nu.
    \label{eq:a3-fast-locking-linear-inequality}
\end{equation}
Variation of constants yields
\begin{equation}
    E_{\perp,\sigma}(\xi)
    \leq
    e^{-c_\gamma(\xi-\xi_0^\sigma)/2}
    E_{\perp,\sigma}(\xi_0^\sigma)
    +
    C\sigma^\nu.
    \label{eq:a3-fast-locking-transverse-decay}
\end{equation}

By the initial off-sector bounds
\eqref{eq:a3-fast-locking-old-initial} and
\eqref{eq:a3-fast-locking-new-h-initial},
we have
$E_{\perp,\sigma}(\xi_0^\sigma)=o(1)$.
Hence the transverse decay estimate
\eqref{eq:a3-fast-locking-transverse-decay} gives,
uniformly on the bootstrap interval,
\[
    E_{\perp,\sigma}(\xi)
    \leq
    E_{\perp,\sigma}(\xi_0^\sigma)
    +
    C\sigma^\nu
    =
    o(1)
    <
    \delta
\]
for sufficiently small $\sigma$. Thus the transverse
bootstrap bound is strictly improved.

Moreover, choosing
$C_{\mathrm{lock}}>2\nu/c_\gamma$
ensures that the exponentially decaying term in
\eqref{eq:a3-fast-locking-transverse-decay}
is $O(\sigma^\nu)$ at the end of the locking
interval. Once the simultaneous continuation
argument is closed, this yields the off-sector
locking conclusion
\eqref{eq:a3-fast-locking-offsector}.

\paragraph{Step 3: contraction of the product errors.}
We now show that the captured-product errors
$e_{r,\sigma}:=\pi_{r,\sigma}-\lambda_{i_r}$,
introduced in Step~1, reach the required $O(\sigma^\nu)$
scale by the end of the locking interval.

Set
\[
    E_0
    :=
    E_{\perp,\sigma}(\xi_0^\sigma)
    =
    o(1),
    \qquad
    t:=\xi-\xi_0^\sigma.
\]
The transverse contraction estimate
\eqref{eq:a3-fast-locking-transverse-decay} gives
\[
    E_{\perp,\sigma}(\xi_0^\sigma+t)
    \leq
    E_0 e^{-c_\gamma t/2}
    +
    C\sigma^\nu.
\]
Consequently, the quadratic product-remainder bound
\eqref{eq:a3-fast-locking-product-forcing} implies
\[
    |\mathcal R_{r,\sigma}^{\pi}
        (\xi_0^\sigma+t)|
    \leq
    C\left(
        \sigma^\nu
        +
        E_0^2 e^{-c_\gamma t}
    \right),
    \qquad 1\leq r\leq m+1,
\]
where we used $\sigma^{2\nu}\leq\sigma^\nu$
for sufficiently small $\sigma$.

Applying variation of constants to the product equation
\eqref{eq:a3-fast-locking-product-equation}, and using
the uniform coercivity
$S_{r,\sigma}\geq c_S>0$ from
\eqref{eq:a3-fast-locking-product-coercivity}, we obtain
\begin{equation}
\begin{aligned}
    |e_{r,\sigma}(\xi_0^\sigma+t)|
    \leq{}&
    e^{-c_S t}
    |e_{r,\sigma}(\xi_0^\sigma)|
    +
    C\sigma^\nu\\
    &+
    CE_0^2
    \int_0^t
        e^{-c_S(t-s)}
        e^{-c_\gamma s}\,ds\\
    \leq{}&
    e^{-c_S t}
    |e_{r,\sigma}(\xi_0^\sigma)|
    +
    C\sigma^\nu
    +
    CE_0^2 t e^{-\beta t},
\end{aligned}
\label{eq:a3-fast-locking-product-decay}
\end{equation}
where
\[
    \beta:=\min\{c_S,c_\gamma\}>0.
\]
The last inequality follows by bounding both
exponential factors in the convolution by their
common decay rate $\beta$. In particular, it also
holds when $c_S=c_\gamma$.

By assumption~1, the initial product errors are
$o(1)$ for the first $m$ captured atoms. For the
new winner, they are uniformly bounded by the
main-coordinate bounds
\eqref{eq:a3-fast-locking-main-bounds}.
Thus
\[
    \max_{1\leq r\leq m+1}
    |e_{r,\sigma}(\xi_0^\sigma)|
    =
    O(1).
\]

Choose the constant $C_{\mathrm{lock}}$ in
\eqref{eq:a3-fast-locking-time} sufficiently large
that
\[
    C_{\mathrm{lock}}
    >
    \max\left\{
        \frac{\nu}{c_S},
        \frac{2\nu}{c_\gamma}
    \right\}.
\]
At the end of the locking interval,
$t=\xi_{\mathrm{lock}}^\sigma
=C_{\mathrm{lock}}\log(1/\sigma)$,
this choice gives
\[
    e^{-c_S t}
    =
    O(\sigma^\nu),
    \qquad
    t e^{-\beta t}
    =
    O(\sigma^\nu).
\]
The second estimate uses
$\beta C_{\mathrm{lock}}>\nu$;
the strict inequality absorbs the logarithmic
factor even when the two decay rates coincide.

Since $E_0=o(1)$, the product-error estimate
\eqref{eq:a3-fast-locking-product-decay} now yields
\[
    \max_{1\leq r\leq m+1}
    |e_{r,\sigma}
        (\xi_0^\sigma+\xi_{\mathrm{lock}}^\sigma)|
    \leq
    C\sigma^\nu.
\]
Recalling the definition of $e_{r,\sigma}$,
this proves the product-locking conclusion
\eqref{eq:a3-fast-locking-product}.

\paragraph{Step 4: the remaining atoms stay microscopic.}
We finally verify the bootstrap bound for the still-unclaimed
atoms. 
Throughout this step, we slightly abuse notation
by writing $\mathbf u_{k,\sigma}^{(a)}(\xi)$ for
$\mathbf u_{k,\sigma}^{(a)}(\Theta_\sigma(\xi))$,
where $\Theta_\sigma$ is defined in
\eqref{eq:a3-fast-slow-time-map} (i.e., we view $\mathbf u_{k,\sigma}^{(a)}$ as a function of local physical time rather than slow time).
 Let $k\notin\mathcal A_{m+1}$.  By assumption~4,
there is a constant $C_{\mathrm{mic}}>0$, independent of
$\sigma$, such that
\[
    \max_{\substack{k\notin\mathcal A_{m+1}\\a\in[\nu]}}
    \|\mathbf u_{k,\sigma}^{(a)}(\xi_0^\sigma)\|
    \leq C_{\mathrm{mic}}\sigma.
\]
We bootstrap the corresponding bound
$2C_{\mathrm{mic}}\sigma$ throughout the locking interval.

By the exact GF equation \eqref{eq:exact-gf}, the
target-driven term for a still-unclaimed atom is
$O(\sigma^{\nu-1})$. The interactions with other
microscopic atoms are $O(\sigma^{2\nu-1})$.
For interactions with a captured atom, the
coarse-annulus bound
\eqref{eq:a3-fast-locking-coarse-annulus}
and the transverse bootstrap
$E_{\perp,\sigma}\leq\delta$ imply that all its factors
have uniformly bounded norms. Each Gram factor
involving one captured and one microscopic atom is
therefore $O(\sigma)$, so these interactions are
also $O(\sigma^{\nu-1})$.

Consequently, throughout the bootstrap interval,
\[
    \max_{\substack{k\notin\mathcal A_{m+1}\\a\in[\nu]}}
    \left\|
        \frac{d\mathbf u_{k,\sigma}^{(a)}}{d\xi}
    \right\|
    \leq C\sigma^{\nu-1}.
\]
Integrating over the logarithmic locking interval
$\xi_{\mathrm{lock}}^\sigma
=C_{\mathrm{lock}}\log(1/\sigma)$
defined in \eqref{eq:a3-fast-locking-time}, we obtain
\begin{equation}
    \max_{\substack{k\notin\mathcal A_{m+1}\\a\in[\nu]}}
    \sup_{\xi_0^\sigma\leq\xi\leq
        \xi_0^\sigma+\xi_{\mathrm{lock}}^\sigma}
    \left\|
        \mathbf u_{k,\sigma}^{(a)}(\xi)
        -
        \mathbf u_{k,\sigma}^{(a)}(\xi_0^\sigma)
    \right\|
    \leq
    C\sigma^{\nu-1}\log\frac1\sigma
    =
    o(\sigma),
    \label{eq:a3-fast-locking-survivor-change}
\end{equation}
because $\nu\geq3$.

Thus every still-unclaimed atom has norm at most
$(C_{\mathrm{mic}}+o(1))\sigma$ in each mode, which
strictly improves the microscopic bootstrap bound
$2C_{\mathrm{mic}}\sigma$ for sufficiently small
$\sigma$. Hence these atoms remain $O(\sigma)$
throughout the locking interval.

Together with the coarse-annulus preservation in
Step~1 and the transverse contraction in Step~2,
this completes the simultaneous continuation argument
and the proof of the lemma.
\end{proof}

\paragraph{Completion of one fast transition.}

\begin{proposition}[Fast capture and renewal]
\label{prop:a3-fast-capture}
Suppose that the induction hypotheses
$\mathrm{(I_m)}$--$\mathrm{(IV_m)}$ hold at $\rho_m^\sigma$, and let
$\eta_{m+1}^\sigma$ be the entrance time supplied by
Proposition~\ref{prop:a3-mesoscopic-entrance}.

Then there exists a reset time $\rho_{m+1}^\sigma$ such that
\begin{equation}
    \rho_{m+1}^\sigma
    -
    \theta_{m+1}^*
    \longrightarrow0,
    \label{eq:a3-fast-new-reset-time}
\end{equation}
and the induction hypotheses
$\mathrm{(I_{m+1})}$--$\mathrm{(IV_{m+1})}$
hold at $\rho_{m+1}^\sigma$.
\end{proposition}

\begin{proof}
Apply Lemma~\ref{lem:a3-fast-growth} and set
\begin{equation}
    \xi_0^\sigma
    :=
    \xi_{\mathrm{grow}}^\sigma.
    \label{eq:a3-fast-locking-start}
\end{equation}
At this time the winning-factor bounds
\eqref{eq:a3-fast-growth-terminal}, the phase alignment
\eqref{eq:a3-fast-growth-phase}, the winner off-sector estimate
\eqref{eq:a3-fast-growth-h}, the old-background estimate
\eqref{eq:a3-fast-growth-old-background}, and the spectator estimate
\eqref{eq:a3-fast-growth-spectators} verify the hypotheses of
Lemma~\ref{lem:a3-fast-locking}.

Define the total physical duration of the fast layer by
\begin{equation}
    T_{\mathrm{fast},\sigma}
    :=
    \xi_{\mathrm{grow}}^\sigma
    +
    \xi_{\mathrm{lock}}^\sigma
    \label{eq:a3-fast-total-duration}
\end{equation}
and set
\begin{equation}
    \rho_{m+1}^\sigma
    :=
    \eta_{m+1}^\sigma
    +
    \sigma^{\nu-2}
    T_{\mathrm{fast},\sigma}.
    \label{eq:a3-fast-reset-definition}
\end{equation}
The growth-time estimate
\eqref{eq:a3-fast-growth-time} and the logarithmic locking time
\eqref{eq:a3-fast-locking-time} give
\begin{equation}
    \begin{aligned}
    \sigma^{\nu-2}T_{\mathrm{fast},\sigma}
    &=
    O\!\left(
        \sigma^{\nu-2}
        r_\sigma^{-(\nu-2)}
    \right)
    +
    O\!\left(
        \sigma^{\nu-2}
        \log\frac1\sigma
    \right)
    \\
    &=
    O\!\left(
        R_\sigma^{-(\nu-2)}
    \right)
    +
    o(1)
    \longrightarrow0.
    \end{aligned}
    \label{eq:a3-fast-slow-width}
\end{equation}
Together with the entrance-time convergence
$\eta_{m+1}^\sigma\to\theta_{m+1}^*$ from
\eqref{eq:a3-entrance-time-limit}, this proves the reset-time convergence
\eqref{eq:a3-fast-new-reset-time}.

It remains to verify the four induction hypotheses at
$\rho_{m+1}^\sigma$.

\paragraph{1. Verification of $\mathrm{(I_{m+1})}$.}
The product-locking conclusion
\eqref{eq:a3-fast-locking-product} of
Lemma~\ref{lem:a3-fast-locking} gives
\begin{equation}
    \max_{1\leq r\leq m+1}
    \left|
        \pi_{r,\sigma}(\rho_{m+1}^\sigma)
        -
        \lambda_{i_r}
    \right|
    \leq
    C_{m+1}\sigma^\nu.
    \label{eq:a3-fast-renew-I-product}
\end{equation}
This is exactly $\mathrm{(I_{m+1})}$.

\paragraph{2. Verification of $\mathrm{(II_{m+1})}$.}
The spectator conclusions of
Lemmas~\ref{lem:a3-fast-growth} and~\ref{lem:a3-fast-locking} show that
every atom whose index is not in $\mathcal A_{m+1}$ remains at the
initialization scale throughout the fast layer. Hence
\begin{equation}
    \max_{\substack{k\notin\mathcal A_{m+1}\\a\in[\nu]}}
    \left\|
        \mathbf u_{k,\sigma}^{(a)}
        (\rho_{m+1}^\sigma)
    \right\|
    \leq
    C_{m+1}\sigma,
    \label{eq:a3-fast-renew-II}
\end{equation}
which is $\mathrm{(II_{m+1})}$.

\paragraph{3. Verification of $\mathrm{(III_{m+1})}$.}
Let $(k,i)\in\mathcal E_{m+1}$. Then
$k\neq k_*$ and $i\neq i_*$, so this candidate is a spectator throughout
the fast transition.

Its physical-time derivative is $O(\sigma^{\nu-1})$ in every mode.
Since
$y_{ki,\sigma}^{(a)}=x_{ki,\sigma}^{(a)}/\sigma$, its microscopic
coordinate therefore satisfies
\begin{equation}
    \left|
        \frac{d}{d\xi}
        y_{ki,\sigma}^{(a)}
        \bigl(\Theta_\sigma(\xi)\bigr)
    \right|
    \leq
    C\sigma^{\nu-2}.
    \label{eq:a3-fast-renew-profile-derivative}
\end{equation}
Integrating over the total fast duration and using the vanishing slow width
\eqref{eq:a3-fast-slow-width} gives
\begin{equation}
    \max_{\substack{(k,i)\in\mathcal E_{m+1}\\a\in[\nu]}}
    \left|
        y_{ki,\sigma}^{(a)}(\rho_{m+1}^\sigma)
        -
        y_{ki,\sigma}^{(a)}(\eta_{m+1}^\sigma)
    \right|
    \longrightarrow0.
    \label{eq:a3-fast-renew-profile-exact-change}
\end{equation}

Every candidate in $\mathcal E_{m+1}$ has clock strictly larger than
$\theta_{m+1}^*$ by the future-clock property
\eqref{eq:a3-available-clocks-future}. Hence its isolated profile is smooth
in a fixed neighborhood of $\theta_{m+1}^*$. Since both
$\eta_{m+1}^\sigma$ and $\rho_{m+1}^\sigma$ converge to this time,
\begin{equation}
    \max_{\substack{(k,i)\in\mathcal E_{m+1}\\a\in[\nu]}}
    \left|
        \zeta_{ki}^{(a)}(\rho_{m+1}^\sigma)
        -
        \zeta_{ki}^{(a)}(\eta_{m+1}^\sigma)
    \right|
    \longrightarrow0.
    \label{eq:a3-fast-renew-profile-isolated-change}
\end{equation}

Finally, the diagonal profile estimate
\eqref{eq:a3-entrance-diagonal-profile-control}, established in the proof
of Proposition~\ref{prop:a3-mesoscopic-entrance}, gives
\begin{equation}
    \max_{\substack{(k,i)\in\mathcal E_{m+1}\\a\in[\nu]}}
    \left|
        y_{ki,\sigma}^{(a)}(\eta_{m+1}^\sigma)
        -
        \zeta_{ki}^{(a)}(\eta_{m+1}^\sigma)
    \right|
    \longrightarrow0.
    \label{eq:a3-fast-renew-profile-at-entrance}
\end{equation}
Combining the exact fast-layer change
\eqref{eq:a3-fast-renew-profile-exact-change}, the isolated-profile change
\eqref{eq:a3-fast-renew-profile-isolated-change}, and the entrance
comparison \eqref{eq:a3-fast-renew-profile-at-entrance} proves
$\mathrm{(III_{m+1})}$.

\paragraph{4. Verification of $\mathrm{(IV_{m+1})}$.}
The off-sector locking conclusion
\eqref{eq:a3-fast-locking-offsector} gives
\begin{equation}
    \max_{\substack{1\leq r\leq m+1\\a\in[\nu]}}
    \left\|
        \mathbf h_{r,\sigma}^{(a)}
        (\rho_{m+1}^\sigma)
    \right\|
    \leq
    C_{m+1}\sigma^\nu,
    \label{eq:a3-fast-renew-IV-captured}
\end{equation}
which is the captured-atom part of
$\mathrm{(IV_{m+1})}$.

For a surviving atom $k\notin\mathcal A_{m+1}$, the microscopic norm bound
$\mathrm{(II_{m+1})}$ just proved implies that every scalar coordinate is
$O(\sigma)$. In particular,
\begin{equation}
    \max_{\substack{
        k\notin\mathcal A_{m+1},\
        i\in\mathcal S_{m+1}\\
        a\in[\nu]
    }}
    \left|
        x_{ki,\sigma}^{(a)}
        (\rho_{m+1}^\sigma)
    \right|
    \leq
    C_{m+1}\sigma,
    \label{eq:a3-fast-renew-IV-blocked}
\end{equation}
or, equivalently, the corresponding rescaled coordinates are uniformly
$O(1)$. Thus $\mathrm{(IV_{m+1})}$ holds.

The renewed product condition
$\mathrm{(I_{m+1})}$ and off-sector locking
$\mathrm{(IV_{m+1})}$ also imply, through the exact balance invariant
\eqref{eq:a3-main-coordinate-balance-from-invariant}, the derived
captured-factor balance
\begin{equation}
    \max_{\substack{1\leq r\leq m+1\\a\in[\nu]}}
    \left|
        \left|
            q_{r,\sigma}^{(a)}
            (\rho_{m+1}^\sigma)
        \right|
        -
        \lambda_{i_r}^{1/\nu}
    \right|
    \longrightarrow0,
    \label{eq:a3-fast-renew-derived-balance}
\end{equation}
in agreement with the general consequence
\eqref{eq:a3-induction-captured-balance}.

We have therefore renewed
$\mathrm{(I_{m+1})}$--$\mathrm{(IV_{m+1})}$ at a reset time satisfying
\eqref{eq:a3-fast-new-reset-time}. This completes the induction step.
\end{proof}

\paragraph{Renewal of the induction.}
The renewed profile-preservation hypothesis
$\mathrm{(III_{m+1})}$ compares every surviving candidate with the same
isolated profile $\boldsymbol{\zeta}_{ki}$ defined from the original
initialization in \eqref{eq:a3-candidate-profile}. Thus earlier captures do
not restart the candidate clocks.

By the future-clock property
\eqref{eq:a3-available-clocks-future}, now with $m$ replaced by $m+1$,
\begin{equation}
    \tau_{ki}
    >
    \theta_{m+1}^*
    \qquad
    \text{for every }(k,i)\in\mathcal E_{m+1}.
    \label{eq:a3-renewal-future-clocks}
\end{equation}
If $m+1<M$, the greedy next-edge rule
\eqref{eq:a3-next-greedy-edge} therefore gives
\begin{equation}
    (k_{m+2},i_{m+2})
    =
    \operatorname*{arg\,min}_{(k,i)\in\mathcal E_{m+1}}
    \tau_{ki},
    \qquad
    \theta_{m+2}^*
    =
    \min_{(k,i)\in\mathcal E_{m+1}}
    \tau_{ki}.
    \label{eq:a3-renewal-next-clock}
\end{equation}
Since
\begin{equation}
    \rho_{m+1}^\sigma
    -
    \theta_{m+1}^*
    \longrightarrow0
    \label{eq:a3-renewal-reset-convergence}
\end{equation}
and the next clock has the fixed positive gap
$\theta_{m+2}^*-\theta_{m+1}^*>0$, we have, for all sufficiently small
$\sigma$,
\begin{equation}
    \rho_{m+1}^\sigma
    <
    \theta_{m+2}^*.
    \label{eq:a3-renewal-reset-before-next-clock}
\end{equation}
Proposition~\ref{prop:a3-slow-propagation} can therefore be applied again
with $m$ replaced by $m+1$. The induction continues with the same original
isolated profiles and candidate clocks.

\subsection{Completion of the proof of the greedy matching limit}
\label{app:a3-completion}

We now assemble the preceding estimates to complete the induction and prove
Theorem~\ref{thm:greedy-matching-limit}. The only additional ingredient is
control of the terminal plateau after the last accepted edge. At that point
there is no subsequent activation clock, so
Proposition~\ref{prop:a3-slow-propagation} is no longer the appropriate
propagation statement. Instead, the exhaustion of one side of the matching
allows us to propagate the final locked state directly.

\begin{lemma}[Propagation after the final capture]
\label{lem:a3-terminal-propagation}
Suppose that the induction hypotheses
$\mathrm{(I_M)}$--$\mathrm{(IV_M)}$ hold at the final reset time
$\rho_M^\sigma$. Then, for every fixed
$\Theta>\theta_M^*$,
\begin{equation}
    \sup_{\theta\in[\rho_M^\sigma,\Theta]}
    \bigl\|
        \widehat T_\sigma(\theta)-T_M
    \bigr\|_F
    \longrightarrow0
    \qquad
    \text{as }\sigma\to0.
    \label{eq:a3-terminal-output}
\end{equation}
\end{lemma}

\begin{proof}
Since $M=\min\{H,p\}$, after the final greedy acceptance one side of the
bipartite graph is exhausted. Equivalently,
\begin{equation}
    \mathcal A_M=[H]
    \qquad\text{or}\qquad
    \mathcal S_M=[p].
    \label{eq:a3-terminal-exhaustion}
\end{equation}
Thus either there are no surviving atoms, or every target sector has already
been claimed.

Fix $\Theta>\theta_M^*$. The reset-time convergence supplied by the final
application of Proposition~\ref{prop:a3-fast-capture},
\begin{equation}
    \rho_M^\sigma
    \longrightarrow
    \theta_M^*,
    \label{eq:a3-terminal-reset-convergence}
\end{equation}
implies that
$\rho_M^\sigma<\Theta$ for all sufficiently small $\sigma$.

We use a continuation argument on
$[\rho_M^\sigma,\Theta]$. For the captured atoms, impose the coarse annulus
bootstrap
\begin{equation}
    \frac12\lambda_{i_r}^{1/\nu}
    <
    \bigl|
        q_{r,\sigma}^{(a)}(\theta)
    \bigr|
    <
    2\lambda_{i_r}^{1/\nu},
    \qquad
    r\leq M,\quad a\in[\nu].
    \label{eq:a3-terminal-bootstrap-captured}
\end{equation}
If surviving atoms remain, then the exhaustion property
\eqref{eq:a3-terminal-exhaustion} forces
$\mathcal S_M=[p]$. In this case we also impose
\begin{equation}
    \max_{\substack{
        k\notin\mathcal A_M\\
        i\in[p],\ a\in[\nu]
    }}
    \bigl|
        y_{ki,\sigma}^{(a)}(\theta)
    \bigr|
    <
    C_M+1.
    \label{eq:a3-terminal-bootstrap-survivors}
\end{equation}
If $\mathcal A_M=[H]$, there are no surviving atoms and the second bootstrap
condition is vacuous.

The captured-factor balance at the reset time,
\eqref{eq:a3-induction-captured-balance} with $m=M$, gives strict initial
slack in the annulus
\eqref{eq:a3-terminal-bootstrap-captured}. If surviving atoms remain, then
all sectors are claimed, and the blocked-sector induction bound
\eqref{eq:a3-induction-blocked-rescaled} with $m=M$ gives strict initial
slack in
\eqref{eq:a3-terminal-bootstrap-survivors}. Hence the bootstrap conditions
hold on a nontrivial interval beginning at $\rho_M^\sigma$.

\paragraph{Step 1: the bootstrap assumptions give the hypotheses of the
locked-background proposition.}
Suppose first that surviving atoms remain. The survivor bootstrap
\eqref{eq:a3-terminal-bootstrap-survivors} and orthonormality of the sector
basis give
\begin{equation}
    \begin{aligned}
    \bigl\|
        \mathbf u_{k,\sigma}^{(a)}(\theta)
    \bigr\|^2
    &=
    \sigma^2
    \sum_{i=1}^{p}
    \left|
        y_{ki,\sigma}^{(a)}(\theta)
    \right|^2
    \\
    &\leq
    p(C_M+1)^2\sigma^2.
    \end{aligned}
    \label{eq:a3-terminal-survivors-small}
\end{equation}
Thus every surviving atom is uniformly $O(\sigma)$ on the bootstrap
interval. If no surviving atoms remain, this condition is automatic.

The captured annulus
\eqref{eq:a3-terminal-bootstrap-captured} supplies the coarse lower and
upper bounds required by
Proposition~\ref{prop:a3-locked-background}. Its quantitative initial-locking
assumption follows from the captured-product induction estimate
\eqref{eq:a3-induction-captured-product} and the captured off-sector
induction estimate
\eqref{eq:a3-induction-captured-locking}. Finally,
$[\rho_M^\sigma,\Theta]$ has uniformly bounded length by the reset
convergence \eqref{eq:a3-terminal-reset-convergence}. Hence all assumptions
of Proposition~\ref{prop:a3-locked-background} hold on the current bootstrap
interval.

\paragraph{Step 2: the captured atoms remain locked and balanced.}
The captured-background stability conclusion
\eqref{eq:a3-prop-captured-stability} of
Proposition~\ref{prop:a3-locked-background} gives
\begin{equation}
    \max_{r\leq M}
    \sup_\theta
    \left|
        \pi_{r,\sigma}(\theta)-\lambda_{i_r}
    \right|
    +
    \max_{\substack{r\leq M\\a\in[\nu]}}
    \sup_\theta
    \bigl\|
        \mathbf h_{r,\sigma}^{(a)}(\theta)
    \bigr\|
    =
    O(\sigma^\nu)
    \label{eq:a3-terminal-captured-lock}
\end{equation}
on the bootstrap interval.

At the same time, the absolute main-coordinate balance estimate
\eqref{eq:a3-prop-balance-stability}, which follows from the exact atom-wise
balance invariant, gives
\begin{equation}
    \max_{\substack{r\leq M\\a,b\in[\nu]}}
    \sup_\theta
    \left|
        |q_{r,\sigma}^{(a)}(\theta)|^2
        -
        |q_{r,\sigma}^{(b)}(\theta)|^2
    \right|
    =
    O(\sigma^2).
    \label{eq:a3-terminal-captured-balance-differences}
\end{equation}
The product part of the locking estimate
\eqref{eq:a3-terminal-captured-lock} also gives
\begin{equation}
    \prod_{a=1}^{\nu}
    \bigl|
        q_{r,\sigma}^{(a)}(\theta)
    \bigr|
    =
    \bigl|
        \pi_{r,\sigma}(\theta)
    \bigr|
    =
    \lambda_{i_r}+o(1)
    \label{eq:a3-terminal-captured-product-magnitude}
\end{equation}
uniformly on the bootstrap interval. Combining the asymptotic equality of
the squared magnitudes
\eqref{eq:a3-terminal-captured-balance-differences} with the product
constraint
\eqref{eq:a3-terminal-captured-product-magnitude} yields
\begin{equation}
    \max_{\substack{r\leq M\\a\in[\nu]}}
    \sup_\theta
    \left|
        \bigl|
            q_{r,\sigma}^{(a)}(\theta)
        \bigr|
        -
        \lambda_{i_r}^{1/\nu}
    \right|
    \longrightarrow0.
    \label{eq:a3-terminal-captured-balanced}
\end{equation}
For sufficiently small $\sigma$, this is a strict improvement of the
captured-annulus bootstrap
\eqref{eq:a3-terminal-bootstrap-captured}.

\paragraph{Step 3: any remaining atoms stay microscopic.}
This step is needed only if surviving atoms remain. In that case,
$\mathcal S_M=[p]$ by the exhaustion property
\eqref{eq:a3-terminal-exhaustion}, so every coordinate of every surviving
atom lies in a claimed sector.

The claimed-sector near-freezing estimate
\eqref{eq:a3-prop-claimed-frozen} from
Proposition~\ref{prop:a3-locked-background} therefore gives
\begin{equation}
    \max_{\substack{
        k\notin\mathcal A_M\\
        i\in[p],\ a\in[\nu]
    }}
    \sup_\theta
    \left|
        \frac{d}{d\theta}
        y_{ki,\sigma}^{(a)}(\theta)
    \right|
    \leq
    C\sigma^\nu
    \label{eq:a3-terminal-survivor-derivative}
\end{equation}
on the bootstrap interval. Since
$\Theta-\rho_M^\sigma$ is uniformly bounded, integrating the derivative
bound \eqref{eq:a3-terminal-survivor-derivative} and using the blocked-sector
reset estimate
\eqref{eq:a3-induction-blocked-rescaled} gives
\begin{equation}
    \max_{\substack{
        k\notin\mathcal A_M\\
        i\in[p],\ a\in[\nu]
    }}
    \sup_{\theta\in[\rho_M^\sigma,\Theta]}
    \left|
        y_{ki,\sigma}^{(a)}(\theta)
    \right|
    \leq
    C_M+o(1).
    \label{eq:a3-terminal-survivor-improvement}
\end{equation}
For sufficiently small $\sigma$, this strictly improves the survivor
bootstrap
\eqref{eq:a3-terminal-bootstrap-survivors}.

\paragraph{Step 4: closure of the continuation argument.}
Steps~2 and~3 strictly improve every nonvacuous bootstrap condition.
The same maximal-interval argument as in
Lemma~\ref{lem:a3-bootstrap-continuation} therefore shows that the bootstrap
interval extends to all of
$[\rho_M^\sigma,\Theta]$.

We may consequently apply the tensor-level conclusion
\eqref{eq:a3-prop-output} of
Proposition~\ref{prop:a3-locked-background} on the full interval, obtaining
\begin{equation}
    \sup_{\theta\in[\rho_M^\sigma,\Theta]}
    \bigl\|
        \widehat T_\sigma(\theta)-T_M
    \bigr\|_F
    =
    O(\sigma^\nu).
    \label{eq:a3-terminal-output-rate}
\end{equation}
This proves the terminal propagation statement
\eqref{eq:a3-terminal-output}.
\end{proof}

\begin{proof}[Proof of Theorem~\ref{thm:greedy-matching-limit}]
We work throughout on the probability-one initialization event fixed at the
beginning of Appendix~\ref{app:greedy}, on which all candidate clocks are
finite and pairwise distinct.

\paragraph{Step 1: completion of the induction.}
The base case $m=0$ was verified after the induction hypotheses were
introduced. At the initial reset time
$\rho_0^\sigma=0$ from
\eqref{eq:a3-initial-reset-time},
$\mathrm{(I_0)}$ and the captured-atom part of
$\mathrm{(IV_0)}$ are vacuous,
$\mathrm{(II_0)}$ follows directly from the Gaussian initialization, and
$\mathrm{(III_0)}$ holds exactly because the exact microscopic coordinates
and the isolated profiles have the same initial data.

Suppose now that
$\mathrm{(I_m)}$--$\mathrm{(IV_m)}$ hold at $\rho_m^\sigma$ for some
$m<M$. Proposition~\ref{prop:a3-slow-propagation} propagates this state
through every fixed interval before the next accepted clock. The
mesoscopic-entrance construction of
Proposition~\ref{prop:a3-mesoscopic-entrance} then produces an entrance time
$\eta_{m+1}^\sigma\to\theta_{m+1}^*$, and
Proposition~\ref{prop:a3-fast-capture} carries the resulting fast transition
through to a new reset time $\rho_{m+1}^\sigma$ at which
$\mathrm{(I_{m+1})}$--$\mathrm{(IV_{m+1})}$ hold. In particular, its
reset-time estimate gives
\begin{equation}
    \rho_{m+1}^\sigma
    -
    \theta_{m+1}^*
    \longrightarrow0.
    \label{eq:a3-completion-reset-convergence}
\end{equation}

If $m+1<M$, the next accepted clock has the fixed positive gap
\begin{equation}
    \theta_{m+2}^*
    -
    \theta_{m+1}^*
    >
    0.
    \label{eq:a3-completion-next-clock-gap}
\end{equation}
Indeed, this follows from the distinctness of the accepted candidate clocks,
or equivalently from the future-clock property
\eqref{eq:a3-available-clocks-future} and the greedy next-edge rule
\eqref{eq:a3-next-greedy-edge}. The reset convergence
\eqref{eq:a3-completion-reset-convergence} therefore implies
\begin{equation}
    \rho_{m+1}^\sigma
    <
    \theta_{m+2}^*
    \label{eq:a3-completion-reset-before-next-clock}
\end{equation}
for all sufficiently small $\sigma$.

Crucially, the renewed profile-preservation property
$\mathrm{(III_{m+1})}$ compares every surviving edge in
$\mathcal E_{m+1}$ with the same isolated profile
$\boldsymbol{\zeta}_{ki}$ defined from the original initialization in
\eqref{eq:a3-candidate-profile}. Thus the candidate clocks are preserved
through the fast transition rather than restarted. The induction can
therefore be repeated with the next edge selected by the same original
greedy clock process.

Iterating this argument constructs reset times
$\rho_1^\sigma,\ldots,\rho_M^\sigma$ such that
\begin{equation}
    \rho_m^\sigma
    \longrightarrow
    \theta_m^*,
    \qquad
    1\leq m\leq M,
    \label{eq:a3-completion-all-reset-convergence}
\end{equation}
and the hypotheses
$\mathrm{(I_m)}$--$\mathrm{(IV_m)}$ hold at every
$\rho_m^\sigma$.

\paragraph{Step 2: uniform convergence on every nonterminal plateau.}
Recall the accepted times
$\theta_1^*,\ldots,\theta_M^*$ from
\eqref{eq:a3-accepted-times}, and define the open plateaus between them by
\begin{equation}
    J_0
    :=
    [0,\theta_1^*),
    \qquad
    J_m
    :=
    (\theta_m^*,\theta_{m+1}^*),
    \quad
    1\leq m<M,
    \qquad
    J_M
    :=
    (\theta_M^*,\infty).
    \label{eq:a3-completion-plateaus}
\end{equation}
By the greedy construction, the selected-sector set is constant on each
plateau and equals $\mathcal S_m$. Hence, using the partial target
$T_m$ from \eqref{eq:a3-partial-target},
\begin{equation}
    \sum_{i\in\mathcal S(\theta)}
    \lambda_i
    \mathbf e_i^{(1)}
    \otimes\cdots\otimes
    \mathbf e_i^{(\nu)}
    =
    T_m,
    \qquad
    \theta\in J_m.
    \label{eq:a3-completion-greedy-tensor-on-plateau}
\end{equation}

Let
\begin{equation}
    K
    \subset
    [0,\infty)\setminus\mathcal T_*
    \label{eq:a3-completion-K}
\end{equation}
be compact, as in the statement of
Theorem~\ref{thm:greedy-matching-limit}, and set
\begin{equation}
    K_m
    :=
    K\cap J_m,
    \qquad
    0\leq m\leq M.
    \label{eq:a3-completion-Km}
\end{equation}
Because $K$ is compact and avoids the finite activation set
$\mathcal T_*=\{\theta_1^*,\ldots,\theta_M^*\}$, every nonempty $K_m$ lies
a positive distance from the activation times bounding its plateau.

Fix $m<M$ with $K_m\neq\varnothing$. Choose
$\delta_m>0$ such that
\begin{equation}
    0<\delta_m<
    \frac12
    \bigl(
        \theta_{m+1}^*-\theta_m^*
    \bigr),
    \qquad
    \sup K_m
    <
    \theta_{m+1}^*-\delta_m.
    \label{eq:a3-completion-delta-choice}
\end{equation}
For $m\geq1$, the positive separation of $K_m$ from the left activation time
also gives
\begin{equation}
    \inf K_m
    >
    \theta_m^*.
    \label{eq:a3-completion-left-gap}
\end{equation}
The reset convergence
\eqref{eq:a3-completion-all-reset-convergence} therefore implies
\begin{equation}
    \rho_m^\sigma
    <
    \inf K_m
    \label{eq:a3-completion-reset-before-Km}
\end{equation}
for all sufficiently small $\sigma$. For $m=0$, the same conclusion is
automatic because $\rho_0^\sigma=0$. Hence
\begin{equation}
    K_m
    \subset
    \bigl[
        \rho_m^\sigma,
        \theta_{m+1}^*-\delta_m
    \bigr]
    \label{eq:a3-completion-Km-in-slow-interval}
\end{equation}
for every nonempty $K_m$ with $m<M$ and all sufficiently small $\sigma$.

Applying the uniform plateau approximation
\eqref{eq:a3-slow-output-convergence} from
Proposition~\ref{prop:a3-slow-propagation} on the containing interval
\eqref{eq:a3-completion-Km-in-slow-interval} gives
\begin{equation}
    \sup_{\theta\in K_m}
    \bigl\|
        \widehat T_\sigma(\theta)-T_m
    \bigr\|_F
    \longrightarrow0,
    \qquad
    0\leq m<M.
    \label{eq:a3-completion-nonterminal-output}
\end{equation}

\paragraph{Step 3: uniform convergence on the terminal plateau.}
If $K_M=\varnothing$, there is nothing to prove. Otherwise, because
$K_M\subset(\theta_M^*,\infty)$ and $K_M$ is separated from
$\theta_M^*$, we have
\begin{equation}
    \inf K_M
    >
    \theta_M^*.
    \label{eq:a3-completion-terminal-left-gap}
\end{equation}
Choose a fixed
\begin{equation}
    \Theta>\sup K_M.
    \label{eq:a3-completion-terminal-Theta}
\end{equation}
The final reset convergence
$\rho_M^\sigma\to\theta_M^*$ from
\eqref{eq:a3-completion-all-reset-convergence} and the left gap
\eqref{eq:a3-completion-terminal-left-gap} imply, for sufficiently small
$\sigma$,
\begin{equation}
    K_M
    \subset
    [\rho_M^\sigma,\Theta].
    \label{eq:a3-completion-terminal-inclusion}
\end{equation}
Applying the terminal propagation lemma,
specifically its output estimate
\eqref{eq:a3-terminal-output}, gives
\begin{equation}
    \sup_{\theta\in K_M}
    \bigl\|
        \widehat T_\sigma(\theta)-T_M
    \bigr\|_F
    \longrightarrow0.
    \label{eq:a3-completion-terminal-output}
\end{equation}

\paragraph{Step 4: conclusion for the tensor trajectory.}
There are only finitely many plateaus. Combining the nonterminal convergence
\eqref{eq:a3-completion-nonterminal-output}, the terminal convergence
\eqref{eq:a3-completion-terminal-output}, and the identification of the
greedy tensor on each plateau
\eqref{eq:a3-completion-greedy-tensor-on-plateau} gives
\begin{equation}
    \sup_{\theta\in K}
    \left\|
        \widehat T_\sigma(\theta)
        -
        \sum_{i\in\mathcal S(\theta)}
        \lambda_i
        \mathbf e_i^{(1)}
        \otimes\cdots\otimes
        \mathbf e_i^{(\nu)}
    \right\|_F
    \longrightarrow0.
    \label{eq:a3-completion-output-limit}
\end{equation}
By the slow-time definitions
\eqref{eq:a3-slow-weight} and
\eqref{eq:a3-slow-output},
$\widehat T_\sigma(\theta)$ is precisely
$\widehat T^\sigma(\theta)$ in the notation of the main theorem. Therefore
\eqref{eq:a3-completion-output-limit} is exactly the tensor convergence
\eqref{eq:greedy-output-limit} asserted in
Theorem~\ref{thm:greedy-matching-limit}.

\paragraph{Step 5: conclusion for the loss trajectory.}
For $\theta\in J_m$, the greedy loss
\eqref{eq:greedy-loss} and orthogonality of the diagonal sector tensors give
\begin{equation}
    L_*(\theta)
    =
    \frac12
    \|T-T_m\|_F^2
    =
    \frac12
    \sum_{i\notin\mathcal S_m}
    \lambda_i^2.
    \label{eq:a3-completion-greedy-loss-on-plateau}
\end{equation}
For $\theta\in K_m$, the elementary difference-of-squares identity yields
\begin{equation}
    \begin{aligned}
    \left|
        L^\sigma(\theta)-L_*(\theta)
    \right|
    &=
    \frac12
    \left|
        \|T-\widehat T^\sigma(\theta)\|_F^2
        -
        \|T-T_m\|_F^2
    \right|
    \\
    &\leq
    \frac12
    \left(
        \|T-\widehat T^\sigma(\theta)\|_F
        +
        \|T-T_m\|_F
    \right)
    \bigl\|
        \widehat T^\sigma(\theta)-T_m
    \bigr\|_F .
    \end{aligned}
    \label{eq:a3-completion-loss-bound}
\end{equation}
The tensor convergence already proved in
\eqref{eq:a3-completion-output-limit} makes the second factor tend to zero
uniformly on each $K_m$, while the first factor remains uniformly bounded.
Since there are only finitely many nonempty plateau pieces $K_m$, we obtain
\begin{equation}
    \sup_{\theta\in K}
    \left|
        L^\sigma(\theta)-L_*(\theta)
    \right|
    \longrightarrow0.
    \label{eq:a3-completion-loss-limit}
\end{equation}
This is precisely the loss convergence
\eqref{eq:greedy-loss-limit}. The proof of
Theorem~\ref{thm:greedy-matching-limit} is complete.
\end{proof}

\appendixsectionwithmain
    {Complex modular addition: Fourier details}
    {app:modular-addition}
    {sec:modular-addition}

We give the Fourier calculation underlying
Section~\ref{sec:modular-addition} and derive
Corollary~\ref{cor:addition-greedy} directly from
Theorem~\ref{thm:greedy-matching-limit}.

\paragraph{Fourier diagonalization.}
For $q\in\mathbb Z_p$, the discrete Fourier identity
\[
    \mathbf 1\{q=0\pmod p\}
    =
    \frac{1}{p}\sum_{r\in\mathbb Z_p}\omega^{rq}
\]
gives, with $q=x+y-z$,
\[
    \mathbf 1\{x+y=z\pmod p\}
    =
    \frac{1}{p}
    \sum_{r\in\mathbb Z_p}\omega^{r(x+y-z)}
    =
    \sqrt p
    \sum_{r\in\mathbb Z_p}
    \boldsymbol{\varphi}_r(x)
    \boldsymbol{\varphi}_r(y)
    \overline{\boldsymbol{\varphi}_r(z)}.
\]
This proves \eqref{eq:addition-fourier}. Moreover,
$\{\boldsymbol{\varphi}_r\}_{r\in\mathbb Z_p}$ and
$\{\overline{\boldsymbol{\varphi}_r}\}_{r\in\mathbb Z_p}$ are orthonormal
bases of $\mathbb C^p$. Hence \eqref{eq:addition-fourier} is exactly of the
diagonal form \eqref{eq:diagonal-target}, with the mode-wise orthonormal
bases
\[
    \mathbf e_r^{(1)}
    =
    \mathbf e_r^{(2)}
    =
    \boldsymbol{\varphi}_r,
    \qquad
    \mathbf e_r^{(3)}
    =
    \overline{\boldsymbol{\varphi}_r},
\]
and equal sector amplitudes $\lambda_r=\sqrt p$. Thus all assumptions of
Theorem~\ref{thm:greedy-matching-limit} concerning the target geometry are
satisfied.

In particular, the sector tensors
$\boldsymbol{\varphi}_r\otimes
 \boldsymbol{\varphi}_r\otimes
 \overline{\boldsymbol{\varphi}_r}$
are orthonormal in the Frobenius inner product. Therefore
\[
    \|T_p^{\mathrm{add}}\|_F^2=p^2.
\]

\paragraph{Gaussian coordinates and activation clocks.}
For atom $k$ and Fourier sector $r$, define the microscopic coordinate triple
\[
    \mathbf g_{kr}
    :=
    \left(
        \langle\boldsymbol{\varphi}_r,\mathbf g_k^{(1)}\rangle,\,
        \langle\boldsymbol{\varphi}_r,\mathbf g_k^{(2)}\rangle,\,
        \langle\overline{\boldsymbol{\varphi}_r},\mathbf g_k^{(3)}\rangle
    \right).
\]
For each mode, the change from the original coordinates to the corresponding
Fourier basis is unitary. Since a standard circularly symmetric complex
Gaussian vector is invariant under unitary transformations, its Fourier
coordinates are again independent standard complex Gaussians. Independence
across atoms and modes is preserved as well. Consequently,
\[
    \{\mathbf g_{kr}:k\in[H],\,r\in\mathbb Z_p\}
    \quad\text{are i.i.d. with}\quad
    \mathbf g_{kr}\sim\mathcal N_{\mathbb C}(0,I_3).
\]
Applying the general clock formula \eqref{eq:candidate-clock} with
$\lambda_r=\sqrt p$ gives
\[
    \tau_{kr}
    =
    \frac{\tau_3(\mathbf g_{kr})}{\sqrt p}.
\]
Thus, after the change of time
$\vartheta=\sqrt p\,\theta$, the candidate activation times are
$\xi_{kr}:=\tau_3(\mathbf g_{kr})$. By
Proposition~\ref{prop:clock-atomless}, the $\xi_{kr}$ are i.i.d.\ and
atomless.

\begin{proof}[Proof of Corollary~\ref{cor:addition-greedy}]
Let $\mathcal T_*$ be the set of accepted activation times in the slow-time
variable $\theta$, and let
\[
    \mathcal T_{*,p}^{\mathrm{add}}
    :=
    \sqrt p\,\mathcal T_*
\]
be the corresponding set in the parametrization
$\vartheta=\sqrt p\,\theta$. Equivalently,
$\mathcal T_{*,p}^{\mathrm{add}}$ consists of the values $\xi_{kr}$ associated
with the edges accepted by the greedy process.

For any compact
$K\subset[0,\infty)\setminus\mathcal T_{*,p}^{\mathrm{add}}$,
the rescaled set $K/\sqrt p$ is a compact subset of
$[0,\infty)\setminus\mathcal T_*$. Therefore
Theorem~\ref{thm:greedy-matching-limit}, applied with
$\nu=3$, $\lambda_r=\sqrt p$, and $\theta=\vartheta/\sqrt p$, gives
\[
    \sup_{\vartheta\in K}
    \left\|
        \widehat T_p^{\mathrm{add},\sigma}(\vartheta)
        -
        \sqrt p
        \sum_{r\in S_p(\vartheta)}
        \boldsymbol{\varphi}_r
        \otimes
        \boldsymbol{\varphi}_r
        \otimes
        \overline{\boldsymbol{\varphi}_r}
    \right\|_F
    \longrightarrow 0.
\]
This is the tensor convergence stated in
Corollary~\ref{cor:addition-greedy}.

Since the Fourier sectors are orthonormal, the limiting residual at time
$\vartheta$ is
\[
    \sqrt p
    \sum_{r\notin S_p(\vartheta)}
    \boldsymbol{\varphi}_r
    \otimes
    \boldsymbol{\varphi}_r
    \otimes
    \overline{\boldsymbol{\varphi}_r},
\]
and therefore
\[
    L_{*,p}^{\mathrm{add}}(\vartheta)
    =
    \frac{p}{2}\bigl(p-M_p(\vartheta)\bigr).
\]
Dividing by $p^2/2$ yields
\[
    \frac{2L_{*,p}^{\mathrm{add}}(\vartheta)}{p^2}
    =
    1-\frac{M_p(\vartheta)}{p},
\]
which proves \eqref{eq:addition-greedy-loss}.
\end{proof}

\paragraph{Finite-$p$ Random Edge ordering.}
Because the $Hp$ variables $\xi_{kr}$ are i.i.d.\ and atomless, every strict
ordering of the edges of $K_{H,p}$ is equally likely. Hence, for fixed
$p$ and $H$, the accepted-edge sequence has exactly the same law as that
obtained by processing the edges of $K_{H,p}$ in a uniformly random order
and accepting an edge whenever both endpoints are still unmatched. The clock
distribution determines when the events occur, but not their combinatorial
ordering. This is precisely the Random Edge process used in
Section~\ref{sec:large-p}.

\appendixsectionwithmain
    {Large-system limit: proofs and asymptotics}
    {app:large-p}
    {sec:large-p}

\subsection{Large-system limit of Random Edge matching}
\label{app:random-edge-limit}

We begin by describing the random matching process used in
Proposition~\ref{prop:random-edge-limit} in more detail. The algorithm is
simply greedy matching in a random edge order. Given the complete bipartite
graph $K_{H_p,p}$ between $H_p$ atom vertices and $p$ sector vertices,
choose a uniformly random ordering of its $H_pp$ edges. Inspect the edges
one at a time in this order. An edge is accepted if neither of its endpoints
has appeared in any previously accepted edge; otherwise it is rejected.
Equivalently, an edge is accepted if both of its endpoints are still
unmatched. Following \citet{arnosti2022greedy}, we call this the
\emph{Random Edge} algorithm.

For the large-system analysis, it is useful to realize the same random
ordering continuously. Equivalently, assign each edge an independent
\emph{random} priority, uniformly distributed on $[0,1]$, and inspect the
edges in increasing order of priority. Since the priority distribution is
continuous, the priorities are almost surely distinct, and their ordering is
a uniformly random permutation of the edges.

\paragraph{Edge exposure.}
For $s\in[0,p]$, we say that an edge is \emph{exposed by level $s$} if its
priority is at most $s/p$. Thus an exposed edge is simply an edge that has
already been reached and processed by the algorithm; it may have been either
accepted or rejected. Each edge is independently exposed with probability
$s/p$, so the graph formed by the exposed edges is the bipartite
Erd\H{o}s--R\'enyi graph $G_{H_p,p,s/p}$, in which each of the $H_pp$
possible edges is present independently with probability $s/p$. We call $s$
the \emph{edge exposure parameter}.

The scaling $s/p$ is the sparse scaling on which the matching process has a
nontrivial large-$p$ evolution. Indeed, every atom vertex has $p$ possible
incident edges, and therefore its expected exposed degree at exposure $s$ is
\[
    p\frac{s}{p}=s.
\]
Likewise, every sector vertex has $H_p$ possible incident edges, so its
expected exposed degree is
\[
    H_p\frac{s}{p}
    =
    \frac{H_p}{p}s
    \longrightarrow
    \alpha s.
\]
Thus fixed $s=O(1)$ corresponds to a sparse graph in which both vertex
classes have $O(1)$ exposed degree and the matching process has a nontrivial
large-system evolution. By contrast, if the priority threshold were
$o(p^{-1})$, almost all vertices would have no exposed edges, whereas a fixed
positive priority threshold would give exposed degrees of order $p$ and
would place the graph in a dense regime.

The factor $p^{-1}$ is the essential scaling. Using $H_p^{-1}$ instead would
produce an equivalent parametrization differing asymptotically only by the
constant factor $\alpha$. Our convention makes $s$ the expected exposed
degree on the atom side and is convenient for the tensor application below.

Let $M_p^{\mathrm{RE}}(s)$ denote the number of edges accepted by the time
all edges with priority at most $s/p$ have been processed. Since each
accepted edge matches exactly one sector vertex,
\[
    1-\frac{M_p^{\mathrm{RE}}(s)}{p}
\]
is the fraction of sector vertices that remain unmatched at exposure $s$.
The large-system limit below identifies the deterministic limit of this
quantity.

The limiting matching trajectory is stated in
Proposition~\ref{prop:random-edge-limit}. We write
$X_p\xrightarrow{\mathbb P}X$ for convergence in probability: for every
$\varepsilon>0$,
$\mathbb P(|X_p-X|>\varepsilon)\to0$ as $p\to\infty$.
The formula for $\ell_\alpha(s)$ in
Proposition~\ref{prop:random-edge-limit} is the sparse bipartite Random Edge
asymptotic derived by \citet[Appendix~B.1]{arnosti2022greedy}; we use this
known random-graph limit as a consequence of the tensor-dynamics reduction,
rather than as a new combinatorial result. Related random-order greedy
matching results include \citet{dyer1993average}, while the
differential-equation method used in the proof below is part of the general
framework developed by
\citet{wormald1995differential,wormald1999differential}.

We now give a self-contained proof of
Proposition~\ref{prop:random-edge-limit}. Besides recovering the known
closed form, the argument directly controls the matching trajectory on every
fixed exposure interval, which is the form needed below.

\begin{proof}[Proof of Proposition~\ref{prop:random-edge-limit}]
Write
\[
    \alpha_p:=\frac{H_p}{p},
    \qquad
    \alpha_p\longrightarrow\alpha.
\]
It is enough to prove uniform convergence on an arbitrary fixed interval
$[0,S]$.

\paragraph{Step 1: the random-order process.}
First ignore the numerical values of the edge priorities and retain only their
relative ordering. Since the priorities are i.i.d.\ and continuous, this
ordering is a uniformly random permutation of the $H_pp$ edges of
$K_{H_p,p}$.

Let $\widetilde M_{p,n}$ denote the number of edges accepted after the first
$n$ edges of this random ordering have been processed, and let
$\mathcal F_{p,n}$ be the history up to that point. If
$\widetilde M_{p,n}=m$, then $H_p-m$ atom vertices and $p-m$ sector vertices
remain unmatched. Moreover, every edge joining two currently unmatched
vertices is still unprocessed: had such an edge appeared earlier, it would
have been accepted when processed. Hence exactly
$(H_p-m)(p-m)$ of the $H_pp-n$ unprocessed edges join two unmatched
vertices. Therefore, if $J_{p,n+1}$ is the indicator that the next edge is
accepted,
\begin{equation}
    \mathbb P\!\left(
        J_{p,n+1}=1
        \,\middle|\,
        \mathcal F_{p,n}
    \right)
    =
    \frac{
        (H_p-\widetilde M_{p,n})
        (p-\widetilde M_{p,n})
    }{
        H_pp-n
    }.
    \label{eq:a5-random-edge-exact-drift}
\end{equation}

Set
$m_{p,n}:=\widetilde M_{p,n}/p$ and
$s_n:=n/H_p$. Then
\eqref{eq:a5-random-edge-exact-drift} can be written as
\begin{equation}
    \mathbb E\!\left[
        m_{p,n+1}-m_{p,n}
        \,\middle|\,
        \mathcal F_{p,n}
    \right]
    =
    \frac{1}{H_p}
    f_p(s_n,m_{p,n}),
    \qquad
    f_p(s,m)
    :=
    \frac{
        (\alpha_p-m)(1-m)
    }{
        1-s/p
    }.
    \label{eq:a5-random-edge-scaled-drift}
\end{equation}
For every fixed $S<\infty$, as $p\to\infty$,
\begin{equation}
    f_p(s,m)
    \longrightarrow
    f(m):=(\alpha-m)(1-m)
    \label{eq:a5-random-edge-drift-limit}
\end{equation}
uniformly for $0\leq s\leq S$ and for all values of $m$ attainable by the
process.

\paragraph{Step 2: concentration around the fluid equation.}
Define the martingale
\begin{equation}
    Z_{p,n}
    :=
    \frac1p
    \sum_{j=0}^{n-1}
    \left(
        J_{p,j+1}
        -
        \mathbb E[
            J_{p,j+1}\mid\mathcal F_{p,j}
        ]
    \right).
    \label{eq:a5-random-edge-martingale}
\end{equation}
Its increments have absolute value at most $1/p$. Since
$H_p/p\to\alpha$, Azuma--Hoeffding gives, for every $\varepsilon>0$,
\begin{equation}
    \mathbb P\!\left(
        \max_{0\leq n\leq H_pS}
        |Z_{p,n}|>\varepsilon
    \right)
    \leq
    2\exp\!\left(
        -\frac{\varepsilon^2p^2}{2H_pS}
    \right)
    \longrightarrow0.
    \label{eq:a5-random-edge-martingale-concentration}
\end{equation}

Define the piecewise-constant interpolation
\begin{equation}
    \widehat m_p(s)
    :=
    \frac{
        \widetilde M_{p,\lfloor H_ps\rfloor}
    }{p}.
    \label{eq:a5-random-edge-interpolation}
\end{equation}
Summing \eqref{eq:a5-random-edge-scaled-drift} and using
\eqref{eq:a5-random-edge-martingale-concentration}, together with the uniform
convergence \eqref{eq:a5-random-edge-drift-limit}, gives
\begin{equation}
    \sup_{0\leq s\leq S}
    \left|
        \widehat m_p(s)
        -
        \int_0^s
        f\bigl(\widehat m_p(r)\bigr)\,dr
    \right|
    \xrightarrow{\mathbb P}
    0.
    \label{eq:a5-random-edge-asymptotic-integral-equation}
\end{equation}
Indeed, apart from the martingale term, the sum is a Riemann sum with mesh
$H_p^{-1}$; on $[0,S]$, the functions $f_p$ are uniformly bounded and
Lipschitz in their second argument.

Let $m$ be the solution of
\begin{equation}
    m'(s)
    =
    (\alpha-m(s))(1-m(s)),
    \qquad
    m(0)=0.
    \label{eq:a5-random-edge-matching-ode}
\end{equation}
The vector field in \eqref{eq:a5-random-edge-matching-ode} is Lipschitz on
the bounded interval containing all possible matching fractions.
Subtracting the integral form of
\eqref{eq:a5-random-edge-matching-ode} from
\eqref{eq:a5-random-edge-asymptotic-integral-equation} and applying
Gronwall's inequality therefore yields
\begin{equation}
    \sup_{0\leq s\leq S}
    \left|
        \widehat m_p(s)-m(s)
    \right|
    \xrightarrow{\mathbb P}
    0.
    \label{eq:a5-random-edge-order-fluid-limit}
\end{equation}
This is the standard differential-equation argument in this particularly
simple one-dimensional setting; compare
\citet{wormald1995differential,wormald1999differential}.

\paragraph{Step 3: from the number of processed edges to edge exposure.}
We now return to the actual uniform priorities used in the definition of
$M_p^{\mathrm{RE}}(s)$. Let
\begin{equation}
    \mathcal N_p(s)
    :=
    \#\{
        \text{edges with priority at most }s/p
    \}.
    \label{eq:a5-random-edge-exposure-count}
\end{equation}
For fixed $s$,
$\mathcal N_p(s)$ is binomial with parameters $H_pp$ and $s/p$, so
$\mathbb E\mathcal N_p(s)=H_ps$. Moreover,
\begin{equation}
    \sup_{0\leq s\leq S}
    \left|
        \frac{\mathcal N_p(s)}{H_p}-s
    \right|
    \xrightarrow{\mathbb P}
    0.
    \label{eq:a5-random-edge-exposure-concentration}
\end{equation}
For completeness, this uniform statement follows by fixing a finite mesh
$0,\delta,2\delta,\ldots,S$, applying a binomial concentration bound at
each mesh point, and then using monotonicity of $\mathcal N_p(s)$ between
successive mesh points. First letting $p\to\infty$ and then
$\delta\downarrow0$ proves
\eqref{eq:a5-random-edge-exposure-concentration}.

Processing all edges with priority at most $s/p$ is the same as processing
the first $\mathcal N_p(s)$ edges of the random priority ordering. Hence,
pathwise,
\begin{equation}
    \frac{M_p^{\mathrm{RE}}(s)}{p}
    =
    \widehat m_p\!\left(
        \frac{\mathcal N_p(s)}{H_p}
    \right).
    \label{eq:a5-random-edge-time-change}
\end{equation}
Combining
\eqref{eq:a5-random-edge-order-fluid-limit},
\eqref{eq:a5-random-edge-exposure-concentration}, and the uniform continuity
of $m$ on compact intervals gives
\begin{equation}
    \sup_{0\leq s\leq S}
    \left|
        \frac{M_p^{\mathrm{RE}}(s)}{p}
        -
        m(s)
    \right|
    \xrightarrow{\mathbb P}
    0.
    \label{eq:a5-random-edge-exposure-fluid-limit}
\end{equation}

\paragraph{Step 4: solution of the limiting equation.}
Let
\begin{equation}
    \ell_\alpha(s)
    :=
    1-m(s),
    \label{eq:a5-random-edge-unmatched-fraction}
\end{equation}
so that $\ell_\alpha$ is the limiting fraction of unmatched sector
vertices. Equation~\eqref{eq:a5-random-edge-matching-ode}, which describes
the limiting matching fraction, becomes
\begin{equation}
    \ell_\alpha'(s)
    =
    -\ell_\alpha(s)
    \bigl(
        \ell_\alpha(s)+\alpha-1
    \bigr),
    \qquad
    \ell_\alpha(0)=1.
    \label{eq:a5-random-edge-unmatched-ode}
\end{equation}
For $\alpha=1$, this gives
$\ell_1'(s)=-\ell_1(s)^2$, and therefore
\begin{equation}
    \ell_1(s)
    =
    \frac{1}{1+s}.
    \label{eq:a5-random-edge-balanced-solution}
\end{equation}
If $\alpha\neq1$, separation of variables in
\eqref{eq:a5-random-edge-unmatched-ode} gives
\begin{equation}
    \frac{
        \ell_\alpha(s)
    }{
        \ell_\alpha(s)+\alpha-1
    }
    =
    \frac1\alpha
    e^{-(\alpha-1)s},
    \label{eq:a5-random-edge-separated-solution}
\end{equation}
and hence
\begin{equation}
    \ell_\alpha(s)
    =
    \frac{
        (\alpha-1)e^{-(\alpha-1)s}
    }{
        \alpha-e^{-(\alpha-1)s}
    }.
    \label{eq:a5-random-edge-explicit-solution}
\end{equation}
Together with
\eqref{eq:a5-random-edge-exposure-fluid-limit}, this is precisely
\eqref{eq:random-edge-limit}. Since $S<\infty$ was arbitrary, the convergence
is locally uniform for $s\geq0$. Together with
\eqref{eq:a5-random-edge-exposure-fluid-limit}, this proves
\eqref{eq:random-edge-limit} and hence
Proposition~\ref{prop:random-edge-limit}.
\end{proof}

\subsection{From Random Edge matching to the equal-amplitude sector tensor limit}
\label{app:equal-sector-large-p}

We now prove Theorem~\ref{thm:universal-large-p}. There is one minor
technical point in passing from the fixed-dimensional greedy limit to the
large-system trajectory. Theorem~\ref{thm:greedy-matching-limit}, specifically
the loss convergence in \eqref{eq:greedy-loss-limit}, controls GF away from
the accepted activation times, whereas
Theorem~\ref{thm:universal-large-p} takes a supremum over an entire exposure
interval, including the fast transition layers. For equal-amplitude sectors,
each jump of the normalized greedy loss has size exactly $1/p$. Together with
monotonicity of the GF loss, this makes the transition layers negligible as
$p\to\infty$.

For the comparison with GF, we realize the Random Edge process on the same
probability space as the Gaussian initialization. For each candidate
atom--sector pair $(k,i)$, let
\[
    \xi_{ki}:=\tau_\nu(\mathbf g_{ki}).
\]
and assign the corresponding edge the priority $F_\nu(\xi_{ki})$, where
$F_\nu$ is the CDF of the scalar activation clock. Since $F_\nu$ is
continuous, the probability-integral transform implies that these priorities
are i.i.d.\ uniform on $[0,1]$. Thus they generate exactly the Random Edge
process of Proposition~\ref{prop:random-edge-limit}, but with its randomness
coupled to the same microscopic initialization that determines the GF
trajectory.

Since $F_\nu$ is nondecreasing and the transformed priorities are almost
surely distinct, their ordering agrees almost surely with the ordering of the
original clock variables $\xi_{ki}$. Hence the two greedy processes accept
exactly the same sequence of atom--sector pairs. Throughout this subsection,
$M_p^{\mathrm{RE}}(s)$ denotes this coupled realization; no additional
randomness is introduced.

For fixed $p$, define the finite-$p$ Random Edge loss trajectory
\begin{equation}
    \ell_p^{\mathrm{RE}}(s)
    :=
    1-\frac{M_p^{\mathrm{RE}}(s)}{p},
    \label{eq:a5-finite-random-edge-loss}
\end{equation}
and define the normalized GF loss in the same exposure parametrization by
\begin{equation}
    \ell_{p,\sigma}^{\mathrm{GF}}(s)
    :=
    \frac{
        2L\!\left(
            U_{p,\sigma}(t_{p,\sigma}(s))
        \right)
    }{
        p\lambda_p^2
    }.
    \label{eq:a5-finite-gf-exposure-loss}
\end{equation}
Here $t_{p,\sigma}(s)$ is the exposure-to-training-time conversion
defined in \eqref{eq:exposure-training-time}. In what follows we will use  the inverse $F_\nu^{-1}$, which is well-defined on $[0,1)$ by Proposition~\ref{prop:clock-atomless}.

The following observation bridges the fast transition layers.

\begin{lemma}[Uniform finite-$p$ comparison]
\label{lem:a5-finite-p-comparison}
Fix $p$ and $S<\infty$, with $S<p$, and couple the Random Edge process to
the Gaussian initialization as above. Then, for almost every realization of
that initialization,
\begin{equation}
    \limsup_{\sigma\downarrow0}
    \sup_{0\leq s\leq S}
    \left|
        \ell_{p,\sigma}^{\mathrm{GF}}(s)
        -
        \ell_p^{\mathrm{RE}}(s)
    \right|
    \leq
    \frac1p.
    \label{eq:a5-finite-p-comparison}
\end{equation}
\end{lemma}

\begin{proof}
Under the coupling above, the Random Edge priorities and the tensor activation
clocks have the same ordering. At an exposure value $s$ that is not one of
the finitely many edge priorities, processing all priorities up to $s/p$
therefore selects exactly the same greedy edges as processing the tensor
activation clocks up to
\begin{equation}
    \theta_p(s)
    :=
    \frac{1}{\lambda_p}
    F_\nu^{-1}\!\left(\frac{s}{p}\right).
    \label{eq:a5-exposure-slow-time}
\end{equation}
Consequently, away from the accepted exposure levels, the normalized greedy
loss associated with the tensor clocks is exactly
$\ell_p^{\mathrm{RE}}(s)$.

By Theorem~\ref{thm:greedy-matching-limit}, specifically
\eqref{eq:greedy-loss-limit}, for every $s$ that is not an accepted exposure
level,
\begin{equation}
    \ell_{p,\sigma}^{\mathrm{GF}}(s)
    \longrightarrow
    \ell_p^{\mathrm{RE}}(s)
    \qquad
    \text{as }\sigma\downarrow0.
    \label{eq:a5-pointwise-exposure-convergence}
\end{equation}

We now use monotonicity to extend this control through the transition layers.
Along GF,
\begin{equation}
    \frac{d}{dt}L(U_{p,\sigma}(t))
    =
    -
    \sum_{\substack{k\in[H_p]\\a\in[\nu]}}
    \left\|
        \nabla_{\mathbf u_k^{(a)}}L(U_{p,\sigma}(t))
    \right\|^2
    \leq0.
    \label{eq:a5-gf-loss-monotonicity}
\end{equation}
Since $t_{p,\sigma}(s)$ is nondecreasing in $s$,
$\ell_{p,\sigma}^{\mathrm{GF}}(s)$ is nonincreasing in $s$.
The finite Random Edge trajectory
$\ell_p^{\mathrm{RE}}(s)$ is also nonincreasing, and every one of its jumps
has size exactly $1/p$, because each accepted edge matches exactly one
additional sector.

There are only finitely many accepted exposure levels in $[0,S]$. Choose
a finite mesh
\begin{equation}
    0=s_0<s_1<\cdots<s_N=S
    \label{eq:a5-exposure-mesh}
\end{equation}
such that no mesh point is an accepted exposure level and every interval
$[s_j,s_{j+1}]$ contains at most one accepted exposure level. For fixed
$S$, the endpoints $0$ and $S$ are themselves almost surely not accepted
exposure levels; if necessary, the same argument may be applied on a slightly
larger interval and then restricted to $[0,S]$.

By \eqref{eq:a5-pointwise-exposure-convergence}, at the finitely many mesh
points,
\begin{equation}
    \max_{0\leq j\leq N}
    \left|
        \ell_{p,\sigma}^{\mathrm{GF}}(s_j)
        -
        \ell_p^{\mathrm{RE}}(s_j)
    \right|
    \longrightarrow0.
    \label{eq:a5-mesh-point-convergence}
\end{equation}

Now fix $s\in[s_j,s_{j+1}]$. Monotonicity gives
\begin{equation}
    \ell_{p,\sigma}^{\mathrm{GF}}(s_j)
    \geq
    \ell_{p,\sigma}^{\mathrm{GF}}(s)
    \geq
    \ell_{p,\sigma}^{\mathrm{GF}}(s_{j+1}),
    \label{eq:a5-gf-monotone-sandwich}
\end{equation}
and similarly
\begin{equation}
    \ell_p^{\mathrm{RE}}(s_j)
    \geq
    \ell_p^{\mathrm{RE}}(s)
    \geq
    \ell_p^{\mathrm{RE}}(s_{j+1}).
    \label{eq:a5-re-monotone-sandwich}
\end{equation}
Because $[s_j,s_{j+1}]$ contains at most one accepted exposure level,
\begin{equation}
    0
    \leq
    \ell_p^{\mathrm{RE}}(s_j)
    -
    \ell_p^{\mathrm{RE}}(s_{j+1})
    \leq
    \frac1p.
    \label{eq:a5-one-jump-mesh}
\end{equation}
Using
\eqref{eq:a5-gf-monotone-sandwich}--\eqref{eq:a5-one-jump-mesh}, we obtain
\begin{equation}
    \left|
        \ell_{p,\sigma}^{\mathrm{GF}}(s)
        -
        \ell_p^{\mathrm{RE}}(s)
    \right|
    \leq
    \frac1p
    +
    \max\left\{
        \left|
            \ell_{p,\sigma}^{\mathrm{GF}}(s_j)
            -
            \ell_p^{\mathrm{RE}}(s_j)
        \right|,
        \left|
            \ell_{p,\sigma}^{\mathrm{GF}}(s_{j+1})
            -
            \ell_p^{\mathrm{RE}}(s_{j+1})
        \right|
    \right\}.
    \label{eq:a5-transition-sandwich-bound}
\end{equation}
Taking the supremum over $s\in[0,S]$ and then
$\limsup_{\sigma\downarrow0}$, the mesh-point convergence
\eqref{eq:a5-mesh-point-convergence} yields
\eqref{eq:a5-finite-p-comparison}.
\end{proof}

Lemma~\ref{lem:a5-finite-p-comparison} is the only additional ingredient
needed to combine the fixed-$p$ tensor limit with the large-$p$ Random Edge
limit. Notice that the error $1/p$ does not estimate the width of a fast
transition in time. Rather, it bounds its possible effect on the
\emph{normalized loss}, because a single equal-amplitude sector contributes
exactly $1/p$ of the total target energy.

\begin{proof}[Proof of Theorem~\ref{thm:universal-large-p}]
Fix a compact set $K\subset[0,\infty)$ and $\varepsilon>0$. Choose
$S<\infty$ such that $K\subset[0,S]$. Since the theorem takes
$p\to\infty$, we may restrict to sufficiently large $p$ so that $S<p$.

By the triangle inequality,
\begin{equation}
    \begin{aligned}
    \sup_{s\in K}
    \left|
        \ell_{p,\sigma}^{\mathrm{GF}}(s)
        -
        \ell_\alpha(s)
    \right|
    &\leq
    \sup_{0\leq s\leq S}
    \left|
        \ell_{p,\sigma}^{\mathrm{GF}}(s)
        -
        \ell_p^{\mathrm{RE}}(s)
    \right|
    \\
    &\quad
    +
    \sup_{0\leq s\leq S}
    \left|
        \ell_p^{\mathrm{RE}}(s)
        -
        \ell_\alpha(s)
    \right|.
    \end{aligned}
    \label{eq:a5-large-p-triangle}
\end{equation}

For sufficiently large $p$, $1/p<\varepsilon/3$. Then
Lemma~\ref{lem:a5-finite-p-comparison}, specifically the finite-$p$
comparison \eqref{eq:a5-finite-p-comparison}, implies
\begin{equation}
    \lim_{\sigma\downarrow0}
    \mathbb P\!\left(
        \sup_{0\leq s\leq S}
        \left|
            \ell_{p,\sigma}^{\mathrm{GF}}(s)
            -
            \ell_p^{\mathrm{RE}}(s)
        \right|
        >
        \frac{\varepsilon}{3}
    \right)
    =
    0.
    \label{eq:a5-finite-p-probability-control}
\end{equation}
On the other hand,
Proposition~\ref{prop:random-edge-limit}, namely the large-system Random Edge
limit \eqref{eq:random-edge-limit}, gives
\begin{equation}
    \mathbb P\!\left(
        \sup_{0\leq s\leq S}
        \left|
            \ell_p^{\mathrm{RE}}(s)
            -
            \ell_\alpha(s)
        \right|
        >
        \frac{2\varepsilon}{3}
    \right)
    \longrightarrow0
    \qquad
    \text{as }p\to\infty.
    \label{eq:a5-random-edge-probability-control}
\end{equation}
Combining
\eqref{eq:a5-large-p-triangle}--\eqref{eq:a5-random-edge-probability-control}
therefore yields
\begin{equation}
    \lim_{p\to\infty}
    \limsup_{\sigma\downarrow0}
    \mathbb P\!\left(
        \sup_{s\in K}
        \left|
            \ell_{p,\sigma}^{\mathrm{GF}}(s)
            -
            \ell_\alpha(s)
        \right|
        >
        \varepsilon
    \right)
    =
    0.
    \label{eq:a5-equal-sector-final-limit}
\end{equation}
Finally, substituting the definition
\eqref{eq:a5-finite-gf-exposure-loss} of
$\ell_{p,\sigma}^{\mathrm{GF}}$ gives exactly
\eqref{eq:equal-diagonal-large-system-limit}, completing the proof.
\end{proof}

\subsection{Lower tail of the complex cubic activation clock}
\label{app:cubic-tail}

We now analyze the lower tail of the scalar activation clock for
$\nu=3$ and complex Gaussian initialization. Let
\begin{equation}
    F_3(x)
    :=
    \mathbb P\!\left(
        \tau_3(\mathbf g)\leq x
    \right),
    \qquad
    \mathbf g\sim\mathcal N_{\mathbb C}(0,I_3).
    \label{eq:a5-cubic-clock-cdf}
\end{equation}
The goal of this subsection is to prove the asymptotic
\begin{proposition}[Lower tail of the complex cubic clock]
\label{prop:a5-cubic-clock-tail}
As $x\downarrow0$,
\begin{equation}
    F_3(x)
    =
    A_3\,\frac{\sqrt3}{x}
    \exp\!\left(-\frac{3}{x^2}\right)
    \bigl(1+o(1)\bigr),
    \qquad
    A_3
    =
    \frac{5\sqrt{21\pi}}{27}.
    \label{eq:a5-cubic-clock-tail}
\end{equation}
\end{proposition}

The exponential factor comes from the most favorable direction of the
microscopic initialization, while the factor $x^{-1}$ and the constant
$A_3$ result from Gaussian fluctuations around this direction.

\paragraph{Radial reduction.}
Write the complex Gaussian vector in polar form
\begin{equation}
    \mathbf g=R\boldsymbol{\omega},
    \qquad
    R=\|\mathbf g\|,
    \qquad
    \|\boldsymbol{\omega}\|=1.
    \label{eq:a5-cubic-radial-decomposition}
\end{equation}
For a standard complex Gaussian vector, $R$ and $\boldsymbol{\omega}$ are independent,
$\omega$ is uniform on the unit sphere in $\mathbb C^3$, and
$R^2$ has the Gamma distribution with shape $3$ and scale $1$. Hence
\begin{equation}
    \mathbb P(R\geq r)
    =
    e^{-r^2}
    \left(
        1+r^2+\frac{r^4}{2}
    \right).
    \label{eq:a5-cubic-radial-tail}
\end{equation}
By the homogeneity
\eqref{eq:clock-homogeneity}, specialized to $\nu=3$,
$\tau_3(R\boldsymbol{\omega})=R^{-1}\tau_3(\boldsymbol{\omega})$. Conditioning on $\boldsymbol{\omega}$ therefore
gives the exact representation
\begin{equation}
    F_3(x)
    =
    \mathbb E_{\boldsymbol{\omega}}
    \left[
        \exp\!\left(
            -\frac{\tau_3(\boldsymbol{\omega})^2}{x^2}
        \right)
        \left(
            1+
            \frac{\tau_3(\boldsymbol{\omega})^2}{x^2}
            +
            \frac{\tau_3(\boldsymbol{\omega})^4}{2x^4}
        \right)
    \right].
    \label{eq:a5-cubic-tail-radial-representation}
\end{equation}
Thus the lower-tail problem reduces to locating and expanding the minimum of
$\tau_3(\boldsymbol{\omega})$ on the unit sphere.

\paragraph{The optimal direction.}
For a unit vector $\boldsymbol{\omega}\in\mathbb C^3$, write
\begin{equation}
    \rho_a:=|{\omega}^{(a)}|^2,
    \qquad
    \rho_1+\rho_2+\rho_3=1,
    \qquad
    \psi
    :=
    \arg\!\left(
        {\omega}^{(1)}{\omega}^{(2)}{\omega}^{(3)}
    \right).
    \label{eq:a5-cubic-direction-variables}
\end{equation}
Only the three magnitudes and the total product phase $\psi$ affect the
scalar clock.

Recall from
\eqref{eq:q-equation} that, with
$B=\operatorname{Im}
({\omega}^{(1)}{\omega}^{(2)}\omega^{(3)})$,
\begin{equation}
    D_{\boldsymbol{\omega}}(q)
    =
    (q+\rho_1)(q+\rho_2)(q+\rho_3)-B^2.
    \label{eq:a5-cubic-D}
\end{equation}
For every $q\geq0$, the arithmetic--geometric mean inequality gives
\begin{equation}
    D_{\boldsymbol{\omega}}(q)
    \leq
    (q+\rho_1)(q+\rho_2)(q+\rho_3)
    \leq
    \left(q+\frac13\right)^3.
    \label{eq:a5-cubic-D-upper-bound}
\end{equation}
If the initial real part of the product is nonnegative, the clock formula
\eqref{eq:clock-integral-positive} therefore gives
\begin{equation}
    \tau_3(\boldsymbol{\omega})
    \geq
    \frac12
    \int_0^\infty
    \left(q+\frac13\right)^{-3/2}\,dq
    =
    \sqrt3.
    \label{eq:a5-cubic-clock-minimum-bound}
\end{equation}
If the initial real part is negative,
\eqref{eq:clock-integral-negative} contains an additional positive
turning-point contribution, so the same lower bound holds strictly.

Equality in \eqref{eq:a5-cubic-clock-minimum-bound} requires equality in
\eqref{eq:a5-cubic-D-upper-bound} for all $q\geq0$. Hence
\begin{equation}
    \rho_1=\rho_2=\rho_3=\frac13,
    \qquad
    B=0,
    \qquad
    \operatorname{Re}
    \bigl(
        \omega^{(1)}\omega^{(2)}\omega^{(3)}
    \bigr)>0.
    \label{eq:a5-cubic-minimizers}
\end{equation}
Equivalently, the minimizing directions have equal magnitudes
$|\omega^{(a)}|=3^{-1/2}$ and total phase
$\psi=0\pmod{2\pi}$. Thus
\begin{equation}
    \min_{\|\boldsymbol{\omega}\|=1}\tau_3(\boldsymbol{\omega})=\sqrt3.
    \label{eq:a5-cubic-minimum}
\end{equation}

\paragraph{Quadratic expansion around the minimizing directions.}
Set
\begin{equation}
    \rho_a
    =
    \frac13+\delta_a,
    \qquad
    \delta_1+\delta_2+\delta_3=0,
    \label{eq:a5-cubic-delta}
\end{equation}
and take $\psi$ to be the representative of the total phase near zero.
For $r=q+1/3$,
\begin{equation}
    \prod_{a=1}^3(q+\rho_a)
    =
    r^3
    -
    \frac r2
    \sum_{a=1}^3\delta_a^2
    +
    \delta_1\delta_2\delta_3,
    \label{eq:a5-cubic-product-expansion}
\end{equation}
while
\begin{equation}
    B^2
    =
    \rho_1\rho_2\rho_3\sin^2\psi
    =
    \frac{\psi^2}{27}
    +
    o\!\left(
        \sum_{a=1}^3\delta_a^2+\psi^2
    \right).
    \label{eq:a5-cubic-B-expansion}
\end{equation}
Consequently,
\begin{equation}
    D_{\boldsymbol{\omega}}(q)
    =
    r^3
    -
    \frac r2
    \sum_{a=1}^3\delta_a^2
    -
    \frac{\psi^2}{27}
    +
    o\!\left(
        \sum_{a=1}^3\delta_a^2+\psi^2
    \right),
    \label{eq:a5-cubic-D-expansion}
\end{equation}
uniformly on compact $q$-intervals, with the resulting integral remainder
also uniform.

In a neighborhood of the minimizers the product has positive real part, so
we may use \eqref{eq:clock-integral-positive}. Expanding
$D_{\boldsymbol{\omega}}(q)^{-1/2}$ to second order and integrating gives
\begin{equation}
    \tau_3(\boldsymbol{\omega})
    =
    \sqrt3
    +
    \frac{9\sqrt3}{20}
    \sum_{a=1}^3\delta_a^2
    +
    \frac{\sqrt3}{14}\psi^2
    +
    o\!\left(
        \sum_{a=1}^3\delta_a^2+\psi^2
    \right).
    \label{eq:a5-cubic-clock-local-expansion}
\end{equation}
Indeed, the two required integrals are
\begin{equation}
    \int_{1/3}^\infty r^{-7/2}\,dr
    =
    \frac25\,3^{5/2},
    \qquad
    \int_{1/3}^\infty r^{-9/2}\,dr
    =
    \frac27\,3^{7/2}.
    \label{eq:a5-cubic-expansion-integrals}
\end{equation}
Squaring \eqref{eq:a5-cubic-clock-local-expansion} yields the form convenient
for Laplace asymptotics:
\begin{equation}
    \tau_3(\boldsymbol{\omega})^2
    =
    3
    +
    \frac{27}{10}
    \sum_{a=1}^3\delta_a^2
    +
    \frac37\psi^2
    +
    o\!\left(
        \sum_{a=1}^3\delta_a^2+\psi^2
    \right).
    \label{eq:a5-cubic-clock-square-expansion}
\end{equation}

\begin{proof}[Proof of Proposition~\ref{prop:a5-cubic-clock-tail}]
For a uniform direction $\boldsymbol{\omega}$ on the unit sphere in $\mathbb C^3$, the
squared magnitudes
$(\rho_1,\rho_2,\rho_3)$ have the
$\operatorname{Dirichlet}(1,1,1)$ distribution. Thus their density on
\begin{equation}
    \Delta_2
    :=
    \left\{
        (\rho_1,\rho_2):
        \rho_1>0,\ \rho_2>0,\ \rho_1+\rho_2<1
    \right\}
    \label{eq:a5-cubic-simplex}
\end{equation}
is the constant $2$. The coordinate phases are independent and uniform, so
the total product phase $\psi$ is uniform on $[0,2\pi)$ and independent of
the magnitudes. Hence the directional expectation in
\eqref{eq:a5-cubic-tail-radial-representation} is integration against
\begin{equation}
    \frac1\pi\,
    d\rho_1\,d\rho_2\,d\psi.
    \label{eq:a5-cubic-direction-measure}
\end{equation}

By \eqref{eq:a5-cubic-minimum}, the exponent
$\tau_3(\boldsymbol{\omega})^2$ has its minimum value $3$ precisely at
$\rho_1=\rho_2=\rho_3=1/3$ and
$\psi=0\pmod{2\pi}$. The integral formulas for the clock imply continuity
near this minimizing set, and the strict inequality in
\eqref{eq:a5-cubic-clock-minimum-bound} shows that the contribution outside
any fixed neighborhood of it is exponentially smaller than
$e^{-3/x^2}$. It therefore suffices to use the local expansion
\eqref{eq:a5-cubic-clock-square-expansion}.

Write
$\delta_3=-\delta_1-\delta_2$. The quadratic part in
\eqref{eq:a5-cubic-clock-square-expansion} is
\begin{equation}
    Q(\delta_1,\delta_2,\psi)
    :=
    \frac{27}{10}
    \left(
        \delta_1^2+\delta_2^2+
        (\delta_1+\delta_2)^2
    \right)
    +
    \frac37\psi^2.
    \label{eq:a5-cubic-quadratic-form}
\end{equation}
Under the rescaling
$\delta_1=xu_1$, $\delta_2=xu_2$, $\psi=xv$, the directional volume
element contributes a factor $x^3$. Moreover, in
\eqref{eq:a5-cubic-tail-radial-representation}, the dominant polynomial
term is
\begin{equation}
    \frac{\tau_3(\boldsymbol{\omega})^4}{2x^4}
    =
    \frac{9}{2x^4}
    \bigl(1+o(1)\bigr)
    \label{eq:a5-cubic-dominant-radial-factor}
\end{equation}
on the $O(x)$ neighborhood contributing to the Laplace integral. The other
two terms in
\eqref{eq:a5-cubic-tail-radial-representation} are lower by powers of $x$.

It follows by the standard Laplace rescaling and dominated convergence that
\begin{equation}
    F_3(x)
    \sim
    \frac{9}{2\pi x}
    e^{-3/x^2}
    \int_{\mathbb R^3}
    e^{-Q(u_1,u_2,v)}
    \,du_1\,du_2\,dv.
    \label{eq:a5-cubic-tail-laplace}
\end{equation}
The Gaussian integral factors. Since
\begin{equation}
    u_1^2+u_2^2+(u_1+u_2)^2
    =
    \begin{pmatrix}u_1&u_2\end{pmatrix}
    \begin{pmatrix}2&1\\1&2\end{pmatrix}
    \begin{pmatrix}u_1\\u_2\end{pmatrix},
    \label{eq:a5-cubic-magnitude-matrix}
\end{equation}
and the matrix has determinant $3$,
\begin{equation}
    \int_{\mathbb R^2}
    \exp\!\left[
        -\frac{27}{10}
        \bigl(
            u_1^2+u_2^2+(u_1+u_2)^2
        \bigr)
    \right]
    du_1\,du_2
    =
    \frac{10\pi}{27\sqrt3}.
    \label{eq:a5-cubic-magnitude-gaussian}
\end{equation}
Similarly,
\begin{equation}
    \int_{\mathbb R}
    e^{-3v^2/7}\,dv
    =
    \sqrt{\frac{7\pi}{3}}.
    \label{eq:a5-cubic-phase-gaussian}
\end{equation}
Substituting
\eqref{eq:a5-cubic-magnitude-gaussian} and
\eqref{eq:a5-cubic-phase-gaussian} into
\eqref{eq:a5-cubic-tail-laplace} gives
\begin{equation}
    F_3(x)
    \sim
    \frac{5\sqrt{7\pi}}{9x}
    e^{-3/x^2}.
    \label{eq:a5-cubic-tail-prefactor}
\end{equation}
Finally,
\begin{equation}
    \frac{5\sqrt{7\pi}}9
    =
    \sqrt3\,
    \frac{5\sqrt{21\pi}}{27}
    =
    \sqrt3\,A_3,
    \label{eq:a5-cubic-A3-identity}
\end{equation}
which proves \eqref{eq:a5-cubic-clock-tail}.
\end{proof}

\subsection{Asymptotic inversion and physical training time}
\label{app:addition-time-inversion}

We now use the cubic clock tail from
Proposition~\ref{prop:a5-cubic-clock-tail} to prove
Theorem~\ref{thm:explicit-addition-loss}. For complex modular addition,
$\nu=3$ and $\lambda_p=\sqrt p$, so the exact edge exposure corresponding
to physical training time $t$ is
\[
    s_{p,\sigma}(t)
    =
    pF_3(\sqrt p\,\sigma t).
\]
The first step is to asymptotically invert this relation for fixed
$O(1)$ exposure levels.

For $s>0$, set
\begin{equation}
    \beta_p(s)
    :=
    \log p
    +\frac12\log\log p
    +\log A_3
    -\log s .
    \label{eq:a5-beta-definition}
\end{equation}
Then the explicit time from
\eqref{eq:addition-explicit-training-time} can be written as
\begin{equation}
    t^{\mathrm{as}}_{p,\sigma}(s)
    =
    \frac{\sqrt3}{
        \sigma\sqrt p\,\sqrt{\beta_p(s)}
    }.
    \label{eq:a5-explicit-time-beta}
\end{equation}

\begin{lemma}[Uniform asymptotic inversion]
\label{lem:a5-asymptotic-inversion}
For every compact $K\subset(0,\infty)$,
\begin{equation}
    \sup_{s\in K}
    \left|
        s_{p,\sigma}\!\left(
            t^{\mathrm{as}}_{p,\sigma}(s)
        \right)
        -s
    \right|
    \longrightarrow0
    \qquad
    \text{as }p\to\infty.
    \label{eq:a5-uniform-asymptotic-inversion}
\end{equation}
The expression on the left is independent of $\sigma$.
\end{lemma}

\begin{proof}
For $\beta_p(s)$ defined in \eqref{eq:a5-beta-definition}, 
the explicit time
\eqref{eq:addition-explicit-training-time} satisfies
\begin{equation}
    x_p(s)
    :=
    \sqrt p\,\sigma
    t^{\mathrm{as}}_{p,\sigma}(s)
    =
    \frac{\sqrt3}{\sqrt{\beta_p(s)}}.
    \label{eq:a5-clock-argument}
\end{equation}
In particular, $x_p(s)$ is independent of $\sigma$.

If $K\subset(0,\infty)$ is compact, then
\begin{equation}
    \sup_{s\in K}
    \left|
        \frac{\beta_p(s)}{\log p}-1
    \right|
    \longrightarrow0,
    \label{eq:a5-beta-logp}
\end{equation}
because $\log s$ is bounded on $K$. Hence
$x_p(s)\to0$ uniformly for $s\in K$.

We may therefore apply the lower-tail asymptotic
\eqref{eq:a5-cubic-clock-tail} uniformly along $x_p(s)$. Thus
\begin{equation}
    F_3(x_p(s))
    =
    A_3\frac{\sqrt3}{x_p(s)}
    \exp\!\left(
        -\frac{3}{x_p(s)^2}
    \right)
    \bigl(1+r_p(s)\bigr),
    \label{eq:a5-tail-substitution}
\end{equation}
where
\begin{equation}
    \sup_{s\in K}|r_p(s)|
    \longrightarrow0.
    \label{eq:a5-tail-uniform-remainder}
\end{equation}
Using
\[
    \frac{\sqrt3}{x_p(s)}
    =
    \sqrt{\beta_p(s)},
    \qquad
    \frac{3}{x_p(s)^2}
    =
    \beta_p(s),
\]
we obtain
\begin{equation}
    pF_3(x_p(s))
    =
    pA_3\sqrt{\beta_p(s)}
    e^{-\beta_p(s)}
    \bigl(1+r_p(s)\bigr).
    \label{eq:a5-tail-evaluated}
\end{equation}
By the definition of $\beta_p(s)$,
\begin{equation}
    e^{-\beta_p(s)}
    =
    \frac{s}{
        pA_3\sqrt{\log p}
    }.
    \label{eq:a5-beta-exponential}
\end{equation}
Therefore
\begin{equation}
    pF_3(x_p(s))
    =
    s
    \sqrt{
        \frac{\beta_p(s)}{\log p}
    }
    \bigl(1+r_p(s)\bigr).
    \label{eq:a5-exposure-expansion}
\end{equation}

Since
\[
    s_{p,\sigma}\!\left(
        t^{\mathrm{as}}_{p,\sigma}(s)
    \right)
    =
    pF_3(x_p(s)),
\]
it follows from \eqref{eq:a5-exposure-expansion} that
\begin{equation}
    \begin{aligned}
    &
    \left|
        s_{p,\sigma}\!\left(
            t^{\mathrm{as}}_{p,\sigma}(s)
        \right)
        -s
    \right|
    \\
    &\qquad
    =
    s\left|
        \sqrt{
            \frac{\beta_p(s)}{\log p}
        }
        \bigl(1+r_p(s)\bigr)
        -1
    \right|.
    \end{aligned}
    \label{eq:a5-exposure-additive-error}
\end{equation}
Because $K$ is compact, $\sup_{s\in K}s<\infty$. Moreover,
\eqref{eq:a5-beta-logp} implies
\[
    \sup_{s\in K}
    \left|
        \sqrt{
            \frac{\beta_p(s)}{\log p}
        }
        -1
    \right|
    \longrightarrow0,
\]
while \eqref{eq:a5-tail-uniform-remainder} gives
$\sup_{s\in K}|r_p(s)|\to0$. Taking the supremum over $s\in K$ in
\eqref{eq:a5-exposure-additive-error} therefore yields
\eqref{eq:a5-uniform-asymptotic-inversion}.
\end{proof}

Lemma~\ref{lem:a5-asymptotic-inversion} proves
\eqref{eq:addition-asymptotic-inverse}. We next transfer the 
large-system loss trajectory from exact exposure to the explicit physical
time.

\begin{proof}[Proof of Theorem~\ref{thm:explicit-addition-loss}]
The explicit formula
\eqref{eq:addition-explicit-training-time} is exactly
\eqref{eq:a5-explicit-time-beta}, so it remains to prove the loss
convergence.

For notational convenience, define the actual exposure reached at the
proposed physical time by
\begin{equation}
    \widehat s_p(s)
    :=
    s_{p,\sigma}\!\left(
        t^{\mathrm{as}}_{p,\sigma}(s)
    \right).
    \label{eq:a5-induced-exposure}
\end{equation}
Although the notation on the right contains $\sigma$, the quantity
$\widehat s_p(s)$ is independent of $\sigma$, because
$\sqrt p\,\sigma t^{\mathrm{as}}_{p,\sigma}(s)=x_p(s)$.
By Lemma~\ref{lem:a5-asymptotic-inversion},
\begin{equation}
    \sup_{s\in K}
    \left|
        \widehat s_p(s)-s
    \right|
    \longrightarrow0
    \label{eq:a5-induced-exposure-convergence}
\end{equation}
for every compact $K\subset(0,\infty)$.

Fix such a compact set $K$. Choose another compact interval
$K'\subset(0,\infty)$ containing both $K$ and
$\widehat s_p(K)$ for all sufficiently large $p$. Since
$\lambda_p=\sqrt p$, the normalized loss at
$t^{\mathrm{as}}_{p,\sigma}(s)$ can be written as the normalized
equal-amplitude sector loss at exposure $\widehat s_p(s)$. Hence
\begin{equation}
    \begin{aligned}
    &\sup_{s\in K}
    \left|
        \frac{
            2L\!\left(
                U_{p,\sigma}\!\left(
                    t^{\mathrm{as}}_{p,\sigma}(s)
                \right)
            \right)
        }{p^2}
        -
        \ell_\alpha(s)
    \right|
    \\
    &\qquad\leq
    \sup_{r\in K'}
    \left|
        \frac{
            2L\!\left(
                U_{p,\sigma}(t_{p,\sigma}(r))
            \right)
        }{p^2}
        -
        \ell_\alpha(r)
    \right|
    +
    \sup_{s\in K}
    \left|
        \ell_\alpha(\widehat s_p(s))
        -
        \ell_\alpha(s)
    \right|.
    \end{aligned}
    \label{eq:a5-explicit-loss-triangle}
\end{equation}

The first term is controlled by
Theorem~\ref{thm:universal-large-p}, specifically the equal-amplitude sector
large-system convergence
\eqref{eq:equal-diagonal-large-system-limit}. The second term converges
deterministically to zero by
\eqref{eq:a5-induced-exposure-convergence} and uniform continuity of
$\ell_\alpha$ on $K'$. Therefore, for every $\varepsilon>0$,
\begin{equation}
    \lim_{p\to\infty}
    \limsup_{\sigma\downarrow0}
    \mathbb P\!\left(
        \sup_{s\in K}
        \left|
            \frac{
                2L\!\left(
                    U_{p,\sigma}\!\left(
                        t^{\mathrm{as}}_{p,\sigma}(s)
                    \right)
                \right)
            }{p^2}
            -
            \ell_\alpha(s)
        \right|
        >
        \varepsilon
    \right)
    =
    0,
    \label{eq:a5-explicit-loss-final}
\end{equation}
which is \eqref{eq:addition-explicit-loss-limit} and completes the proof.
\end{proof}

\paragraph{Location and width of the learning transition.}
The explicit inverse also gives the physical-time scales stated after
Theorem~\ref{thm:explicit-addition-loss}. For every fixed $s>0$,
$\beta_p(s)\sim\log p$, and therefore
\begin{equation}
    t^{\mathrm{as}}_{p,\sigma}(s)
    \sim
    \frac{\sqrt3}{
        \sigma\sqrt{p\log p}
    },
\end{equation}
which gives the physical location of every fixed exposure level.

The separation between two fixed exposure levels is smaller. Let
$0<s_-<s_+<\infty$. Writing
\begin{equation}
    b_p
    :=
    \log p
    +\frac12\log\log p
    +\log A_3,
    \label{eq:a5-bp-definition}
\end{equation}
we have
\begin{equation}
    t^{\mathrm{as}}_{p,\sigma}(s_\pm)
    =
    \frac{\sqrt3}{\sigma\sqrt p}
    \bigl(
        b_p-\log s_\pm
    \bigr)^{-1/2}.
    \label{eq:a5-two-exposure-times}
\end{equation}
Applying the mean value theorem to $x\mapsto x^{-1/2}$ gives
\begin{equation}
    t^{\mathrm{as}}_{p,\sigma}(s_+)
    -
    t^{\mathrm{as}}_{p,\sigma}(s_-)
    =
    \frac{
        \sqrt3\,\log(s_+/s_-)
    }{
        2\sigma\sqrt p\,(\log p)^{3/2}
    }
    \bigl(1+o(1)\bigr),
\end{equation}
because every intermediate point between
$b_p-\log s_+$ and $b_p-\log s_-$ is asymptotic to $\log p$.

Thus the physical location of a fixed exposure level is of order
$(\sigma\sqrt{p\log p})^{-1}$, whereas the separation between two fixed
exposure levels is smaller by an additional factor of order
$(\log p)^{-1}$. At balanced width, $\alpha=1$ and
$\ell_1(s)=(1+s)^{-1}$, so the exposure level $s=1$ corresponds to the
asymptotic half-loss time.

\begin{figure}[t]
    \centering
    \includegraphics[scale=0.74]{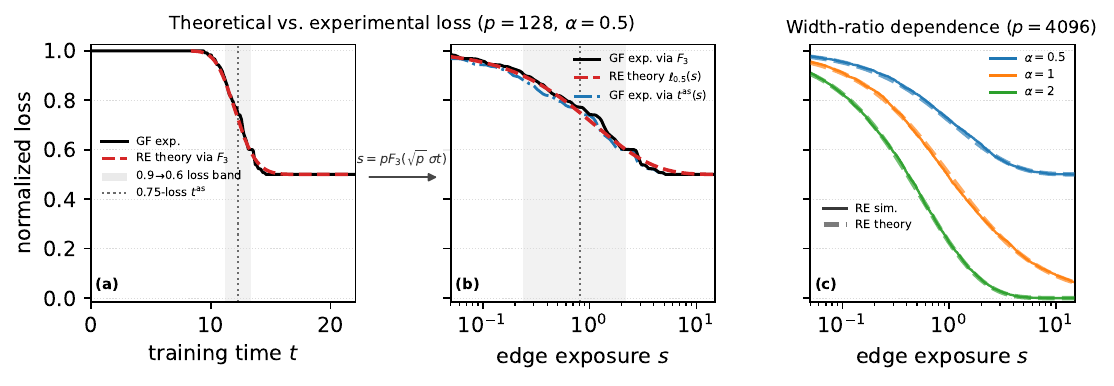}
    \includegraphics[scale=0.74]{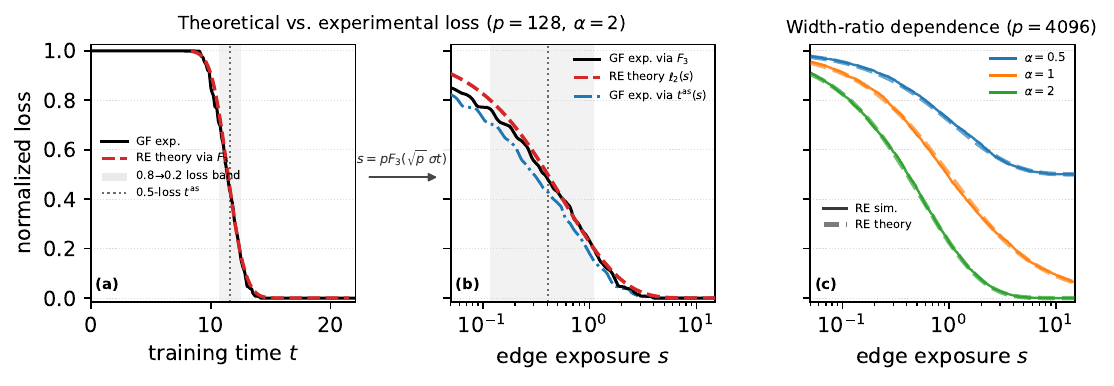}
    \caption{
\textbf{Large-system loss dynamics for complex modular addition: additional experiments with $\alpha=0.5$ and $\alpha=2$.} See Figure \ref{fig:large-system} in the main text for the case $\alpha=1$.
}
    \label{fig:large-system2}
\end{figure}

\paragraph{Experimental verification.} See Figure \ref{fig:large-system} in the main text for the case $\alpha=1$ and Figure \ref{fig:large-system2} for the cases $\alpha=0.5$ and $\alpha=2$.

\appendixsectionwithmain{Beyond scalar sectors: real modular addition details}
{app:real-addition}{sec:real-addition}

\subsection{From complex Fourier sectors to real Fourier blocks}
\label{app:real-fourier-blocks}

We begin by deriving the real Fourier decomposition underlying
Section~\ref{sec:real-addition}. Recall from
\eqref{eq:addition-fourier} that
\begin{equation}
    T_p^{\mathrm{add}}
    =
    \sqrt p
    \sum_{r\in\mathbb Z_p}
    \boldsymbol{\varphi}_r
    \otimes
    \boldsymbol{\varphi}_r
    \otimes
    \overline{\boldsymbol{\varphi}_r},
    \qquad
    \boldsymbol{\varphi}_r(x)
    =
    p^{-1/2}e^{2\pi i rx/p}.
    \label{eq:a6-complex-fourier-decomposition}
\end{equation}
Over $\mathbb C$, each summand is a scalar rank-one sector. Over
$\mathbb R$, however, the sectors indexed by $r$ and $-r$ must be combined,
because
\[
    \boldsymbol{\varphi}_{-r}
    =
    \overline{\boldsymbol{\varphi}_r}.
\]
We now make the resulting real block structure explicit.

For a frequency satisfying $r\not\equiv-r\pmod p$, define the normalized
real cosine and sine vectors
\begin{equation}
    \mathbf c_r
    :=
    \sqrt2\,\operatorname{Re}\boldsymbol{\varphi}_r,
    \qquad
    \mathbf s_r
    :=
    \sqrt2\,\operatorname{Im}\boldsymbol{\varphi}_r.
    \label{eq:a6-cosine-sine-vectors}
\end{equation}
Then $\mathbf c_r,\mathbf s_r$ are orthonormal and
\begin{equation}
    \boldsymbol{\varphi}_r
    =
    \frac{\mathbf c_r+i\mathbf s_r}{\sqrt2},
    \qquad
    \boldsymbol{\varphi}_{-r}
    =
    \frac{\mathbf c_r-i\mathbf s_r}{\sqrt2}.
    \label{eq:a6-complex-to-real-basis}
\end{equation}
For the three tensor modes, choose the real bases
\begin{equation}
    \mathbf e_r^{(1)}
    =
    \mathbf e_r^{(2)}
    =
    \mathbf e_r^{(3)}
    :=
    \mathbf c_r,
    \qquad
    \mathbf f_r^{(1)}
    =
    \mathbf f_r^{(2)}
    :=
    \mathbf s_r,
    \qquad
    \mathbf f_r^{(3)}
    :=
    -\mathbf s_r.
    \label{eq:a6-real-block-bases}
\end{equation}
The sign in the third mode reflects the conjugation of the third Fourier
factor in \eqref{eq:a6-complex-fourier-decomposition}. With this convention,
\begin{equation}
    \boldsymbol{\varphi}_r
    \otimes
    \boldsymbol{\varphi}_r
    \otimes
    \overline{\boldsymbol{\varphi}_r}
    =
    \frac{1}{2\sqrt2}
    \bigl(
        \mathbf e_r^{(1)}+i\mathbf f_r^{(1)}
    \bigr)
    \otimes
    \bigl(
        \mathbf e_r^{(2)}+i\mathbf f_r^{(2)}
    \bigr)
    \otimes
    \bigl(
        \mathbf e_r^{(3)}+i\mathbf f_r^{(3)}
    \bigr),
    \label{eq:a6-one-complex-sector}
\end{equation}
while the $-r$ term is its complex conjugate. Adding the two terms therefore
gives
\begin{equation}
    \begin{aligned}
    &
    \sqrt p
    \left(
        \boldsymbol{\varphi}_r
        \otimes
        \boldsymbol{\varphi}_r
        \otimes
        \overline{\boldsymbol{\varphi}_r}
        +
        \boldsymbol{\varphi}_{-r}
        \otimes
        \boldsymbol{\varphi}_{-r}
        \otimes
        \overline{\boldsymbol{\varphi}_{-r}}
    \right)
    \\
    &\qquad
    =
    \sqrt{\frac p2}\,
    B_r,
    \end{aligned}
    \label{eq:a6-conjugate-pair-block}
\end{equation}
where
\begin{equation}
    B_r
    :=
    \mathbf e_r^{(1)}
    \mathbf e_r^{(2)}
    \mathbf e_r^{(3)}
    -
    \mathbf e_r^{(1)}
    \mathbf f_r^{(2)}
    \mathbf f_r^{(3)}
    -
    \mathbf f_r^{(1)}
    \mathbf e_r^{(2)}
    \mathbf f_r^{(3)}
    -
    \mathbf f_r^{(1)}
    \mathbf f_r^{(2)}
    \mathbf e_r^{(3)}.
    \label{eq:a6-real-fourier-block}
\end{equation}
Thus every conjugate pair $\{r,-r\}$ becomes a two-dimensional real Fourier
block, and after suppressing the mode and frequency indices,
\eqref{eq:a6-real-fourier-block} is exactly the canonical tensor
\[
    B=eee-eff-fef-ffe
\]
used in \eqref{eq:real-fourier-block}.

The only exceptions are the self-conjugate frequencies
$r\equiv-r\pmod p$. The frequency $r=0$ is always self-conjugate, and when
$p$ is even so is $r=p/2$. Their Fourier vectors are real, so each gives a
one-dimensional scalar sector. Let
\begin{equation}
    \mathcal S_p
    :=
    \{r\in\mathbb Z_p:2r=0\pmod p\},
    \label{eq:a6-self-conjugate-frequencies}
\end{equation}
and let $\mathcal R_p$ contain one representative from each remaining pair
$\{r,-r\}$. Then the full real decomposition can be written as
\begin{equation}
    T_p^{\mathrm{add}}
    =
    \sqrt p
    \sum_{r\in\mathcal S_p}
    \boldsymbol{\varphi}_r
    \otimes
    \boldsymbol{\varphi}_r
    \otimes
    \boldsymbol{\varphi}_r
    +
    \sqrt{\frac p2}
    \sum_{r\in\mathcal R_p}
    B_r.
    \label{eq:a6-full-real-fourier-decomposition}
\end{equation}

The blocks in \eqref{eq:a6-full-real-fourier-decomposition} are mutually
orthogonal. Indeed, distinct conjugate pairs span orthogonal real Fourier
subspaces, while the four rank-one terms in each $B_r$ are themselves
mutually orthogonal. Consequently,
\begin{equation}
    \|B_r\|_F^2=4,
    \qquad
    \left\|
        \sqrt{\frac p2}\,B_r
    \right\|_F^2
    =
    2p.
    \label{eq:a6-real-block-energy}
\end{equation}
Since $\|T_p^{\mathrm{add}}\|_F^2=p^2$, every two-dimensional block carries
a fraction $2/p$ of the total target energy, whereas each scalar
self-conjugate sector carries a fraction $1/p$.

This decomposition makes precise why the use of complex weights was
essential in Sections~\ref{sec:modular-addition}--\ref{sec:large-p}.
Complex weights keep the frequencies $r$ and $-r$ as separate scalar
rank-one sectors. Restricting the weights to be real forces them into the
coupled block $B_r$, whose internal rank geometry is analyzed next.

\subsection{Geometry of the canonical two-dimensional block}
\label{app:real-block-geometry}

We now study the canonical tensor
\begin{equation}
    B
    =
    eee-eff-fef-ffe
    \in
    \mathbb R^2\otimes\mathbb R^2\otimes\mathbb R^2,
    \label{eq:a6-canonical-block}
\end{equation}
where $(\mathbf e,\mathbf f)$ is an orthonormal basis in each mode. As noted in
\eqref{eq:a6-real-block-energy},
\begin{equation}
    \|B\|_F^2=4.
    \label{eq:a6-canonical-block-norm}
\end{equation}
The block has a continuous phase symmetry that will account for the family
$(E_\Theta,F_\Theta)$ appearing in
Proposition~\ref{prop:real-block-rank-cascade}.

For $\theta\in\mathbb R$, define the rotated basis
\begin{equation}
    \mathbf e_\theta
    :=
    \cos\theta\,\mathbf e+\sin\theta\,\mathbf f,
    \qquad
    \mathbf f_\theta
    :=
    -\sin\theta\,\mathbf e+\cos\theta\,\mathbf f.
    \label{eq:a6-rotated-basis}
\end{equation}
Then
\begin{equation}
    \mathbf e_\theta+i\mathbf f_\theta
    =
    e^{-i\theta}(\mathbf e+i\mathbf f).
    \label{eq:a6-rotated-complex-vector}
\end{equation}
Since
\begin{equation}
    B
    =
    \operatorname{Re}
    \bigl[
        (\mathbf e+i\mathbf f)
        \otimes
        (\mathbf e+i\mathbf f)
        \otimes
        (\mathbf e+i\mathbf f)
    \bigr],
    \label{eq:a6-block-complex-form}
\end{equation}
independent rotations of the three modes leave $B$ unchanged whenever their
angles sum to zero modulo $2\pi$.

More precisely, let
\begin{equation}
    \Theta=(\theta_1,\theta_2,\theta_3),
    \qquad
    \theta_1+\theta_2+\theta_3
    \equiv0\pmod{2\pi},
    \label{eq:a6-admissible-phase}
\end{equation}
and define
\begin{align}
    E_\Theta
    &:=
    \mathbf e_{\theta_1}\otimes
    \mathbf e_{\theta_2}\otimes
    \mathbf e_{\theta_3},
    \label{eq:a6-Etheta}
    \\
    F_\Theta
    &:=
    \mathbf e_{\theta_1}\otimes
    \mathbf f_{\theta_2}\otimes
    \mathbf f_{\theta_3}
    +
    \mathbf f_{\theta_1}\otimes
    \mathbf e_{\theta_2}\otimes
    \mathbf f_{\theta_3}
    +
    \mathbf f_{\theta_1}\otimes
    \mathbf f_{\theta_2}\otimes
    \mathbf e_{\theta_3}.
    \label{eq:a6-Ftheta}
\end{align}
Expanding \eqref{eq:a6-block-complex-form} in the rotated bases gives
\begin{equation}
    B=E_\Theta-F_\Theta.
    \label{eq:a6-phase-decomposition}
\end{equation}
The four rank-one tensors on the right-hand side are mutually orthogonal.
Consequently,
\begin{equation}
    \|E_\Theta\|_F^2=1,
    \qquad
    \|F_\Theta\|_F^2=3,
    \qquad
    \langle E_\Theta,F_\Theta\rangle=0.
    \label{eq:a6-EF-norms}
\end{equation}

We next record two elementary properties of $B$ that will be useful in the
rank calculation. Define its spectral norm by
\begin{equation}
    \|B\|_{\mathrm{sp}}
    :=
    \sup_{\|\mathbf x_1\|=\|\mathbf x_2\|=\|\mathbf x_3\|=1}
    \left|
        \left\langle
            B,
            \mathbf x_1\otimes\mathbf x_2\otimes\mathbf x_3
        \right\rangle
    \right|.
    \label{eq:a6-block-spectral-norm-definition}
\end{equation}

\begin{lemma}[Basic geometry of the real Fourier block]
\label{lem:a6-real-block-geometry}
For the tensor $B$ in \eqref{eq:a6-canonical-block}:

\begin{enumerate}
    \item $\|B\|_{\mathrm{sp}}=1$.

    \item For every unit vector $\mathbf x$ in any one of the three modes, the
    contraction of $B$ against $\mathbf x$ is a $2\times2$ orthogonal matrix.
    In particular, both of its singular values are equal to $1$.
\end{enumerate}
\end{lemma}

\begin{proof}
Every unit vector in $\mathbb R^2$ can be written as
\[
    \mathbf x(\alpha)
    =
    \cos\alpha\,\mathbf e+\sin\alpha\,\mathbf f.
\]
A direct expansion of \eqref{eq:a6-canonical-block} gives
\begin{equation}
    \left\langle
        B,
        \mathbf x(\alpha)\otimes
        \mathbf x(\beta)\otimes
        \mathbf x(\gamma)
    \right\rangle
    =
    \cos(\alpha+\beta+\gamma).
    \label{eq:a6-block-trilinear-form}
\end{equation}
Its absolute value is at most $1$, and equality is attained, for example,
at $\alpha=\beta=\gamma=0$. Hence
$\|B\|_{\mathrm{sp}}=1$.

For the second claim, by symmetry it is enough to contract the first mode.
Writing
$\mathbf x(\alpha)=\cos\alpha\,\mathbf e+\sin\alpha\,\mathbf f$,
the resulting bilinear form has matrix
\begin{equation}
    B(\mathbf x(\alpha),\cdot,\cdot)
    =
    \begin{pmatrix}
        \cos\alpha & -\sin\alpha \\
        -\sin\alpha & -\cos\alpha
    \end{pmatrix}
    \label{eq:a6-block-contraction}
\end{equation}
in the bases $(\mathbf e,\mathbf f)$ of the second and third modes. This matrix satisfies
\[
    B(\mathbf x(\alpha),\cdot,\cdot)^\top
    B(\mathbf x(\alpha),\cdot,\cdot)
    =
    I_2,
\]
so it is orthogonal and has singular values $1,1$.
\end{proof}

The phase decomposition \eqref{eq:a6-phase-decomposition} already identifies
a family of best rank-one approximants to $B$. Indeed,
\eqref{eq:a6-canonical-block-norm} and
Lemma~\ref{lem:a6-real-block-geometry} imply that the smallest squared error
of a rank-one approximation is
\begin{equation}
    \|B\|_F^2-\|B\|_{\mathrm{sp}}^2
    =
    4-1
    =
    3,
    \label{eq:a6-best-rank-one-error}
\end{equation}
and each $E_\Theta$ attains this value, with residual
\begin{equation}
    B-E_\Theta=-F_\Theta.
    \label{eq:a6-first-cascade-residual}
\end{equation}
The next subsection shows that this residual lies on the boundary of the
real rank-two tensors and uses the contraction property above to determine
the complete rank cascade.

\subsection{Rank and border-rank cascade}
\label{app:real-block-rank-cascade}

We now prove Proposition~\ref{prop:real-block-rank-cascade}. Recall that
\[
    d_q(B)^2
    =
    \inf_{\operatorname{rank}_{\mathbb R}(X)\leq q}
    \|B-X\|_F^2 .
\]
The values for ranks zero and one follow directly from the geometry
established above. The rank-two case is more subtle: its optimal error is an
infimum that is not attained.

\begin{proof}[Proof of Proposition~\ref{prop:real-block-rank-cascade}]
~ \paragraph{Ranks zero and one.}
By \eqref{eq:a6-canonical-block-norm},
\[
    d_0(B)^2=\|B\|_F^2=4.
\]
For a unit rank-one tensor
$\mathbf x_1\otimes\mathbf x_2\otimes\mathbf x_3$, optimizing also over its
scalar coefficient gives squared approximation error
\[
    \|B\|_F^2
    -
    \left|
        \langle
            B,
            \mathbf x_1\otimes\mathbf x_2\otimes\mathbf x_3
        \rangle
    \right|^2.
\]
Since Lemma~\ref{lem:a6-real-block-geometry} gives
$\|B\|_{\mathrm{sp}}=1$, we obtain
\begin{equation}
    d_1(B)^2=4-1=3.
    \label{eq:a6-rank-one-error}
\end{equation}
Moreover, for every admissible phase $\Theta$ from
\eqref{eq:a6-admissible-phase}, the rank-one tensor $E_\Theta$ attains this
value. By \eqref{eq:a6-phase-decomposition},
\begin{equation}
    B-E_\Theta=-F_\Theta,
    \label{eq:a6-rank-one-residual}
\end{equation}
so $-F_\Theta$ is the residual after the first stage of the cascade.

\paragraph{Rank-two lower bound.}
Let $X$ be any real tensor of CP rank at most two. Write
\begin{equation}
    X
    =
    \mathbf a_1\otimes\mathbf b_1\otimes\mathbf c_1
    +
    \mathbf a_2\otimes\mathbf b_2\otimes\mathbf c_2.
    \label{eq:a6-rank-two-decomposition}
\end{equation}
Choose a unit vector $\mathbf u$ in the first mode orthogonal to
$\mathbf a_1$. Then
\[
    X(\mathbf u,\cdot,\cdot)
    =
    \langle \mathbf u,\mathbf a_2\rangle\,
    \mathbf b_2\otimes\mathbf c_2
\]
has matrix rank at most one. On the other hand,
Lemma~\ref{lem:a6-real-block-geometry} shows that
$B(\mathbf u,\cdot,\cdot)$ is a $2\times2$ orthogonal matrix, whose two
singular values are both equal to one. Its Frobenius distance from the set of
rank-one matrices is therefore exactly one. Hence
\begin{equation}
    \|B-X\|_F
    \geq
    \|B(\mathbf u,\cdot,\cdot)-X(\mathbf u,\cdot,\cdot)\|_F
    \geq1.
    \label{eq:a6-rank-two-lower-bound}
\end{equation}
Thus
\begin{equation}
    d_2(B)^2\geq1.
    \label{eq:a6-rank-two-lower-bound-squared}
\end{equation}

In fact, equality in \eqref{eq:a6-rank-two-lower-bound} cannot hold for any
rank-two tensor $X$. To see this, suppose that
$\|B-X\|_F=1$ and write $R:=B-X$.

If $\mathbf a_1$ and $\mathbf a_2$ in
\eqref{eq:a6-rank-two-decomposition} are linearly dependent, choose a unit
vector $\mathbf u$ orthogonal to their common span. Then
$X(\mathbf u,\cdot,\cdot)=0$, whereas
$B(\mathbf u,\cdot,\cdot)$ is orthogonal and therefore has Frobenius norm
$\sqrt2$. This would give $\|R\|_F\geq\sqrt2$, a contradiction.

Suppose instead that $\mathbf a_1$ and $\mathbf a_2$ are linearly
independent, and let $\mathbf u_j$ be a unit vector orthogonal to
$\mathbf a_j$, $j=1,2$. Applying
\eqref{eq:a6-rank-two-lower-bound} with $\mathbf u_1$ and $\mathbf u_2$
gives
\[
    \|R(\mathbf u_j,\cdot,\cdot)\|_F\geq1=\|R\|_F,
    \qquad j=1,2.
\]
Contraction by a unit vector cannot increase the Frobenius norm, so equality
must hold in both cases. If $\mathbf v_j$ is a unit vector orthogonal to
$\mathbf u_j$, then
\[
    \|R\|_F^2
    =
    \|R(\mathbf u_j,\cdot,\cdot)\|_F^2
    +
    \|R(\mathbf v_j,\cdot,\cdot)\|_F^2
\]
implies $R(\mathbf v_j,\cdot,\cdot)=0$. Thus the first-mode support of $R$
is contained in $\operatorname{span}\{\mathbf u_j\}$ for both $j=1$ and
$j=2$. Since $\mathbf u_1$ and $\mathbf u_2$ are not parallel, this forces
$R=0$, again a contradiction. Therefore the rank-two lower bound is never
attained.

\paragraph{Rank-two upper bound and border rank.}
It remains to show that the lower bound can be approached. It is enough to
consider $\Theta=0$, for which
\begin{equation}
    E=eee,
    \qquad
    F=eff+fef+ffe,
    \qquad
    B=E-F.
    \label{eq:a6-canonical-EF}
\end{equation}
For $\varepsilon\neq0$, define
\begin{equation}
    X_\varepsilon
    :=
    -\frac{
        (\mathbf f+\varepsilon\mathbf e)^{\otimes3}
        -
        \mathbf f^{\otimes3}
    }{\varepsilon}.
    \label{eq:a6-border-rank-sequence}
\end{equation}
Each $X_\varepsilon$ has real CP rank at most two. Expanding the numerator
gives
\begin{equation}
    X_\varepsilon
    =
    -F
    -
    \varepsilon
    \bigl(
        eef+efe+fee
    \bigr)
    -
    \varepsilon^2 eee,
    \label{eq:a6-border-rank-expansion}
\end{equation}
and therefore
\begin{equation}
    X_\varepsilon\longrightarrow-F
    \qquad
    \text{as }\varepsilon\to0.
    \label{eq:a6-border-rank-limit}
\end{equation}
Consequently,
\[
    B-X_\varepsilon
    \longrightarrow
    B+F
    =
    E,
\]
so
\begin{equation}
    d_2(B)^2
    \leq
    \|E\|_F^2
    =
    1.
    \label{eq:a6-rank-two-upper-bound}
\end{equation}
Together with \eqref{eq:a6-rank-two-lower-bound-squared}, this proves
\begin{equation}
    d_2(B)^2=1.
    \label{eq:a6-rank-two-error}
\end{equation}
Since the lower bound is not attained, the infimum in
\eqref{eq:a6-rank-two-error} is not attained.

The same construction applies to every admissible phase $\Theta$ after
rotating the three mode-wise bases. Hence $-F_\Theta$ lies in the closure
of the real rank-two tensors, and the corresponding limiting residual is
\begin{equation}
    B-(-F_\Theta)=E_\Theta.
    \label{eq:a6-rank-two-residual}
\end{equation}

\paragraph{Degeneration of the rank-two representation.}
For completeness, we verify that the border-rank approximation above
necessarily involves diverging rank-one terms. First observe that
$F_\Theta$ has real CP rank three. By the phase symmetry, it is enough to
consider $F$ from \eqref{eq:a6-canonical-EF}. Its first-mode slices are
\begin{equation}
    F(\mathbf e,\cdot,\cdot)
    =
    \begin{pmatrix}
        0&0\\
        0&1
    \end{pmatrix},
    \qquad
    F(\mathbf f,\cdot,\cdot)
    =
    \begin{pmatrix}
        0&1\\
        1&0
    \end{pmatrix}.
    \label{eq:a6-F-slices}
\end{equation}
A general matrix in their two-dimensional span is
\[
    \begin{pmatrix}
        0&y\\
        y&x
    \end{pmatrix},
\]
whose determinant is $-y^2$. Hence this slice space contains only one
rank-one direction, namely the multiples of
$F(\mathbf e,\cdot,\cdot)$. If $F$ had CP rank at most two, its
two-dimensional slice space would be spanned by two rank-one matrices, which
is impossible. Thus
\begin{equation}
    \operatorname{rank}_{\mathbb R}(F_\Theta)=3.
    \label{eq:a6-F-rank-three}
\end{equation}

Now let
\[
    Y_n=A_n+C_n\longrightarrow-F_\Theta,
\]
where $A_n$ and $C_n$ are rank-one tensors. If both sequences of rank-one
summands were bounded in Frobenius norm, then after passing to a subsequence
they would converge to tensors $A$ and $C$ of rank at most one, because the
rank-one cone is closed in finite dimension. This would give
$-F_\Theta=A+C$ and hence
$\operatorname{rank}_{\mathbb R}(F_\Theta)\leq2$, contradicting
\eqref{eq:a6-F-rank-three}.

In fact, both summand norms must diverge. If, for example,
$\|A_n\|_F$ did not tend to infinity, then along some subsequence it would
remain bounded. Since $Y_n\to-F_\Theta$ is bounded, the identity
$C_n=Y_n-A_n$ would make $\|C_n\|_F$ bounded along the same subsequence,
contradicting the preceding argument. The same reasoning applies with
$A_n$ and $C_n$ interchanged. Therefore
\[
    \|A_n\|_F\longrightarrow\infty,
    \qquad
    \|C_n\|_F\longrightarrow\infty.
\]
Thus every rank-two sequence converging to $-F_\Theta$ must degenerate
through cancellation of two unbounded rank-one terms. The explicit sequence
\eqref{eq:a6-border-rank-sequence} exhibits this behavior directly, with
both summands of order $|\varepsilon|^{-1}$.

\paragraph{Rank three.}
Finally, the identity
\begin{equation}
    B
    =
    2\mathbf e^{\otimes3}
    -
    \frac12(\mathbf e+\mathbf f)^{\otimes3}
    -
    \frac12(\mathbf e-\mathbf f)^{\otimes3}
    \label{eq:a6-rank-three-decomposition}
\end{equation}
shows that $\operatorname{rank}_{\mathbb R}(B)\leq3$. Since
$d_2(B)^2=1>0$, the rank cannot be at most two. Hence
\[
    \operatorname{rank}_{\mathbb R}(B)=3,
    \qquad
    d_3(B)^2=0.
\]
Together with \eqref{eq:a6-rank-one-error} and
\eqref{eq:a6-rank-two-error}, this proves
\eqref{eq:real-block-rank-errors} and completes the proof.
\end{proof}

The proposition therefore gives the sequence of infimal residual states
\[
    B
    \longrightarrow
    -F_\Theta
    \longrightarrow
    E_\Theta
    \longrightarrow
    0.
\]
The second transition is fundamentally different from an ordinary
finite-rank approximation step: $E_\Theta$ is reached only as the residual
of a degenerating rank-two representation. This algebraic feature provides
a natural geometric explanation for the large parameter growth observed
during the middle stage of the real-sector cascades in
Figure~\ref{fig:cascades}.

\clearpage
\phantomsection
\label{app:glossary}
\addcontentsline{toc}{section}{Glossary}
\section*{Glossary}

This glossary summarizes the terminology used throughout
the paper. It is intended as a conceptual guide; precise
mathematical definitions are given in the referenced
sections and equations.

\begin{description}

\item[Accepted edge.]
An atom--sector pair $(k,i)$ selected by the greedy
matching process. A pair is accepted if neither its
atom nor its sector has been claimed by an earlier
accepted pair. The accepted edges are denoted by
$(k_r,i_r)$; see \eqref{eq:a3-accepted-edges}
and \eqref{eq:a3-next-greedy-edge}.

\item[Activation clock.]
The candidate activation time of an atom--sector
pair, measured in the slow time
$\theta=\sigma^{\nu-2}t$.
For pair $(k,i)$, it is defined by
$\tau_{ki}=\tau_\nu(\mathbf g_{ki})/\lambda_i$,
where $\tau_\nu$ is the blow-up time of the
isolated scalar flow; see \eqref{eq:scalar-clock}
and \eqref{eq:candidate-clock}.
An activation clock predicts when the pair
would become macroscopic in the
vanishing-initialization limit if its atom and
sector remained available. Not every candidate
clock results in an actual learning event.

\item[Atom.]
A single rank-one summand
$\mathbf u_k^{(1)}\otimes\cdots\otimes
\mathbf u_k^{(\nu)}$
of the CP model. Each atom consists of $\nu$
trainable vectors, one for each mode.
The model contains $H$ atoms indexed by
$k\in[H]$; see \eqref{eq:cp-model}.

\item[Atom--sector pair.]
A possible assignment of model atom $k$ to target
sector $i$, represented by an edge $(k,i)$
of the bipartite graph $K_{H,p}$.
Each pair has a candidate activation clock
$\tau_{ki}$; see \eqref{eq:candidate-clock}.

\item[Available edge.]
An atom--sector pair $(k,i)$ whose atom and
sector have not yet been claimed.
After $m$ accepted edges, the set of available
pairs is
$\mathcal E_m=([H]\setminus\mathcal A_m)
\times([p]\setminus\mathcal S_m)$;
see \eqref{eq:a3-available-edges}.

\item[Bipartite graph.]
A graph whose vertices are divided into two
disjoint sets, with edges connecting vertices
from different sets only. In the scalar-sector
theory, we consider the complete bipartite graph
$K_{H,p}$, whose two vertex sets consist of
$H$ model atoms and $p$ target sectors.
Its $Hp$ edges represent all possible
atom--sector pairs $(k,i)$.
See Section~\ref{sec:greedy}.

\item[Captured atom and sector.]
An atom and a target sector associated with an
accepted edge $(k_r,i_r)$. After the corresponding
fast transition, the captured atom approximately
represents the target component in sector $i_r$.
The sets of captured atoms and sectors after $m$
events are denoted by $\mathcal A_m$ and
$\mathcal S_m$, respectively; see
\eqref{eq:a3-claimed-sets}.

\item[Diagonal target.]
A tensor admitting a decomposition
$T=\sum_{i=1}^p\lambda_i
\mathbf e_i^{(1)}\otimes\cdots\otimes
\mathbf e_i^{(\nu)}$,
where the vectors $\mathbf e_i^{(a)}$ form
an orthonormal basis in each mode $a$.
The bases may differ between modes and need
not coincide with the original coordinate bases.
See \eqref{eq:diagonal-target}.

\item[Edge exposure.]
The parameter $s$ describing how many candidate
edges have been processed in the large-system
Random Edge process. At exposure $s$, each edge
is independently exposed with probability $s/p$.
Thus $s$ is the expected number of exposed edges
incident to each atom vertex. The conversion
from GF training time to edge exposure is given
in \eqref{eq:exposure-training-time}.

\item[Equal-amplitude sectors.]
Scalar sectors whose target coefficients
$\lambda_i$ are all equal. In this case,
the candidate activation clocks $\tau_{ki}$
are independent and identically distributed,
so their ordering is equivalent to a uniformly
random ordering of the edges of $K_{H,p}$.
See Section~\ref{sec:large-p}.

\item[Fast transition.]
A learning stage during which a previously
microscopic atom grows to macroscopic size and
captures a target sector. Its duration vanishes
in the slow time $\theta$ as $\sigma\to0$.
The fast-capture analysis is given in
Appendix~\ref{app:fast-capture}. In Figure \ref{fig:scalar-sectors}, fast transitions correspond to the approximately vertical pieces of the loss staircase. 

\item[Greedy matching.]
The procedure that processes atom--sector pairs
in increasing order of their candidate activation
clocks $\tau_{ki}$, accepting a pair whenever
both endpoints are still available.
Theorem~\ref{thm:greedy-matching-limit} proves
that this procedure determines the limiting
learning trajectory of ordinary CP gradient flow
for diagonal targets. See also
\eqref{eq:a3-next-greedy-edge}.

\item[Locking.]
The stabilization of a captured atom near its
assigned scalar sector. Quantitatively, its
main-coordinate product
$\pi_{r,\sigma}=\prod_a q_{r,\sigma}^{(a)}$
approaches the target coefficient $\lambda_{i_r}$,
while its off-sector components
$\mathbf h_{r,\sigma}^{(a)}$ become small.
See \eqref{eq:a3-locked-q-pi},
\eqref{eq:a3-locked-h}, and
Lemma~\ref{lem:a3-fast-locking}.

\item[Macroscopic, mesoscopic, and microscopic scales.]
These terms describe the magnitude of weights
as the initialization scale $\sigma\to0$.
Microscopic weights have magnitude $O(\sigma)$.
Mesoscopic weights have an intermediate magnitude
$r_\sigma$ satisfying
$\sigma\ll r_\sigma\ll1$.
Macroscopic weights have magnitude of order one,
bounded away from zero.
See \eqref{eq:a3-fast-mesoscopic-properties}
and Appendix~\ref{app:fast-capture}.

\item[Mode.]
One of the $\nu$ tensor factors in an order-$\nu$
tensor. Modes are indexed by $a\in[\nu]$.
Each CP atom has one trainable vector
$\mathbf u_k^{(a)}$ in each mode, and the
diagonal target has a corresponding orthonormal
basis $\{\mathbf e_i^{(a)}\}_{i=1}^p$.
See \eqref{eq:cp-model} and
\eqref{eq:diagonal-target}.

\item[Physical time and slow time.]
Physical time $t$ is the original, unrescaled
time variable of gradient flow in
\eqref{eq:loss-gf}. Slow time is
$\theta=\sigma^{\nu-2}t$, the rescaled variable
in which the candidate activation clocks remain
finite as $\sigma\to0$.
During a fast transition, the local physical-time
increment $\xi$ is related to slow time by
\eqref{eq:a3-fast-slow-time-map}.
Physical time does not mean computational
wall-clock time.

\item[Random Edge.]
The random-order greedy matching algorithm:
edges of a bipartite graph are processed in
uniformly random order and accepted whenever
both endpoints are unmatched.
For equal-amplitude sectors, the greedy matching
process induced by vanishing-initialization GF
has the same distribution as Random Edge.
Its large-system limit is given in
Proposition~\ref{prop:random-edge-limit}.
Random Edge is not a separate algorithm
used to train the tensor model.

\item[Real Fourier block.]
A tensor component obtained by combining a
pair of conjugate Fourier frequencies in the
real representation of modular addition.
Its underlying mode-wise subspaces are
two-dimensional, unlike those of a scalar sector.
Such a block has a nontrivial internal tensor-rank
structure; see \eqref{eq:real-fourier-block}
and Proposition~\ref{prop:real-block-rank-cascade}.

\item[Reset time.]
A checkpoint $\rho_m^\sigma$ introduced in the
induction proof after the $m$th fast transition.
At this time, the first $m$ accepted atoms are
locked to their sectors, while the remaining
atoms are microscopic.
The GF dynamics itself is not restarted;
see \eqref{eq:a3-reset-times}.

\item[Scalar sector.]
A sector that is one-dimensional in every mode.
In the diagonal-target setting, sector $i$
is spanned by
$\mathbf e_i^{(1)}\otimes\cdots\otimes
\mathbf e_i^{(\nu)}$
and carries the target component
$\lambda_i\mathbf e_i^{(1)}\otimes\cdots
\otimes\mathbf e_i^{(\nu)}$.
See \eqref{eq:diagonal-target}.
Theorem~\ref{thm:greedy-matching-limit}
concerns precisely such scalar sectors.

\item[Sector.]
A tensor-product subspace
$V^{(1)}\otimes\cdots\otimes V^{(\nu)}$,
where $V^{(a)}$ is a subspace of the vector
space associated with mode $a$.
A scalar sector has
$\dim V^{(a)}=1$ for every mode.
More generally, the subspaces may have
higher dimensions, as in the two-dimensional
real Fourier blocks discussed in
Section~\ref{sec:real-addition}.
The rigorous greedy matching theory developed
in this paper applies to scalar sectors.

\item[Width ratio.]
The ratio $H_p/p$ between the number of model
atoms and the number of scalar target sectors.
In the large-system limit,
$H_p/p\to\alpha\in(0,\infty)$.
The limiting normalized loss depends on
$\alpha$ through the function
$\ell_\alpha$; see
Proposition~\ref{prop:random-edge-limit}
and Theorem~\ref{thm:universal-large-p}.

\end{description}

\end{document}